\documentclass[11pt,titlepage,oneside,openany]{book}
\usepackage{times}

\usepackage{graphicx}
\usepackage{latexsym}
\usepackage{amsmath}
\usepackage{amssymb}
\usepackage{microtype}
\usepackage{libertine}
\usepackage{subcaption}
\usepackage{caption}
\usepackage{tikz}
\usepackage{pgfplots}

\usetikzlibrary{external}
\usepgfplotslibrary{external}
\usepackage{xcolor}
\usepackage{comment}
\usetikzlibrary{positioning}

\makeatletter
\edef\@figdir{pictures}

\catcode`\#=11
\catcode`\|=6
\newcommand{\@basicpgfpreamble}[1]{%
	\unexpanded{%
		\documentclass{standalone}^^J
		\usepackage{pgf}^^J
		\let\oldpgfimage\pgfimage^^J
		\renewcommand{\pgfimage}[2][]{\oldpgfimage[#1]{|1/#2}}^^J
	}%
}
\catcode`\#=6
\catcode`\|=12

\let\@pgfpreamble\@basicpgfpreamble

\newcommand{\setpgfpreamble}[1]{%
	\renewcommand{\@pgfpreamble}[1]{\@basicpgfpreamble{##1}\unexpanded{#1}}
}

\newcounter{@pgfcounter}
\newwrite\@pgfout
\newread\@pgfin

\newcommand{\importpgf}[3][]{%
	\edef\@figfile{temppic-\the@pgfcounter}%

	\IfFileExists{\@figdir/\@figfile.pdf}%
	{\includegraphics[#1]{\@figdir/\@figfile.pdf}}%
	{\errmessage{Error during compilation of figure #2/#3}}%
	\stepcounter{@pgfcounter}%
}
\makeatother

\setpgfpreamble{%
	\usepackage{amsmath}
}

\usepackage{etoolbox} 
\usepackage{listofitems} 

\usepackage{mathtools}

\usepackage{acronym}

\tikzset{>=latex} 
\colorlet{myred}{blue!80!black}
\colorlet{myblue}{blue!40!red}
\colorlet{mygreen}{green!60!black}
\colorlet{mydarkred}{myred!40!black}
\colorlet{mydarkblue}{myblue!40!black}
\colorlet{mydarkgreen}{mygreen!40!black}
\colorlet{MyColorOne}{blue!50}
\newcommand{\lightercolor}[3]{
    \colorlet{#3}{#1!#2!white}
}
\definecolor{beaublue}{rgb}{0.74, 0.83, 0.9}
\definecolor{blanchedalmond}{rgb}{1.0, 0.92, 0.8}
\definecolor{khaki}{rgb}{0.76, 0.69, 0.57}
\definecolor{battleshipgrey}{rgb}{0.52, 0.52, 0.51}
\definecolor{asparagus}{rgb}{0.53, 0.66, 0.42}
\definecolor{palechestnut}{rgb}{0.87, 0.68, 0.69}
\definecolor{ashgrey}{rgb}{0.7, 0.75, 0.71}
\definecolor{royalblue}{rgb}{0.25, 0.41, 0.88}
\definecolor{amber}{rgb}{1.0, 0.49, 0.0}
\definecolor{inchworm}{rgb}{0.7, 0.93, 0.36}
\definecolor{anti-flashwhite}{rgb}{0.95, 0.95, 0.96}

\lightercolor{MyColorOne}{50}{MyColorOneLight}
\tikzstyle{node}=[thick,circle,draw=myblue,minimum size=22,inner sep=0.5,outer sep=0.6]
\tikzstyle{connect}=[->,thick,mydarkblue,shorten >=1]
\tikzset{ 
  node 1/.style={node,mydarkgreen,draw=black,fill=mygreen!25},
  node 2/.style={node,mydarkblue,draw=black,fill=blanchedalmond},
  node 3/.style={node,mydarkred,draw=black,fill=beaublue},
}
\def\nstyle{int(\lay<\Nnodlen?min(2,\lay):3)} 

\usepackage{ntheorem}

\usepackage{tabularx}

\usepackage{arydshln}

\usepackage{algorithm}
\usepackage{algorithmic}
\usepackage{csquotes}

\usepackage[hidelinks]{hyperref}

\newcommand\norm[1]{\lVert#1\rVert}
\newcommand{\abs}[1]{\left\lvert #1 \right\rvert}

\begin{document}

\pagenumbering{roman}
\begin{titlepage}
	\vspace*{2cm}
  \begin{center}
   {\Large Dynamic interval restrictions on action spaces \\in deep reinforcement learning for obstacle avoidance\\}
   \vspace{2cm} 
   {Master Thesis\\}
   \vspace{2cm}
   {presented by\\
    Tim Grams \\
    Matriculation Number 1722252\\
   }
   \vspace{1cm} 
   {submitted to the\\
    Institute for Enterprise Systems\\
    Dr.\ Christian Bartelt\\
    University of Mannheim\\} \vspace{2cm}
   {April 2023}
  \end{center}
\end{titlepage} 

\chapter*{Abstract}
Deep reinforcement learning algorithms typically act on the same set of actions. However, this is not sufficient for a wide range of real-world applications where different subsets are available at each step. In this thesis, we consider the problem of interval restrictions as they occur in pathfinding with dynamic obstacles. When actions that lead to collisions are avoided, the continuous action space is split into variable parts. Recent research learns with strong assumptions on the number of intervals, is limited to convex subsets, and the available actions are learned from the observations. Therefore, we propose two approaches that are independent of the state of the environment by extending parameterized reinforcement learning and ConstraintNet to handle an arbitrary number of intervals. We demonstrate their performance in an obstacle avoidance task and compare the methods to penalties, projection, replacement, as well as discrete and continuous masking from the literature. The results suggest that discrete masking of action values is the only effective method when constraints did not emerge during training. When restrictions are learned, the decision between projection, masking, and our MPS-TD3 approach seems to depend on the task at hand. We compare the results with varying complexity and give directions for future work. 

\tableofcontents
\newpage

\listofalgorithms

\listoffigures

\listoftables

\chapter*{List of Abbreviations}
\begin{acronym}[MPS-TD3]
\acro{DDPG}{Deep Deterministic Policy Gradients}
\acro{DL}{Deep Learning}
\acro{DQN}{Deep Q-Network}
\acro{MDP}{Markov Decision Process}
\acro{MPS-TD3}{Multi-Pass Scaled Twin Delayed Deep Deterministic Policy Gradients}
\acro{MSBE}{Bellman Mean Squared Error}
\acro{PAM}{Parameterized Action Masking}
\acro{PPO}{Proximal Policy Optimization}
\acro{RL}{Reinforcement Learning}
\acro{TD3}{Twin Delayed Deep Deterministic Policy Gradients}
\end{acronym} 


\newpage

\pagenumbering{arabic}

\chapter{Introduction}
\label{cha:intro}
Deep reinforcement learning (RL) has shown success in various fields such as the internet of things \cite{chen2021deep}, finance \cite{hu2019deep} and autonomous driving \cite{kiran2021deep}.
An example is AlphaGo surpassing professional players and defeating the European champion in the board game Go in 2016 \cite{silver2016mastering}. The objective of RL is to maximize a cumulative reward in an unknown environment by trial and error \cite{sutton2018reinforcement}. This can range from the monetary return of a trading algorithm \cite{zhang2020deep} to an artificial reward for controlling robots in a factory \cite{ibarz2021train}. Applications include discrete as well as continuous problems and the algorithms have proven to learn effectively in environments with mostly static action spaces. 
In doing so, the agent acts consistently on the same set of actions \cite{sutton2018reinforcement}. 
However, this is not sufficient for a wide range of real-world applications where different sets are available at each step. The action space results from restrictions that occur naturally \cite{ozhamaratli2022deep} or are given by security constraints \cite{achiam2017constrained} and second authorities \cite{pernpeintner2021governing}. For example, spending more money than an agent owns is not possible and the budget may vary between time steps \cite{ozhamaratli2022deep}. Restrictions might change depending on the state of the environment and the temporary action space can have an arbitrary form.  

\section{Problem Statement}
\begin{figure}[t]
\centering
\begin{tikzpicture}[scale=1.25]
    \begin{scope}[xshift = -0.5cm]
        \draw(-3.75,0) rectangle (0,2.5);
    
        \filldraw[color=asparagus, fill=asparagus!10, very thick](-0.5,2) circle (0.25);
    
        \filldraw[fill = battleshipgrey!15, thick, rotate around = {15:(-3,0.8)}] (-3,0.3) arc(-90:90:0.5);
        \node[inner sep=0pt] (russell) at (-3,0.8) {\includegraphics[width=.04\textwidth]{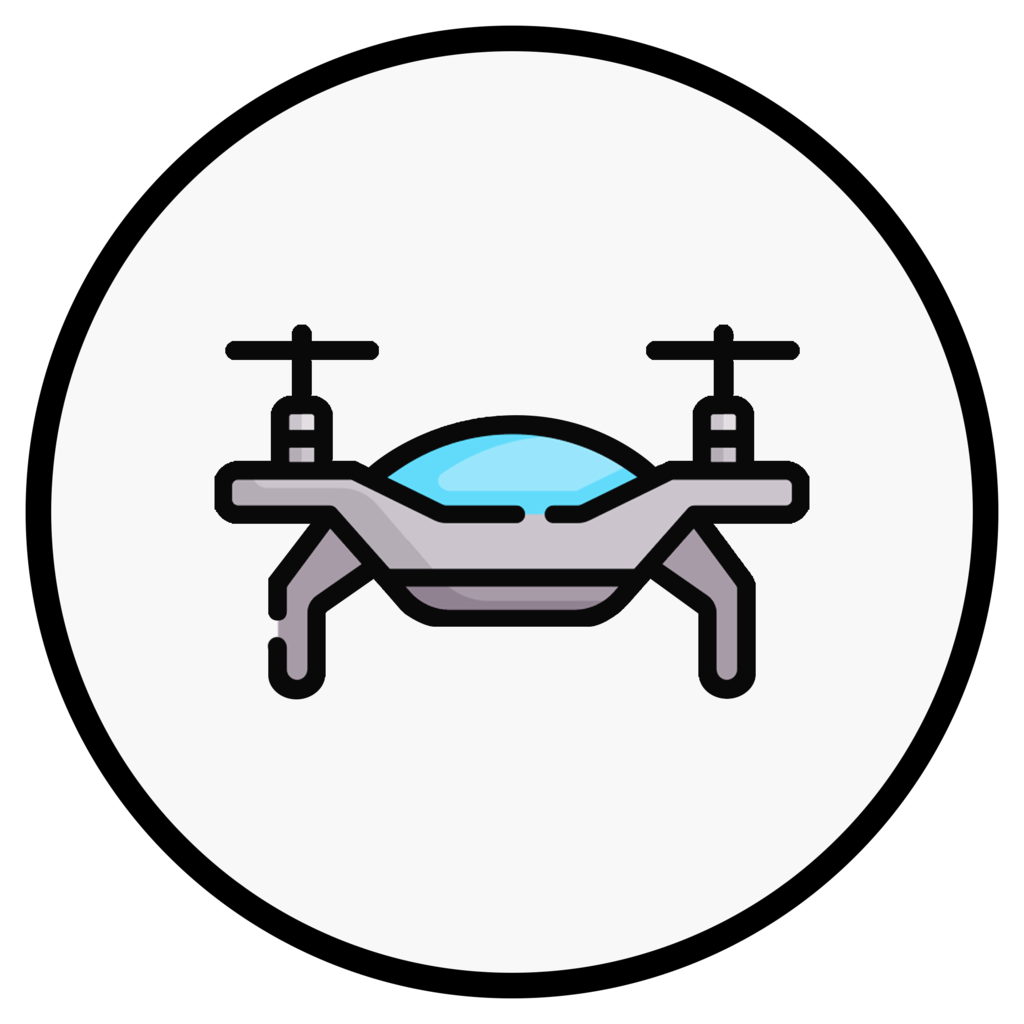}};
    
        \filldraw[fill = battleshipgrey](-2.2,0.25) rectangle (-2.7,1.5);
        \filldraw[fill = battleshipgrey](-1.2,0.75) rectangle (-1.7,2);
    \end{scope}

    \draw (1,1.25) -- (0.5,2);
    \draw (1,1.25) -- (0.5,0.5);
    
    \begin{scope}[xshift = 0.5cm]
        \filldraw[inchworm, fill = battleshipgrey!15, very thick] (4.5,0.5) arc(0:180:1.5);
        \draw[very thick, rotate around = {25:(3,0.5)}] (4.5,0.5) arc(0:105:1.5);
        \draw[inchworm, very thick] (4.5,0.5) arc(0:25:1.5);
        \node[inner sep=0pt] (russell) at (3,0.5) {\includegraphics[width=.06\textwidth]{images/drone.png}};
    
        \node(hi) at (1.33,1.25) {\footnotesize (1)};
        \node(hi) at (3.6,2.2) {\footnotesize (2)};
        \node(hi) at (4.8,0.85) {\footnotesize (3)};
    \end{scope}
\end{tikzpicture}
\caption[Obstacle avoidance with dynamic interval restrictions]{Obstacle avoidance with dynamic interval restrictions \cite{deus}}
\label{fig:intro-obstacle-avoidance}
\end{figure}

RL algorithms with interval action space restrictions constitute one of the research challenges. The sets produce subsets of disjoint intervals and are of particular importance since they arise in many real-world applications. 
The significance is best demonstrated by considering a pathfinding example. Figure \ref{fig:intro-obstacle-avoidance} illustrates a drone trying to reach the circular goal in the top right corner while avoiding the grey rectangular obstacles. For simplicity reasons, we assume that the drone has a fixed step size and can only move in a forward direction at each time step. Therefore, the next location will be at any point on the black semicircle. However, when the drone is not supposed to collide with an obstacle, the available actions reduce to the two disjoint intervals (1) and (3) on the right side of the figure. The concept is not far from reality. Barriers have been shown to be effectively detectable by sensors \cite{meyer2020taming}.
In the case of varying obstacles, the restrictions can be considered dynamic: The same position leads to different allowed subsets.
Nevertheless, such a variable action space is not covered by state-of-the-art RL algorithms.
Recent research kept all actions available and invalid selections were handled in different ways such as ignoring \cite{huang2019comparing} or penalizing them with reward shaping \cite{hu2021obstacle,bouhamed2020autonomous}.
However, the fixed input of models such as neural networks hinders observations in including constraints without strong assumptions on the number of possible intervals. An example is defining the length of zero-padding before training. Algorithms with architectures respecting dynamic restrictions would be more useful: The output range can change but the observation is the same. In the example, this would mean that the drone does not observe obstacles but finds optimal paths over varying subsets of the action space.
Discrete masking has shown high effectiveness \cite{huang2020closer} so that naturally continuous problems have been discretized using domain knowledge \cite{choi2022marl}. 
Fields such as safe RL restrict the action space but focus on the derivation of constraints, and do not agree on a method to implement the restrictions \cite{taylor2020learning}. Some search for the closest allowed action, others sample from the available space, and applications were limited to convex subsets \cite{saferloverview}. 
Another downside of the continuous methods is that the available parts are learned implicitly from the observations. Therefore, the feasible actions are fixed. For example, a safety controller always finds the same safe subset for a state \cite{saferloverview}.
A common solution to handle multiple intervals dynamically is missing.  
To fill this gap in the literature, existing approaches have to be compared and, if necessary, extended. The results can help in suggesting techniques and overcoming the limitation of convex and static restrictions. 
Since obstacle avoidance naturally produces action spaces of dynamic and disjoint intervals, it can provide a starting point and serve as the foundation for other fields.

\section{Research Questions}
The aim of this thesis is to compare and find techniques for continuous deep RL algorithms to work with dynamic interval action space restrictions. The focus will be on obstacle avoidance. This includes finding answers to the following research questions: 

\begin{enumerate}
    \item How can continuous deep RL algorithms follow an optimal policy with dynamic interval restrictions on their action space?
    \item How can the performance of such methods be evaluated?
\end{enumerate}

\section{Related Work}
This work is part of a number of research papers. The following section gives an overview of related fields and methods dealing with constrained action spaces and obstacle avoidance.

\subsection{Constrained Action Space}
The subsection starts with an overview of techniques that strictly follow action space restrictions. The methods can be broadly classified into discrete and continuous problems. 
Afterward, we introduce the fields of constraint RL and output-constraint deep learning (DL). Constraint RL allows violations of restrictions and models a fundamentally different problem by simultaneously learning a cost function. However, it is similar in imposing restrictions on actions. Constrained DL is comparable in handling constraints through neural architectures and deals with an almost identical problem in a wider context. 

\subsubsection{Discrete Restrictions}
A number of research papers incorporate discrete restrictions by doing nothing when an invalid action is chosen \cite{berner2019dota, chen2017deep, kuhnle2021designing} or penalizing the agent \cite{yoo2021dynamic, ran2022optimizing, wu2020autonomous}. More sophisticated methods include action masking \cite{huang2020closer}, embeddings \cite{dulac2015deep,ammanabrolu2020graph,ardon2022inapplicable} and elimination networks \cite{zahavy2018learn}. 
For example, Kuhnle et al. \cite{kuhnle2021designing} design an adaptive production control system and simply proceed with the next time step when an invalid action is taken. Contrary, Ran et al. \cite{ran2022optimizing} optimize a data center with constraints on the resources per machine and the airflow in a cooling system. A negative penalty is applied when actions go outside the range. 
Huang et al. \cite{huang2020closer} are the first to investigate the theoretical foundations of invalid action masking and compare the technique with penalty-based learning. The authors analyze the results of a real-time strategy game on discretized maps with a maximum of $24 \times 24 = 576$ cells. The empirical results suggest that action masking is superior and agents keep some behavior when the mask is removed since the available actions for an observation stay the same.
These outcomes are similar to those reported by Kanervisto et al. \cite{kanervisto2020action}. In this case, Starcraft 2 experiments are run and masking turns out to be crucial for learning. 
Zahavy et al. \cite{zahavy2018learn} follow a different approach. The authors combine deep Q-networks (DQN) with contextualized bandits which learn to eliminate actions and take the maximum of action values only over feasible subsets. 
The same holds for Dulac et al. \cite{dulac2015deep} who create embeddings for actions and search only for valid choices with a k-nearest neighbor search. However, the embeddings are limited to discrete actions and do not allow continuous representations. 
This is a major disadvantage since Dulac et al. \cite{dulac2015deep} demonstrate that the optimal discretization strategy is crucial. This is done by obtaining error bounds in extensive-form games. The results are related to this thesis by amplifying the importance of continuous restrictions. The authors caution that continuous domains, such as wifi-jamming, are often discretized which imposes errors and lowers the final performance.

\subsubsection{Continuous Restrictions}
Methods to restrict continuous action spaces mainly emerged in the areas of constrained and safe RL.
Krasowski et al. \cite{saferloverview} classify algorithms that provide hard constraints into three categories: Masking maps the entire action space to convex subsets, 
replacement replaces invalid actions, and projection searches for the closest possible value. The following gives an overview of these fields:
\begin{itemize}
    \item \textbf{Projection}: As an example of projection techniques, Pham et al. \cite{pham2018optlayer} project any action in the unconstrained domain to the feasible space by minimizing the euclidean distance with a quadratic program.
    Sanket et al. \cite{sanket2020solving} suggest a more efficient solution. Instead of minimizing the euclidean distance, they search for the closest action on a line to the center of a polygon action space. The technique is applied to find an online solution for the threat screening game while maintaining a hard bound of acceptable risk. Taylor et al. \cite{taylor2020learning} locate the closest point with constraints as control barrier functions and control affine systems. However, a barrier function is often not trivial to find. Thirugnanam et al. \cite{thirugnanam2022safety} show how the approach could be transferred to collision avoidance with polygons but rely on model predictive control. Cheng et al. \cite{cheng2019end} propose a way to use the constraints with model-free deep RL algorithms. As a downside, Brosowski et al. \cite{brosowsky2021sample} argue that projection does not always lead to the optimal action since the region maximizing the action value might not lie directly at the border of the feasible space.  

    \item \textbf{Replacement}: Replacement techniques substitute invalid actions often by sampling uniformly from the allowed action space \cite{brosowsky2021sample}. However, the value could also come from a supervisor \cite{zhou2021learning, li2018safe} or human feedback \cite{marta2021human}. Hsu et al. \cite{hsu2022improving} suggest an additional replay buffer that stores actions leading to feasible behavior from different states. When the agent encounters a similar situation, it replaces the action with the one leading to the highest reward in this replay buffer. Another variant of Li et al. \cite{li2018safe} uses a meta-learned neural network that can replace actions by adding values to the outputs. This way, the supervisor is supposed to hold the agent inside the feasible set.
    Ozhamaratli et al. \cite{ozhamaratli2022deep} take a different approach. The authors develop an algorithm for learning optimal portfolio allocation and saving strategies for heterogeneous individuals. Whenever an action outside the feasible space is taken, the agent receives a penalty proportional to the violation distance and the action is replaced by one that does nothing.

    \item \textbf{Masking}: Action masking for continuous spaces has been proposed such that actions are scaled into convex subsets. Authors differ mostly in the specific scaling function. For example, Brosowski et al, \cite{saferloverview} use a standardization function from statistics for a one-dimensional, continuous range and Brosowski et al. \cite{brosowsky2021sample} interpolate between vertices of different geometric types. Similarly, Tabas et al. \cite{tabas2022computationally} use a gauge function to map actions to a feasible space of a convex polytope. Other approaches ensure specific properties of multi-dimensional outputs. Liu et al. \cite{liu2020constrained} and Ale et al. \cite{ale2022d3pg} make actions sum to one with a softmax layer 
\end{itemize}
Only a few authors compare multiple continuous restriction techniques with each other. 
Krasowski et al. \cite{saferloverview} explore replacing, projecting with control barrier functions and action masking but no difference in the final performance could be found. The main result is that learning with Proximal Policy Optimization (PPO) on the replaced actions performs worse. The environment is a pendulum-balancing task. Contrary to Tang et al. \cite{tang2020discretizing}, continuous reach the same performance as discrete algorithms. The environment makes a similar assumption to most research. In all presented works, agents learn constant restrictions for the same observation. 

\subsubsection{Constrained Reinforcement learning}
Besides hard constraints, the problems are formulated using constrained Markov decision processes (MDP) \cite{yang2022cup, xu2021crpo, paternain2019constrained}. The allowed actions are  assumed to be unknown and learned together with the execution costs through interacting with the environment. Violations are therefore inevitable. Next to techniques from above, Yu et al. \cite{yu2022reachability} differentiate penalty-based \cite{guan2022integrated}, Lagrangian \cite{xu2018semantic, paternain2019constrained}, and trust-region methods \cite{achiam2017constrained}, as well as others, such as conservative updates \cite{bharadhwaj2020conservative}. The Lagrangian approaches are categorized into primal and primal-dual methods.
A primal method example from Xu et al. \cite{xu2021crpo} alternatingly updates the policy. When no constraints are infringed, the update maximizes the expected return and otherwise minimizes constraint violations. 
In contrast, Paternain et al. \cite{paternain2019constrained} transform the constrained objective into an unconstrained optimization problem with Lagrangian multipliers. Afterward, the primal-dual variables are updated simultaneously with the policy which achieves a zero duality gap. 
Liu et al. \cite{liu2021policy} theoretically compare Lagrangian relaxation and projection. 

\subsubsection{Constrained Deep Learning}
Constrained DL is not RL-specific but suggests ways of restricting neural network outputs. Recently, the field focused on adding loss terms to the objective function \cite{xu2018semantic}, modifying the optimizer \cite{ravi2019explicitly} and constraint-specific architectures \cite{hoernle2022multiplexnet,brosowsky2021sample}. 
Brosowski et al. \cite{brosowsky2021sample} are closest to our work. The authors suggest an input-dependent parameterization of the output layer. This way, outputs are scaled to satisfy different convex geometries. The approach reaches a better performance in facial landmark detection and adaptive cruise control compared to a clipped Twin Delayed Deep Deterministic Policy Gradient (TD3) algorithm. The  allowed subsets are represented as a polytope. The review paper also proposes a way to handle multiple constraints. In this case, the network produces outputs within each feasible region and softmax scores decide on the action.  
Hoernle et al. \cite{hoernle2022multiplexnet} follow a similar approach and introduce separate scaling layers. A fixed latent categorical variable decides on the constraint. Both parts are learned simultaneously for logical formulas in disjunctive normal form.

\subsection{Pathfinding and Obstacle Avoidance}
Besides action space restrictions, this thesis is located in the field of obstacle avoidance. In this context, the applicability of algorithms depends on the available information and the environment. Agents either find paths based on local data or global knowledge. Some assume static and others deal with dynamic obstacles. 

\subsubsection{Local Methods}
Local information usually consists of an agent's location together with images or sensor data. The following depicts recent results in both fields: 
\begin{itemize}
    \item \textbf{Images}: Guo et al. \cite{guo2020deep} train a DQN algorithm with images from a robot's front camera. The resulting path comes close to the most efficient way. However, experiments are only simulation-based. Therefore, Xie et al. \cite{xie2017towards} argue that estimating the distance to obstacles in real-life environments is complicated. The authors introduce an architecture with a depth prediction network. Since the algorithm is only applicable in discrete applications, Cimurs et al. \cite{cimurs2020goal} extend the method with temporal image embeddings to continuous spaces and evaluate it in goal-oriented experiments. The authors also show success with dynamic obstacles. A different approach is taken by Zhou et al. \cite{zhou2022hybrid} who rely on obstacle detection for collision-free path planning with model predictive control. 

    \item \textbf{Laser}: In terms of laser data, Tai et al. \cite{tai2017virtual} train the Deep Deterministic Policy Gradient (DDPG) algorithm to avoid collisions reward-driven. 
    The authors rely on 10-dimensional laser findings together with the distance and angle to the goal. Actions are the directional and angular velocities. Nevertheless, the distances are short and the task focuses on avoiding static obstacles.
    Contrary, Wang et al. \cite{wang2018learning} propose a modular architecture for dynamic environments. The task of the navigation module is to find the shortest path while the second component is responsible for avoiding obstacles. A downside is that the environment is relatively simple.
    In complex environments, Wang et al. \cite{wang2019autonomous} report problems of DDPG getting stuck in a dead end. Incorporating long short-term memory network updates based on full historical trajectories resolves the problem. Choi et al. \cite{choi2021reinforcement} encounter the same problem for the soft actor-critic algorithm and suggest that the behavior is due to exploration problems. The authors use consecutive sensor data to predict an obstacle's movement direction and speed. Recently, Zhang et al. \cite{zhang2022autonomous} explore multiple dynamic obstacles in a three-dimensional environment. On one hand, the results claim that their two-stream network, TD3, and DDPG can still reach the target in most cases. On the other hand, crashes occur and methods are not collision-free. 
\end{itemize}

\subsubsection{Global Methods}
Global methods assume that the agent is fully aware of its surroundings and has perfect knowledge of the environment. In this case, multiple algorithms exist which can find an optimal path. Examples include $\text{A}^*$ \cite{hart1968formal} and the rapidly exploring random tree \cite{li2021rrt}. However, the methods can not deal with dynamic obstacles. A solution is given by the $\text{D}^*$ \cite{stentz1997optimal} algorithm which simply recalculates the path when the rest of the environment is still considered known.
Deep RL algorithms often rely on waypoints and predefined paths \cite{lei2018dynamic, balakrishnan2021curriculum}. Other forms are additional observations such as artificial potential fields \cite{hu2021obstacle, bouhamed2020autonomous,yao2020path,battocletti2021rl,ren2021potential} and penalties for close distances to obstacles \cite{hu2021obstacle, roghair2022vision}.
Meyer et al. \cite{meyer2020taming} steer a vessel over an ocean surface simulation. A predefined path is available but obstacles occur randomly. 255 sensors deliver fine-grained information and are partitioned into 25 clusters. Each cluster is an approximation of the maximum feasible distance until an obstacle occurs. The evaluation is in a static environment. Hu et al. \cite{hu2021obstacle} include moving obstacles and show that PPO surrounds barriers with observations and rewards integrating information about obstacles' locations and directions. 
Choi et al. \cite{choi2022marl} are the most related to our approach. The reason is that the paper puts emphasis on masking actions that can lead to conflicts between agents. Multiple agents navigate in a discretized warehouse and are trained in a centralized way using QMIX with an additional local loss. However, the evaluation focuses on the comparison with independent learning and masking is part of every algorithm. 

\subsection{Parameterized Reinforcement learning}
We propose a neural architecture based on parameterized reinforcement learning to solve dynamic interval restrictions in obstacle avoidance. Combining discrete actions with a continuous parameter is necessary for several domains. For example, Dorokhova et al. \cite{dorokhova2021deep} optimize the charging behavior of electric vehicles. At each time step, the agent selects an electricity source with an amount of power. Bouktif et al. \cite{bouktif2021traffic} optimize the traffic flow by assigning traffic lights a duration time. 

The problem has been solved on- and off-policy. To the best of our knowledge, PA-DDPG \cite{hausknecht2015deep} was the foremost off-policy model for parametric action spaces and relaxes the discrete actions into the continuous space. The architecture produces two continuous values for each combination and updates the weights with the action values from a second neural network. Contrary, Xiong et al. \cite{xiong2018parametrized} make use of the hierarchical architecture and select the next action directly with the action values. This dispenses the relaxation into the continuous space. However, a major downside is that gradients may affect unrelated actions due to the shared critic. Therefore, Bester et al. \cite{bester2019multi} suggest performing multiple forward passes for each parameterization separately. Next to DQN and DDPG, Delalleau et al. \cite{delalleau2019discrete} provide an extension for the Soft Actor-Critic algorithm. 

In the on-policy case, Fan et al. \cite{fan2019hybrid} utilize multiple PPO actors with shared layers. The first actors decide on a discrete action while the others predict associated continuous parameters. A softmax layer over the discrete outputs decides the next step in the environment. In contrast to predicting discrete and continuous values simultaneously, Wei et al. \cite{wei2018hierarchical} take advantage of the reparameterization trick and develop a two-step architecture for trust region policy optimization. Hu et al. \cite{hu2021hierarchical} argue that the value estimation can be improved by respecting the factorization and forecast action values for discrete actions with an additional network separately. 

\section{Contributions and Thesis Outline}
In this thesis, the issue of RL algorithms following dynamic interval restrictions on the action space is discussed. Restrictions are independent of observations and can vary between time steps. The experiments are based on finding a path while avoiding obstacles in an unknown environment. We construct evaluation criteria and a task in which existing approaches, modifications, and novel algorithms are tested. Our main contributions are summarized as follows:

\begin{itemize}
    \item To the best of our knowledge, we are the first to analyze learning methods that follow given but time-specific interval action space restrictions which can be significantly different for the same observation.
    \item We compare existing approaches and introduce novel parameterized and multi-pass architectures to deal with disjoint interval action spaces.
    \item We provide a new action-space-centric perspective on obstacle avoidance.
\end{itemize}

The rest of this thesis is structured as follows:
The second chapter introduces the theoretical foundations of RL, DL, and selected algorithms. It ends with the necessary background on action space restrictions. 
The third chapter modifies algorithms to work on dynamic and multiple intervals. Furthermore, we present details of the obstacle avoidance task, evaluation metrics, implementation, and hyperparameter optimization.
The fourth chapter deals with empirical analysis. We suggest different experiments and discuss their results. 
Finally, the work concludes with a summary, limitations, and an outlook on future research.

\chapter{Theoretical Background}
\label{cha:ba}
This chapter summarizes the fundamentals necessary to understand the research in this thesis. RL is introduced before the combination with DL is discussed. The chapter ends with the theoretical background of action space restrictions.

\section{Reinforcement Learning}
\label{sec:rl}
RL is one of the three machine learning paradigms, alongside supervised and unsupervised learning. It does not deal with patterns in labeled or unlabeled data but tries to find optimal actions in an unknown environment. Formally, the field addresses optimal control of a MDP \cite{sutton2018reinforcement}. This section introduces the general RL problem followed by definitions of necessary concepts. The latter includes the MDP, Bellman equation, and value functions. 

\subsection{Problem Formulation}
\begin{figure}[tb]
\centering
\begin{tikzpicture}[
    >=stealth,
    rect/.style={
        rectangle,
        draw,
        rounded corners=5, 
        inner sep = 0.2cm,
        text depth=0.1em
    }]
    \node[rect] (agent) at (0,0) {Agent};
    \node[rect] (environment) at (0,-1.6) {Environment};
    
    \draw[->, thick] (agent) -- ++(2.5,0) |- 
        node[pos=0.25, right] {$a_t$} (environment);
    
    \coordinate (lower west) at ([shift={(-2cm, -4pt)}]environment.center);
    \coordinate (upper west) at ([shift={(-2cm, 4pt)}]environment.center);
    
    \draw[densely dotted] ([yshift=5pt]upper west) -- ([yshift=-5pt]lower west);
    
    \draw[->, thick] (environment.west |- upper west) -- (upper west)
        node[midway, above] {$r_{t+1}$};
        
    \draw[->, thick] (environment.west |- lower west) -- (lower west)
        node[midway, below] {$s_{t+1}$};

    \draw[->, thick] (upper west) -- ++({-0.5cm + 4pt},0) |- 
        node[pos=0.25, right] {$r_t$} ([yshift=-4pt]agent.west);
        
    \draw[->, thick] (lower west) -- ++({-0.5cm - 4pt},0) |- 
        node[pos=0.25, left] {$s_t$} ([yshift=4pt]agent.west);
\end{tikzpicture}
\caption[Agent-environment interaction]{Agent-environment interaction \cite{sutton2018reinforcement}}
\label{fig:agent-environment}
\end{figure}
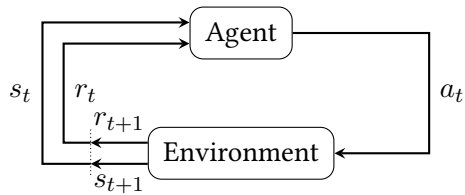

 RL is concerned with agents interacting with an environment to maximize a cumulative reward by trial and error \cite{sutton2018reinforcement}. Figure \ref{fig:agent-environment} gives an overview of the process.

At each time step t, the agent receives a representation of the state $s_t \in \mathcal{S}$ from the environment. Afterward, the agent uses a policy to decide on the next action $a_t \in \mathcal{A}$ to execute. The environment transitions and the agent receives the next state $s_{t+1}$ with reward $r_t$. Take the game Tik Tak Toe for example. The state at each time step is the grid with markers from both players and the agent chooses the next field. The policy can be deterministic $a_t = \mu(s_t)$ or stochastic $a_t {\sim} \pi(a_t {\mid} s_t)$. The stochastic case defines a mapping from state $s_t$ and action $a_t$ to the probability of choosing $a_t$. If the environment is episodic, the procedure repeats until a terminal state $s_T$ is reached and the environment is reset to a starting state $s_0$. The sequence of state, action, and reward triples $\tau$ until $s_T$ is called a trajectory and describes an episode \cite{sutton2018reinforcement}: 
\begin{equation}
    \tau=(s_0,a_0,r_0,s_1,a_1,r_1,s_2...)
\end{equation}

In the long run, the agent tries to maximize the expected return $G_t$. The cumulative reward is often deducted by a discount factor $\gamma \in [0,1]$ such that
\begin{equation}
    G_t = r_{t+1} + \gamma r_{t+2} + \gamma^2 r_{t+3} + ... 
    = \sum_{k=0}^{\infty} \gamma^k r_{t+k+1}
\end{equation}
The discount factor determines the worth of future rewards. If $\gamma = 0$, the agent only tries to maximize the immediate reward. As $\gamma \rightarrow 1$, future rewards are taken into account more strongly \cite{sutton2018reinforcement}. A challenge for algorithms is sparse feedback. For example, in the board game Go the reward is only received at the end of a game. However, artificial and dense intermediate rewards can replace the original. Finding an appropriate reward function is known as reward shaping \cite{li2017deep}. 

\subsection{Markov Decision Process}
The MDP is a mathematical framework for describing the environment of RL problems as a 5-tuple $(\mathcal{S}, \mathcal{A}, \mathcal{P}, \mathcal{R}, \gamma)$ \cite{sutton2018reinforcement}: 

\begin{itemize}
    \item A set of states $\mathcal{S}$, containing all possible states of an environment
    \item A set of actions $\mathcal{A}$, containing all possible actions of an agent
    \item A transition model $s_{t+1} {\sim} P(s_{t+1} {\mid} s_t, a_t)$, describing the probability of transitioning to state $s_{t+1}$ when action $a_t$ is performed in state $s_t$
    \item A reward function $r_t = R(s_t, a_t, s_{t+1})$, determining the reward when executing action $a_t$ in state $s_t$ and transitioning to state $s_{t+1}$
    \item A reward discount factor $\gamma$
\end{itemize}

The environment is assumed to obey the Markov property. In this case, the state $s_t$ retains all information to predict the next state $s_{t+1}$ and reward $r_t$. The transition depends only on $s_t$ and is independent of the prior history of actions. Even if this is rarely the case, the task should be designed as close to the Markov property as possible. Otherwise learning is usually more inefficient but agents might still be able to increase the reward \cite{sutton2018reinforcement}. 

\subsection{Value Functions and Bellman Equation}
\label{sec:vbel}
Almost all RL algorithms require an estimate of the value to be in a specific state. The state-value function $V_\pi(S)$ gives the expected return when the agent starts in state $s$ and always acts according to policy $\pi$:
\begin{equation}
\label{eq:value}
    V_\pi(s) = \mathbb{E} \Bigl[ G_t {\mid} s_t = s \Bigr]
\end{equation}
Similarly, the action-value function $Q_\pi(s,a)$ is the expected return for following policy $\pi$ 
 after performing action $a$ in state $s$. The return can be interpreted as how good it is to take an action in a specific state: 
\begin{equation}
    Q_\pi(s, a) = \mathbb{E} \Bigl[ G_t {\mid} s_t = s, a_t = a \Bigr]
\end{equation}
Both functions can be extended to yield the Bellman equations which are a key concept in many reinforcement learning algorithms and dynamic programming. The equations can be used to approximate the state and action value functions. The complex problem is broken down into smaller and simpler subproblems. For the state value function, the value of a state is written as the sum of the immediate reward and the discounted expected future return:
\begin{subequations}
\begin{align}
    V_\pi(s) 
    &= \mathbb{E}_{\pi} \left[ \sum_{k=0}^{\infty} \gamma^k r_{t+k+1} {\mid} s_t = s \right] \\
    &= \mathbb{E}_{\pi} \left[ R_{t+1} + \gamma \sum_{k=0}^{\infty} \gamma^k r_{t+k+2} {\mid} s_t = s \right] \\
    &= \sum_a \pi(a {\mid} s) \sum_{s',r} p\left(s',r {\mid} s,a\right) \Bigl[ r + \gamma V_\pi(s') \Bigr] \label{eq:bellman_state}
\end{align}
\end{subequations}
Similarly, the action value function leads to 
\begin{equation}
\label{eq:bellman_action}
    Q_\pi(s,a) 
    = \sum_{s'} p\left(s' {\mid} s,a\right) \Bigl[ r + \gamma \sum_{a'} \pi(a' {\mid} s') Q_\pi(s',a') \Bigr]
\end{equation}
Instead of the absolute value, many algorithms, such as DQN from chapter \ref{sec:drl}, learn more stable on the advantage of an action $a$ over sampling from a policy $\pi$. The concept is formulated as the advantage function \cite{sutton2018reinforcement}:
\begin{equation}
    A_\pi(s,a) = Q_\pi(s,a) - V_\pi(s)
\end{equation}

The problem of maximizing the expected return can be solved with optimal value functions. A policy $\pi$ is better than another policy $\pi'$ if the reward of $\pi$ is equal to or better than that of $\pi'$ in every state. There exists at least one optimal policy $\pi^*$ for every MDP. The optimal state-value function $V_\pi^*(s)$ is defined as the expected return when the agent performs actions from the optimal policy $\pi$ starting from state $s$:
\begin{equation}
    V_\pi^*(s) = \max_{\pi} V_\pi(s)
\end{equation}
Correspondingly, the optimal action-value function returns the expected return for taking action $a$ in state $s$ and acting in terms of the optimal policy afterward: 
\begin{equation}
    Q_\pi^*(s,a) = \max_{\pi} Q_\pi(s,a)
\end{equation}
The optimal action-value function is connected to the optimal action. When starting in state $s$, the optimal policy will always select the action which maximizes the expected return. Therefore, the optimal action $a^*$ can be obtained when the optimal action-value function $Q_\pi^*$ is known:
\begin{equation}
\label{eq:optimal_action}
    a^* = \arg \max_a Q^*\left(s,a\right)
\end{equation}
Likewise, the optimal state-value function selects the action with the maximum return:
\begin{equation}
\label{eq:optimal_state}
    V_\pi^*(s) = \max_a Q_\pi^*(s,a)
\end{equation}
Expanding equation \ref{eq:optimal_state} with \ref{eq:value} leads to the Bellman optimality equation for the state value function:
\begin{subequations}
\begin{align}
    V_\pi^*(s) 
    &= \max_a \mathbb{E}_{\pi^*} \Bigl[ G_t {\mid} s_t = s \Bigr] \\
    &= \max_a \sum_{s'} p\left(s' {\mid} s,a\right) \Bigl[ R(s_t, a_t, s') + \gamma V_\pi^*(s') \Bigr] \label{eq:bellman_optimal_state}
\end{align}
\end{subequations}
The expression for the action-value function $Q_\pi^*$ is
\begin{equation}
\label{eq:bellman_optimal_action}
    Q_\pi^*(s,a) = \sum_{s'} p(s' {\mid} s,a) \Bigl[ R(s_t, a_t, s') + \gamma \max_{a'} Q_\pi^*(s',a') \Bigr]
\end{equation}
Similarly to the Bellman equations from equations \ref{eq:bellman_state} and \ref{eq:bellman_action}, they form the basis for a number of algorithms to approximate $V_\pi^*$ and $Q_\pi^*$ \cite{sutton2018reinforcement}. Fields making use of the Bellman equation include dynamic programming \cite{sutton2018reinforcement} and deep RL \cite{xiong2018parametrized}. DQN and DDPG are described in section \ref{sec:drl}.

\section{Deep Learning}
DL is a family of machine learning models allowing computers to learn from experience and provides a framework for approximating non-linear functions from training examples \cite{goodfellow2016deep}. This section gives an introduction to artificial neural networks and their optimization with backpropagation. 

\subsection{Artificial Neural Networks}
\label{sec:ann}
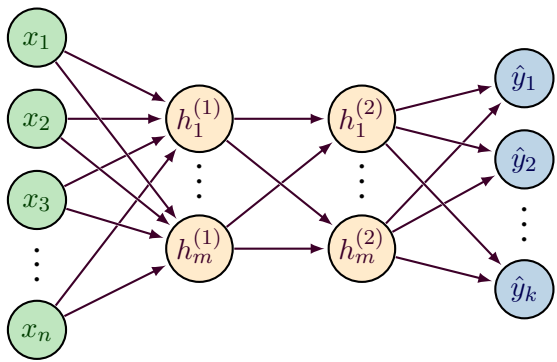
\begin{figure}
    \centering
    \begin{tikzpicture}[x=2.4cm,y=1.2cm,scale=0.9]
      \readlist\Nnod{4,2,2,3} 
      \readlist\Nstr{n,m,k} 
      \readlist\Cstr{x,h^{(\prev)},\hat{y}} 
      \def\yshift{0.6} 
      
      \foreachitem \N \in \Nnod{
        \def\lay{\Ncnt} 
        \pgfmathsetmacro\prev{int(\Ncnt-1)} 
        \foreach \i [evaluate={\c=int(\i==\N); \y=\N/2-\i-\c*\yshift;
                     \x=\lay; \n=\nstyle;
                     \index=(\i<\N?int(\i):"\Nstr[\n]");}] in {1,...,\N}{ 
          \node[node \n] (N\lay-\i) at (\x,\y) {$\strut\Cstr[\n]_{\index}$};
          
          \ifnumcomp{\lay}{>}{1}{ 
            \foreach \j in {1,...,\Nnod[\prev]}{ 
              \draw[white,line width=1.2,shorten >=1] (N\prev-\j) -- (N\lay-\i);
              \draw[connect] (N\prev-\j) -- (N\lay-\i);
            }

          }
          
        }
        \path (N\lay-\N) --++ (0,0.9+\yshift) node[midway,scale=1.4] {$\vdots$}; 
      }
  
    \end{tikzpicture}
    \caption{Neural network with one hidden layer}
    \label{fig:fnn}
\end{figure}

Given a set of training instances with features $x$ and output label $y$, an artificial neural network approximates the data generating function $f^*$ by defining a mapping $\hat{y}=f(x,\theta)$ from inputs $x$ to outputs $\hat{y}$. For example, in handwriting recognition, an image must be assigned a letter and the output vector represents the probability for each choice. The network is a chain of layers in the form of a directed acyclic graph and can include arbitrary operations.
Each layer transforms its input to a more selective and invariant representation which is used by subsequent operations. The transformations are mostly non-linear and might be semantic. In the previous example, the first may indicate the presence of an edge while the second layer combines available edges to primitive objects. The output layer uses the last representation and computes scores for each category. Operations of a layer are usually based on parameters $\theta$ which are optimized to find the best approximation \cite{goodfellow2016deep}. An example network is illustrated in figure \ref{fig:fnn}. 
The input layer $f^1$ transforms the input $x$ to a hidden representation $h^{(1)}$. The hidden layer $f^2$ further transforms $h^{(1)}$ to $h^{(2)}$ which is used by the output layer $f^3$ to produce the final prediction $\hat{y}$. All layers are chained sequentially:
\begin{equation}
    \hat{y} = f^3(f^2(f^1(x)))
\end{equation}

Figure \ref{fig:fnn} is an example of a feedforward neural network. Each layer consists of a number of artificial neurons which are connected to the ones in the previous layer. A single neuron $i$ can be seen as a linear classifier of the previous representation. The intermediate output $s_i$ is given by the weighted sum of the inputs $x_j$ with weights $w_{i,j}$ and a bias term $b_i$. Afterward, the output is fed through an arbitrary activation function which is usually chosen to provide non-linearity. In this case, the universal approximation theorem states that a neural network with at least one hidden layer but a sufficient number of neurons can approximate any non-linear function. However, the optimum is usually hard to find in practice. 
A common choice for an activation function is the rectified linear unit $f(x)=max(x,0)$ since the linear properties make gradient-based optimization easier. 
Overall, the final output of a layer can be obtained by
\begin{equation}
    \hat{y} = f(W^T x + b) \quad \text{with} \quad \hat{y_i} = f(\sum_j x_j w_{i,j} + b)
\end{equation}
Output layers are usually tightly coupled with the task to solve. Different units represent different distributions. For example, a linear neuron can represent a regression task and produce a Gaussian distribution by using $\hat{y}=W^T x + b$ to initialize $p(y{\mid}x) = \mathcal{N}(y;\hat{y},I)$. For a probability distribution over $k$ categorical variables, a softmax function is applied to $k$ linear output neurons. It follows 
\begin{equation}
    P(y=i {\mid} x) = softmax(y)_i = \frac{exp(s_i)}{\sum_j exp(s_j)}
\end{equation}
where $s_i$ is the intermediate result of neuron $i$ \cite{goodfellow2016deep}.

\subsection{Optimization}
\label{sec:optimization}
The optimization of a neural network is based on a cost function $J(\theta; X, y)$ which usually describes the average loss together with a regularization term. The loss $\mathcal{L}(f(x{\mid}\theta), y)$  measures the error between outputs and true values. An example is the mean squared error. Besides, the regularization term $\Omega$ limits the capacity of the network and drives the weights of each neuron closer to the origin. The hyperparameter $\alpha$ determines the regularization strength and is used to reduce overfitting. The most common regularization term is the $L^2$ parameter norm penalty for the value of weights $w$. Then, the regularized objective function is 
\begin{equation}
    J(\theta; X, y) = \frac{1}{N} \sum_{i=1}^N \mathcal{L}(\hat{y}_i, y_i) + \alpha \Omega(\theta) \quad \text{with} \quad \Omega(\theta) = \frac{1}{2} \norm{w}_2^2
\end{equation}
over a number of $N$ training examples \cite{goodfellow2016deep}. 

Training refers to adjusting the weights of the network to minimize the cost. To do so, the gradient vector $g_t = \nabla_\theta J$ indicates for each parameter the amount of which the error would decrease if the weight is increased by a small value. Gradient descent is the most common optimization algorithm and computes the gradient on all training examples. Alternatives, such as mini-batch gradient descent, compute an unbiased estimator by using only a sample of training examples. Mini-batches increase gradient variance but accelerate computation time and can help escape local minima. Both algorithms update the parameters incrementally by taking a step in the direction of the negative gradient:
\begin{equation}
    \theta_t = \theta_{t-1} + \epsilon g_t
\end{equation}
The parameter $\epsilon$ refers to the learning rate and determines the step size. If it is too small, the algorithm converges slowly and can get stuck in local minima while it might jump over the optimum if it is too large. When using the same data for multiple iterations, an epoch is a single pass through the dataset \cite{goodfellow2016deep}.

The Adam optimizer implements several improvements over vanilla gradient descent. One of them is momentum to accelerate convergence speed by tackling poor conditioning and high gradient variance. By using an exponentially decaying average of past gradients, the algorithm continues to move in a similar direction. $\beta_1 \in [0,1]$ is the corresponding hyperparameter which sets the impact of previous gradients. Finally, $m_t$ gives the first-order moment of the speed and direction in the parameter space:
\begin{equation}
    m_t = \beta_1 \cdot m_{t-1} + (1 - \beta_1) \cdot g_t \quad
\end{equation}
Additionally, Adam uses an individual and adaptive learning rate for each parameter since sensitivities can vary. The learning rate shrinks faster for weights with a history of large squared gradients. Similar to the first-order moment, the second-order $v_t$ is described using an exponentially decaying moving average with hyperparameter $\beta_2$:
\begin{equation}
    v_t = \beta_2 \cdot v_{t-1} + (1 - \beta_2) \cdot g_t^2
\end{equation}
Finally, bias correction is implemented by 
\begin{equation}
    \theta_t = \theta_{t-1} - \alpha \cdot \frac{\hat{m_t}}{\sqrt{\hat{v_t}} + \epsilon} 
    \quad \text{with} \quad
    \hat{m_t} = \frac{m_t}{1 - \beta_1^t}
    \quad \text{and} \quad
    \hat{v_t} = \frac{v_t}{1 - \beta_2^t}
\end{equation}
for both parameters $m_t$ and $v_t$ to account for their initialization before the parameters are updated \cite{kingma2014adam}.

\subsection{Backpropagation}
The backpropagation algorithm presents a computationally inexpensive procedure to compute the gradient for variables of an arbitrary function $f$. In DL, the gradient of the cost function is required. The method recursively applies the chain rule when $f$ is composed of other functions. This way, the gradient can be computed for every parameter involved in its computation \cite{goodfellow2016deep}. 
Coming back to the network from figure \ref{fig:fnn}, the gradient can be derived for example with respect to weight $w_{1,1}^{(3)}$, connecting the first neurons of the second hidden and the output layer. When the output before the activation function is denoted as $s_3$, the gradient is given as
\begin{subequations}
\begin{align}
    \frac{\partial y}{\partial w_{1,1}^{(3)}} 
    &= \frac{\partial y}{\partial f^4} \frac{\partial f^4}{\partial w_{1,1}^{(3)}} \\
    &=  {f'}^4(s_3) h_1^{(2)}
\end{align}
\end{subequations}

The algorithm alleviates gradient computation by passing the gradient of each layer with respect to its inputs backward. This denotes the first component of the chain rule expression and can be used to derive the gradient for the parameters of the previous layer. Simultaneously, the gradient with respect to inputs of this layer can be calculated and propagated. By this means, the method avoids the exponential explosion by repeatedly computing the sub-expressions. Assuming that the gradient with respect to inputs $g_t = \nabla_{h^{(k)}} J$ is known, both calculations per layer simplify to
\begin{equation}
    \nabla_{h^{(k-1)}} J = W^{(k)T} g_t
    \quad \text{and} \quad
    \nabla_{W^{(k)}} J = g_t h^{(k-1)T}
\end{equation}
Afterward, an optimization algorithm, such as Adam, can be used to optimize the weights of the network \cite{goodfellow2016deep}. 

\section{Deep Reinforcement Learning}
\label{sec:drl}
Deep RL uses neural networks to approximate at least one of the policy, state-value, or action-value functions \cite{li2017deep}. After the previous chapters have introduced RL and DL, the algorithms DQN, PPO, and DDPG are presented in the following sections. 

\subsection{Deep Q-Learning}
\label{sec:dqn}
DQN selects the action maximizing the expected return through approximating the action value function $Q_\pi$ with the parameters $\theta$ of a neural network. The algorithm is model-free which means that the policy is directly used to determine the next action instead of planning future steps with a model of environmental dynamics. 
A dilemma of the agent is to balance exploitation and exploration \cite{mnih2013playing}. Exploitation refers to taking actions to maximize the immediate reward for which the return is certainly known. Exploration tries unknown actions to find a better policy in the long term \cite{sutton2018reinforcement}. DQN typically addresses this problem by using an $\epsilon$-greedy strategy: The algorithm selects the action maximizing the current action value function $Q_\theta$ with a probability of $(1- \epsilon)$ and draws actions from a uniform distribution with probability $\epsilon$ \cite{mnih2013playing}.  

The learning procedure of DQN is off-policy. Off-policy indicates that the policy can be optimized with experiences generated from a version with different parameters \cite{mnih2013playing}. Contrary, on-policy algorithms only learn from actions taken with the current policy \cite{li2017deep}. While interacting with the environment, each time step $t$ is saved as an experience $(s_t,a_t,r_t,s_{t+1})$ in a replay buffer. This buffer has a maximum capacity $M$ and the oldest experiences are overridden when the number is reached. 
Mini-batches are sampled from this buffer and used to train the network. Since experiences can be reused, off-policy algorithms have a higher data efficiency and smoother learning progress because training samples are not dependent on previous policy parameters. Mini-batches would be highly correlated without a replay buffer and increase the variance of updates \cite{mnih2013playing}. 

The action value function $Q_\theta$ is learned iteratively with the Bellman optimality equation defined in equation \ref{eq:bellman_optimal_action} \cite{mnih2013playing}. The loss function of the neural network is implemented as the Bellman mean squared error (MSBE) \cite{xiong2018parametrized}. The MSBE is the mean squared difference between the output of the approximated action value function and the true action value function according to the optimality equation
\begin{equation}
\label{eq:MSBE}
    \mathcal{L}(\theta) = \mathop{\mathbb{E}}_{s,a {\sim} p(\cdot)} \Bigl( y_i - Q_\theta(s,a) \Bigr)^2
\end{equation}
where $y_i = \mathbb{E}_{s' {\sim} \varepsilon} [ r_t + \gamma \max_{a'} Q_{\theta'}(s',a') ]$ is the target for training iteration $i$ based on environment $\varepsilon$. The probability distribution $\rho(a,s)$ is known as the behavior distribution and refers to the distribution of actions taken by the agent during its interactions with the environment. The action values $Q_{\theta'}$ used in the training target $y_i$ come from a separate target network with parameters $\theta'$ to improve training stability. Without a target network, the action values $Q(S_{t+1}, a)$ for all actions $a$ in the next time step $t+1$ would increase simultaneously with $Q(S_t, a)$. The problem comes from similarities between the two states which leads to overestimating the target. The target network is updated every $C$ step with the parameters of the behavior policy being directly copied or set as the exponentially decaying average $\theta' = r \theta + (1-r) \theta'$ \cite{mnih2013playing}. All steps of DQN are summarized in algorithm \ref{alg:dqn}. 
\begin{algorithm}[t]
\caption[Deep Q-Learning]{Deep Q-Learning \cite{mnih2013playing}} \label{alg:dqn}
    Initialize replay buffer $\mathcal{D}$ \\
    Initialize action-value network $Q$ with random weights $\theta$ \\
    Initialize target network $\theta' \gets \theta$
    \begin{algorithmic}[1]
    \FOR{$\text{episode} = 1,2$}
        \STATE Initialize environment and receive initial state $s_0$;
        \FOR{$t=1$ \TO{T}} 
        \STATE With probability $\epsilon$ select random action $a_t$ \\
        otherwise $a_t = \max_a Q_{\theta}(s,a)$
        \STATE{Execute action $a_t$, receive reward $r_t$ and observe next state $s_{t+1}$}
        \STATE{Store experience $(s_t,a_t,r_t,s_{t+1})$ in replay buffer $\mathcal{D}$}
        \STATE{Sample random minibatch $(s_i,a_i,r_i,s'_i)$ of size $N$ from $\mathcal{D}$}
        \STATE{Set $y_i = \begin{cases} r_i, & \text{for terminal }s'_i \\
        r_i + \gamma \max_{a'} Q_{\theta'}(s'_i,a'), & \text{for non-terminal }s'_i \end{cases}$}
        \STATE Update action value network $\theta \gets \frac{1}{N} \bigl( y_i - Q_\theta(s,a) \bigr)^2$
        \STATE Update target network every $C$ step 
        $\theta' \gets \tau \theta + (1-\tau) \theta'$
        \ENDFOR
    \ENDFOR
    \end{algorithmic}
\end{algorithm}

Wang et al. \cite{wang2016dueling} propose dueling networks to further improve learning stability which is particularly useful with many similar valued actions. The motivation is that the choice of actions might be less important in certain steps. Consider the case when a car is in the middle way of a three-lane highway. It does not matter whether the car goes left or right until other drivers appear. Action values $Q(s,a)$ are split into their state value $V(s)$ and action advantage $A(s,a)$ parts. The individual advantage must be subtracted from the average over all action advantages. Otherwise neither $V$ nor $A$ would be identifiable given $Q$:
\begin{equation}
\label{eq:updateq}
    Q(s,a) = V(s) + ( A(s,a) - \frac{1}{{\mid} A {\mid}} \sum_{a'} A(s,a') )
\end{equation}
This approach helps to reduce the correlation between updates of different actions. The agent better understands the relative advantage of taking a particular action over other actions in the same state. The state value and action advantage are typically learned through separate neural networks or one that splits up into two streams \cite{wang2016dueling}. 
\newpage
An improvement to the sampling strategy is prioritized experience replay. The central idea is to replay experiences that contain a lot of information more frequently with a new sampling probability $P$ for experience $i$
\begin{equation}
    P(i) = \frac{p_i^\alpha}{\sum_k p_k^\alpha}
\end{equation}
In this context, $p_i$ is a priority value for experience $i$ and $\alpha$ is a hyperparameter for the prioritization strength where $\alpha = 0$ corresponds to uniform sampling. The priority value is usually chosen to be the sum of the absolute error of the value function $\abs{\delta} = \abs{r_t + \gamma V(s_{t+1}) - V(s_t)}$ and a small value $\epsilon$ for numerical stability. The absolute error is  known as the temporal difference error. It is the difference between the current estimate of the value function and the estimate that would be obtained by following the Bellman equation for the state value function in equation \ref{eq:bellman_state}. Since the importance sampling breaks with the idea of uniform sampling to avoid the correlation between training experiences, the method adopts importance-sampling weights $w_i = ( MP(i))^{-\beta}$ during training. The hyperparameter $\beta$ determines the strength of the weights \cite{schaul2015prioritized}. 

\subsection{Proximal Policy Optimization}
PPO samples the next action directly from a parameterized policy and is considered a gradient-based method. The policy is a neural network with parameters $\theta$ that takes an observation as input and maps it to the probability of choosing each possible action. In contrast to value-based algorithms, such as DQN, policy neural networks can naturally represent continuous action spaces and stochastic policies. As shown in section \ref{sec:ann}, one output neuron could represent the mean and another the standard deviation for continuous tasks.  Since no model of environmental dynamics is used, the algorithm is model-free. The exploitation and exploration trade-off is handled through the randomness of the policy itself. The variance of actions should decrease while learning and the agent learns to exploit. 
Contrary to DQN, the weights are directly optimized to increase the expected long-term return instead of approximating the value of taking an action \cite{Zhang2020}. The gradient estimate for the objective function $J = \mathbb{E}_{\tau {\sim} \pi} [\sum_k \gamma^k r_{t+k}]$ is defined as
\begin{equation}
\label{eq:vgrad}
    \nabla J_\theta(\theta) = \mathbb{E}_{\tau {\sim} \pi_\theta} \Bigl[ 
    \nabla_\theta \log \pi_\theta(a{\mid}s) \hat{A}_{\pi_\theta}(a,s)
    \Bigr]
\end{equation}
including an advantage estimate $\hat{A}_{\pi_\theta}(a,s)$ for more stable training \cite{schulman2017proximal}.

The training of PPO is on-policy. This means that in every update iteration $k$, the latest policy is used to sample a number of trajectories $\mathcal{D}_k = \{\tau_1, ..., \tau_i\}$ from the environment. Afterward, the collected data  is used for multiple epochs.
However, large update steps are not well-justified and can be destructive. The reason is that the learning rate is fixed beforehand and might overshoot the reward peak. This behavior is particularly problematic when the surface of the objective function is narrow with sharp transitions between high- and low-performing policy parameters \cite{schulman2017proximal}. 

PPO constructs a trust region around the initial policy to prevent destructively large parameter updates through multiple epochs. To do so, the probability ratio $r_t(\theta)$ between the old and current policy is utilized
\begin{equation}
    r_t(\theta) = \frac{\pi_\theta(a_t,s_t)}{\pi_{\theta_{\text{old}}}(a_t,s_t)}
\end{equation}
where $\pi_\theta$ corresponds to the probability of action $a_t$ under the latest policy and $\pi_{\theta_{old}}$ is the probability with the parameters before the current training iteration. Instead of directly optimizing the gradient of equation \ref{eq:vgrad}, PPO maximizes a clipped surrogate objective $\mathcal{L}^{CLIP}$.  The divergence between the two policies is clipped with hyperparameter $\epsilon$ so that the ratio stays within the trust region $[1-\epsilon, 1+\epsilon]$. Additionally taking the minimum of the clipped and unclipped surrogate provides a lower bound so that the ratio is only replaced when the objective gets larger
\begin{equation}
\label{eq:ppo_l_clip}
    \mathcal{L}^{CLIP} = \mathbb{E}_{\tau {\sim} \pi_\theta} \Bigl[ \min(r_t(\theta)A_t,\text{clip}(r_t(\theta), 1- \epsilon, 1+\epsilon)\hat{A}_t) \Bigr]
\end{equation}
When clipping applies, the objective results in a zero gradient since the policy parameters are not part of $\mathcal{L}^{CLIP}$ anymore. Therefore, parameter updates in the same direction stop after a maximum change has been reached \cite{schulman2017proximal}. For example, assuming that the probability for an action $a$ was $60\%$ and the advantage is positive, the parameters are updated so that the action becomes more likely in the future. After a number of training iterations, the probability has increased to $80\%$. In the case of $\epsilon = 0.2$, $r_t(\theta)=\frac{0.8}{0.6}=1.33 > 1.2 = 1 + \epsilon$ and the probability ratio is replaced. No further updates which make the action more likely are performed. 

The surrogate objective $L^{CLIP}$ requires an estimate of the advantage $\hat{A}_t$. PPO is therefore considered an actor-critic algorithm. 
\begin{algorithm}[t]
\caption[Proximal Policy Optimization]{Proximal Policy Optimization \cite{schulman2017proximal}} \label{alg:ppo}
    Initialize policy $\pi$ with random weights $\theta$ \\
    Initialize state value function $V$ with random weights $\phi$
    \begin{algorithmic}[1]
    \FOR{$\text{episode} = 1,2$}
        \STATE Collect set of trajectories $\mathcal{D}$ by running policy $\pi_\theta$ in the environment
        \STATE Compute advantage estimate $\hat{A}_t$ using state-value network $V_\psi$ and \\ generalized advantage estimation
        \STATE Update policy $\pi_\theta$ by maximizing the clipped surrogate objective \\
        $\mathbb{E}_{\tau {\sim} \pi_\theta} \bigl[ \min(r_t(\theta)A_t,\text{clip}(r_t(\theta), 1- \epsilon, 1+\epsilon)\hat{A}_t) \bigr]$
        \STATE Update state value function $V_\phi$ based on the MSBE \\
        $(V_\phi(s_t) - y_i)^2$ with $y_i = r_t + \gamma V_{\phi}(s_{t+1})$
    \ENDFOR
    \end{algorithmic}
\end{algorithm}
This means that the training involves two neural networks. The critic is a separate network with parameters $\phi$ and gives an estimate of the value function $V_\phi$ which is required to compute the advantage $\hat{A}_t$. The advantage is then used to update the parameters of the actor which is the policy network. A commonly used method to approximate the advantage is generalized advantage estimation (GAE) \cite{schulman2017proximal}. In this case, the advantage can be computed based on experienced rewards in a trajectory and a value function estimator
\begin{equation}
    A_t^{(n)} = r_0 + \gamma r_1 + ... + \gamma^{n-1}r_{n-1} + \gamma^nV_\phi(s_n) - V_\phi(s_0)
\end{equation}
When only the immediate reward is considered, the equation is similar to the temporal difference error. Incorporating more rewards increases the variances due to environmental dynamics. The estimate $V_\phi$ is smoother but might introduce bias. Generalized advantage estimation balances bias and variance with an exponentially weighted average between estimates based on an increasing number of rewards
\begin{align}
    A^{GAE(\gamma,\lambda)} &= 
    (1-\lambda) (A^{1} + \lambda A^{(2)} + \lambda^2 A^{(3)} + ...)
\end{align}
so that when $\lambda=0$ only the reward of the time step itself is taken into account \cite{schulman2015high}.
The neural network of $V_\phi$ is updated based on the MSBE $\mathbb{E}_{s,a} (V_\phi(s_t) - y_i)^2$ with $y_i = \mathbb{E}_{s'} [r_t + \gamma V_{\phi}(s_{t+1})]$ \cite{schulman2017proximal}. All steps are summarized in algorithm \ref{alg:ppo}.

In order to encourage exploration, an entropy bonus $\mathcal{H}$ multiplied with a hyperparameter $h$ can be appended to the objective of the policy network $\mathcal{L}^{CLIP}$. Entropy is a measure of the randomness in the policy. The bonus encourages taking actions that increase the entropy which in turn encourages exploration. The entropy term improves learning stability and prevents too deterministic policies. The hyperparameter $h$ weights the relative importance of the entropy in the optimization \cite{mnih2016asynchronous}. 

\subsection{Twin-Delayed DDPG}
\label{sec:td3}
DDPG provides an extension of DQN to continuous spaces. However, it is not trivial to find the action maximizing the action-value function with large and unconstrained function approximators. The algorithm, therefore, uses an actor-critic approach. The actor $\mu$ is a neural network with parameters $\theta^Q$ which maps observations to deterministic actions. The critic is an action value network $Q_\mu$ with parameters $\theta^\mu$ and is used to estimate how good the actions of the actor are. The critic is learned based on the Bellman equation for the action value function of equation \ref{eq:bellman_action} \cite{lillicrap2015continuous}. When the action value function is deterministic, the equation can be written as 
\begin{equation}
    Q_\mu(s_t,a_t) = 
    \mathbb{E}_{s_{t+1} {\sim} \varepsilon} 
    \Bigl[ r(s_t,a_t) + \gamma Q_\mu(s_{t+1}, \mu(s_{t+1})) \Bigr]
\end{equation}
where the next state is sampled from the environment $\varepsilon$. Assuming that the action value function $Q_\mu$ corresponds to the value of the optimal action $ \arg \max_a Q(s,a)$, the critic weights can be updated using the MSBE $\mathbb{E}_{s,a} (V_\theta(s_t) - y_i)^2$ with the target $y_i = r(s_t,a_t) + \gamma Q_\mu (s_{t+1}, \mu(s_{t+1}))$. At the same time, the weights of the actor are adjusted to maximize the expected long-term return. This is done by taking a step in the direction of the gradient of the action value function with respect to the actor parameters. By this means, the actor tries to maximize the output of the critic. The gradient is calculated with the chain rule which leads to
\begin{subequations}
\begin{align}
    \nabla_{\theta^\mu} J 
    &\approx \mathbb{E}_{s_t {\sim} p^\beta}
    \Bigl[ \nabla_{\theta^\mu} Q(s_t, \mu(s_t)) \Bigr] \\
    &= \mathbb{E}_{s_t {\sim} p^\beta}
    \Bigl[ \nabla_a Q(s_t, \mu(s_t)) \nabla_{\theta^\mu} \mu(s_t) \Bigr]
\end{align}
\end{subequations}
with $p^\beta$ being the discounted state visitation distribution which is the probability distribution over states that an agent visits while interacting with an environment \cite{silver2014deterministic}. Silver et al. \cite{silver2014deterministic} provide the full proof for the policy gradient. 

As DQN, DDPG is considered off-policy and makes use of experience replay. At each training iteration, a random minibatch of size $N$ is sampled from the replay buffer. The algorithm further employs target networks which are updated every $C$ steps using the exponentially decaying average. This has been shown to be necessary to stabilize the training of the action value function. Instead of the $\epsilon$-greedy strategy for exploration, DDPG adds action noise sampled from a noise process $\mathcal{N}$ to the deterministic action
\begin{equation}
    \pi(s_t) = \mu(s_t) + \mathcal{N}
\end{equation}
The noise process $\mathcal{N}$ can be chosen to fit the environment. The original paper suggests the Ornstein-Uhlenbeck process but independent Gaussian noise \cite{lillicrap2015continuous} with mean zero has also been shown to work \cite{fujimoto2018addressing}.  
For the evaluation, noise is discarded and the deterministic action is executed \cite{lillicrap2015continuous}. 

Fujimoto et al. \cite{fujimoto2018addressing} introduce TD3 and propose several improvements over DDPG. All TD3 steps are described in algorithm \ref{alg:ddpg}. 
\begin{algorithm}[t]
\caption[Twin-Delayed DDPG]{Twin-Delayed DDPG \cite{fujimoto2018addressing}} \label{alg:ddpg}
    Initialize critic networks $Q_{\theta_1}$, $Q_{\theta_2}$ with random parameters $\theta_1$, $\theta_2$ \\ Initialize actor network $\mu_\phi$ with random parameters $\phi$ \\
    Initialize target networks $\theta'_1 \gets \theta_1$, $\theta'_2 \gets \theta_2$, $\phi' \gets \phi$ \\
    Initialize replay buffer $\mathcal{D}$ to capacity $N$
    \begin{algorithmic}[1]
    \FOR{$t=1$ \TO{T}} 
        \STATE Select action with exploration noise $a_t {\sim} \mu(s_t) + \mathcal{N}(0,\sigma)$ \\ 
        and observe reward $r_t$ and new state $s_{t+1}$
        \STATE{Store experience $(s_t,a_t,r_t,s_{t+1})$ in replay buffer $\mathcal{D}$}
        \STATE{Sample random minibatch $(s_i,a_i,r_i,s'_i)$ of size $N$ from $\mathcal{D}$}
        \STATE $\Tilde{a} = \mu_{\theta_{targ}}(s) + \text{clip}(\mathcal{N}(0, \Tilde{\sigma}), -c, c))$
        \STATE $y_t = r + \gamma \min_{i=1,2} Q_i'(s_t, \mu(s_t))$
        \STATE Update critics $\theta_i \gets \min_{\theta_i} \frac{1}{N} \sum (y - Q_{\theta_i}(s,a))^2$
        \IF{$t \mod delay = 0$}
            \STATE Update $\phi$ using the policy gradient \\
            $\frac{1}{N} \sum \nabla_a Q_{\theta_1}(s,\mu_\phi(s)) \nabla_\phi \mu_\phi(s)$
            \STATE Update target networks \\
            $\theta'_i \gets \tau \theta + (1-\tau) \theta'_i $ \\
            $\phi' \gets \tau \phi + (1-\tau) \phi'$
        \ENDIF
    \ENDFOR
    \end{algorithmic}
\end{algorithm}
First, Target smoothing tackles the problem of overfitting to peaks in the action value function which are hard to escape. The technique serves similarly to a regularizer and adds noise to each dimension of the action. Afterward, the value is clipped so that it stays in a small range. The target actions are given by
\begin{equation}
    \Tilde{a} = \mu_{\theta'}(s) + 
    \text{clip}(\mathcal{N}(0, \Tilde{\sigma}), -c, c))
\end{equation}
Hyperparameter $c$ denotes the smoothing range. Secondly, two action value functions are learned simultaneously to prevent overestimating the value of an action. A single target value $y_t$ is used to update both of the  critic networks. The loss is  based on the action value network with the smaller value
\begin{equation}
    y_t = r + \gamma \min_{i=1,2} Q_i'(s_t, \mu(s_t))
\end{equation}
The actor is constantly learned using $\max_\theta \mathbb{E}_{s {\sim} \mathcal{D}} \bigl[ Q_{\phi_1} (s, \mu(s)) \bigr]$ which is similar to DDPG. However, the network is updated less frequently than the critic to further improve stability \cite{fujimoto2018addressing}. 

\subsection{Parameterized Deep Q-Learning}
\label{sec:p-dqn}
Parameterized DQN (P-DQN) is an algorithm for discrete-continuous action spaces which is associated with a high- and low-level action.
\begin{algorithm}[t]
\caption[Parameterized Deep Q-Learning]{Parametrized Deep Q-Learning \cite{xiong2018parametrized}} \label{alg:p-dqn}
    Initialize replay buffer $\mathcal{D}$ \\
    Initialize action-value network $Q_\theta$ with random weights $\theta$ \\
    Initialize policy network $x_{k, \phi}$ with random weights $\phi$
    \begin{algorithmic}[1]
    \FOR{$t=1$ \TO{T}} 
        \STATE Compute action parameters $x_{k, \phi} \gets x_{k, \phi}(s_t)$
        \STATE With probability $\epsilon$ select random action $a_t = (k_t,x_{k_t, \phi})$ \\
        otherwise $a_t = (k_t,x_{k_t, \phi})$ with $k_t = \max_a Q_{\theta}(s,k,x_{k, \phi})$
        \STATE{Execute action $a_t$, receive reward $r_t$ and observe next state $s_{t+1}$}
        \STATE{Store experience $(s_t,a_t,r_t,s_{t+1})$ in replay buffer $\mathcal{D}$}
        \STATE{Sample random minibatch $(s_i,a_i,r_i,s'_i)$ of size $N$ from $\mathcal{D}$}
        \STATE{Set $y_i = \begin{cases} r_i, & \text{for terminal }s'_i \\
        r_i + \gamma \max_{k} Q_{\theta}(s_{i+1},k,x_{k, \phi}(s_{i+1})), & \text{for non-terminal }s'_i \end{cases}$}
        \STATE Update action value network $\frac{1}{N} \bigl( y_i - Q_\theta(s,k,x_{k, \phi}) \bigr)^2$
        \STATE Update policy network $ - \sum_k Q(s,k,x_{k, \phi}(s))$
    \ENDFOR
    \end{algorithmic}
\end{algorithm}
First, the agent chooses a discrete action $k$ out of a finite set $K$. Afterward, a continuous parameter $x_k$ is selected based on the discrete choice. The action space becomes $\mathcal{A} = \{ (k,x_k) \mid x_k \in X_k \text{ for all } k \in K \}$ with action value function $Q(s,k,x_k)$. When this function is assumed to be fixed, the new Bellman equation can be written as 
\begin{equation}
    Q(s_t,k_t,x_{t_t}) = 
    \mathbb{E}_{s_t,r_t} \Bigl[ r_t + \gamma \max_k Q(s_{t+1}, k, x_k(s_{t+1})) \mid s_t = s \Bigr]
\end{equation}
with $x_k(s_{t+1})$ being a policy function that maps states to continuous parameters. Both, $Q(s,k,x_k)$ and $x_k(s_{t+1})$ can be represented by a neural network with parameters $\theta$ and $\phi$ respectively. The action value network is updated similarly to DQN using the MSBE while the policy network searches for parameters maximizing $Q(s,k,x_k)$. The two loss functions are
\begin{equation}
    \mathcal{L}^Q = \frac{1}{2} \Bigl( y_i - Q_\theta(s,k, x_{k, \phi}) \Bigr)^2 \quad \text{and} \quad 
    \mathcal{L}^{x_k} = - \sum_{k=1}^K Q(s,k,x_{k, \phi}(s_t))
\end{equation}
The algorithm further employs the $\epsilon$-greedy strategy for exploration and a replay buffer which can be combined with prioritized experience replay \cite{xiong2018parametrized}. All steps are summarized in algorithm \ref{alg:p-dqn}.

\section{Action Space Restrictions}
\label{sec:res}
Various approaches exist to restrict the action space of an agent. 
The agent-environment interaction loop allows two intervention points: Action replacement and projection let the agent operate in the unrestricted domain and alter the action as a postprocessing step afterward. Action masking maps the action space directly to the allowed subset $A^\varphi$.
\begin{figure}[t]
\centering
\begin{subfigure}{.6\linewidth}
  \centering
  \begin{tikzpicture}[scale=1.0]
        \node(environment)[draw, rounded corners=5, inner sep = 0.2cm, text depth=0.1em] at (-2.5,0) {Environment};
        \node(agent)[draw, rounded corners=5, inner sep = 0.2cm, text depth=0.1em] at (2.5,0) {Agent};
        \node(rep)[draw, rounded corners=5, inner sep = 0.2cm, text depth=0.1em, fill = beaublue] at (-0.9,-1.5) {Replacement/Projection};

        \draw[->, thick] ([yshift=4pt]environment.east) -- ([yshift=4pt]agent.west) node[midway, above] {$r_t$};
        \draw[->, thick] ([yshift=-4pt]environment.east) -- ([yshift=-4pt]agent.west) node[midway, below] {$s_{t+1}$};
        \draw[->, thick] (agent.south) -- (2.5, -1.5) node[pos=0.5, right] {$a_t$} |- ([yshift=-4pt]rep.east);
        \draw[->, thick] ([yshift=-23pt]environment.south) -- node[pos=0.5, left] {$a_t^\varphi$} (environment.south);
        \draw[->, thick] (1.475,-4pt) |- ([yshift=4pt]rep.east);
    \end{tikzpicture}
  \caption{Postprocessing}
  \label{fig:postprocessing}
\end{subfigure}%
\begin{subfigure}{.4\linewidth}
  \centering
  \begin{tikzpicture}[scale=1.0]
        \node(environment)[draw, rounded corners=5, inner sep = 0.2cm, text depth=0.1em] at (-1.5,0) {Environment};
        \node(agent)[draw, rounded corners=5, inner sep = 0.2cm, text depth=0.1em] at (1.5,0) {Agent};
        \node(mask)[draw, rounded corners=5, inner sep = 0.2cm, text depth=0.1em, fill = beaublue] at (1.5,1.5) {Mask};

        \draw[->, thick] ([yshift=4pt]environment.east) -- ([yshift=4pt]agent.west) node[midway, above] {$s_{t+1}$};
        \draw[->, thick] ([yshift=-4pt]environment.east) -- ([yshift=-4pt]agent.west) node[midway, below] {$r_t$};
        \draw[->, thick] (mask.south) -- (agent.north) node[midway, right] {$\mathcal{A}^\varphi$};
        \draw[->, thick] (-0.11,4pt) |- (mask.west);
        \draw[->, thick] (agent.south) |- (0,-1) -- (-1.5,-1) node[pos = 1, left] {$a_t^\varphi$} -- (environment.south);
    \end{tikzpicture}
  \caption{Preventive masking}
  \label{fig:preventive}
\end{subfigure}
\caption[Intervention points to incorporate action space restrictions]{Intervention points to incorporate action space restrictions \cite{saferloverview}}
\label{fig:intervention_points}
\end{figure}
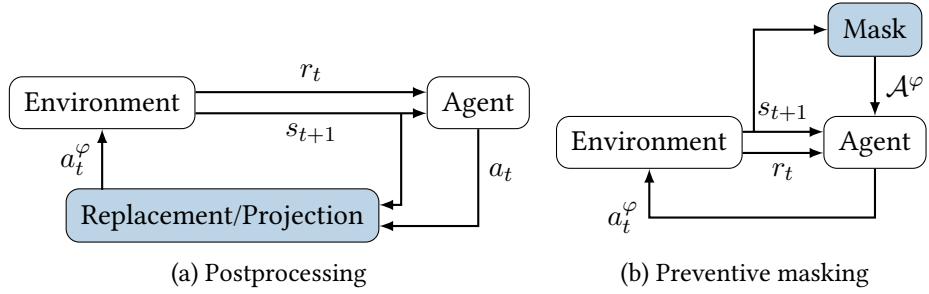
Figure \ref{fig:intervention_points} compares the two methods visually. At each time step $t$, a postprocessing step receives state information $s_t$ which allows replacing an invalid action $a_t$ with a valid action $a_t^\varphi$. Afterward, it is sent to the environment. Contrary, action masking alters the action space based on the state $s_t$ before the agent chooses an action $a_t^\varphi$ out of a set $A^\varphi$ containing only valid actions \cite{saferloverview}.
After an introduction to deep reinforcement has been given in previous chapters, the action space restriction techniques, that this thesis builds on, are described in the following chapters. 

\subsection{Discrete Masking}
\label{sec:discrete-masking}
Invalid action masking is a technique for discrete action spaces to avoid taking unavailable actions by "masking" them out. The prerequisite is that the final layer of a neural network performs some action selection. This could be taking the maximum from the action value function or sampling from a discrete output distribution of a stochastic policy. The technique is therefore applicable to DQN and PPO. Take PPO as an example and consider the case when the parameters of the policy network directly define the logits $\theta = [l_0,l_1,l_2,l_3] = [1.0, 1.0, 1.0, 1.0]$. The logits are passed through a softmax layer to obtain action probabilities:
\begin{subequations}
\begin{align}
    \pi_0(\cdot {\mid} s_0) 
    &= softmax([l_0,l_1,l_2,l_3]) \\
    &= [0.25,0.25,0.25,0.25]
\end{align} 
\end{subequations}
Typically, the policy would sample an action from this distribution and execute it in the environment. However, discrete masking adds a large negative number $M$ to the logits of the invalid actions. Hereby, a common value is $M=-10^8$. The $mask: \mathbb{R} \mapsto \mathbb{R}$ defines the masking process and allows to calculate the re-normalized probability distribution. The updated logits suppress the action probabilities to zero in the case of at least one remaining valid action. The $mask$ is either a constant or piecewise function depending on the availability of the specific actions. Assume that the MDP in the example reaches the terminal state directly after an action was taken and the reward is $r=+1$. Further, assume that action $a_2$ with the corresponding logit $l_2$ is masked out. Then, the action probabilities after the $mask$ has been applied become
\begin{subequations}
\begin{align}
    \pi'_\theta(\cdot{\mid}s_0) 
    &= softmax(mask([l_0,l_1,l_2,l_3])) \\
    &= softmax([l_0,l_1,M,l_3]) \\
    &= [0.33,0.33,0.00,0.33]
\end{align}
\end{subequations}
Note that the probability of the invalid action is not exactly but only virtually zero when a sufficiently large negative number $M$ is chosen.
The mask can be considered a state-dependent, differential function since the identity and constant parts are both differentiable on their own. The technique accelerates learning with action space restrictions by making the gradient that is passed backward from a masked selection zero. Consider the previous example again. The policy gradient without action masking is given by
\begin{subequations}
\begin{align}
    g_{\text{policy}}
    &= \nabla_\theta \log \pi_\theta(a_0 {\mid} s_0 G_0)  \\
    &= [0.75,0.25,0.25,0.25]
\end{align}
\end{subequations}
The corresponding invalid action policy gradient with masking is
\begin{subequations}
\begin{align}
    g_{\text{invalid action policy}} 
    &= \nabla_\theta \log \pi'_\theta(a_0{\mid}s_0) G_0 \\
    &= [0.67,-0.33,0.00,-0.33]
\end{align}
\end{subequations}
By this means, the gradient of the invalid action corresponds to zero and the logit of the mask is not updated \cite{huang2020closer}. 

In the case of DQN, the agent samples uniformly from the allowed set $\mathcal{A}_\varphi$ during exploration. For exploitation, the maximum action value is chosen out of $\mathcal{A}_\varphi$. The training needs to be adapted to incorporate the allowed actions as well. The new target $y_i$ considers only valid actions and can be calculated as 
\begin{equation}
    y_i = r_t + \gamma \max_{a' \in \mathcal{A}_\varphi} Q_{\theta'}(s',a')
\end{equation}
so that only valid actions are selected. The effect is similar to that of PPO since the gradient for actions that have not been selected does not exist in DQN \cite{saferloverview}. 

\subsection{Continuous Masking}
Continuous action masking can be implemented in several ways. However, all approaches add an additional layer after the output of a neural network. The final layer maps the unconstrained action space $\mathcal{A}$ to a subset $\mathcal{A}^\varphi$ through a scaling operation \cite{saferloverview}. 

This thesis builds on ConstraintNet. The architecture adds multiple layers to a neural network that maps input features $\mathcal{X}$ to output $\mathcal{Y}$. One of these layers performs scaling $\phi: \mathcal{Y} \times \mathcal{S} \mapsto \mathcal{C}$ but does not include learnable parameters. The scaled output is calculated based on the output of the original network $y \in \mathcal{Y}$ and a parameter vector $s \in \mathcal{S}$. The parameter $s$ defines the valid subset $\mathcal{C}(s)$ of the full output space $\mathcal{Y}$.  Nevertheless, since the scaling layer $\phi$ contains no learnable parameters, the logic must be given by the original network. Therefore, the scaling operation $\phi$ is chosen to be differentiable with respect to the parameters and the parameterization $s$ should be part of the input. The neural network has to learn to map the same datapoint $x$ to the true output $y$ under different constraints. If the restriction would be missing in the input, the same features would be mapped to a different location when the constraints change. Typically, a separate neural network $g(s)$ learns to transform the parameter to a suitable representation that is concatenated with an intermediate output $z$ of the original network. All operations can be summarized as 
\begin{equation}
    f_\theta(x,s) = \phi(h_{\theta_1}(z, g_{\theta_2}(s)), s)
\end{equation}
where $h_\theta$ denotes the layers of the neural network receiving the concatenated hidden feature and parameter representation as input. The scaling layer is designed to support convex polytopes $\mathcal{P}$ which are given by a convex hull of $M$ vertices $\{v^{(i)}\}_{i=1}^M$ of dimension N:
\begin{equation}
    \mathcal{P}(\{v^{(i)}\}_{i=1}^M) = \Bigl\{ \sum_i p_i v^{(i)}: p_i \leq 0, \sum_i p_i = 1 \Bigr\}
\end{equation}
When the vertices are functions of the parameter $s$, the scaling layer can be constructed such that representation $z$ is scaled into the polytope with
\begin{equation}
\label{eq:scale_a}
    \phi(z,s) = \sum_i \sigma_i(z_i) v^{(i)}(s)
\end{equation}
where $\sigma_i(\cdot)$ is the ith component of a softmax function. Multiple, non-convex constraints can be considered by incorporating softmax scores into the neural network. The highest probability decides on the constraint containing the ground truth. 
The scaling layer can be designed arbitrarily and application-dependent \cite{brosowsky2021sample}. An alternative to equation \ref{eq:scale_a} is to scale the unrestricted action into the feasible domain with equation \ref{eq:scale_b}. The $min(\cdot)$ and $max(\cdot)$ operations correspond to the minimum and maximum value of the full action space. 
\begin{equation}
\label{eq:scale_b}
    a^\varphi = (a - \min(\mathcal{A}))
    \frac{\max(\mathcal{A}^\varphi) - \min(\mathcal{A}^\varphi)}
    {\max(\mathcal{A}) - \min(\mathcal{A})} + \min(A^\varphi)
\end{equation}
The agent can explore the entire action space and the action probabilities are not influenced by the constraint \cite{saferloverview}. The action space perspective of continuous action masking is shown in figure \ref{fig:masking}. 

\subsection{Replacement and Projection}
\label{sec:repandproj}
Another way to restrict the action space of an agent is to replace an invalid action heuristically. Among other methods, this could be achieved by randomly sampling from the allowed action space or projecting the action from the unrestricted domain to the feasible space.
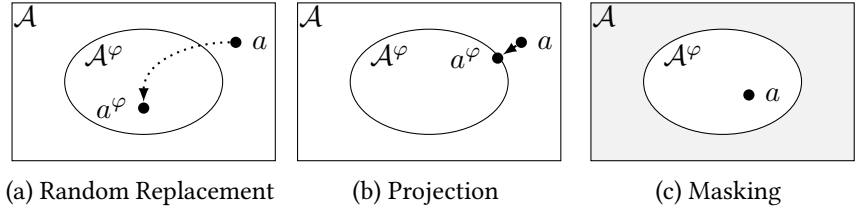
\begin{figure}[t]
\centering
\begin{subfigure}{.3\linewidth}
  \centering
  \begin{tikzpicture}[scale=0.7]
        \node at (1.75,2) {$\mathcal{A}^\varphi$};
        \node(original) at (4.25,2.25)[circle,fill,inner sep=1.5pt, label=right:{$a$}]{};
        \node(replaced) at (2.5,1)[circle,fill,inner sep=1.5pt, label=left:{$a^\varphi$}]{};
        \node at (0.25,2.75) {$\mathcal{A}$};
        
        \draw (0,0) rectangle (5,3);
        \draw (2.5,1.5) ellipse (1.5cm and 1cm);
        \draw[->, dotted, thick] (original) to[out=180,in=90] (replaced);
    \end{tikzpicture}
  \caption{Random Replacement}
  \label{fig:replacement}
\end{subfigure}%
\begin{subfigure}{.3\linewidth}
  \centering
  \begin{tikzpicture}[scale=0.7]
        \node at (1.75,2) {$\mathcal{A}^\varphi$};
        \node(original) at (4.25,2.25)[circle,fill,inner sep=1.5pt, label=right:{$a$}]{};
        \node(replaced) at (3.8,1.95)[circle,fill,inner sep=1.5pt, label=left:{$a^\varphi$}]{};
        \node at (0.25,2.75) {$\mathcal{A}$};
        
        \draw (0,0) rectangle (5,3);
        \draw (2.5,1.5) ellipse (1.5cm and 1cm);
        \draw[->, thick] (original) -- (replaced);
    \end{tikzpicture}
  \caption{Projection}
  \label{fig:projection}
\end{subfigure}
\begin{subfigure}{.3\linewidth}
  \centering
    \begin{tikzpicture}[scale=0.7]
        \draw (0,0) rectangle (5,3)[fill=gray!10];
        \draw (2.5,1.5) ellipse (1.5cm and 1cm)[fill=white];
        
        \node at (1.75,2) {$\mathcal{A}^\varphi$};
        \node(original) at (3,1.25)[circle,fill,inner sep=1.5pt, label=right:{$a$}]{};
        \node at (0.25,2.75) {$\mathcal{A}$};
    \end{tikzpicture}
  \caption{Masking}
  \label{fig:masking}
\end{subfigure}
\caption[Action space perspective of replacement, projection, and masking]{Action space perspective \cite{saferloverview}}
\label{fig:replacement_projection}
\end{figure}
The latter usually involves solving the following optimization problem 
\begin{subequations}
\label{eq:proj}
\begin{alignat}{2}
&\!\arg\min_{x}        &\qquad& \norm{a - a^\varphi}_p\label{eq:optProb}\\
&\text{subject to} &      & f_i(a^\varphi,s) \leq 0, \forall i \in 1,...,n
\end{alignat}
\end{subequations}
where $\norm{\cdot}_p$ is any p-norm with $p \geq 1$, $p \in \mathbb{R}$ and $f_i(a^\varphi,s)$ describes restrictions. 
The agents can be trained either with the tuple $(s,a,s',r)$ containing the original value or the tuple $(s,a^\varphi,s',r)$ with the replaced action. The unaltered action makes sense when the agent receives a penalty for violating the restriction. This way, the constraints are learned in a reward-driven way. Contrary, learning on the replaced action, updates the agent with a value that does not stem from its policy $\pi$. While this is expected in off-policy learning, it violates the assumption of on-policy algorithms. Figures \ref{fig:replacement} and \ref{fig:projection} depict replacement and projection from the action space perspective \cite{saferloverview}. 

\addtocontents{toc}{\protect\newpage}
\chapter{Methodology}
This thesis aims to investigate deep RL with dynamic interval restrictions on the action space. Chapter \ref{cha:ba} introduced the general framework with different algorithms and reviewed existing solutions. In order to find answers to the research questions, this chapter will first formally define what we consider interval restriction before their integration into RL is presented. Afterward, an evaluation environment is outlined together with the training cycle for experiments.

\section{Interval Restrictions}
\label{sec:interval_restrictions}
\begin{figure}[t]
\centering
\begin{subfigure}{.5\linewidth}
    \centering
    \begin{tikzpicture}[scale=1.2]
    \node(restriction_beginning) [circle,fill = white,inner sep=1.5pt, label=left:{$\mathcal{A}^\varphi$}] at (-2,-1) {};
    \node(restriction_beginning) [circle,fill = white,inner sep=1.5pt, label=left:{$\mathcal{C}$}] at (-2,-0.5) {};
    
    \node(hi) at (-2,-1.4) {\footnotesize -10};
    \node(hi) at (-1,-1.4) {\footnotesize -5};
    \node(hi) at (0,-1.4) {\footnotesize 0};
    \node(hi) at (1,-1.4) {\footnotesize 5};
    \node(hi) at (2,-1.4) {\footnotesize 10};

    \draw[-, dotted, thick] (-2, 0.2) -- (-2,-1.2);
    \draw[-, dotted, thick] (-1, 0.2) -- (-1,-1.2);
    \draw[-, dotted, thick] (0, 0.2) -- (0,-1.2);
    \draw[-, dotted, thick] (1, 0.2) -- (1,-1.2);
    \draw[-, dotted, thick] (2, 0.2) -- (2,-1.2);
    
    \node(action_space_beginning) [circle,fill = white,inner sep=1.5pt, label=left:{$\mathcal{A}$}] at (-2,0) {};
    \node(restriction_between_3) [circle,fill = red!80!black,inner sep=1.5pt] at (-1.5,-0.5) {};
    \node(restriction_between_1) [circle,fill = red!80!black,inner sep=1.5pt] at (-1.1,-0.5) {};
    \node(restriction_between_2) [circle,fill = red!80!black,inner sep=1.5pt] at (0.3,-0.5) {};
    \node(restriction_end) [circle,fill = red!80!black,inner sep=1.5pt] at (1.3,-0.5) {};

    \node(allowed_beginning) [circle,fill = green!70!black,inner sep=1.5pt] at (-2,-1) {};
    \node(allowed_end) [circle,fill = green!70!black,inner sep=1.5pt] at (-1.5,-1) {};

    \draw[very thick, red!80!black] (restriction_between_3) -- (restriction_between_1);
    \draw[very thick, red!80!black] (restriction_between_2) -- (restriction_end);
    \draw[very thick, asparagus] (allowed_beginning) -- (allowed_end);

    \draw[-, very thick] (-2,0) -- (2,0);
    \draw[-, very thick, asparagus] (-1.1,-1) -- (0.3,-1);
    \draw[-, very thick, asparagus] (1.3,-1) -- (2,-1);

    \fill (-2,0) circle (0.075);
    \fill[fill = asparagus] (-2,-1) circle (0.075);
    \fill[fill = asparagus] (-1.1,-1) circle (0.075);
    \fill[fill = asparagus] (1.3,-1) circle (0.075);
    \draw[color = red!80!black, fill=white] (-1.5,-0.5) circle (0.075);
    \draw[color = red!80!black, fill=white] (0.3,-0.5) circle (0.075);

    \fill[color = asparagus] (-1.5,-1) circle (0.075);
    \fill[color = asparagus] (0.3,-1) circle (0.075);
    \fill[color = asparagus] (2,-1) circle (0.075);
    \draw[color = red!80!black, fill=white] (-1.1,-0.5) circle (0.075);
    \draw[color = red!80!black, fill=white] (1.3,-0.5) circle (0.075);
    \fill[color = black] (2,0) circle (0.075);
    \node at (0,-1.6) {};

\end{tikzpicture}
    \caption{Exemplary action space}
    \label{fig:exemplary_action_space}
\end{subfigure}%
\begin{subfigure}{.5\linewidth}
    \centering
    \begin{tikzpicture}[
        >=stealth,
        rect/.style={
            rectangle,
            draw,
            rounded corners=5, 
            inner sep = 0.2cm,
            text depth=0.1em
        }]
        \node[rect] (agent) at (0,0) {Agent};
        \node[rect] (environment) at (0,-1.6) {Environment};
        
        \draw[->, thick] (agent) -- ++(2.5,0) |- 
            node[pos=0.25, right] {$a_t$} (environment);
        
        \coordinate (lower west) at ([shift={(-2cm, -4pt)}]environment.center);
        \coordinate (upper west) at ([shift={(-2cm, 4pt)}]environment.center);
        \coordinate (north) at ([shift={(0, 0.425cm)}]environment.north);
    
        \draw[densely dotted] ([yshift=5pt]upper west) -- ([yshift=-5pt]lower west);
        \draw[densely dotted] ([yshift=5pt]upper west) -- ([yshift=-5pt]lower west);
        \draw[densely dotted] ([xshift=5pt]north) -- ([xshift=-5pt]north);
        \draw[densely dotted] ([xshift=5pt]north) -- ([xshift=-5pt]north);
        
        \draw[->, thick] (environment.west |- upper west) -- (upper west)
            node[midway, above] {$r_{t+1}$};
            
        \draw[->, thick] (environment.west |- lower west) -- (lower west)
            node[midway, below] {$s_{t+1}$};
    
        \draw[->, thick] (upper west) -- ++({-0.5cm + 4pt},0) |- 
            node[pos=0.25, right] {$r_t$} ([yshift=-4pt]agent.west);
            
        \draw[->, thick] (lower west) -- ++({-0.5cm - 4pt},0) |- 
            node[pos=0.25, left] {$s_t$} ([yshift=4pt]agent.west);
    
        \draw[->, thick] (environment.north) -- (north) node[pos = 0.5, left] {$\mathcal{A}^\varphi_{t+1}$};
        \draw[->, thick] (north) -- (agent.south) node[pos = 0.5, right] {$\mathcal{A}^\varphi_t$};
        
    \end{tikzpicture}
    \caption{Restricted agent-environment interaction}
    \label{fig:restricted_loop}
\end{subfigure}
\caption{Dynamic interval restrictions framework}
\label{fig:interval_overview}
\end{figure}
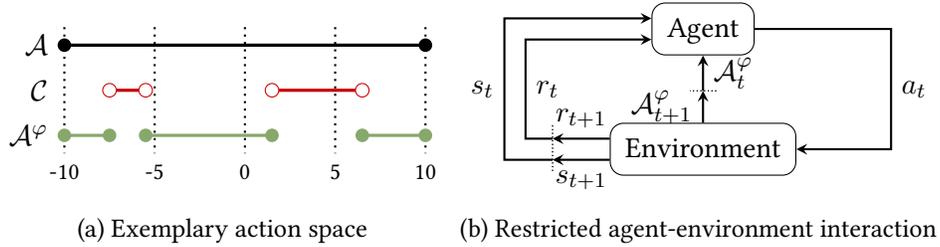

This thesis builds on an action space $\mathcal{A} = [a,b] \in \mathbb{R}$ as a one-dimensional, bounded, and closed interval. Both ends are closed and bounded to comply with the OpenAI Gym framework \cite{gym} which has become standard in most RL implementations \cite{liang2018rllib, raffin2019stable}. We consider restrictions $\mathcal{C}_t$ and allowed subsets $\mathcal{A}^\varphi_t$ of the form
\begin{equation}
    \mathcal{C}_t = \bigcup_{i=1}^{[\mathcal{C}]} (l_i, u_i) \subseteq \mathbb{R} \quad \text{and} \quad 
    \mathcal{A}^\varphi_t = \bigcup_{i=1}^{[\mathcal{A}^\varphi]} [a_i,b_i] = \mathcal{A} - \mathcal{C}_t \in \mathcal{A}
\end{equation}
where the size of the available and unavailable action space is defined analogously as $\abs{\mathcal{C}_t} = \sum_i (u_i - l_i)$ and $\abs{\mathcal{A}^\varphi_t} = \sum_i (b_i - a_i)$.
Representing restrictions as the union of open intervals has the advantage that the set is disjoint since $l_1 < u_1 < ... < l_i < u_i$. Furthermore, actions contained in one of the subsets are unavailable which, we believe, corresponds to the most natural representation of interval restrictions. A finite number of intervals is required so that the set is pre-computable. 
The valid action space is the difference between the entire action space and the restrictions. 
The resulting allowed actions are therefore also a possibly empty set of intervals. Each of the subsets is closed and bounded which is again in line with recent RL implementations. The property allows seamless integration of dynamically changing action spaces in existing environments.
When no constraints apply, the set of restrictions is empty $\mathcal{C}_t = \varnothing$, and the restricted equals the unrestricted action space $\mathcal{A}^\varphi_t = \mathcal{A}$. The restrictions are not limited to a subset of the action space, so constraints overlapping the boundaries can make all actions unavailable. Figure \ref{fig:exemplary_action_space} shows an exemplary scenario and valid subsets.
We consider that the intervals in both sets $\mathcal{C}_t$ and $\mathcal{A}^\varphi_t$ are dynamic. Dynamic means that the restrictions and allowed actions can change when the environment transitions. The sets only hold for a single time step $t$. Contrary to most literature \cite{saferloverview, paternain2019constrained}, we update the agent-environment loop: The agent receives the valid action space for the next time step $t+1$ together with the state from the environment. Therefore, the restrictions  are assumed to be provided and do not have to be derived from the recent observation. The updated interactions are illustrated in figure \ref{fig:restricted_loop}. Note that the restrictions for the same state representation are not necessarily equal, i.e., $s_t = s_{t'} \nRightarrow \mathcal{C}_t = \mathcal{C}_{t'}$. 

\section{Algorithms}
After the previous section introduced the necessary notations, the following is intended to answer the first research question and describes algorithms to handle dynamic interval restrictions. To evaluate the performance of our modifications, we compare them with different methods in the literature. We first elaborate on details of the baselines before our proposed improvements are presented. 

\subsection{Baselines}
In the experiments, we include a reward-driven approach and the replacement, projection, and masking methods from section \ref{sec:res}. Masking is applied both, on a continuous and discrete action space since discretization turned out to be a common procedure in the literature \cite{huang2020closer}. The reason is that discrete masking is theoretically well-founded with promising empirical results. We compare our algorithms with all practices as it is unclear which baseline achieves the highest performance. To the best of our knowledge, the approaches have only been applied to convex continuous subsets. Therefore, we extend the techniques to multiple intervals and combine them with PPO and TD3. Since TD3 is limited to continuous spaces, DQN is used for discrete actions. We select the algorithms to include diverse methods. We believe TD3 and PPO are the state of the art in deterministic off-policy and stochastic on-policy learning. Our list of algorithms is presented below. It would also be possible to compare our methods with others in the field of obstacle avoidance, such as observations from lasers with a higher reach or artificial potential fields. However, we consider only information to look for one step ahead and assume that no global information or image data is available. 

\begin{enumerate}
    \item \textbf{Penalty-Based Learning}. The penalty-based approach is implemented such that the agent receives a constant penalty when a restriction is violated and the episode is reset. We experimented with different scales of rewards $r_{coll} \in \{0, -1, -5, -10, -20\}$ but perform the experiments with a constant value of $-20$. Contrary to Marchesini et al. \cite{marchesini2020discrete}, who report varying results with different rewards, our algorithms showed a consistent behavior. Therefore, we rely on the maximum scale. Similar to the literature \cite{huang2020closer}, we expect the restriction awareness to increase with higher values. Since the technique must be provided with information about obstacles, the observations are extended to include the restrictions. This is done by defining a maximum number of intervals and zero-padding them. For example, the set of allowed actions $[(2.0, 8.0)]$ with a maximum number of three inputs would be represented as $(2.0, 8.0, 0.0, 0.0, 0.0, 0.0)$. Of course, this assumption cannot be made in most real-life applications. Nevertheless, the general concept is similar to most solutions in obstacle avoidance: Information about close surroundings is given and obstacles are avoided by learning to recognize them in observations with a constant collision penalty. Examples are Choi et al. \cite{choi2021reinforcement} and Tai et al. \cite{tai2017virtual}. Note that the baseline is not violation free and not of direct interest to this thesis. We are interested in methods that are continuously following restrictions. However, we believe the approach is the best practice to address our research questions and has shown good results, for example in \cite{wang2018learning}, when navigating around dynamic obstacles. 

    \item \textbf{Replacement and Projection}. Replacement and projection techniques follow the implementations of Brosowski et al. \cite{saferloverview} and are considered in two forms: First, in case of an invalid action, a random replacement baseline samples uniformly from the allowed action space. Second, the projection baseline finds the closest valid action by minimizing the euclidean distance. The reward is unaltered and determined by the executed valid action. The loss for PPO is computed based on the experience tuple containing the invalid action to not break with the algorithm assumptions. Similarly, DQN and DDPG learn on the tuple with the original action as no difference could be found in previous research \cite{saferloverview}. Likewise to the above, we include the zero-padded intervals in the observation. The projection baseline can be described using equation \ref{eq:proj} from section \ref{sec:repandproj}: This is achieved by setting $p=2$ and introducing a constraint $f_{[u_i,l_i]}(a) = (a-u_i)(l_i-a) < 0$ for each restriction $i$. The constraint is smaller than zero whenever action $a$ is not contained in the interval $[u_i, l_i]$. 

    \item \textbf{Discrete Masking}. Discrete masking is done similarly to many review papers on a discretized continuous action space. However, we do not consider domain knowledge for discretization: The action space is split into $k$ equally spaced atomic actions. We follow the suggestions of Tang et al. \cite{tang2020discretizing} and calculate the specific continuous action for a discrete choice $i$ as 
    \begin{equation}
        \mathcal{A}_i = \frac{i(\max(\mathcal{A}) - \min(\mathcal{A}))}{k-1} - 1
    \end{equation}
    The authors discuss that when the number of atomic actions is too small, the agent might not be able to reach every continuous position. Likewise, if it is too large, learning might be slower and more difficult. Nevertheless, to the best of our knowledge, most papers on obstacle avoidance still consider rather coarse-grained action spaces in their evaluation. Therefore, we also test masking only with nine discrete actions. 
    To prevent the agent from taking unavailable actions, the mask is applied with $M = -3.4^{38}$. In the case of PPO, $M$ is added to the logits before the softmax layer and for TD3 to the action values. The observation contains no information about restrictions. 

    \item \textbf{Continuous Masking}. Continuous masking is implemented likewise to ConstraintNet \cite{brosowsky2021sample} with the scaling function from Brosowski et al. \cite{saferloverview} which is defined in equation \ref{eq:scale_b}. However, we add small modifications to support multiple intervals since the authors only propose an architecture to scale actions to convex subsets.  First, we think of multiple disjoint intervals as a concatenated version which is in turn convex. This allows the final layer to scale the output in the full action space to the valid subset with only a small change: The size of the action space in the numerator is calculated over all intervals. This leads to the final equation
    \begin{equation}
        a' = 
        (a - \min(\mathcal{A}) )
        \frac{\sum_i^{[\mathcal{A^\varphi}]} (b_i - a_i)}
        {\max(\mathcal{A}) - \min(\mathcal{A})}
    \end{equation}
    After scaling, the minimum of the original interval belonging to $a'$ is added to get the valid action $a^\varphi$. The step can be interpreted as splitting the concatenated intervals again and placing them at their specific location. Second, we do not process the observation and constraints first separately and pass the concatenated representations through shared layers. We directly add zero-padded intervals to the input observation similar to the above baselines. 
\end{enumerate}

\subsection{Parameterized Action Masking}
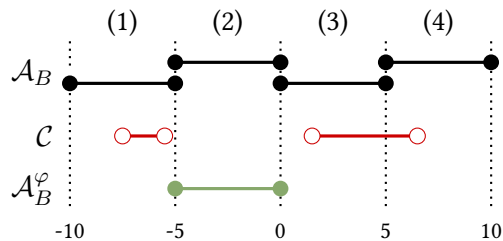
\begin{figure}[b]
\centering
\begin{tikzpicture}[scale=1.4]
    \node(restriction_beginning) [circle,fill = white,inner sep=1.5pt, label=left:{$\mathcal{A}^\varphi_B$}] at (-2,-1) {};
    \node(restriction_beginning) [circle,fill = white,inner sep=1.5pt, label=left:{$\mathcal{C}$}] at (-2,-0.5) {};

    \draw[-, dotted, thick] (-2, 0.4) -- (-2,-1.2);
    \draw[-, dotted, thick] (-1, 0.4) -- (-1,-1.2);
    \draw[-, dotted, thick] (0, 0.4) -- (0,-1.2);
    \draw[-, dotted, thick] (1, 0.4) -- (1,-1.2);
    \draw[-, dotted, thick] (2, 0.4) -- (2,-1.2);
    
    \node(a1) [circle,fill = white,inner sep=1.5pt] at (-1,0) {};
    \node(a2) [circle,fill = black,inner sep=1.5pt] at (0,0) {};
    \node(a3) [circle,fill = black,inner sep=1.5pt] at (1,0) {};

    \node(action_space_beginning) [circle,fill = white,inner sep=1.5pt, label=left:{$\mathcal{A}_B$}] at (-2,0.1) {};
    \node(restriction_between_3) [circle,fill = red!80!black,inner sep=1.5pt] at (-1.5,-0.5) {};
    \node(restriction_between_1) [circle,fill = red!80!black,inner sep=1.5pt] at (-1.1,-0.5) {};
    \node(restriction_between_2) [circle,fill = red!80!black,inner sep=1.5pt] at (0.3,-0.5) {};
    \node(restriction_end) [circle,fill = red!80!black,inner sep=1.5pt] at (1.3,-0.5) {};
    \node(action_space_beginning) [circle,fill = white,inner sep=1.5pt, label=above:{(1)}] at (-1.5,0.3) {};
    \node(action_space_beginning) [circle,fill = white,inner sep=1.5pt, label=above:{(2)}] at (-0.5,0.3) {};
    \node(action_space_beginning) [circle,fill = white,inner sep=1.5pt, label=above:{(3)}] at (0.5,0.3) {};
    \node(action_space_beginning) [circle,fill = white,inner sep=1.5pt, label=above:{(4)}] at (1.5,0.3) {};

    \node(hi) at (-2,-1.4) {\footnotesize -10};
    \node(hi) at (-1,-1.4) {\footnotesize -5};
    \node(hi) at (0,-1.4) {\footnotesize 0};
    \node(hi) at (1,-1.4) {\footnotesize 5};
    \node(hi) at (2,-1.4) {\footnotesize 10};

    \node(allowed_beginning) [circle,fill = green!70!black,inner sep=1.5pt] at (-1,-1) {};
    \node(allowed_end) [circle,fill = green!70!black,inner sep=1.5pt] at (0,-1) {};

    \draw[very thick, red!80!black] (restriction_between_3) -- (restriction_between_1);
    \draw[very thick, red!80!black] (restriction_between_2) -- (restriction_end);
    \draw[very thick, asparagus] (allowed_beginning) -- (allowed_end);

    \draw[-, very thick] (1,0.2) -- (2,0.2);
    \draw[-, very thick] (0,0) -- (1,0);
    \draw[-, very thick] (-1,0.2) -- (0,0.2);
    \draw[-, very thick] (-2,0) -- (-1,0);

    \fill (1,0.2) circle (0.075);
    \fill (0,0) circle (0.075);
    \fill (-1,0.2) circle (0.075);
    \fill (-2,0) circle (0.075);
    \fill[fill = asparagus] (-1,-1) circle (0.075);
    \draw[color = red!80!black, fill = white] (-1.5,-0.5) circle (0.075);
    \draw[color = red!80!black, fill = white] (0.3,-0.5) circle (0.075);

    \fill(a1)[color = asparagus] (0,-1) circle (0.075);
    \draw(a1)[color = red!80!black, fill=white] (-1.1,-0.5) circle (0.075);
    \draw(a1)[color = red!80!black, fill=white] (1.3,-0.5) circle (0.075);
    \fill(a1)[color = black] (2,0.2) circle (0.075);
    \fill(a1)[color = black] (1,0) circle (0.075);
    \fill(a1)[color = black] (0,0.2) circle (0.075);
    \fill(a1)[color = black] (-1,0) circle (0.075);

\end{tikzpicture}
\caption[Binned action space]{Binned action space}
\label{fig:binned-action-space}
\end{figure}

All of the baselines from the previous section bring their own limitations: Either the maximum number of restriction intervals must be predefined or atomic, discrete actions selected. Since our definition of restrictions does not include an upper limit, we propose parameterized action masking (PAM) which translates the continuous into a discrete-continuous action space and can deal with restrictions of an arbitrary number of intervals.
The motivation is to reuse the well-justified discrete action masking but lower the discretization error caused by naively choosing a small number of atomic actions. Consider the example of an agent following a curved path with minor angle changes. Even if the action space is not restricted, the agent cannot learn optimal steps due to the coarse actions. Therefore, an improvement could be that all continuous actions stay available when no restrictions apply and discretization for masking is only used when necessary. For this reason, we divide the action space into a predefined number $\abs{k}$ of bins $[a_k,b_k] \in \mathcal{A}$. In this context, every action is assigned to at least one of the ranges, i.e., $\bigcup_{k}[a_k,b_k] = \mathcal{A}$. At the next time step, the agent selects a bin $k$ and a continuous action $x_k \in [a_k,b_k]$ in this range. The advantage of the discrete choice of bins is that the representation allows to apply discrete action masking. This makes a range unavailable when invalid actions are contained. Figure \ref{fig:binned-action-space} shows an exemplary action space $\mathcal{A}_B$ divided into four bins. With the illustrated restrictions, only actions in the second interval are available. The other bins contain unavailable subsets. 

The procedure can be implemented based on P-DQN by adding a scaling and masking layer to the architecture. We rely on the architecture as it is simple to implement, most comparable to DQN and action value masking has shown promising results in similar works \cite{choi2022marl}. Although the method can indicate the effect of fine-grained actions, future research can consider other parameterization techniques. 
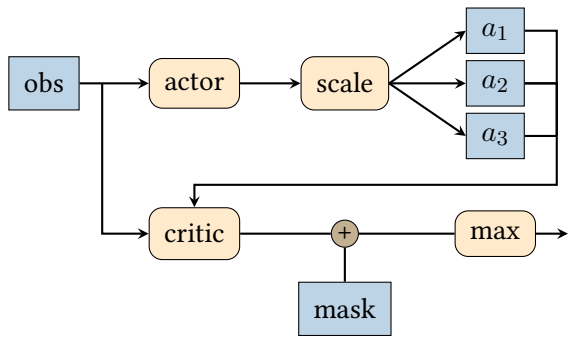
\begin{figure}[t]
\centering
\begin{tikzpicture}[
    >=stealth,
    rect/.style={
        rectangle,
        draw,
        inner sep = 0.2cm,
        text depth=0.1em
    }]
    \node[rect, fill = beaublue] (obs) at (0,0) {obs};
    \node[rect, rounded corners=5, fill = blanchedalmond] (actor) at (2,0) {actor};
    \node[rect, rounded corners=5, fill = blanchedalmond] (scale) at (4,0) {scale};
    
    \node[rect, fill = beaublue] (a1) at (6,0.7) {$a_1$};
    \node[rect, fill = beaublue] (a2) at (6,0) {$a_2$};
    \node[rect, fill = beaublue] (a3) at (6,-0.7) {$a_3$};

    \coordinate (a1_coord) at ([shift={(0.425cm, 0)}]a1.east);
    \coordinate (a2_coord) at ([shift={(0.425cm, 0)}]a2.east);
    \coordinate (a3_coord) at ([shift={(0.425cm, 0)}]a3.east);
    \coordinate (actions_coord) at ([shift={(0.425cm, -0.65)}]a3.east);

    \node[rect, rounded corners=5, fill = blanchedalmond] (critic) at (2,-2) {critic};

    \node[rect, fill = beaublue] (mask) at (4,-3) {mask};

    \coordinate (critic_coord) at ([shift={(0, 0.65)}]critic.center);
    \coordinate (critic_coord2) at ([shift={(-0.425cm, 0)}]critic.west);

    \draw (4,-2) node[align=center,minimum size=4pt,draw,circle,inner sep = 0.03cm, fill =khaki] (test) {+};

    \node[rect, rounded corners=5, fill = blanchedalmond] (max) at (6,-2) {max};

    \draw[->, thick] (obs.east) -- (actor.west);
    \draw[->, thick] (actor.east) -- (scale.west);
    \draw[->, thick] (scale.east) -- (a1.west);
    \draw[->, thick] (scale.east) -- (a2.west);
    \draw[->, thick] (scale.east) -- (a3.west);

    \draw[-, thick] (a1.east) -- (a1_coord) -- (a2_coord);
    \draw[-, thick] (a2.east) -- (a2_coord) -- (a3_coord);
    \draw[->, thick] (a2.east) -- (a2_coord) -- (actions_coord) -- (critic_coord) -- (critic.north);
    \draw[-, thick] (a3.east) -- (a3_coord);
    \draw[-, thick] (critic.east) -- (test);
    \draw[-, thick] (test.east) -- (max.west);
    \draw[-, thick] (mask.north) -- (test.south);
    \draw[->, thick] (max.east) -- ([shift={(0.425cm, 0)}]max.east);
    \draw[->, thick] ([shift={(0.3cm, 0)}]obs.east) --++ (0,-2) -- (critic.west);
\end{tikzpicture}
\caption[PAM architecture]{PAM architecture}
\label{fig:pas}
\end{figure}
The choice of bins is the categorical variable $k$, and the action the corresponding parameterization $x_k$. Figure \ref{fig:pas} illustrates an exemplary architecture with three predefined action bins.
As proposed by Xiong et al. \cite{xiong2018parametrized}, the algorithm uses an actor-critic approach: The actor receives the recent observation as input and outputs a continuous value for each bin. In our case, the output is an action in the full action space. Next, the added scaling layer maps the actions to the associated interval using equation \ref{eq:scale_b}. Now, the actions can be interpreted as giving the highest expected cumulative reward within each range. Note that some of them can be unavailable at this point.
The observation is concatenated with the actions and passed through the critic to obtain action values for each choice. Afterward, the masking layer restricts the action space by adding a large negative number $M=-3.4^{38}$ to the action values of unavailable bins. The mask is calculated such that intervals containing at least a single unavailable action are "masked out". An alternative is to remove only those for which the corresponding parameterization is invalid. We experiment with the first approach since training in partially available bins can be difficult. Whenever the continuous action moves into an unavailable subset, it is not updated anymore because the gradient through the mask shrinks to zero. However, the alternative approach leaves space for future research. 
Finally, the action selection is performed by choosing the maximum of the action values or selecting a value uniformly sampled from the available bins according to the $\epsilon$-greedy strategy. 
The weights are updated using the same loss functions as the original architecture which are described in section \ref{sec:p-dqn}. This is possible since no more learnable parameters are introduced. Only the actor and critic neural networks contain weights and the additional layers are differentiable. The training is compatible with all improvements such as target networks. 

An advantage over the discrete masking baseline is that all actions in the unrestricted action space stay available. Compared to continuous masking, the behavior of the scaling layer must not be learned. However, the number of bins is a hyperparameter that must be specified and the discretization error persists to some extent. Whenever only a small subset of a bin is unavailable, all included actions are taken out of the action space. Similar to discretization, we expect to see better performance with a higher number of bins. We test the same bin count of nine as discrete actions in the baseline for a fair comparison.

\subsection{Multi-Pass Scaled TD3}
Overcoming the discretization error and avoiding hyperparameters, such as a number of predefined bins or an upper limit on the interval count, motivated multi-pass scaled TD3 (MPS-TD3). Additionally, we wanted to integrate learning constraints through the observation space because of ConstraintNet's recent promising results. The idea is to split the non-convex, allowed action space into single intervals which are in turn convex. This way, previously suggested convex scaling methods can be applied and the discretization error eliminated. The agent operates on each subset separately and outputs an action for each interval through multiple forward passes. This procedure can be applied to any number of intervals and dispenses zero-padding. The critic can be used to select the final action by maximizing the action value. Relying on the critic has the advantage that the approach does not introduce additional components as it is part of actor-critic algorithms. 

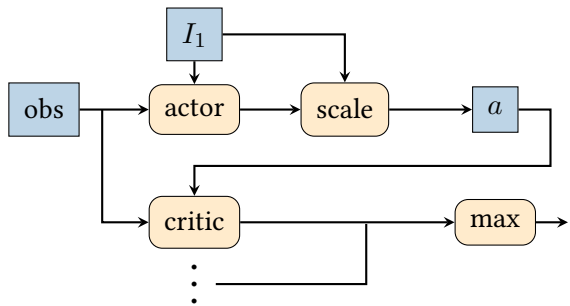
\begin{figure}[t]
\centering
\begin{tikzpicture}[
    >=stealth,
    rect/.style={
        rectangle,
        draw,
        inner sep = 0.2cm,
        text depth=0.1em
    }]
    \node[rect, fill = beaublue] (interval) at (2,1) {$I_1$};
    \node[rect, fill = beaublue] (obs) at (0,0) {obs};
    \node[rect, rounded corners=5, fill = blanchedalmond] (actor) at (2,0) {actor};
    \node[rect, rounded corners=5, fill = blanchedalmond] (scale) at (4,0) {scale};
    
    \node[rect, fill = beaublue] (a2) at (6,0) {$a$};

    \coordinate (a2_coord) at ([shift={(0.425cm, 0)}]a2.east);
    \coordinate (actions_coord) at ([shift={(0.425cm, -0.75)}]a2.east);

    \node[rect, rounded corners=5, fill = blanchedalmond] (critic) at (2,-1.5) {critic};

    \coordinate (critic_coord) at ([shift={(0, 0.75)}]critic.center);
    
    \node[rect, rounded corners=5, fill = blanchedalmond] (max) at (6,-1.5) {max};

    \draw[->, thick] (obs.east) -- (actor.west);
    \draw[->, thick] (actor.east) -- (scale.west);
    \draw[->, thick] (scale.east) -- (a2.west);

    \draw[->, thick] (a2.east) -- (a2_coord) -- (actions_coord) -- (critic_coord) -- (critic.north);
    \draw[->, thick] (critic.east) -- (max.west);
    \draw[->, thick] (max.east) -- ([shift={(0.425cm, 0)}]max.east);
    \draw[->, thick] (interval.south) -- (actor.north);
    \draw[->, thick] (interval.east) -- (4,1) -- (scale.north);
    \path (critic.south) -- ([shift={(0, -0.6cm)}]critic.south) node(dots)[midway,scale=1.6] {$\vdots$};
    \draw[-, thick] ([shift={(0, -0.17)}]dots.east) --++ (2,0) --++ (0,0.8);
    \draw[->, thick] ([shift={(0.3cm, 0)}]obs.east) --++ (0,-1.5) -- (critic.west);
\end{tikzpicture}
\caption[MPS-TD3 architecture]{MPS-TD3 architecture}
\label{fig:mps-ddpg}
\end{figure}

The concept is realized based on TD3 to ensure comparability with the baselines. Furthermore, the algorithm is chosen as it follows an actor-critic approach which does not require an additional value network and the deterministic action needs no extra sampling step. However, the method can still be applied to other approaches either. In this case, the value function might be an additional neural network or linear regression. The architecture of MPS-TD3 is illustrated in figure \ref{fig:mps-ddpg}. The allowed action space is split into single intervals in the first step. Each resulting subset is concatenated with the observation and passed through the architecture separately.  
Both, the observation and allowed interval are necessary inputs for the same reason as in ConstraintNet: Even if the action with the highest reward stays the same and is output correctly by the actor in the unscaled space, different intervals would map the output to different points in the valid action space. For this reason, the agent is supposed to learn the scaling behavior.
Based on the concatenated inputs, the actor first outputs an action in the unrestricted space. Next, the action is mapped to the interval by parameterizing equation \ref{eq:scale_b} in the scaling layer. Contrary to PAM, the value can only represent a valid action.
Afterward, the critic receives the scaled action with the observation as input. The output is the corresponding action value. The procedure is repeated for all intervals and the action with the highest expected return is selected. When the allowed action space is empty and an action must be executed, we suggest sampling uniformly from the unrestricted domain. However, in other cases, it might be more useful to replace the action and do nothing since everything is forbidden. In terms of exploration, a mixture of Gaussian noise and the $\epsilon$-greedy strategy results in the most stable training. We hypothesize that the latter explores across and action noise within intervals. The training steps are similar to TD3 as described in section \ref{sec:td3}. Note that the weights are only updated based on the action which was executed in the environment since no training tuples are available in the other cases. The training can be performed with all proposed improvements. 

MPS-TD3 can be compared to the continuous action masking baseline. The difference is that the disjoint intervals are not concatenated but split. As previously, the zero-padding can be dispensed this way. 
Nonetheless, multiple forward passes drastically increase runtime. However, we can make use of the suggestion from Bester et al. \cite{bester2019multi}. The authors use the parallel mini-batch processing capabilities of the PyTorch library to process multiple forward passes as a single, parallel pass. The actions for intervals are computed similarly to a mini-batch of observations. 

\section{Environment}
\label{sec:env}
After presenting different approaches, the following illustrates an environment to compare algorithms in dealing with interval action space restrictions. We first define fundamental design properties before the actual task is described. 

\subsection{Desiderata}
\label{sec:properties}
In order to answer the research question of how the performance with different algorithms can be measured, we developed five fundamental criteria for evaluation environments:

\begin{enumerate}
    \item \textbf{Interval restrictions must naturally arise}. The constraints are an inevitable part of solving the specific task. This property shows the relevance of the research questions and allows the comparison to other methods in the field. Contrary, the introduction of artificial restrictions would be rather complex, must be well justified and the transferability is not guaranteed. 
    \item \textbf{Strategic restriction handling gives an advantage}. Varying allowed subsets lead to different optimal actions which are predictable. This is necessary to evaluate algorithms. Otherwise, arbitrary actions from the allowed subsets can be executed and the differences are due to chances.
    \item \textbf{Restrictions are dynamic}. The constraints for the same observation can differ. Agents must learn to find the optimal action over arbitrary subsets. When the restriction is coupled with the observation, the problem reduces to finding an optimal policy with restrictions as part of the state. 
    \item \textbf{The number of restrictions can be adjusted}. The environment setup can be designed to produce more or fewer subsets. This is necessary to check the scalability of algorithms. The integration allows investigating if the same methods show a high reward under different conditions.
    \item \textbf{The proportion of the available actions can be changed} Similarly to the number of restrictions, the ratio of the available to the unavailable actions can have an impact on the reward of algorithms and results may differ.
\end{enumerate}
We assume that the above points must hold to measure the effect of previously presented restrictions.

\subsection{Overview}
\begin{figure}[t]
\centering
\begin{subfigure}{.5\linewidth}
  \centering
  \fbox{\includegraphics[width=5cm]{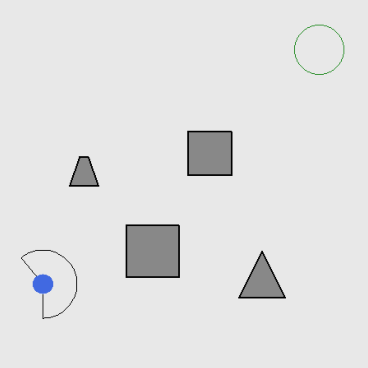}}
  \caption[Exemplary setup]{Exemplary setup}
	\label{fig:environment-example}
\end{subfigure}%
\begin{subfigure}{.5\linewidth}
  \centering
  \begin{tikzpicture}[scale=1.4]
    \fill[fill = anti-flashwhite] (1.5,0) arc(0:180:1.5);
    \draw[black, fill = ashgrey] (-0.1,1.2) rectangle (1.4,1.4);
    
    \begin{scope}
      \clip(-0.5,0.8) rectangle (1.8,1.8);
      \filldraw[palechestnut, color=palechestnut] (1.5,0) arc(0:180:1.5);
    \end{scope}
    
    \draw[inchworm, thick, rotate around = {130:(0,0.0)}] (1.5,0) arc(0:35:1.5);
    \draw[black, thick, rotate around = {165:(0,0.0)}] (1.5,0) arc(0:15:1.5);
    \draw[black, thick, rotate around = {25:(0,0.0)}] (1.5,0) arc(0:105:1.5);
    \draw[inchworm, thick, rotate around = {15:(0,0.0)}] (1.5,0) arc(0:10:1.5);
    \draw[black, thick] (1.5,0) arc(0:15:1.5);
    \draw[black, thick] (-0.5,0.8) rectangle (1.8,1.8);

    \draw[-, rotate around = {15:(0,0.)}] (1.4,0) -- (1.6,0);
    \draw[-, rotate around = {165:(0,0.)}] (1.4,0) -- (1.6,0);

    \node at (1.6,0.95) {\footnotesize $D$};
    \node at (0.9,0.95) {\footnotesize $I$};
    \node at (-1,1.3) {\footnotesize $d^{\text{safe}}$};
    \node at (0.05,0.45) {\footnotesize $s^{\text{agent}}$};
    \node[circle,fill = black,inner sep=1.5pt] at (0,0) {};
    \node at (1.3,1.6) {\footnotesize $r^{\text{agent}}$};
    \node at (1.75,0.3) {\footnotesize $a_{\text{max}}$};
    \node at (-1.75,0.3) {\footnotesize $a_{\text{min}}$};
    \node at (1.15,0.15) {\footnotesize $\mathcal{P}^\text{step}$};

    \draw[-, thick, rotate around = {32:(0,0)}] (0,0) -- (1.5,0);
    \draw[-, thick, rotate around = {25:(0,0)}] (0,0) -- (1.5,0);
    \draw[-, thick, rotate around = {123:(0,0)}] (0,0) -- (1.5,0);
    \draw[-, thick, rotate around = {130:(0,0)}] (0,0) -- (1.5,0);

    \draw[<->] (1,1.4) -- (1,1.8);

    \node at (0,-0.2) {\footnotesize $(x^{\text{agent}}, y^{\text{agent}})$};
    \node at (0,-0.5) {};
    \end{tikzpicture}
  \caption{Restriction derivation}
  \label{fig:restriction-derivation}
\end{subfigure}
\caption[Obstacle avoidance environment overview]{Obstacle avoidance environment overview}
\label{fig:env_overview}
\end{figure}
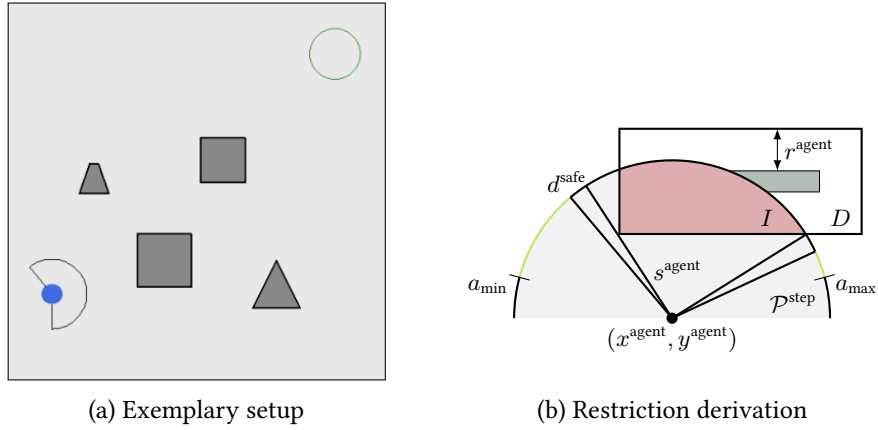

We designed an environment based on aspects (1) - (5) from the previous subsection: The task of the agent is to find the optimal path to a goal while avoiding random obstacles. The experiment corresponds to what is often considered mapless navigation in the literature \cite{choi2021reinforcement, marchesini2020discrete}. In the following, we will show that this naturally produces interval restrictions (1).
An exemplary setup of the environment is illustrated in figure \ref{fig:environment-example}.

The main components are the agent, goal, and obstacles on a map of a fixed width $w$ and height $h$. 
The agent is represented as a blue circle and associated with a radius $r^{\text{agent}}$, current location $(x^{\text{agent}}_t, y^{\text{agent}}_t)$, perspective $p^{\text{agent}}_t \in [0,360)$ and fixed step size $s^{\text{agent}}$. The perspective is the angle with respect to the x-axis. At each time step $t$, the agent chooses an angle $a^{\text{agent}}_t$ in which the subsequent step is taken. This leads to the next state with $p^{\text{agent}}_{t+1} = p^{\text{agent}}_t + a^{\text{agent}}_t$ and $(x^{\text{agent}}_{t+1},y^{\text{agent}}_{t+1})$. The agent can only choose from a limited range of actions $[a_{\text{min}}, a_{\text{max}}]$ to avoid rapid and rather unrealistic direction changes. The goal is to reach the target which is illustrated in the top right corner. The location of the goal $(x^{\text{goal}}, y^\text{goal})$ is reached when the distance $d^{\text{goal}}_t$ between the agent and goal is less than a threshold $r^{\text{goal}}$. Hereby, the distance between the agent and the target location is calculated as the euclidean distance 
\begin{equation}
    d_{\text{goal}} = \sqrt{(x^{\text{goal}}_t - x^{\text{agent}}_t)^2 + (y^{\text{goal}}_t - y^{\text{agent}}_t)^2}
\end{equation}
Similarly, the obstacles are defined by a location $(x^{\text{obstacle}}, y^{\text{obstacle}})$, radius $r^{\text{obstacle}}$ and form $f^{\text{obstacle}} \in \{\text{rectangle}, \text{trapeze}, \text{triangle}, \text{octagon}\}$. Different shapes have the advantage that it is harder for the agent to generalize. This assures the need for strategic restriction handling (2). A shared strategy that works for all obstacles must be developed. Cimurs et al. \cite{cimurs2020goal} give examples of different behaviors. Each obstacle is represented as a polygon associated with a set of coordinates $\mathcal{P}^{\text{obstacle}}$. To allow dynamic restrictions (3), multiple waypoints $(x^{\text{waypoint}}, y^{\text{waypoint}})$ can be optionally added to a list associated with an obstacle. The obstacle then moves with step size $s^{\text{obstacle}}$ between the coordinates from the initial location to the end of the list. When the last waypoint is reached, the path is traveled backward. 

The environment characteristic (3) is further implemented by determining the position and behavior of some obstacles randomly before each episode by using algorithm \ref{alg:obstacle-generation}. The algorithm is inspired by Meyer et al. \cite{meyer2020taming}. The procedure takes a number of obstacles and waypoints as input and uses multiple probability distributions to sample a setup. We first draw a form $f^i$ and radius $r^i$ for obstacle $i$ before a random location on the map is generated. Sampling is repeated until a valid location is found. This way, collisions with existing obstacles, the starting position, and the goal are avoided. Locations must also keep a minimum distance $r^{\text{agent}}$ to avoid dead ends. Afterward, the waypoints are defined by finding valid distances and directions from the last location in the list of coordinates. The parameters for the distributions can be individually set. This allows, for example, to bias the location of obstacles to the center of the map to raise the probability that the agent encounters them. Furthermore, varying the obstacle count and their radius $r^{\text{obstacle}}$ complies with properties (4) and (5): By increasing $r^{\text{obstacle}}$, the ratio of the available to the unavailable action space should shrink. Likewise, the number of disjoint subsets is expected to increase with more but smaller obstacles. 

When assuming that the actions are not limited to specific angles and without considering collisions, the set of possible locations in the next time step forms a circle with radius $s^{\text{agent}}$ and center $(x_t^{\text{agent}},y_t^{\text{agent}})$.
\begin{algorithm}[t]
\caption[Obstacle Generation]{Obstacle Generation} \label{alg:obstacle-generation}
    Initialize obstacles $\gets \{ \}$
    \begin{algorithmic}[1]
    \WHILE{$obstacles \neq num\_obstacles$}
        \STATE Draw form $f^i$ uniformly with probability $\frac{1}{\abs{f}}$
        \STATE Draw radius $r^i \sim \mathcal{N} (\mu_1, \sigma^2_1)$
        \STATE Draw location $(x^i, y^i) \sim \mathcal{N} (\mu_2, \Sigma)$ until valid
        \IF{$valid$}
        \STATE obstacles.insert($(f^i,r^i,x^i)$)
            \WHILE{$obstacles.waypoints \neq num\_waypoints$}
                \STATE Draw distance $d^k \sim \mathcal{N} (\mu_3, \sigma^2_2)$
                \STATE Draw direction $a^k \sim \mathcal{N} (\mu_4, \sigma_3)$ until valid
                \STATE obstacles.waypoints.insert($d^k,a^k$)
            \ENDWHILE
        \ENDIF
    \ENDWHILE
    \end{algorithmic}
\end{algorithm}
Take figure \ref{fig:restriction-derivation} as an example. We make use of this property and calculate restrictions with the intersection $I$ between the cyclic polygon $\mathcal{P}^{\text{step}}$ and a collision zone $D$. The collision zone describes the area in which an agent would collide with an obstacle due to the radius $r^{\text{agent}}$. The area can be computed by adding a margin of $r^{\text{agent}}$ to each side of the obstacle polygon. Subsequently, we compute all angles between the agent and the corner points of intersection $I$. Hereby, an important property of this set is that the maximum and minimum define the range of actions which would lead to a collision. This way, it is ensured that the agent can not jump over an obstacle between time steps. Since it is not possible to stand exactly on the border of an obstacle, we further add a small safety distance $d^{\text{safe}}$ with which the final restriction $\mathcal{C}_i = (l_i,u_i)$ is computed. The procedure is repeated for every intersection geometry until the whole set of restrictions $\mathcal{C}_t$ is obtained. Finally, the difference with the action space $\mathcal{A}$ gives the allowed subsets $\mathcal{A}^\varphi_t$ for time step $t$. Figure \ref{fig:restriction-derivation} is also an example of an interval restriction that arises naturally in the context of environment assumption (1). 
Note that we neglect the exact movement of the obstacles and calculate restrictions only based on the location at the time steps. By this means, motions can be interpreted as rather jumping with $s^{\text{obstacle}}$ but the increasing complexity can be avoided. The advantage outweighs as the research goal focuses on restrictions instead of real-world simulations. When an obstacle moves into the agent between time steps, we use the difference instead of the intersection to determine the allowed angles but follow the same procedure. 

The motivation for the presented concept stems from the fact that fine-grained restrictions are also not a concern in reality. The literature research has demonstrated that sensors with short spacing have already been used, for example, to integrate external knowledge into a reward function. Therefore, we assume the data can also be used to safely cut obstacles out of the action space. In summary, the environment fulfills all properties (1) - (5) from the previous subsection.

\subsection{Observation Space}
Similar to the general design of the environment, the observation space also supports the properties from section \ref{sec:properties}. Obstacles are not included to decouple the restrictions from the observations. This way, the constraints are dynamic and the agent must learn to handle them with action space restrictions (3). The full observation space corresponds to
\begin{equation}
    o^{\text{agent}}_t = (x^{\text{agent}}_t, y^{\text{agent}}_t, \delta^{\text{goal}}_t, d^{\text{goal}}_t, p^{\text{agent}}_t, t)
\end{equation}
Next to the current location, the agent receives the time step $t$, distance $d^{\text{goal}}_t$, perspective $p^{\text{agent}}_t$, and angle to the target $\delta^{\text{goal}}_t$. The latter is necessary to determine the action minimizing the angle to the target. The perspective has been shown to improve the convergence although the target angle is partly correlated. The time step $t$ must be contained to keep the Markov property of the environment since the reward is connected to the number of steps. 

\subsection{Reward Function}
The reward function is inspired by the field of mapless navigation. For example, Choi et al. \cite{choi2021reinforcement} and Marchesini et al. \cite{marchesini2020discrete} take a similar approach. The reward for a time step $t$ consists of three parts:
\begin{equation}
    r_t = 
    \begin{cases}
        r_{\text{goal}} & \text{if goal reached} \\
        -r_{\text{collision}} & \text{if collision} \\
        c \cdot (d^{\text{agent}}_t - d^{\text{agent}}_{t-1}) - p \cdot r_{\text{step}} & \text{otherwise}
    \end{cases}
\end{equation}
The first component describes a large, positive reward when the agent reaches the goal. Similarly, a negative penalty is returned for a collision with an obstacle. We assume a collision occurs whenever a restriction is violated even if this is not always the case due to the safety distance. Otherwise, it would not be possible to learn the constraints. Since the presented models are collision-free by design, this part is only relevant for the reward-based baseline. Besides the two components, the agent receives a reward proportional to the improvement to the target minus a penalty for already taken steps. The improvement is calculated as the difference between the last two distances $d^{\text{agent}}_t$ and $d^{\text{agent}}_{t+1}$ and is scaled by a hyperparameter $c$. The scale of the penalty is set by another hyperparameter $p$. The reward for the same improvement gets less the more steps have already been carried out. The intention behind the two parts is that the agent has the incentive to find a way to the target and at the same time come up with the shortest path. 

\section{Evaluation}
\label{sec:evaluation}
Each experiment in this thesis is designed to explore specific properties of the algorithms. All methods are trained for the same maximum number of time steps. The number is chosen empirically to make sure that all policies converge by that time.
We investigate each algorithm with respect to the terminology defined by Chan et al. \cite{chan2019measuring}. The authors introduce three axes of variability: During training - within runs, during training - across runs, and after training. 
We differentiate training runs and evaluation rollouts to analyze these axes. 
A training run involves a number of time steps in which an algorithm updates its policy and explores the environment. Analyzing the performance during training can reveal the difficulty of learning with restrictions. Evaluating across multiple runs allows quantifying the impact of the initialization of optimizers, the environment, and random seeds.
Contrary, an evaluation rollout corresponds to a single episode in which deterministic actions are selected in a possibly new environment. This allows checking the ability to deal with unseen dynamic action space restrictions. Similarly, evaluating across multiple rollouts allows for assessing the stochasticity of the environment and algorithm optimization. Assessing the evaluation in two parts has the advantage to consider that some algorithms might take more time to converge but provide more robustness to new constraints than others. The specific setup is described with the corresponding experiment in the following chapter. 
Our metrics are designed to assess an algorithm's ability to reach the goal. Details are described below. We report each metric with the standard deviation as a measure of dispersity to ascertain all three axes of variability.

\begin{enumerate}
    \item \textbf{Average returns}: This is the average return that the agent receives.  We report the return two times: In between two consecutive training iterations and across all evaluation rollouts. The training return is expected to be noisier due to exploration. However, the reward curve should indicate that the agent is learning. Together with a measure of dispersity, the curve explores both training axes of variability. The evaluation return ascertains after training variability. Note that we only consider finished episodes in the calculation. 
    
    \item \textbf{Environments solved}: This is the fraction of finished episodes in which the goal is reached. It is the proportion of solved episodes between two training iterations or across all evaluation environments. The evaluation environments ascertain how often the agent reaches the target location with new restrictions and if it is generally possible to find paths with different allowed action space subsets. Note that the path might not be the optimal one. The metric assesses all three axes of variability. 
    
    \item \textbf{Average episode length}: This is the number of steps an agent needs to reach the goal and can be interpreted as a measure of path optimality. Fewer steps indicate a shorter path since the episode ends when the goal is reached. If the agent does not succeed, the episode length corresponds to the maximum number of allowed steps. This is why the metric is more useful after some training steps and the agent already learned to reach the goal. The average episode length is best suited to measure after training variability. 
    
    \item \textbf{Fraction of allowed actions} This is the ratio of the allowed action space size compared to unavailable subsets. The metric is useful to explore the performance of algorithms when more or fewer actions are allowed and is a reliability measure. We report the average between two iterations during training runs and plot the fraction over time across all evaluation rollouts. A constant and non-zero average indicates that the restrictions actually occur. Therefore, the obstacles are not surrounded without an effect on the actions. Further, the proportion confirms that adding obstacles really leads to more restrictions on the action space. 
    
    \item \textbf{Allowed interval length}: This metric considers the length of each allowed interval at a time step separately. The metric is reported as the average, minimum, and maximum and is a control variable. We also note the variance within the action space. The length of the single intervals can reveal an effect on the reward and episode length.
    
    \item \textbf{Number of intervals}: This is the absolute number of disjoint intervals in the action space at each time step. The metric serves the same purpose as the length of the intervals and is a control variable. We also report it as the average over entire episodes and evaluation time steps. 
    
    \item \textbf{Time steps required to reach a certain performance} This is the number of time steps an algorithm requires to reach a certain performance level. If more than 80\% of the episodes between two training iterations are solved, then the time step is recorded. The metric assesses the during training - across runs variability and the sample efficiency. 
\end{enumerate}

Additionally, we follow the suggestions of Colas et al. \cite{colas2019hitchhiker} and compare the average returns, solved environments, and episode length in the evaluation with a Welch's t-test \cite{welch1947generalization}. We consider results below the 0.05 level as significant. The test has shown the lowest false positive error and does not assume equal variances. This is particularly helpful since the sensitivity of algorithms to factors such as random seeds differs \cite{henderson2018deep}. All models are trained with rectified linear units using the Adam optimizer which was described in section \ref{sec:optimization}. We use $\beta_1 = 0.9$, $\beta_2=0.999$ and set $\gamma = 0.99$. Videos\footnote{https://www.youtube.com/@dynamicintervalrestrictions/playlists} of all evaluation rollouts are made available online to improve traceability. 

To improve the readability of this thesis, we round all values to two decimal points. In the tables and figures, we denote discrete masking with the extension \emph{-masked} and continuous masking as \emph{-masking}. 

\section{Hyperparameter Optimization}
\label{sec:hpo}
Henderson et al. \cite{henderson2018deep} show evidence that small changes in hyperparameters can significantly affect the final performance. Hence, to compare our models with the optimal configuration, we use Hyperopt \cite{bergstra2015hyperopt} with asynchronous Hyperband \cite{li2018massively} for hyperparameter optimization. 
Hyperopt uses Bayesian optimization to suggest promising configurations. Asynchronous Hyperband is based on successive halving. The idea is to randomly sample a number of configurations from a predefined search space and group them in a rung. The rung has a predefined budget $b$ which is uniformly distributed across samples. After training each algorithm for the allocated number of iterations, the top $\frac{1}{n}$ candidates are promoted to the next rung. This next rung considers fewer configurations and assigns an increased budget by a factor of $n$. The process is repeated until the maximum number of training iterations is reached. Asynchronous hyperband successively adds new algorithm samples and checks if configurations can be promoted when a setup reaches the end of learning. We set $n=2$ and run each  approach for at least 20 iterations. The reason is that exploration can mask the true performance at the beginning. The procedure has the advantage that runs which are stuck with low rewards are stopped at an early stage. The maximum number of time steps per configuration is set to 150.000. All training runs are executed over four different seeds and the results are averaged. This avoids overfitting to the randomness due to the environment and algorithm. The search space for each algorithm can be found in appendix \ref{sec:hp-spaces} and the final configurations are described in appendix \ref{sec:hp-configuration}. We optimize the training reward since the experiments in the next chapter aim on finding well-trained algorithms to explore their behavior with new restrictions. Environment variations are part of the training setup. 

\section{Implementation}
The algorithms are implemented with PyTorch \cite{pytorch} and Ray Rllib \cite{liang2018rllib}. We make use of the predefined PPO, TD3, and DQN models. Similarly, the hyperparameters are tuned with the implementations of Hyperopt and asynchronous Hyperband in Ray Tune \cite{liaw2018tune}. The environment is based on OpenAI Gym \cite{gym}. Moreover, we use Shapely \cite{shapely2007} for geometry computations and pygame\footnote{https://www.pygame.org} for visualization. Results are fully documented, reproducible, and available on the author's GitHub repository\footnote{https://www.github.com/timg339/Dynamic-Interval-Restrictions}. 

\chapter{Experiments and Analysis}
In this chapter, we conduct different experiments which use path optimality to explore interval restrictions on action spaces. In terms of answers to the questions of this thesis, we differentiate when agents are trained in different forms: The first experiment learns without obstacles but measures the performance when constraints occur in the evaluation environments. Afterward, we compare the results to training algorithms with restrictions. We place dynamic obstacles that shift between waypoints and are set at different locations before each episode. Details of the experiments are described in the following subsections.

\section[Learning Without Obstacles]{Learning Without Obstacles}
\label{sec:without}
In the optimal case, agents should still find the action maximizing the expected return in different subsets of the action space although no dynamic restrictions emerged during training. However, all presented methods, except for discrete masking, require a constraint representation in the observations to learn the scaling function and optimal action for restrictions. Therefore, we assume that the different subsets must already emerge during training to maximize the reward. 

To test each algorithm's performance, we conduct experiments where the entire action space is constantly available on the optimal path during training, but random restrictions are applied in the evaluation. The experiments are conducted in three parts: First, we train each algorithm to reach the target and measure the performance in an environment without obstacles. Then, we introduce simple restrictions during the evaluation before moving on to more complex setups. This is done because we could not find any prior research on unseen dynamic action space restrictions, so it is unclear whether agents can find alternative paths in different action spaces, even in low-complexity environments.

\subsection{Training}
\label{sec:training}
In order to analyze changes resulting from restrictions, it is crucial to ensure that algorithms demonstrate equivalent performance in an unrestricted domain. When comparing methods with different behavior, it can be difficult to discern to what extent subsequent experimental results are due to the imposed restrictions. Moreover, it is important to investigate whether any modifications hinder learning. For instance, replacing invalid actions with a random sample may lead to slow convergence. Therefore, this section will present the training outcomes by first providing a detailed description of the environment and learning procedure, and then discussing and comparing the final performance of each algorithm. Working towards the goal of this thesis, the experiment is supposed to give insight into which algorithms can maximize the reward in environments with static restrictions and confirms the conditions for forthcoming experiments. This is similar to how constraints commonly appear in the literature.

\subsubsection{Setup}
All algorithms are trained in the same environment outlined in section \ref{sec:env}. The specific design choices are summarized in table \ref{tab:env-setup}.
As mentioned above, training is conducted without obstacles. The motivation for the environmental parameters is that the goal can be reached within a small number of steps without making it a trivial task.  This is done to increase the difficulty of restriction handling by offering several alternative paths. We assume that significant differences between algorithms would be more likely to be found in such a scenario.

In this context, the size of the map is closely related to the step size of the agent. The minimum number of steps on a large map with a simultaneously large step size is equivalent to a small environment with few steps. Based on a step size of 1.0, we set the width and height to $15$. This makes a minimum number of 15 steps necessary to reach the target location. Accordingly, we set the maximum length of an episode to $40$ which accounts for longer paths and exploration. 
On one hand, if the maximum number of steps is too small, agents might not be able to reach the goal, and investigating the episode length loses effect. On the other hand, if it is too large, the probability rises that algorithms randomly drop into the goal.
A relatively larger step size has the advantage that restrictions increase since more obstacles fall into the radius. 

The agent is placed in the bottom left of the environment, and the goal is located in the top right corner. The range of actions is set to $[-110,110]$ to not allow sharp turns in the opposite direction but still keep enough flexibility so that not the entire space can frequently get unavailable. 
The reward function has been determined empirically. Each algorithm is trained for 50.000 steps. Note that static restrictions occur in this setup at the border of the environment to keep the agent on the map. Therefore, the training is not completely restriction-free and comparable to the usual approach in the literature \cite{dalal2018safe}. It must also be taken into account that, because of an error during the training process, PPO with discrete was trained with parameters optimized for continuous masking. It was impossible to repeat the optimization due to the time constraints of this thesis. 
We set the zero-padding for the baseline algorithms to support up to eight disjoint intervals. Due to the simplicity of our task, we do not expect the count to rise higher. 

To test the fitness of our models, we evaluate each without exploration after the last training iteration. The same environment is used for the rollouts. We run the algorithms on six different seeds, including 40, 41, 42, 43, 44, and 45.
\begin{table}[t]
\begin{center}
\begin{tabular}{lc}
\hline
\multicolumn{1}{c}{\textbf{Parameter}}                       & \textbf{Value}       \\ \hline
\textbf{Agent}                                               & \multicolumn{1}{l}{} \\
step size $s^{\text{agent}}$                                 & $1.0$                \\
radius $r^{\text{agent}}$                                    & $0.4$                \\
action space range $[a_\text{min},a_\text{max}]$             & $[-110,110]$         \\
starting position $(x^{\text{agent}}_t, y^{\text{agent}}_t)$ & $(1.0,1.0)$          \\
starting perspective $p^{\text{agent}}_0$                    & $90.0$               \\
safety distance $d^{\text{safe}}$                            & $0.05$               \\ \hline
\textbf{Environment}                                         & \multicolumn{1}{l}{} \\
height $h$                                                   & $15.0$               \\
width $w$                                                    & $15.0$               \\
maximum steps $T$                                            & $40$                 \\
goal position $(x^{\text{goal}},y^{\text{goal}})$            & $12.0$               \\
goal distance threshold $r^{\text{goal}}$                    & $0.5$                \\ \hline
\textbf{Reward}                                              & \multicolumn{1}{l}{} \\
on goal $r_{\text{goal}}$                                    & $50.0$               \\
on collision $r_{\text{collision}}$                          & $20.0$               \\
improvement scale $c$                                        & $5.0$                \\
step penalty scale $p$                                       & $0.05$               \\ \hline
\end{tabular}
\caption{Environmental parameters for the experiments}
\label{tab:env-setup}
\end{center}
\end{table}

\subsubsection{Results}
In line with the axes of variability, the following first depicts outcomes within and across training runs before we continue with the evaluation rollouts.

\begin{figure}[t]
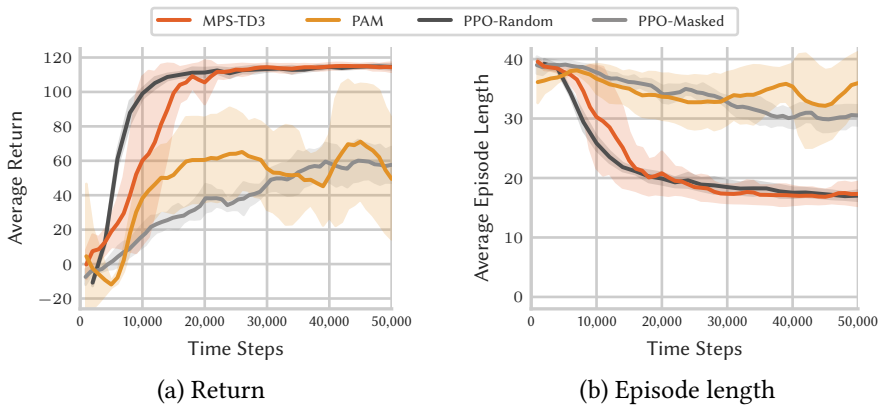

\centering
\begin{subfigure}{0.8\linewidth}
  \centering
  \input{experiment_1/legend.pgf}
\end{subfigure}
\begin{subfigure}{.49\linewidth}
  \centering
  \input{experiment_1/text/return.pgf}
  \caption{Return}
  \label{fig:exp1_training_return}
\end{subfigure}%
\begin{subfigure}{.49\linewidth}
  \centering
  \input{experiment_1/text/steps.pgf}
  \caption{Episode length}
  \label{fig:exp1_training_steps}
\end{subfigure}
\caption[Training progress without dynamic restrictions]{Training progress without dynamic restrictions}
\label{fig:exp1_training_results}
\end{figure}

\paragraph{Within and across training runs}
The results indicate a smooth learning curve for MPS-TD3 and most baseline algorithms that converges close to the maximum reward. Due to a large number of methods, figure \ref{fig:exp1_training_return} shows exemplarily the results of MPS-TD3 and PPO with random replacement. They represent the optimal behavior by having a steady return increase without a drop in performance. The episodes between the last two training iterations achieve average cumulative rewards of 114.34 and 114.54 respectively. 

However, PAM and PPO with discrete masking fail to maximize the expected return and converge to a sub-optimal policy. Both learning curves are also depicted in figure \ref{fig:exp1_training_return}.
The progress of PAM's average return continuously has a high standard deviation with up- and downward trends. At the end of the training, the approach reaches a relatively low mean cumulative reward of $49.46$. Another fact that caught our attention is that the action values explode. They increase simultaneously to the reward and come close to the maximum achievable return after about 20.000 time steps. Then, the model becomes increasingly unstable and does not improve further. 
In terms of PPO with discrete masking, the average return settles around 60. Additional time steps did not increase the performance. The final plateau is reached at approximately $40.000$ time steps although the learning curve in figure \ref{fig:exp1_training_results} does not show the progress with more training iterations. 

Moreover, our analysis reveals that MPS-TD3 and the off-policy baselines have a higher standard deviation than on-policy learning. Specifically, TD3 with projection demonstrates this trend in appendix \ref{app:exp1_0_td3} most clearly but the dispersity can also be observed from MPS-TD3 in figure \ref{fig:exp1_training_results}. Across separate runs, the metrics show more variation in the first 40,000 time steps, but the differences decrease as the agent nears the end of training. The higher standard deviation aligns with other results such that off-policy algorithms take longer to converge. For instance, random replacement requires a median time of five training iterations to solve 80\% of the episodes when combined with PPO, while it takes 11.50 with MPS-TD3.

Furthermore, we discover that the average episode length follows a similar pattern to the reward. This means that with more training iterations, the agents find shorter ways to the target. All methods start with the maximum number of steps and gradually reduce them to approach the optimal value. The exceptions are penalty-based agents since they are not collision free from the beginning. Additionally, discrete PPO and PAM terminate with episode lengths above 30. The described behavior is depicted in Figure \ref{fig:exp1_training_steps}. The lines show for PPO with discrete masking and random replacement, MPS-TD3,  and PAM the step counts with their standard deviation over time. The episode lengths of the other algorithms are illustrated in appendix \ref{app:exp1_0_episode_length_ppo} and \ref{app:exp1_0_episode_length_td3}. Also regarding the number of steps, off-policy has a higher standard deviation than on-policy learning. 

Likewise, the average restrictions get lower as training continues. The trajectories show that the agents become more proficient at avoiding the borders of the environment and taking a direct path toward the goal. As a result, they have access to a wider range of actions. For example, the TD3 algorithm with continuous masking starts in appendix \ref{app:exp1_0_allowed_action_space_td3} with an average allowed action space of less than 150 but the metric increases until almost the entire action space is available. 

Similarly, the number of intervals decreases in appendixes \ref{app:exp1_0_count_ppo}, \ref{app:exp1_0_count_td3}, and \ref{app:exp1_0_count_own} over time. Most agents deal with a single subset by the end of training. Only PAM, MPS-TD3, and discrete PPO maintain a higher average number of intervals. However, MPS-TD3 simultaneously increases the size of the available action space. A larger fraction is unrestricted but split over more intervals. Again our algorithms and the off-policy baselines have a higher standard deviation than the on-policy algorithms. Note that all results and metrics can be found in appendix \ref{app:exp1_0}.

\paragraph{After training}
In the evaluation rollouts, MPS-TD3 and the baselines achieve a high average return which is similar to the end of training. Table \ref{tab:exp1_training_ev_results} describes the mean cumulative reward, episode length, and fraction of solved episodes without exploration.
\begin{table}[t]
\centering
\begin{tabular}{llll}
\hline
\multicolumn{1}{c}{\textbf{Approach}} & \multicolumn{1}{c}{\textbf{Return}} & \multicolumn{1}{c}{\textbf{Steps}} & \multicolumn{1}{c}{\textbf{Solved}} \\ \hline
TD3                                   & $114.57 \pm 3.67$                     & $17.33 \pm 2.88$                     & $100.00\% \pm 0.00\%$                   \\
TD3-Projection                        & $91.46 \pm 55.30$                     & $17.60 \pm 4.22$                     & $83.33\% \pm 40.82\%$                  \\
TD3-Masking                           & $114.27 \pm6.21$                      & $18.83 \pm 5.53$                     & $100.00\% \pm 0.00\%$                   \\
TD3-Random                            & $114.13 \pm 4.54$                     & $18.17 \pm 3.87$                     & $100.00\% \pm 0.00\%$                   \\
DQN-Masked                            & $113.80 \pm 4.83$                     & $18.83 \pm 4.22$                     & $100.00\% \pm 0.00\%$                   \\ \hline

PPO                                   & $115.16 \pm 0.13$                     & $15.00 \pm 0.00$                     & $100.00\% \pm 0.00\%$                   \\
PPO-Projection                        & $115.12 \pm 1.03$                     & $15.33 \pm 0.52$                     & $100.00\% \pm 0.00\%$                  \\
PPO-Masking                           & $114.81 \pm 0.38$                     & $15.00 \pm 0.00$                     & $100.00\% \pm 0.00\%$                   \\
PPO-Random                            & $114.87 \pm 0.50$                     & $15.00 \pm 0.00$                     & $100.00\% \pm 0.00\%$                   \\
PPO-Masked                            & $73.06 \pm 64.29$                     & $18.00 \pm 1.83$                    & $66.67\% \pm 51.64\%$                 \\ \hline

MPS-TD3                               & $115.53 \pm 1.70$                     & $16.83 \pm 1.47$                     & $100.00\% \pm 0.00\%$                   \\
PAM                                   & $11.88 \pm 41.37$                     & $38.00 \pm 0.00$                     & $16.67\% \pm 40.8\%$                 \\ \hline
\end{tabular}
\caption[Average return, steps, and solved episodes - learned without obstacles]{Average return, steps, and solved episodes without obstacles}
\label{tab:exp1_training_ev_results}
\end{table}
That the methods are comparable in the final return is supported by the Welch's t-test in which no mean, except for PAM, is significantly different from another ($p \geq 0.05$). However, we found that discrete PPO reaches only 73.06, and TD3 with projection also made a lower average return of $91.46$ contrary to the training convergence. We discovered that the reason for the latter is an episode in which the agent misses the goal and follows a sub-optimal policy. This can be seen in the corresponding videos. 
\newpage
In line with training results, PAM finishes with the lowest mean cumulative reward of $11.88$. It is the only difference that is significant at the 0.01 level to all algorithms with the exception of PPO with discrete masking (p = 0.08). The trajectories indicate that the actions have a random pattern. PAM has an average step count of $38.00$ and ends closer to the starting point than other algorithms.

Additionally, table \ref{tab:exp1_training_ev_results} shows that MPS-TD3 has a significantly different average episode length compared to the on-policy baselines (p $\leq$ 0.05). However, the effect size can be considered small since the mean is 16.4 and the smallest among all TD3 and DQN variants. Generally, off-policy algorithms take slightly more steps than PPO. However, the differences between all other algorithms are not statistically significant at the 0.05 level.

Except for PAM, TD3 with projection and the discrete PPO, MPS-TD3, and the baselines solve all of the evaluation episodes. The single run which misses the goal in projection with TD3 has been described earlier and leads to a decrease in the fraction of solved environments to $83.33\%$. The same holds for two runs of PPO with discrete action masking. The target location is slightly missed and the border of the environment is followed away from the goal. However, contrary to the random behavior of PAM, both still pass the midpoint of the map when moving toward the target. The specific paths can be again followed in the videos. 

\begin{figure}[t]
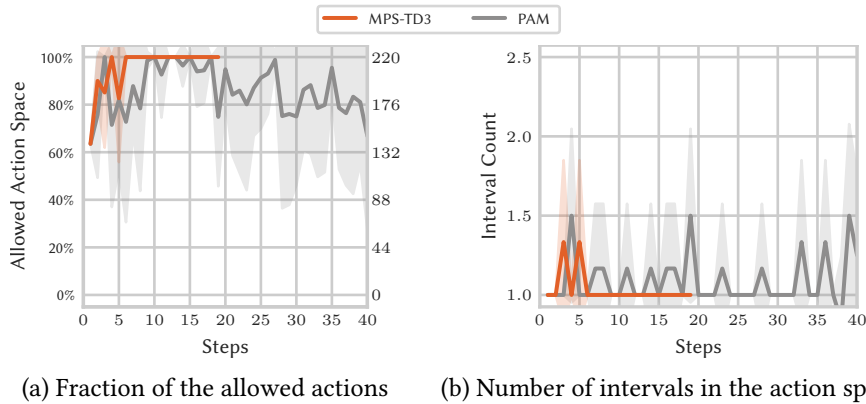

\centering
\begin{subfigure}{0.8\linewidth}
  \centering
  \input{experiment_1/text/legend_pam_mps.pgf}
\end{subfigure}
\begin{subfigure}{.49\linewidth}
  \centering
  \input{experiment_1/text/allowed.pgf}
  \caption{Fraction of the allowed actions}
  \label{fig:exp1_training_ev_allowed}
\end{subfigure}%
\begin{subfigure}{.49\linewidth}
  \centering
  \input{experiment_1/text/count.pgf}
  \caption{Number of intervals in the action space}
  \label{fig:exp1_training_ev_count}
\end{subfigure}
\caption[MPS-TD3 and PAM action space behavior]{MPS-TD3 and PAM action space behavior}
\label{fig:exp1_training_ev}
\end{figure}

In addition to the above, the restrictions throughout the episodes are at a minimum. This is shown by the average fraction of the allowed actions and the number of intervals in the action space. The entire set is consistently available for most algorithms. Next to PAM, appendixes \ref{app:exp1_0_ppo}, \ref{app:exp1_0_td3}, and \ref{app:exp1_0_own} show only PPO with discrete masking and TD3 with random replacement and projection having fluctuations after 20 steps. Figure \ref{fig:exp1_training_ev_allowed} illustrates the average fraction of the action space and \ref{fig:exp1_training_ev_count} the number of disjoint intervals at each time step. The lines depict MPS-TD3 and PAM to contrast optimal and sub-optimal behavior. The larger restrictions of PAM are due to when the episode is not solved and the agent runs into the border of the environment again. Contrary, MPS-TD3 represents the behavior when an optimal path is followed. Only a single rollout comes close to the border at the beginning of the episode and causes fluctuations in the graph. All metrics, visualizations, and significance values are contained for each algorithm in appendix \ref{app:exp1_1}. 

\subsubsection{Discussion}
The results of the experiment have several implications for the research directions in this thesis. First, the study found that PPO, TD3, and DQN can navigate a mapless environment, which is consistent with previous research \cite{wang2018learning, marchesini2020discrete}. 
Due to the task's simplicity, we do not need additional information such as artificial potential fields. MPS-TD3 and the baseline modifications do not seem to hinder learning. Specifically, the outcomes of MPS-TD3 show that utilizing the critic to decide between multiple intervals seems to work. We achieved comparable results to the continuous masking baseline that scales the action into concatenated intervals. However, the effect might be due to that the rollouts are almost constantly based on a single interval. In this case, MPS-TD3 and the continuous masking baseline work similarly. In terms of convergence speed, the unmodified algorithms reached the final performance as fast as the methods strictly respecting restrictions. A similar result was found by Krasowski et al. \cite{saferloverview} in an inverted pendulum task with a safety controller and Dalal et al. \cite{dalal2018safe} who train DDPG to reach a goal while staying on a similar map to this experiment with euclidean projection. 

Nonetheless, on- and off-policy learning have small discrepancies. Changes in results with restrictions may be due to the algorithms' unconstrained performances. Nevertheless, we believe differences are caused by hyperparameters and less task suitability. TD3 has turned out very sensitive to small changes as has been already reported by Henderson et al. \cite{henderson2018deep}. No model stands out as clearly superior in unrestricted training performance. Future studies can aim for more suitable hyperparameters while testing harder tasks. Differences might be visible then. 

Third, the study found that more research is needed to stabilize the PAM algorithm. Contrary to previous applications in different domains \cite{dorokhova2021deep, bouktif2021traffic}, the approach has poor performance when used to divide the action space into bins. It may be useful to control the explosion of action values to improve its performance. A starting point could be using more sophisticated parameterized models such as the architectures by Bester et al. \cite{bester2019multi} or Hu et al. \cite{hu2021hierarchical}, inverted gradients \cite{hausknecht2015deep}, and learning rates that satisfy the Robbins-Monro condition \cite{robbins1951stochastic} as suggested by Xiong et al. \cite{xiong2018parametrized}. The comparison in subsequent experiments is not fair, as PAM performs inadequately in the unrestricted domain. The algorithm hence already fails in view of the first research question. 
The same holds for PPO with discrete masking which is probably due to the configuration.

Nevertheless, the similar performance of MPS-TD3 and most methods allows us to measure the effect of restrictions in the majority of cases. 

\subsection{Simple Restrictions}
\label{sec:simple_restrictions_without}
Following the aim of this thesis, we study next how the presented algorithms perform with simple and dynamic restrictions on their action space. This is done by incorporating random obstacles into the environment. 

\subsubsection{Setup}
\label{sec:exp1_simple_setup}
The properties of the environment are the same as in the previous section but with one obstacle located at the center of the map. This obstacle produces a single restriction interval at a time. When the midpoint is reached, the agent receives the same observation as during training but a different subset of the action space is available. The possible actions for the same state change between rollouts, i.e. $s_t = s_{t'} \nRightarrow \mathcal{C}_t = \mathcal{C}_{t'}$. Hence, the restrictions can be considered dynamic according to section \ref{sec:interval_restrictions}. We select the center as the position for obstacles since the previous subsection found that most algorithms follow a straight path to the goal.

The prerequisite that agents must pass the midpoint is also examined visually and runs with incorrect behavior are discarded. This way ensures a fair comparison among the algorithms.
Accordingly, projection with TD3 on seed two, and seeds two and three of PAM are removed. TD3 with random projection moves to the top left corner before taking the straight way. This reduces the effect of midpoint restrictions. On one hand, this procedure reduces the power of the experiments. On the other hand, only in this way, validity is ensured.

Another question that arises is if models not reaching the goal without obstacles should be included. Although they narrowly miss the target, we believe they can still provide valuable insight into reactions to restrictions. Specifically, we are interested in whether these models would still make it close to the goal or get stuck when an obstacle is introduced. This information can help us better understand what adjustments may be needed to improve their performance. As a consequence, we decided to use them in the following experiment.

Additionally, the size of restrictions can have a significant impact. Larger obstacles increase the distance between the new optimal path and the learned behavior. We sample the obstacle dimensions from a uniform distribution of $U(0.2, 6.8)$ to avoid bias towards specific scales. Note that we still consider different forms as well. Our results must generalize over a variety of restrictions. 

In total, we experiment with 40 evaluation environments created with seeds from 0 to 39. Except for projection with TD3 and PAM, every setup is tested with all six training runs, resulting in 240 evaluations for each method. 

\subsubsection{Results}
\label{sec:without-simple-results}
\begin{figure}[t]
\centering
  \centering
  \input{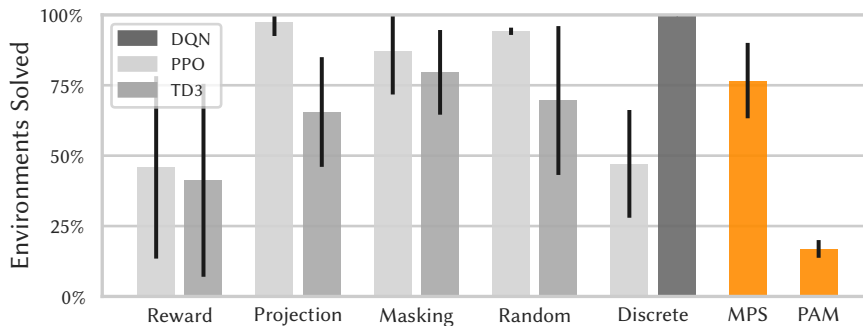}
\caption[Portions of solved episodes with simple restrictions]{Portions of solved episodes with simple restrictions}
\label{fig:exp1_simple_solved}
\end{figure}

The results show that DQN with discrete masking achieves the highest performance in terms of both, reward and fraction of reached goals. It is the only method that successfully solves all of the evaluation environments. This can be read from Figure \ref{fig:exp1_simple_solved} comparing the percentages of all algorithms. The success of DQN with discrete masking is statistically confirmed. The difference to every other algorithm, except for PPO with projection and masking, is statistically significant at the 0.05 level. In particular the first comes close to DQN, achieving a 97.50\% success rate.

In addition, trajectories exhibit DQN following a smooth path in most cases. The average episode length is 20.52, which is relatively low compared to other off-policy baselines. For instance, MPS-TD3 requires 24.72 steps on average. However, the variability is high and the differences are not statistical (p $\geq$ 0.05). The only exceptions are TD3 with random replacement and PAM (p = 0.02). The on-policy baselines also take a similar number of steps to DQN (p $\geq$ 0.05). For instance, PPO with projection has an average episode length of 20.54 (p = 0.99). 
\newpage
Within the same training paradigm, MPS-TD3, projection, and continuous masking tend to perform similarly. By training paradigm, we mean that a learning type is applied with different modifications. We find no significant differences between the reward, solved fraction, and steps, of the same algorithm, such as MPS-TD3 compared to TD3 with continuous masking or projection (p $\geq$ 0.05). Only PPO with random replacement has less steps than masking (p $\leq$ 0.01) and fewer returns compared to projection (p = 0.03). 

Overall, on-policy learning performs better than MPS-TD3 and the off-policy baselines in environments with continuous action spaces, while the opposite is true with discrete actions. For example, PPO with projection achieves an average reward of 109.43, compared to TD3's 78.09 (p = 0.02). Although the differences are not always significant, the mean is consistently lower, pointing to a trend. Similarly, MPS-TD3 has a lower average return than PPO with projection (p $\leq$ 0.01), masking (p = 0.16), and random replacement (p $\leq$ 0.01). 

The PAM algorithm achieves the lowest level of performance in all fields. The mean return is 16.76 and the agent reaches the goal in 16.88\% of the cases (p $\leq$ 0.01 except for penalty-based). On average 30.34 steps are required. The trajectories indicate that even when the agent passes an obstacle, it misses the goal. 

\begin{figure}[t]
\centering
\begin{subfigure}{.3\linewidth}
  \centering
  \fbox{\includegraphics[width=3.5cm]{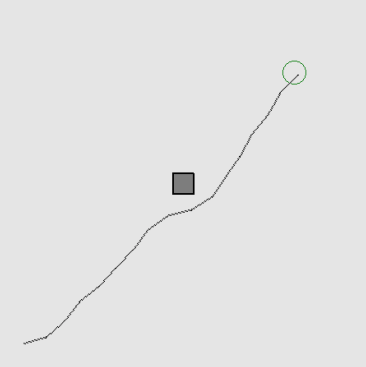}}
  \caption{Smaller obstacle}
  \label{fig:small}
\end{subfigure}%
\begin{subfigure}{.3\linewidth}
  \centering
  \fbox{\includegraphics[width=3.5cm]{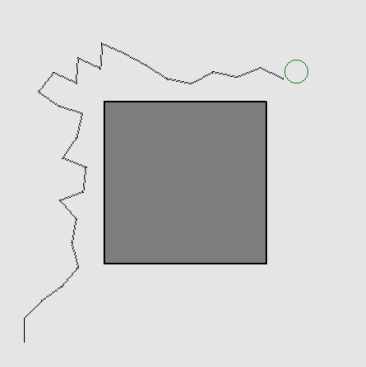}}
  \caption{Larger obstacle}
  \label{fig:large}
\end{subfigure}
\begin{subfigure}{.3\linewidth}
  \centering
  \fbox{\includegraphics[width=3.5cm]{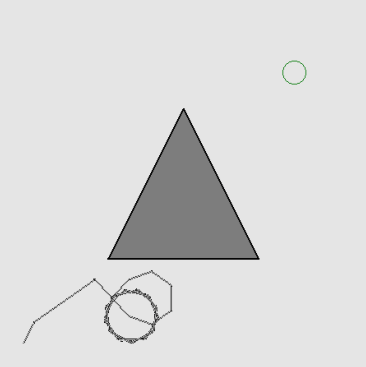}}
  \caption{Circular behavior}
  \label{fig:circular}
\end{subfigure}
\label{fig:exp1_simple_trajectories}
\caption[PPO trajectory examples with a single obstacle]{PPO trajectory examples with a single obstacle}
\end{figure}

Further investigating trajectories suggests for our algorithms and the baselines that the larger the restriction, the less the reward, and the path gets less smooth. Examples are shown in figures \ref{fig:small} and \ref{fig:large}. More trajectories are available from appendixes \ref{app:exp1_2_trajectory_start} to \ref{app:exp1_2_trajectory_end}. When PPO diverges minimally from its policy, the path continues smoothly and straight. When the alternative route gets further away, random oscillations occur and it is less likely that the agent makes it to the goal. 

Another finding is circular behavior. The pattern is best visualized by PPO in figure \ref{fig:circular} and occurs for MPS-TD3, PAM, and all baselines except for DQN. The situation arises when the agent selects the same invalid actions which bring back similar states. From these states, the same choices are made. This leads to repeated and inefficient behavior. However, the situation can usually be escaped after a few rounds. We found it happening less frequently with random replacement compared to methods that project an action always to the same value. 
The whole results and visualizations for the algorithms can be found in appendix \ref{app:exp1_2}. 

\subsubsection{Action Space}
\begin{figure}[t]
\centering
  \centering
  \input{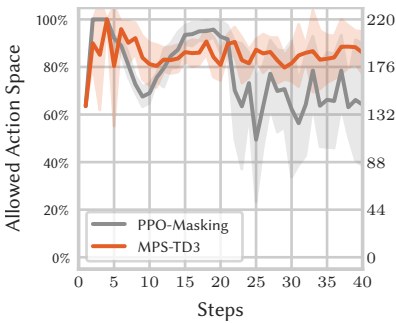}
\caption{Fraction of the allowed actions for MPS-TD3 and PPO with masking}
\label{fig:exp1_simple_allowed}
\end{figure}

The mean size of the available action space is between 180.38 and 194.47. For instance, MPS-TD3 has an average of 187.90, indicating that around 85.41\% of the action space is unrestricted. In the beginning, more actions are allowed but the portion declines between 5 and 10 steps when an obstacle occurs. Then, the restrictions shrink until most algorithms reach the goal at around 20 steps. The fluctuations afterward are caused by rollouts that do not find a straight path to the target location but run into the border of the environment first. Figure \ref{fig:exp1_simple_allowed} illustrates exemplary the size of allowed actions with respect to the time steps for MPS-TD3, DQN with discrete, and PPO with continuous masking. The course of the other approaches in appendixes \ref{app:exp1_2_length_ppo}, \ref{app:exp1_2_length_td3}, and \ref{app:exp1_2_length_own} is similar. 

The same pattern holds for the single intervals which are the smallest at around 10 steps. The lowest average of simultaneously available subsets is 1.06 in the case of DQN. However, this number can increase up to 1.14 for TD3, PPO, and MPS-TD3. The mean standard deviation of the subset sizes across all algorithms is 47.01. Appendixes \ref{app:exp1_2_count_ppo}, \ref{app:exp1_2_count_td3}, and \ref{app:exp1_2_count_own} indicate that the allowed space usually consists of a single interval. The average length of the subsets comes close to the mean of the total available action space. For example, the different parts have an average size of 177.55 in the case of PPO with continuous masking. The mean minimum length across all intervals in an action space is 174.59 and the maximum is 180.50. The control variables for all algorithms are contained in appendix \ref{app:exp1_2}. 

\subsubsection{Discussion}
Aligning with our research aim, we proposed a first way to evaluate algorithms with dynamic interval action space restrictions. 
From the perspective of this experiment, we see partial evidence for our hypothesis that constraint-learning methods require different action space subsets during training. On one hand, MPS-TD3 and the off-policy baselines performed worse with respect to the previous subsection. Larger fluctuations occurred and randomly replacing actions works similarly well. We hypothesize deviations from the policy can generally not be handled well. On the other hand, on-policy differences are relatively small. Note that most authors test algorithms with restrictions when the constraints occur during training. Therefore, we added to, for example, Krasowski et al. \cite{saferloverview} and Huang et al. \cite{huang2020closer}, how the approaches perform with unseen restrictions.  

In our case, DQN was the best method for previously unknown restrictions, MPS-TD3 was similar to the continuous off-policy baselines, and restrictions learned with penalties failed in the majority of cases. The latter is similar to when agents are trained with restrictions since the performance drop was also found by Brosowski et al. \cite{brosowsky2021sample} and Kochdumper et al. \cite{kochdumper2022provably}. Likewise to the previous subsection, off-policy continuous masking achieved the same performance as MPS-TD3. Therefore, considering the intervals separately does not seem to have an advantage when the action values for subsets are not properly learned. Nevertheless, our method adds an alternative to zero-padding. It must be investigated in subsequent experiments if the methods are also comparable when more disjoint intervals occur. The success of DQN is contrary to the discretization error in the literature \cite{dulac2015deep} and promotes the use of advanced discretization techniques \cite{tang2020discretizing,sinclair2019adaptive}. However, similarly to Krasowski et al. \cite{saferloverview}, the reason could be that the task is too simple to exaggerate differences between algorithms and it is not necessary to choose from a fine-grained action space to bypass an obstacle efficiently.  

As previously, the on-policy tend to perform better than the continuous off-policy approaches and come close to DQN. We believe the difference is due to the performance in the unrestricted environment although the discrepancies are considered relatively large. Likewise to the previous subsection, the effect could come from hyperparameters and less task suitability since TD3 was built for high dimensional action spaces in complex environments \cite{lillicrap2015continuous}. The lower performance of PPO on the discretized action space is probably caused by the learned sub-optimal policy during training. It is also important to mention that not all algorithms follow a completely straight path and have an advantage with small obstacles. Nevertheless, we believe the bias is rather small and negligible. 

As expected, PAM fails to find appropriate paths. The performance was anticipated since only about 16\% of the environments were initially solved in the unrestricted domain. Following the suggestions from the previous section, it is unclear to what extent the performance changed due to the restrictions. However, future research should aim at stabilizing PAM. We still believe the procedure is promising because of the connection to DQN which reached remarkable results. 

Overall, the experiment seems to have worked effectively even though not a single method turned out as superior. The control variables indicated similar restrictions on the action space on the straight path to the goal between all methods. 

\subsection{Complex Restrictions}
\label{sec:complex_res_exp_1}
Since we did not find clear answers to the research questions in a simple environment, a second experiment is conducted with more complex restrictions. This is intended to make differences more clearly recognizable. 

\subsubsection{Setup}
The complex scenario is again the same environment as in previous experiments but with 14 obstacles randomly generated using algorithm \ref{alg:obstacle-generation}. Examples can be found in appendixes \ref{app:exp1_3_trajectory_start} to \ref{app:exp1_3_trajectory_end}. We assume that differences in steps are more significant when agents have to bypass multiple obstacles with narrow paths in between. Since more barricades are spread on the map, we expect restrictions to occur constantly and to have more disjoint intervals on average. Locations are sampled from a normal distribution with mean $\mu = (7.5,7.5)$ and covariance matrix $\Sigma = \big(\begin{smallmatrix}
  4.0 & 0.0\\
  0.0 & 4.0
\end{smallmatrix}\big)$ according to algorithm \ref{alg:obstacle-generation}. The size of the obstacles is yielded from $\mathcal{N}(1.0, 0.25)$ and clipped to fit into the range $[0.5,1.5]$. The covariance matrix is set to bias obstacles to occur on the line between the agent starting position and the goal more frequently. As in the previous experiment, we use the seeds from 0 to 39 to generate environments which make 240 evaluation rollouts per algorithm. 

\subsubsection{Results}
\begin{figure}[t]
\centering
  \centering
  \input{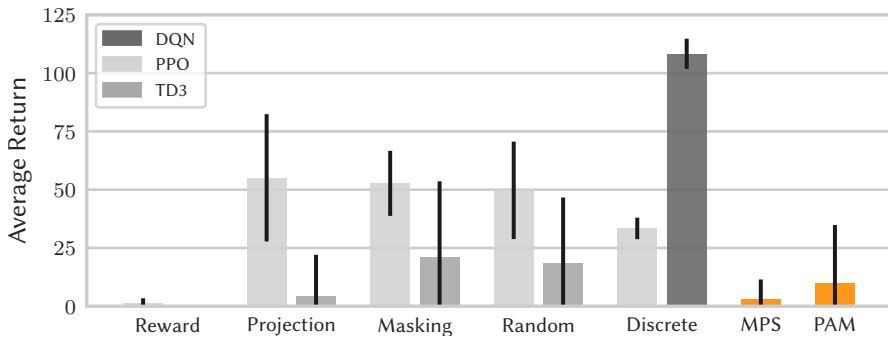}
\caption{Average return with complex restrictions}
\label{fig:exp1_complex_return}
\end{figure}

DQN achieves the highest rewards and fraction of solved environments. The mean return is 108.31 with 97.50\% solved environments and on average 21.51 steps are taken to the goal. The difference in the return and solved episodes is significant with respect to MPS-TD3, PAM, and the other algorithms at the 0.01 level. Figure \ref{fig:exp1_complex_return} illustrates the mean cumulative reward and standard deviation for each approach.
As a counterexample, PPO with projection has an average return of 55.10 with 48.75\% solved episodes and a mean of 28.60 steps. At the same time, PPO with projection reaches the highest rewards among all on-policy baselines. 

This example shows that also in terms of path optimality DQN outperforms most algorithms (p $\leq$ 0.01). Only discrete (p = 0.21) and continuous masking (p = 0.07) come closer to DQN with average episode lengths of 24.22 and 25.81 respectively. Appendix \ref{app:exp1_3_plot_steps} illustrates that all methods, with DQN excluded, solve less than half of the episodes and need relatively long paths to reach the goal. Exemplary routes are illustrated for traceability in appendixes \ref{app:exp1_3_trajectory_start} to \ref{app:exp1_3_trajectory_end}. Most of the DQN trajectories are straight without temporary runaways.

Furthermore, the performance of our proposed algorithms, MPS-TD3 and PAM, falls. The average cumulative reward is 5.35 and 10.04 respectively. The trajectories reveal that the methods already struggle with leaving the first obstacles behind. A high number of steps is necessary even when the goal is reached. In the case of MPS-TD3, the average episode length is 35.16 while it is 28.27 with PAM. MPS-TD3 solves 5.83\% and PAM 12.08\% of the episodes. Additionally, we find circular behavior in some rollouts. This also holds for the baselines except for DQN. 

In general, the results suggest that there are no significant differences within on- or off-policy learning with strict continuous restrictions (p $\geq$ 0.05). For instance, PPO has a mean return of 52.71, 55.10, and 49.73 for masking, projection, and random replacement. On the off-policy side, a similar picture emerges. Both of our proposed algorithms reach a likewise average return. The trend holds for the length of an episode and the fraction of solved environments either. The only difference is that masking requires slightly fewer steps than random replacement (p $\leq$ 0.01). However, note that the cumulative reward is off-policy constantly lower than on-policy although the significance is only found between the projection baselines.  
All results, visualizations, and significance values can be found in appendix \ref{app:exp1_3}. 

\subsubsection{Action Space}
\begin{figure}[t]
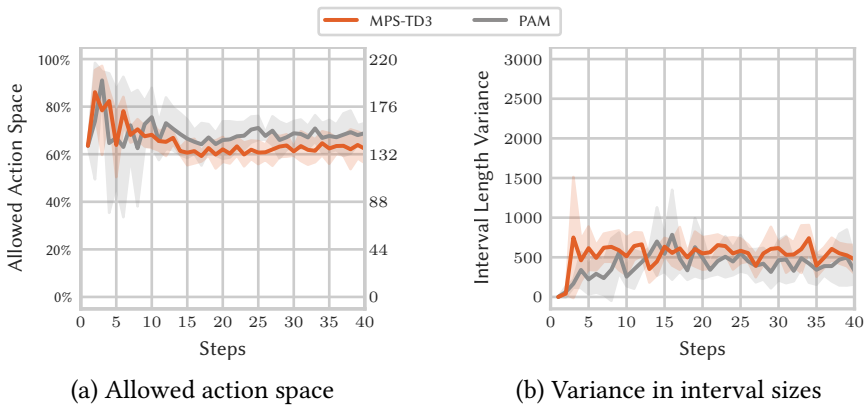

\centering
\begin{subfigure}{0.8\linewidth}
  \centering
  \input{experiment_1/text/legend_pam_mps.pgf}
\end{subfigure}
\begin{subfigure}{.49\linewidth}
  \centering
  \input{experiment_1/text/allowed_9.pgf}
  \caption{Allowed action space}
  \label{fig:exp1_complex_allowed}
\end{subfigure}%
\begin{subfigure}{.49\linewidth}
  \centering
  \input{experiment_1/text/variance_3.pgf}
  \caption{Variance in interval sizes}
  \label{fig:exp1_complex_variance}
  \centering
\end{subfigure}
\label{exp1_complex_actions}
\caption{MPS-TD3 and PAM action space behavior with multiple obstacles}
\end{figure}

The average allowed action space is similar between all algorithms and ranges from 64.67\% with MPS-TD3 to 68.73\% in the case of PAM. The portions make average sizes of 142.27 and 151.20 respectively. Only the unmodified TD3 and PPO can select with 75.04\% and 74.70\% from larger fractions. As an example, the average portion of available actions is visualized for MPS-TD3 and PAM in figure \ref{fig:exp1_complex_allowed} over time. The courses of the other algorithms behave likewise and can be found in appendixes \ref{app:exp1_3_length_ppo}, \ref{app:exp1_3_length_td3}, and \ref{app:exp1_3_length_own}. The restrictions are slightly lower at the beginning and the trajectories show that the only constraints are caused by the border of the environment. Afterward, most algorithms fail to find a way to the goal. They maintain large restrictions by running into obstacles again. 

Between 1.32 and 1.44 disjoint intervals are on average available at the same time. Only MPS-TD3 has to choose with a mean of 1.50 from significantly more intervals. The average size for a single subset for MPS-TD3 is 114.46 and 130.00 in the case of PAM. Figure \ref{fig:exp1_complex_variance} illustrates the variance in the interval sizes within an action space for both methods over time. The results show that the variability is approximately constant. More metrics are described in appendix \ref{app:exp1_3}.

\subsubsection{Discussion}
Overall, the outcomes are similar to the previous experiment but the differences are more clear. 
The experiment indicates that DQN with masking is the only method to find alternative paths which maximize the reward. 
Estimating a finite number of action values in parallel and then removing unavailable parts performs better than replacing actions heuristically or scaling the output. We hypothesize the advantage comes from the learning procedure. The implicit handling makes the task less complex since no scaling needs to be learned. However, the procedure is problematic in continuous action spaces since the number of actions is infinite. More research is necessary to develop techniques without discretization.

All other algorithms, including MPS-TD3, perform significantly worse than in simple environments and our architecture is not an alternative compared to on-policy baselines. However, the outcomes are partly expected. The previous subsection predicted the trend and all experience tuples on the optimal path are based on the full action space. The agent has no reference which could indicate an advantage for an interval or an action and may confuse the obstacle with the border of the map. Therefore, imposing artificial restrictions could improve performance. Nevertheless, we expect arbitrary action space subsets to impact the difficulty of the task. Future experiments must explore the influence of artificial restrictions with varying properties and can try MPS-TD3 with different models, such as PPO. Further, we encourage integrating the performance without trained restrictions in future works from other authors. For example, when using pre-trained models, it might not always be possible to incorporate restrictions from the beginning.

In line with previous results, no superior method for continuous action spaces could be found. The outcomes only suggest that penalty-based approaches are worse. The result is problematic since reward shaping is a common practice to handle restrictions in the literature \cite{ran2022optimizing, wang2019autonomous}. Future works should consider replacing their penalty with methods such as random replacement or using DQN with a discretization of the action space. Note that results are only transferable when the agents face dynamic restrictions in their application that they did not encounter before. As previously mentioned, research in this field is sparse and it is difficult to compare our results to previous works. 

In obstacle avoidance, the results promote using local data to restrict the action space when unseen obstacles occur and suggest relying on discrete actions until better continuous methods are found. For example, the sensor data of Tai et al. \cite{tai2017virtual} or Meyer et al. \cite{meyer2020taming} could be used to exclude intervals in which an obstacle is detected too close instead of returning less reward. We hypothesize unseen restriction handling can be improved this way. It would also be interesting to compare the performance when global information becomes available.  
\newpage
Overall, the experiment indicated the anticipated effect. Compared to the previous subsection, the average fraction of the allowed action space decreased and the number of simultaneous disjoint intervals was increasing. Subsequent experiments can work on more restrictions and explore the limits of DQN. 

\subsection{Summary}
In the context of this chapter, three experiments with dynamic restrictions were carried out. We included 10 baseline algorithms and our proposed improvements MPS-TD3 and PAM. However, we failed to stabilize PAM which makes implications difficult. We explored the case when agents are not trained with restrictions but face them when applied. Related to the research question of how the performance of RL algorithms can be measured with dynamic interval action space restrictions, the environments are used to provide the first benchmarks. For answers on how an optimal alternative policy can be found, our results suggest the following guidelines. Note that we do not consider PAM further since it failed to learn even in unrestricted episodes.

\begin{enumerate}
    \item \textbf{The techniques do not seem to have a negative effect on learning}. The baseline algorithms and MPS-TD3 converged close to the maximum reward. Discrete PPO was an exception. However, we believe this behavior comes from the sub-optimal configuration and may not due to the restriction-handling procedure.

    \item \textbf{Penalty-based restrictions should be avoided}. Returning a negative reward continuously performed the worst. While in simple environments, knowledge from the minimal restrictions could be used to surround at least some of the obstacles, the method collapsed completely in the complex domain. 

    \item \textbf{Any other continuous algorithm works for simple restrictions}. When we assume that the baselines are compared in combination with PPO, since it was generally more suitable for the environment, then the differences were only marginal. Nevertheless, projection and DQN had a slight advantage.

    \item \textbf{Discretization and DQN masking is necessary for complex subsets}. DQN with discrete masking was the only method to achieve good results in an environment with multiple obstacles. Despite the discretization error, it is therefore inevitable for previously unseen and more complex dynamic restrictions. 
\end{enumerate}

\newpage
\section{Learning With Dynamic Obstacles}
When dynamic restrictions emerge during training, architectures parameterizing some of their layers should have an advantage. Domain knowledge is directly incorporated into the structure of the neural network. In this case, experience tuples allow learning the scaling behavior and interval selection. Therefore, we assume that performances change compared to models trained without restrictions. For example, MPS-TD3 is supposed to surround obstacles more smoothly than postprocessing actions with random replacement in which an invalid selection is replaced with an arbitrary feasible value. 

To explore each agent's behavior, we split the experiment again into three parts: First, we train the agent to reach the goal. This is similar to the previous section except that random restrictions can already appear at this point. Afterward, we repeat the evaluations in the simple and complex domains. 

\subsection{Training}
\label{sec:exp2_training}
As in earlier analyses, assessing the performance in an unrestricted episode is crucial when it comes to measuring the effect of obstacles on algorithms. In the context of this experiment, different learning outcomes must be considered since dynamic restrictions already occur on the optimal path during training. Hence, this subsection first presents the new training results. This is again done by describing the updated learning procedure before continuing with the final performance. The experiment is supposed to explore two folds: First, we are interested in whether agents can learn under dynamic restrictions and still find the optimal path when the environment is obstacle-free. Second, it measures the new baseline performance of each algorithm which can be compared with episodes containing restrictions. 

\subsubsection{Setup}
\begin{figure}[t]
\centering
\fbox{\includegraphics[width=3.5cm]{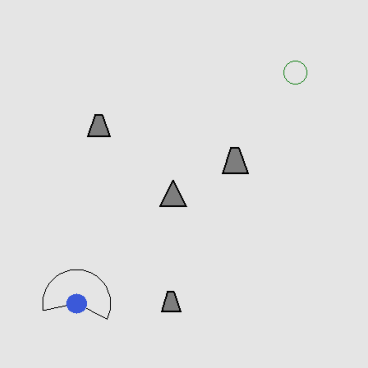}}
\caption{Example of a training episode with four dynamic obstacles}
\label{fig:exp2_env_example}
\end{figure}

The following experiments are based on the same environment as in previous subsections. The specific parameters can be read from table \ref{tab:env-setup} and are described in section \ref{sec:training}. Before each episode, we place four random obstacles on the map. Each of them moves between two waypoints. An exemplary setup is illustrated in figure \ref{fig:exp2_env_example}. On one hand, more obstacles might make learning more difficult and exhibit differences more clearly. On the other hand, the influence on the training behavior of even small restrictions is currently unknown. Four obstacles have the advantage, that, due to the spread, the probability rises that the agent actually encounters one of them without making the task too complex. This is important at the beginning of training when the path of the agent varies. The covariance matrix is set to be independent and identically distributed with $\Sigma = \big(\begin{smallmatrix}
  7.5 & 0.0\\
  0.0 & 7.5
\end{smallmatrix}\big)$ and describes the variance around mean $\mu = (7.5, 7.5)$. The size of the obstacles is sampled from $\mathcal{N}(1.0, 0.25)$ and clipped to fit into the range $[0.25,1.75]$. The waypoints have an average distance of $\mu = 4.0$ with variance $\sigma^2 = 0.25$. The step size is 0.1. This way, more than 40 episode steps are on average required to reach the second waypoint. The movement speed of the obstacles is relatively low compared to the agent. This is necessary so that the algorithms can always escape restrictions. Note that we train the discrete PPO with manually tuned hyperparameters. To ensure the reproducibility of our results, we sample the training environments from a fixed order of seeds. We start with seed 0 and gradually increase the value by one after an episode ends. 
The algorithms are trained with seeds 40, 41, 42, 43, 44, and 45. We run each for 800.000 time steps because of the increased complexity. 

\subsubsection{Results}
Similarly to the previous experiments, we start presenting the results within and across training runs and then continue with the outcomes of the evaluation rollouts according to the axes of variability.

\begin{figure}[t]
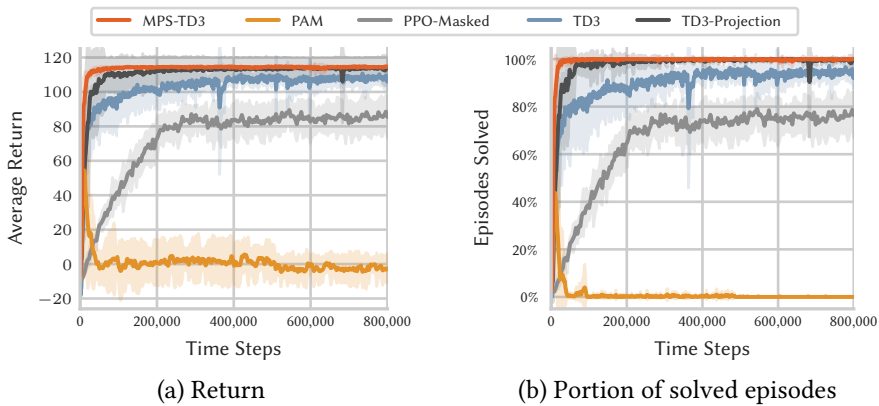

\centering
\begin{subfigure}{0.8\linewidth}
  \centering
  \input{experiment_2/text/legend.pgf}
\end{subfigure}
\begin{subfigure}{.49\linewidth}
  \centering
  \input{experiment_2/text/return.pgf}
  \caption{Return}
  \label{fig:exp2_training_return}
\end{subfigure}%
\begin{subfigure}{.49\linewidth}
  \centering
  \input{experiment_2/text/solved.pgf}
  \caption{Portion of solved episodes}
  \label{fig:exp2_training_solved}
\end{subfigure}
\caption{Training progress with dynamic restrictions}
\label{fig:exp2_training}
\end{figure}

\paragraph{Within and across training runs}
Most methods show a smooth and sharp learning curve. This behavior is illustrated in appendix \ref{app:exp2_0_ppo}, \ref{app:exp2_0_td3}, \ref{app:exp2_0_own}, and figure \ref{fig:exp2_training_return} which illustrates MPS-TD3 and TD3 with projection exemplary. The lines depict the average return with the standard deviation. MPS-TD3 and TD3 achieve mean rewards of 114.63 and 113.95 after the last training iteration. Near the end of the training, almost all episodes are solved. The fraction of goals that are reached improves likewise to the reward and shows a similar steady rise. The solved portion while training progresses is depicted in figure \ref{fig:exp2_training_solved}. That the task is learned quickly is also indicated by the number of time steps necessary to solve more than 80\% of the episodes on average. MPS-TD3 reaches the plateau of convergence after 8.000 and TD3 with projecting invalid actions requires 15.000 time steps. Afterward, the variability stays small and the algorithms keep a constant reward. The learning behavior holds for all methods with the exception of a few. 

For example, PAM does not learn a proper policy. The reward constantly fluctuates around zero and the majority of episodes are not solved throughout all time steps. Both patterns are likewise depicted in figure \ref{fig:exp2_training}. The shaded area additionally shows a larger standard deviation. It must be taken into account that, similarly to the previous section, we found that the action values are rising constantly. The average return increases at the beginning until the action value estimates get larger than the true values. Afterward, the model gets unstable and unlearns the previous. The inability to find a path is also indicated by the mean episode length in appendix \ref{app:exp2_0_solved_own}, which sticks near the maximum of 40 steps. 

Furthermore, PPO with discrete masking and penalty-based restrictions find a sub-optimal policy. The lower rewards and fewer solved episodes can be read from figure \ref{fig:exp2_training} as well. The plot example shows the unmodified TD3 algorithm which behaves similarly to PPO. Compared to the other baselines, the progress has not the same sharpness and the across training runs variability is higher. The latter is indicated by the larger standard deviation. 

However, the average episode length of penalty-based methods does not seem different from algorithms with a higher reward after convergence. Since collisions occur more frequently at the beginning, the average number of steps is increasing instead of declining. In appendixes \ref{app:exp2_0_steps_ppo} and \ref{app:exp2_0_steps_td3}, the rise can be considered as strong as the progress of algorithms solving all episodes. Note that this does not hold for the discrete PPO which converges to approximately 26 steps on average. The decline in the average episode length is also less steady.
\newpage
Generally, a smaller portion of actions is available near the beginning of training which increases simultaneously with the reward. For example, MPS-TD3 could only choose from approximately 70\% of the unrestricted action space during the first iterations. After convergence, the average available fraction is constantly oscillating around 80\%. Throughout the training, the actions are split over an average of approximately two intervals. Appendix \ref{app:exp2_0} indicates that this is similar between all algorithms and includes the other metrics from section \ref{sec:evaluation}. 

\paragraph{After training}
In the evaluation rollouts without obstacles, all runs, except for PAM, find proper paths. This is shown by the average return, episode length, and fraction of solved environments which are described in table \ref{tab:exp2_training_ev_results}. 
\begin{table}[t]
\centering
\begin{tabular}{llll}
\hline
\multicolumn{1}{c}{\textbf{Approach}} & \multicolumn{1}{c}{\textbf{Return}} & \multicolumn{1}{c}{\textbf{Steps}} & \multicolumn{1}{c}{\textbf{Solved}} \\ \hline
TD3                                   & $115.02 \pm 2.18$                   & $16.50 \pm 1.76$                   & $100.00\% \pm 0.00\%$               \\
TD3-Projection                        & $114.15 \pm 6.11$                   & $17.83 \pm 5.49$                   & $100.00\% \pm 0.00\%$               \\
TD3-Masking                           & $117.54 \pm 0.27$                   & $16.00 \pm 0.00$                   & $100.00\% \pm 0.00\%$               \\
TD3-Random                            & $115.56 \pm 5.11$                   & $17.50 \pm 3.67$                   & $100.00\% \pm 0.00\%$               \\
DQN-Masked                            & $117.61 \pm 0.09$                   & $16.00 \pm 0.00$                   & $100.00\% \pm 0.00\%$               \\ \hline

PPO                                   & $114.82 \pm 1.07$                   & $15.17 \pm 0.41$                   & $100.00\% \pm 0.00\%$               \\
PPO-Projection                        & $115.17 \pm 0.08$                   & $15.00 \pm 0.00$                   & $100.00\% \pm 0.00\%$               \\
PPO-Masking                           & $115.73 \pm 1.91$                   & $16.00 \pm 0.00$                   & $100.00\% \pm 0.00\%$               \\
PPO-Random                            & $115.03 \pm 0.23$                   & $15.00 \pm 0.00$                   & $100.00\% \pm 0.00\%$               \\
PPO-Masked                            & $115.04 \pm 1.54$                   & $17.00 \pm 0.89$                   & $100.00\% \pm 0.00\%$               \\ \hline

MPS-TD3                               & $115.76 \pm 1.68$                   & $15.67 \pm 0.52$                   & $100.00\% \pm 0.00\%$               \\
PAM                                   & $0.20 \pm 13.78$                    & $-$                                & $0.00\% \pm 0.00\%$                 \\ \hline
\end{tabular}
\caption[Average return, steps, and solved episodes - learned with obstacles]{Average return, steps, and solved episodes without obstacles}
\label{tab:exp2_training_ev_results}
\end{table}
Although according to appendixes \ref{app:exp2_1_test_reward} and \ref{app:exp2_1_test_steps} some differences in the metrics are statistically significant, the effect size can be considered small. An example is TD3 with continuous masking and PPO projecting invalid actions. The rewards are 117.54 and 115.17 respectively so the difference is only 2.37 (p $\leq$ 0.01). The same pattern can be found in the average episode length where the mean number of steps is 16.00 and 15.00 (p $\leq$ 0.01). An equivalent fit is additionally supported by the fact that MPS-TD3 and the baselines reach the goal in all of the environments. It is only necessary to take into account that PAM fails to solve a single episode and reaches mean returns of 0.20 which is significantly lower (p $\leq$ 0.01). All results, visualizations, and significance values are illustrated in appendix \ref{app:exp2_1}.

\begin{figure}[t]
\centering
  \centering
  \input{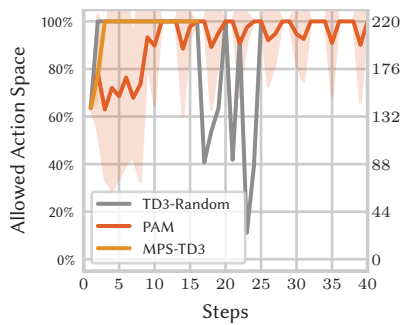}
\caption[Allowed action space over time]{Allowed action space over time}
\label{fig:exp2_training_ev_allowed}
\end{figure}

The action space behaves likewise to the direct path found in most trajectories. Appendixes \ref{app:exp2_1_count_ppo}, \ref{app:exp2_1_count_td3}, and \ref{app:exp2_1_count_own} indicate that typically only a single interval is available. The restrictions occur at the beginning of an episode for the same reason as in the previous section: The waypoints show that the agent starts next to a border of the environment. As an example, figure \ref{fig:exp2_training_ev_allowed} illustrates the fraction of the allowed actions for MPS-TD3, PAM, and TD3 with random projection over time. Near the end of an episode, single runs of the latter cause fluctuations.  By this means, the algorithms pass the center of the map but hit the border right next to the goal before entering it. The action space is summarized with all introduced metrics in appendixes \ref{app:exp2_1_control} and \ref{app:exp2_1_single}. Further, the variables are illustrated over time in appendixes \ref{app:exp2_1_control_ppo}, \ref{app:exp2_1_control_td3}, and \ref{app:exp2_1_control_own}.

\subsubsection{Discussion}
The results indicate that strictly respecting restrictions can improve learning progress and final performance. The conclusion is similar to Brosowski et al. \cite{brosowsky2021sample} with static and convex, as well as Huang et al. \cite{huang2019comparing} with discrete restrictions.
We hypothesize that the reason is the more accurate action space. Likewise, Kanervisto et al. \cite{kanervisto2020action} could improve PPO by experimenting with the number of actions. 
When the actions contain invalid subsets, agents learn from more actions. Otherwise, they focus on the available parts from the beginning, making the task less complex.

The advantage seems to be connected to improved exploration. Specifically, in the presented environment this might have a significant impact. Since an episode ends when an invalid action is selected, it is less likely that the agent explores states further away from the starting location at the early stages of training. Restrictions have to be learned before approaching more distant locations. This is not the case for algorithms respecting restrictions by design and the agent might reach the goal earlier by chance. 
This means that the penalty-based approaches find the target location less frequently at the beginning which could slow learning. The goal might have a greater impact than small improvements toward the target. 
\newpage
Our proposed algorithm MPS-TD3 comes close to the optimal path within the first training iterations. The outcome is similar to all baselines except for the penalty-based methods and discrete PPO. The unmodified algorithms confirm Karasowski et al. \cite{saferloverview} who also report lower rewards near the beginning. We assume that the discrete PPO results are again due to the sub-optimal configuration. We add to the literature that MPS-TD3 can learn to choose between different intervals. However, the performance is the same as when the subsets are concatenated. It is unclear how the outcomes differ when more intervals occur. 

As in the previous section, our results promote strictly respecting restrictions in obstacle avoidance as opposed to reward-shaping. Sensor or image data could cut dangerous angles out of the action space. By this means, the convergence time may improve and routes would be, contrary to Zhang et al. \cite{zhang2022autonomous}, completely collision-free. Future experiments can also try to combine the two methods: Likewise to Meyer et al. \cite{meyer2020taming}, the reward increases as agents approach barriers but actions are replaced if they lead to a collision. While Choi et al. \cite{choi2022marl} improved the performance by masking discrete actions that lead to collisions of every algorithm, we were investigating the actual effect of restrictions in a continuous domain.

Overall, the similar results of the algorithms ease the comparison in subsequent experiments. The desired impact of obstacles during training could be achieved. However, PAM could not solve any of the environments. In line with obstacle-free episodes, we anticipate the action values to be responsible for the instability. 

\subsection{Simple Restrictions}
The following part of this thesis uses the agents from the previous subsection and investigates how the approaches perform with simple constraints when dynamic restrictions already occurred during the learning process. 

\subsubsection{Setup}
Similarly to before, the evaluation environment contains a single obstacle in the center of the map. The implementation is the same as in section \ref{sec:exp1_simple_setup}. Note that for some episodes, the obstacle has a comparable size and form to what the agents already encountered during training. For this reason, we expect generally higher performance. Nevertheless, algorithms might handle the restrictions in various ways revealing differences. The similarity to section \ref{sec:exp1_simple_setup} allows comparing the results to agents trained without dynamic restrictions. As formerly, we check each algorithm to pass the midpoint of the map visually and remove the runs of seed four for the unmodified TD3 as well as seed five for TD3 with projection and seed two with random replacement. The setups are sampled using seeds from 0 to 39. 

\subsubsection{Main Results}
\begin{figure}[t]
\centering
  \centering
  \input{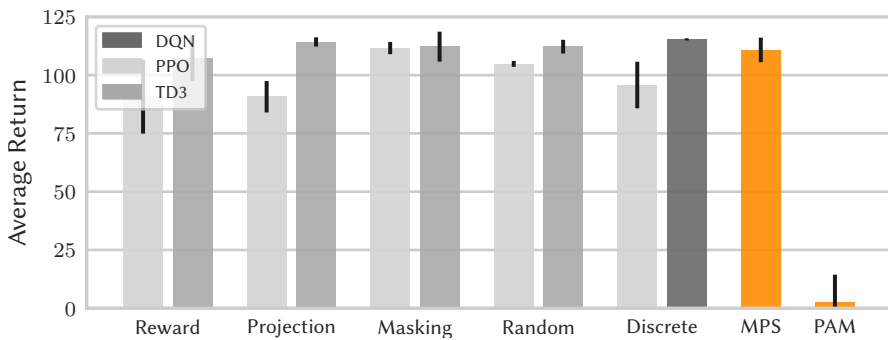}
\caption[Average return with simple restrictions]{Average return with simple restrictions}
\label{fig:exp2_simple_return}
\end{figure}

The results show that DQN yields the highest average return and is the only approach reaching the goal in all episodes. Figure \ref{fig:exp2_simple_return} compares the mean cumulative rewards of MPS-TD3 and PAM to the baseline algorithms. Illustrations of the other metrics can be found in appendixes \ref{app:exp2_2_plot_solved} and \ref{app:exp2_2_plot_steps}. The difference of DQN is significant with respect to all on-policy methods (p $\leq$ 0.05) also because of the lower standard deviation as indicated by the black bars. The mean return is 115.34 suggesting a straight path. DQN shows relatively low episode lengths with an average number of 17.90 steps. In contrast, MPS-TD3 takes a mean of 19.99. 

However, no difference in the rewards could be found between off-policy algorithms (p $\geq$ 0.05). For example, MPS-TD3 is close to DQN with an average cumulative reward of 110.82. All methods reach mean returns of more than 100. A similar picture emerges for the fraction of solved episodes (p $\geq$ 0.05). MPS-TD3 reaches the goal in 97.92\% of the cases which is only 2.09\% more than masking.

Appendix \ref{app:exp2_2_results} shows that projection has the lowest average episode length on- and off-policy.
The approach needs off-policy a mean of 17.23 and therefore significantly fewer steps than all algorithms. Exceptions are the unmodified TD3 (p = 0.06), both masking approaches (p $\geq$ 0.05), and PPO with projection (p $\leq$ 0.01). The mean episode length of the on-policy projection is with 16.39 significantly the lowest (p $\geq$ 0.05). However, note that only 80\% of the episodes are solved. DQN, as the approach with the highest reward, requires 17.90 actions on average. 

Overall, on-policy learning shows marginally less performance compared to the off-policy counterparts. The average return is lower for penalty-based restrictions, projection, discrete masking, and random replacement (p $\leq$ 0.01). The figure above shows that the return drops on average below 100 for the unmodified PPO and the combination with projection, and discrete masking. For example, penalty-based restrictions reach 90.77 while TD3 makes an average return of 107.03. 

Out of all on-policy baselines, continuous masking achieves 111.61 and therefore the highest reward (p $\leq$ 0.05). The same pattern holds for the fraction of solved episodes. The method solves 97.08\% of the environments while the second highest percentage is a portion of only 91.67\% in the case of random replacement. 

Additionally, trajectory analysis reveals that circular behavior happens less frequently in this experiment than in agents trained without dynamic restrictions. This can be followed in the videos and paths for each approach in appendixes \ref{app:exp2_2_trajectory_start} to \ref{app:exp2_2_trajectory_end}.
The results for all algorithms and metrics which are not listed in this subsection are summarized in appendix \ref{app:exp2_2}. 

\subsubsection{Action Space}
\begin{figure}[t]
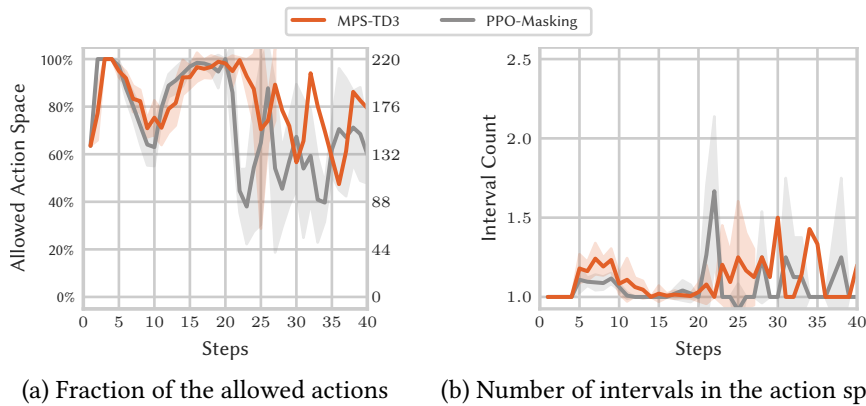

\centering
\begin{subfigure}{0.8\linewidth}
  \centering
  \input{experiment_2/text/legend_2.pgf}
\end{subfigure}
\begin{subfigure}{.49\linewidth}
  \centering
  \input{experiment_2/text/allowed_3.pgf}
  \caption{Fraction of the allowed actions}
  \label{fig:exp2_simple_allowed}
\end{subfigure}%
\begin{subfigure}{.49\linewidth}
  \centering
  \input{experiment_2/text/count.pgf}
  \caption{Number of intervals in the action space}
  \label{fig:exp2_simple_count}
\end{subfigure}
\caption[Action space behavior of MPS-TD3 and PPO with masking]{Action space behavior of MPS-TD3 and PPO with masking}
\label{fig:exp2_simple_actions}
\end{figure}

The action space follows a similar behavior to section \ref{sec:exp1_simple_setup}.  We illustrate the allowed fraction and number of available intervals for MPS-TD3 and PPO with continuous masking in figure \ref{fig:exp2_simple_actions}. Appendixes \ref{app:exp2_2_control_ppo}, \ref{app:exp2_2_control_td3}, and \ref{app:exp2_2_control_own} show that the patterns are similar between all algorithms. The trajectories reveal that the agents start with constraints due to the border of the environment. Afterward, no restrictions occur until the obstacle blocks the path. On the straight way to the goal, the action space is the smallest on the 10th step. The average number of disjoint intervals increases since the obstacle could split the action space into multiple parts. As the previous subsection indicated that most algorithms reach the target location at approximately 20 steps, larger fractions of the actions are available until then. In most cases, the straight path to the goal without restrictions is taken after the obstacle has been passed. The variability after approximately 20 steps is due to runs that miss the target and follow a sub-optimal policy.
\newpage
Over entire episodes, the average size of the allowed action space is between 165.93 for PPO with projection and 192.99 in the unmodified on-policy case. This makes 75.42\% and 87.72\% respectively. The mean of other algorithms is in between these two values. The single intervals are with lengths between 162.15 and 187.94 on average only slightly smaller. This is already indicating the low variance within an action space. Since the average number of disjoint intervals is relatively low, the dispersity does not increase unexpectedly high. The reason is that most of the time only a single interval is available causing the variance to be zero. For example, in the case of PPO, the average size of the smallest interval in the action space is 185.29 while the largest is 190.58. This pattern can also be observed in appendix \ref{app:exp2_2_control_ppo} over time. Until the 20th step, the mean standard deviation of the interval sizes is constantly below 22 and the lengths are close. From appendixes \ref{app:exp2_2_variance_ppo}, \ref{app:exp2_2_variance_td3}, and \ref{app:exp2_2_variance_own} follows that the standard deviation oscillates for most algorithms at around 15. Appendix \ref{app:exp2_2} further contains all metrics for every algorithm. 

\subsubsection{Discussion}
Likewise to the previous section, the answer is unclear when looking for a preferred method to handle interval action space restrictions. Similarly to the agents trained without dynamic obstacles, DQN was the most stable and achieved the highest results. However, differences are only marginal, and it is uncertain if the results hold for more complex environments due to the discretization error. 

MPS-TD3 is comparable to DQN but works in continuous action spaces. More tests can show if our approach is preferable to the other baselines due to the ability to handle a possibly infinite number of subsets. In contrast to section \ref{sec:without-simple-results}, the unmodified algorithms perform similarly to strictly following restrictions. These findings support that sensor \cite{tai2017virtual}, image \cite{guo2020deep}, or global data \cite{meyer2020taming} is sufficient to pass simple dynamic obstacles. Future works could explore the setups of Wang et al. \cite{wang2019autonomous} to see how all approaches deal with dead ends. 

We hypothesize the general high performance of all approaches might be due to overfitting. The presented scenario could have appeared likewise during training in which all algorithms have learned proper paths. The exception is obstacles larger and smaller than the ones during training. Nevertheless, it is unlikely that the obstacle was found exactly in the middle of the map and the same restrictions occurred. Therefore, the constraints can still be considered new. Due to the equal results, it is questionable if the presented environment is useful in terms of the question of how the performance of algorithms with dynamic restrictions can be measured or if the performance is really almost equivalent. Future experiments are necessary. 
\newpage
During the analysis, it has been shown that continuous masking and projection achieved lower step counts than MPS-TD3. This could suggest better restriction handling and would mean that scaling into the concatenation of the multiple disjoint intervals performs better than considering each separately. However, the projection results could be related to the fewer solved episodes. These episodes might be the ones with larger obstacles requiring longer paths. Subsequent works should consider controlling for the size of obstacles more specifically. Note again that some algorithms do not follow a completely straight path and have an advantage with small obstacles. 

However, in summary, the obstacles have shown the desired effect of restrictions similar to the agents trained without obstacles in section \ref{sec:simple_restrictions_without}. 

\subsection{Complex Restrictions}
After the previous subsection presented the outcomes with a simple environment, the following repeats the evaluation in the more complex domain. 

\subsubsection{Setup}
Again, the setup is the same as in section \ref{sec:complex_res_exp_1}. We use the previously presented implementation since we intend to compare the results with the earlier agents. The environments are sampled with seeds from 0 to 39 which makes a total of 240 evaluation rollouts per algorithm. 

\subsubsection{Results}
The outcomes show that MPS-TD3, projection, masking, and DQN are similar in average return. All of them achieve high rewards with only a few minor differences being significant at the 0.05 level. For example, TD3 with continuous masking reaches 113.88 compared to MPS-TD3 with 110.73 (p $\leq$ 0.01). Contrarily, DQN, MPS-TD3, PPO with continuous masking, and both projection approaches are not significantly different from each other (p $\geq$ 0.05). 

Likewise to the return, the average fraction of solved episodes can be considered generally high. MPS-TD3 and TD3 with continuous masking solve most episodes. Both reach the goal in 99.17\% of the rollouts. This implies that the algorithms miss the target in only two out of the 240 environments. Exceptions are the unmodified TD3 and PPO that attain with 33.75\% and 6.67\% the target location less frequently. Additionally, PPO with discrete masking and random replacement find a way in only 82.01\% and 83.33\% of the cases. Our proposed PAM algorithm does not solve a single episode. Nevertheless, even randomly replacing invalid actions with TD3  arrives 95.42\% of the time. 

\begin{table}[t]
\centering
\begin{tabular}{llll}
\hline
\multicolumn{1}{c}{\textbf{Approach}} & \multicolumn{1}{c}{\textbf{Return}} & \multicolumn{1}{c}{\textbf{Steps}} & \multicolumn{1}{c}{\textbf{Solved}} \\ \hline

TD3                                   & $43.94 \pm 19.87$                   & $22.89 \pm 2.17$                   & $33.75\% \pm 17.23\%$               \\
TD3-Projection                        & $107.36 \pm 6.57$                   & $19.80 \pm 4.90$                   & $95.00\% \pm 4.47\%$                \\
TD3-Masking                           & $113.88 \pm 1.28$                   & $17.59 \pm 0.81$                   & $99.17\% \pm 1.29\%$                \\
TD3-Random                            & $106.85 \pm 4.65$                   & $21.36 \pm 2.34$                   & $95.42\% \pm 4.31\%$                \\
DQN-Masked                            & $112.14 \pm 2.40$                   & $17.88 \pm 0.28$                   & $96.25\% \pm 2.62\%$                \\ \hline

PPO                                   & $12.70 \pm 7.96$                    & $22.02 \pm 5.80$                   & $6.67\% \pm 4.92\%$                 \\
PPO-Projection                        & $110.61 \pm 3.06$                   & $16.45 \pm 0.23$                   & $95.00\% \pm2.74\%$                 \\
PPO-Masking                           & $106.80 \pm 4.92$                   & $18.15 \pm 0.75$                   & $86.03\% \pm 2.86\%$                \\
PPO-Random                            & $94.56 \pm 4.39$                    & $21.83 \pm 1.61$                   & $83.33\% \pm 4.38\%$                \\
PPO-Masked                            & $67.56 \pm 1.45$                    & $22.86 \pm 1.76$                   & $82.01\% \pm 1.66\%$                \\ \hline

MPS-TD3                               & $110.73 \pm 1.42$                   & $21.10 \pm 0.10$                   & $99.17\% \pm 2.04\%$                \\
PAM                                   & $-10.87 \pm 3.94$                   & $-$                                & $0.00\% \pm 0.00\%$                 \\ \hline
\end{tabular}
\caption[Average return, steps, and solved episodes - complex restrictions]{Average return, steps, and solved episodes}
\label{tab:exp2_complex}
\end{table}

In terms of path optimality, we found that MPS-TD3, random replacement, discrete PPO, and unmodified algorithms have higher average episode lengths. For example, while having relatively high rewards and many solved episodes, MPS-TD3 has an average episode length of 21.10 with a low standard deviation. Randomly replacing invalid actions takes mean steps of 21.36 and 21.83. Therefore the required actions are significantly higher than the ones of both continuous masking algorithms, PPO with projection and DQN (p $\leq$ 0.05).   

Overall, PAM, discrete PPO, and unmodified algorithms perform worse across all metrics. For instance, TD3, and PAM achieve significantly lower mean returns of 43.94 and -10.87 compared to the 113.88 of TD3 with continuous masking (p $\leq$ 0.01). Also, the discrete PPO attains with an average of 67.56 lower cumulative rewards (p $\leq$ 0.01). Table \ref{tab:exp2_complex} shows that the pattern holds across metrics. 

Just like in the previous sections, we observe that on-policy approaches tend to achieve lower values than off-policy methods for all the metrics being considered. For example, PPO with random replacement solves only 83.33\% of the episodes and achieves a relatively low mean return of 94.56 compared to 95.42\% and 106.85 (p $\leq$ 0.01). Although statistical significance is missing for the on- and off-policy difference between the continuous masking algorithms, the mean cumulative reward and fraction of solved episodes is higher. Only projection reaches in combination with PPO an average return of 110.61 as opposed to 107.36 with TD3 (p $\geq$ 0.05). The general trend is best illustrated in appendixes \ref{app:exp2_3_plot_rerward}, \ref{app:exp2_3_plot_solved}, and \ref{app:exp2_3_plot_steps}. All visualizations of the metrics comparing our algorithms and the baselines can be found in appendix \ref{app:exp2_3}. 

\subsubsection{Action Space}
\begin{figure}[t]
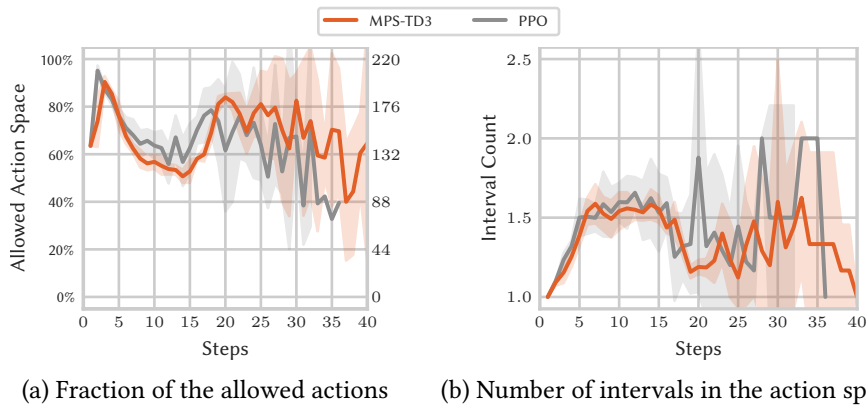

\centering
\begin{subfigure}{0.8\linewidth}
  \centering
  \input{experiment_2/text/legend_4.pgf}
\end{subfigure}
\begin{subfigure}{.49\linewidth}
  \centering
  \input{experiment_2/text/allowed_4.pgf}
  \caption{Fraction of the allowed actions}
  \label{fig:exp2_complex_allowed}
\end{subfigure}%
\begin{subfigure}{.49\linewidth}
  \centering
  \input{experiment_2/text/count_4.pgf}
  \caption{Number of intervals in the action space}
  \label{fig:exp2_complex_count}
\end{subfigure}
\caption[Action space behavior for MPS-TD3 and PPO]{Action space behavior for MPS-TD3 and PPO}
\label{fig:exp2_complex_actions}
\end{figure}

The action space is similar to section \ref{sec:complex_res_exp_1} but restrictions get fewer as the agent approaches the goal. The total average available length is between 140.90 for TD3 with projection and 160.93 for the unmodified PPO. Percentage-wise these are 64.04\% and 73.15\%. All other algorithms, including MPS-TD3 and PAM, have a mean in between the two values. On average, the actions are split over 1.28 and 1.39 disjoint intervals. Figure \ref{fig:exp2_complex_actions} shows the allowed fraction and number of available intervals for MPS-TD3 and PPO over time. The algorithms are selected as exemplary. Appendix \ref{app:exp2_3_control_ppo}, \ref{app:exp2_3_control_td3}, and \ref{app:exp2_3_control_own} demonstrate an equivalent course for the other methods. The restrictions are smallest near the beginning and then increase until the size of the allowed action space is smallest at approximately the 15th step. Afterward, more actions become available before most agents reach the goal. The standard deviation in the figure suggests that the variability is relatively small. Only the second half of the maximum episode length shows larger fluctuations caused by single runs missing and reaching the goal later. 

The average size of a single interval is approximately 20 smaller than the sum of all subsets. The appendix shows that this is similar between all algorithms. For example, the available actions of MPS-TD3 make an average of 144.93 while the individual intervals have a mean length of 124.01. The smallest interval has an average size of 113.56 and the largest 134.53. The course behaves likewise to figure \ref{fig:exp2_complex_allowed}. The fewest actions are available before the 15th step. Correspondingly, the standard deviation of the interval sizes within the action space is highest between the 5th and 15th actions with means oscillating around approximately 22. The average and pattern over time of each metric are contained in appendix \ref{app:exp2_3}. 

\subsubsection{Discussion}
The outcomes provide valuable insights that are relevant to the research question of this thesis.
In line with our expectations, MPS-TD3, continuous masking, and projection significantly improved their performance compared to section \ref{sec:complex_res_exp_1} and are an alternative to DQN. Each of the algorithms comes with strengths and weaknesses. For example, while MPS-TD3 is worse in finding the shortest path, the approach is generally more likely to reach the goal. Training with obstacles seems to allow an understanding of the scaling behavior and action values of intervals. Likewise, to Krasowski et al. \cite{saferloverview}, the decision for an algorithm depends on the task at hand. However, the results suggest that the impact of the choice can be minor. Further tests may explore if differences emerge in more complex environments. We also believe that projection highly depends on the implementation. In our case, the projected action is close to the most efficient way to pass an obstacle. Hence, the performance may be overestimated. Subsequent studies can analyze the effect with specific environment layouts and obstacle forms. 

Second, random replacement, PAM, and unmodified algorithms took more steps or solved significantly fewer episodes. The results are contrary to the experiments with a single obstacle indicating that the approaches are more suitable for simple domains. 
Randomly replacing invalid actions was only slightly worse. We assume strategies must be learned or well-designed as in the projection case. It would be interesting if sophisticated techniques such as extra replay buffers \cite{hsu2022improving} improve results. However, the approach is preferable over penalty-based restrictions and PAM. The unmodified algorithms should be avoided after they have also shown slower convergence. PAM can not be evaluated as the training failed.

In the context of obstacle avoidance, the outcomes are similar to section \ref{sec:complex_res_exp_1} and suggest strictly avoiding obstacles in complex environments when the information is available. Contrary to the simple rollouts, the unmodified approaches have problems when many obstacles have to be passed. However, discretization for DQN is not required as opposed to training without restrictions.
 
Overall, the action space had similar sizes as in the previous subsection but restrictions emerged continuously and with more intervals. Nonetheless, experiments in different tasks are necessary to control the transferability of results.

\subsection{Summary}
In this part of the thesis, we repeated the three experiments from section \ref{sec:without}. Contrary to before, we optimized MPS-TD3. PAM and each of the 10 baseline algorithms with dynamic interval restrictions during training. By this means, the procedure explored the research questions more similarly to recent literature. Nevertheless, we found that experiments are not complete and need to be expanded in subsequent works. As previously, the results suggest specific recommendations that can be used to address the research questions and guide future directions. Note that we again do not consider PAM since the method could not be stabilized in this section either. 

\begin{enumerate}
    \item \textbf{Strictly restricting restrictions can improve return and convergence}. The original TD3 and PPO converged slower than the other approaches and reached fewer returns between the last training iterations. Note that it is questionable if this holds for the discrete PPO due to the sub-optimal hyperparameters in our experiments. 

    \item \textbf{All techniques can work properly with simple restrictions}. When only off-policy algorithms are considered, the differences were only marginal and all achieved a high reward. However, this does not hold in combination with PPO. Overall, DQN had slight but not significant advantages and a lower variability on all metrics. 

    \item \textbf{Penalty-based restrictions should be avoided with complex subsets}. Returning a negative reward achieved significantly fewer returns when restrictions were more complex compared to the simple domain. The method seems not suitable when restrictions increasingly differ from training. 

    \item \textbf{Complex replacement strategies must be learned or well defined}. Randomly replacing actions achieved high rewards but took longer paths than more sophisticated techniques. However, we also believe that projecting to fixed valid points must be well-designed to reach similar or better results as in this experiment. 
\end{enumerate}

\chapter{Conclusion}
After presenting the experimental results, the next sections conclude the paper with a summary, a description of limitations, and an outlook for future work. 

\section{Summary}
This thesis applied dynamic interval action space restrictions in deep RL.
The first goal was to find algorithms following an optimal policy with varying allowed actions. We developed two approaches that comply with an arbitrary number of subsets. 
First, PAM builds on the success of parameterized RL and uses P-DQN to divide the action space into predefined bins. The bins allow applying the theoretically well-founded discrete action masking but lower the discretization error. The trick is to combine each bin with a fine-grained action. Second, MPS-TD3 modifies TD3 to dispense the discretization and considers each of the intervals separately through multiple forward passes. The actions are scaled into the specific subset by parameterizing the output layer of the neural network. Afterward, the critic decides on the action to execute in the environment. By this means, we added an approach to support disjoint subsets to ConstraintNet. The algorithms were compared to common approaches to handle restrictions in the literature. The baselines included projection, replacement, as well as discrete and continuous masking. We applied PPO and TD3 unmodified and with each extension. 

The second goal was to find methods for evaluation algorithms with dynamic interval restrictions. We started by developing fundamental criteria. Afterward, we presented an obstacle avoidance environment with metrics to compare policies and monitor the effect of restrictions. We analyzed the learning behavior and tested agents in simple and complex environments. The experiments differentiated when algorithms were trained with and without obstacles. 

The outcomes had several implications for future research directions.
In terms of training progress, our study demonstrated that also with dynamic constraints, strictly adhering to restrictions can improve convergence and final performance. MPS-TD3 performed better than the unmodified algorithms but worked similarly to projection, masking, and random replacement. However, the advantage disappeared without dynamic restrictions. Only PAM turned out to overestimate action values and requires further modifications. Therefore, the method can currently not be regarded as a solution for the first research question of this thesis. Nevertheless, we presented recommendations that might help in stabilizing the algorithm. Since suggestions between the other approaches could not be found, we believe that our tasks were too simple to exaggerate differences clearly and leave complicated environments for future work.

We also conducted experiments with already converged agents. The outcomes indicated that discretizing a continuous action space and masking action values with DQN is inevitable when similar restrictions did not emerge in the training episodes. All continuous algorithms suffered from major performance drops as the unseen constraints got more complex. 
Afterward, we repeated the experiments with agents that already encountered obstacles during training. The results revealed that MPS-TD3, continuous masking, and projection can be used likewise to DQN. By this means, MPS-TD3 can be a continuous alternative without defining a maximum number of possible intervals prior to training. Nevertheless, the final decision between the algorithms seems to depend on the task at hand. We assume differences might become visible in other environments but no further experiments could be conducted due to the scope of this work. 

Besides, our results provided a new perspective on obstacle avoidance. While former works focused on shaping the reward, action, and observation space prior to training, we assume that the available information can restrict the agent to subsets of the actions at each time step individually. Contrary to studies that projected invalid actions into static and convex intervals, we evaluated disjoint subsets that can vary for the same observation. Our approach was collision-free from the beginning and has shown evidence that the flexible but more precise action space may improve pathfinding around obstacles. 

In summary, we could not conclusively clarify the question regarding RL algorithms for optimal policies with dynamic interval action space restrictions but gave insights for future work. We presented an obstacle avoidance task that provided first benchmarks and motivate extending the environment for further experiments. 
\newpage
\section{Limitations}
Although we monitored the effect of obstacles on the action space and varied the difficulty of our tasks, our study is due to several limitations.
A list of aspects that must be considered regarding the validity of our results is presented below. 

\begin{description}
    \item[One-Dimensional Action Space] The insights only hold for actions of a single dimension. As soon as the action space becomes multi-dimensional, the complexity of the task itself and MPS-TD3 rises drastically as the number of convex subsets increases. The performances might change and algorithms may be better suited for low- or high-dimensional problems. Further, actions of different forms can be combined. For example, the gym framework allows the integration of continuous, discrete, binary, and textual actions into composites such as dictionaries, tuples, and graphs. The spaces can consist of arbitrary subsets in different forms. Generally, the proposed approaches are capable of dealing with more complex structures as the actions are translated to discrete or continuous outputs but combinations of approaches are necessary which was not explored in this thesis. 
    
    \item[Simple Restriction Properties] Obstacles in our experiments produce relatively simple restrictions so that often a single interval is available. In practice, the number of disjoint subsets can get large and vary drastically. It might be inefficient or impossible to apply extensive zero-padding to the input. In other cases, restrictions may not be appropriately represented by intervals at all. For example, when the continuous output of a neural network in a highly non-linear embedding space should be limited. Then storing all intervals for each dimension can be infeasible. Additionally, note that restrictions might be interdependent. For example, a range of a dimension might be only allowed when another action lies in a specific interval. Only the projection and replacement baselines could handle the dependencies as no input representation is required. The projection would need a distance function that projects to the closest valid combination. Contrary to the other methods, the allowed subset is not learned and masking is not performed separately for each dimension. However, the results are based on heuristics.  
    
    \item[Generalization] Our outcomes only hold for the presented domain of obstacle avoidance in mapless navigation. No claims can be made about the transferability of the results to other tasks or algorithms. For example, TD3 is generally more suitable for high-dimensional problems and performance may change in other environments. Specifically, real-world transfer has been previously shown to be difficult \cite{zhu2021deep}. Learning physics for a robot to surround obstacles with extraneous impacts is more demanding and differences between the baselines and MPS-TD3 might become visible. Therefore, we expect the performance of each algorithm to be strongly task-dependent. Furthermore, properties, such as that the valid action of projection is close to the optimum, can not be assumed. By this means, the limitation includes different obstacle avoidance environments as well as unrelated fields.  

    \item[Availability of Restrictions] We assume that the restrictions can be calculated precisely and are provided with each step from the environment. However, it is not always possible to obtain fast and reliable constraints in practice. The computation time can differ and the interval boundaries might be not exact. For example, an autonomous driving situation might have already passed until the allowed motor parameters are approximated. Noisy restrictions simultaneously increase task complexity and would allow collisions even when the constraints are correctly followed. Besides, it might be possible that a finite set of restrictions can not be pre-computed at all. In this case, ways to limit an agent to unknown allowed subsets must be found. 
    
    \item[Hyperparameter Sensitivity] In this work, all algorithms were optimized with respect to the first 150.000 time steps. We believe that the differences in the performance without restrictions are due to the configurations. The initial differences between the approaches may change when hyperparameters are tuned with respect to more training iterations and seeds which was not possible due to the time scope in this thesis. The PPO with discrete masking was also learned with sub-optimal settings because of an error during the training process and the search could not be repeated. It is unclear if the approach would perform similarly to its off-policy counterpart. 
\end{description}

\section{Research Outlook}
The limitations from the previous section can provide starting points for future research. Besides adding more algorithms from the literature and performing more hyperparameter optimization, we consider the following points as the next steps. 

\begin{description}
    \item[PAM Stability] We believe more effort should be spent on stabilizing the PAM algorithm. The approach masks action values similar to DQN. The discrete method was the only effective algorithm when restrictions were applied to the agents trained without obstacles. Therefore, PAM could be a promising continuous alternative with fine-grained actions. The stabilization may be achieved by inverting the gradients and using learning rates that satisfy the Robbins-Monro condition. However, more sophisticated parameterized on- and off-policy architectures, such as H-PPO proposed by Fan et al \cite{fan2019hybrid}, can also be considered.  

    \item[Action Space Properties] The developed environment can be used to control for different properties of restrictions more specifically. For example, smaller and larger paths between obstacles can manipulate the fraction of the allowed action space during an episode. Our experiments indicated that the performance gets less when the alternative deviates more from the learned policy. However, future experiments should aim to implement a metric that controls for the deviation of the optimal restricted policy. It would also be interesting to compare the results to outcomes when the action space continuously consists of more disjoint intervals on average. The effect can be achieved by setting a large number of small obstacles on the map. Afterward, the impact can be measured and compared to the action space variables of this thesis. Moreover, a combination of static training and dynamic evaluation obstacles could make the selection of the interval actually matter. For example, an inefficient choice might make the agent take a longer path around a known static obstacle. Finally, future work can add more dimensions to the action space to approach more complex restrictions. The step size may be selected by the agent instead of being defined beforehand. 
    
    \item[Transferability] The generalization can be tested in different tasks. For example, interval restrictions arise in variations of multi-step bandit problems where taking an action blocks a subset temporarily in the next steps. Consider an action space $\mathcal{A} = [1,1]$ with an unknown underlying reward function and assume that a selection makes an interval of size 0.1 unavailable. Then action $a = 0.5$ produces the dynamic action space $\mathcal{A}_\varphi = \{[-1,0.4], [0.4,1.0]\}$ for the next time step.  An implementation of such a multi-step bandit problem based on maximizing oil extraction is implemented on the author's GitHub repository. Other applications are the cake-cutting problem, multi-agent systems in which agents depend on each other, and a governance that imposes external restrictions.

    \item[Restriction Correctness] Different noise processes can be used to disturb the interval boundaries of restrictions. This way, approximations are simulated and the combination of collisions and dynamic restrictions is analyzed. The algorithms have to learn when to rely on the provided constraints and may handle perturbations differently well. This way, the approaches take a step toward real-life environments. 
    
\end{description}


\bibliographystyle{plain}
\bibliography{thesis-ref}

\appendix
\captionsetup{list=no}

\addtocontents{toc}{\protect\newpage}
\chapter{Hyperparameters}
\label{cha:hp-configurations}
The following appendices describe the setup for hyperparameter optimization and the final configurations.

\section{Search Spaces}
\label{sec:hp-spaces}
We start by presenting the search spaces. Note that continuous parameters are denoted as floating points, discrete ranges as integers, and categorical choices in curly brackets. For example, $[5,50]$ means all integer values from $5$ to $50$.

\begin{itemize}
    \item \textbf{PPO}:
    \begin{table}[h]
    \small
    \begin{center}
    \begin{tabular}{ll}
    \hline
    \multicolumn{1}{c}{\textbf{Hyperparameter}} & \multicolumn{1}{c}{\textbf{Search Space}} \\ \hline
    clip parameter                               & $[0.1,0.4]$                               \\
    value function clip parameter                & $[5.0,15.0]$                                  \\
    value function loss coefficient              & $[0.3,2.0]$                               \\
    entropy coefficient                          & $[0.0,0.2]$                               \\
    learning rate                                & $[5 \cdot 10^{-8},0.005]$                 \\
    lambda                                       & $[0.8,1.0]$                               \\
    hidden neurons                               & $\{[128, 128], [256, 128], [256, 256],$   \\
                                                 & $[512, 256], [256, 512], [256]$           \\
                                                 & $[512], [128], [64]\}$                    \\
    mini-batch size                              & $[4,200]$                                 \\
    stochastic gradient descent iterations       & $[3,60]$                                  \\
    batch size                                   & $[200,6000]$                              \\ \hline
    \end{tabular}
    \caption{PPO hyperparameter search spaces}
    \label{tab:ppo-hp-search-spaces}
    \end{center}
    \end{table}
    
    \newpage
    \item \textbf{DQN and PAM}: Hyperparamers below the dashed line are PAM-specific.
    \begin{table}[!h]
    \small
    \begin{center}
    \begin{tabular}{ll}
    \hline
    \multicolumn{1}{c}{\textbf{Hyperparameter}} & \multicolumn{1}{c}{\textbf{Search Space}} \\ \hline
    learning rate                                & $[5 \cdot 10^{-8},0.1]$                  \\
    sigma                                        & $[0.0,2.0]$                               \\
    hidden neurons                               & $\{[128, 128], [256, 128], [256, 256],$   \\
                                                 & $[512, 256], [256, 512], [256]$           \\
                                                 & $[512], [128], [64]\}$                    \\
    initial epsilon                              & $[0.1,0.8]$                               \\
    final epsilon                                & $[0.005,0.1]$                             \\
    epsilon reduction time steps                 & $[1000,50000]$                            \\
    replay buffer capacity                       & $[1500,70000]$                            \\
    prioritized replay alpha                     & $[0.3,0.7]$                               \\
    prioritized replay beta                      & $[0.2,0.5]$                               \\
    batch size                                   & $[28,400]$                               \\ \hdashline
    categorical distribution temperature         & $[0.0,2.0]$                               \\
    parameter network hidden neurons             & $\{[128, 128], [256, 128], [256, 256],$   \\
                                                 & $[512, 256], [256, 512], [256]$           \\
                                                 & $[512], [128], [64]\}$                    \\
    parameter network learning rate              & $[5 \cdot 10^{-8},0.1]$                   \\ 
    dueling                                      & $\{True, False\}$                         \\ 
    double q learning                            & $\{True, False\}$                         \\ \hline
    \end{tabular}
    \caption{DQN and PAM hyperparameter search spaces}
    \label{tab:dqn-hp-search-spaces}
    \end{center}
    \end{table}

    \newpage
    \item \textbf{TD3 and MDS-TD3}:
    \begin{table}[!h]
    \small
    \begin{center}
    \begin{tabular}{ll}
    \hline
    \multicolumn{1}{c}{\textbf{Hyperparameter}} & \multicolumn{1}{c}{\textbf{Search Space}} \\ \hline
    policy delay                                 & $[1, 4]$                                  \\
    target policy noise                          & $[0.0,0.5]$                               \\
    target noise clip factor                     & $[0.0,0.5]$                               \\
    hidden neurons actor                         & $\{[128, 128], [256, 128], [256, 256],$   \\
                                                 & $[512, 256], [256, 512], [256]$           \\
                                                 & $[512], [128], [64]\}$                    \\
    hidden neurons critic                        & $\{[128, 128], [256, 128], [256, 256],$   \\
                                                 & $[512, 256], [256, 512], [256]$           \\
                                                 & $[512], [128], [64]\}$                    \\
    learning rate actor                          & $[5 \cdot 10^{-8}, 0.001]$                \\
    learning rate critic                         & $[5 \cdot 10^{-8}, 0.001]$                \\
    tau                                          & $[5 \cdot 10^{-4}, 0.005]$                \\
    l2 regularization                            & $[10^{-8}, 0.001]$                        \\
    initial epsilon                              & $[0.05,0.4]$                              \\
    final epsilon                                & $[0.005,0.1]$                             \\
    epsilon reduction time steps                 & $[1000,30000]$                            \\
    initial noise scale                          & $[0.2,30.0]$                              \\
    final noise scale                            & $[0.01,3.0]$                             \\
    noise reduction time steps                   & $[1000,50000]$                            \\
    replay buffer capacity                       & $[1500,70000]$                            \\
    prioritized replay alpha                     & $[0.3,0.7]$                               \\
    prioritized replay beta                      & $[0.2,0.5]$                               \\
    target update frequency                      & $[1,4]$                                   \\
    batch size                                   & $[50,512]$                               \\ \hline
    \end{tabular}
    \caption{TD3 and MDS-TD3 hyperparameter search spaces}
    \label{tab:td3-hp-search-spaces}
    \end{center}
    \end{table}
\end{itemize}

\newpage
\section{Final Configurations}
\label{sec:hp-configuration}
After having defined the search spaces, the following tables describe the final hyperparameter values for each experiment and algorithm. We round each value to two decimal points. The specific setting can be found in the corresponding GitHub repository. 
    \begin{table}[h]
    \footnotesize
    \begin{center}
    \begin{tabular}{lll}
    \hline
    \multicolumn{1}{c}{\textbf{Hyperparameter}} & \multicolumn{1}{c}{\textbf{Without Obstacles}} & \multicolumn{1}{c}{\textbf{With Obstacles}} \\ \hline
    clip parameter                               & $0.38$                & $0.34$                           \\
    value function clip parameter                & $32.52$                   & $7.15$               \\
    value function loss coefficient              & $1.36$                   & $1.51$            \\
    entropy coefficient                          & $1.23 \cdot 10^{-2}$      & $1.55 \cdot 10^{-3}$                         \\
    learning rate                                & $1.11 \cdot 10^{-4}$       & $1.21 \cdot 10^{-4}$          \\
    lambda                                       & $0.84$                   & $0.92$            \\
    hidden neurons                               & $[256, 256]$              & $[256, 256]$      \\
    mini-batch size                              & $84$                       & $87$          \\
    gradient descent iterations       & $43$                       & $4$           \\
    batch size                                   & $1998$                    & $2112$          \\ \hline
    \end{tabular}
    \caption{PPO configuration}
    \label{tab:ppo-hp-search-spaces}
    \end{center}
    \end{table}

    \begin{table}[h]
    \footnotesize
    \begin{center}
    \begin{tabular}{lll}
    \hline
    \multicolumn{1}{c}{\textbf{Hyperparameter}} & \multicolumn{1}{c}{\textbf{Without Obstacles}} & \multicolumn{1}{c}{\textbf{With Obstacles}} \\ \hline
    clip parameter                               & $0.30$                & $0.38$                           \\
    value function clip parameter                & $15.75$                   & $5.98$               \\
    value function loss coefficient              & $0.91$                   & $1.33$            \\
    entropy coefficient                          & $1.44 \cdot 10^{-3}$      & $7.06 \cdot 10^{-3}$                         \\
    learning rate                                & $5.29 \cdot 10^{-5}$       & $5.22 \cdot 10^{-6}$          \\
    lambda                                       & $0.85$                   & $0.87$            \\
    hidden neurons                               & $[256, 256]$              & $[256, 256]$      \\
    mini-batch size                              & $31$                       & $116$          \\
    gradient descent iterations       & $31$                       & $24$           \\
    batch size                                   & $2022$                    & $4868$          \\ \hline
    \end{tabular}
    \caption{PPO with projection configuration}
    \label{tab:ppo-hp-search-spaces}
    \end{center}
    \end{table}

    \newpage
    \begin{table}[h]
    \footnotesize
    \begin{center}
    \begin{tabular}{lll}
    \hline
    \multicolumn{1}{c}{\textbf{Hyperparameter}} & \multicolumn{1}{c}{\textbf{Without Obstacles}} & \multicolumn{1}{c}{\textbf{With Obstacles}} \\ \hline
    clip parameter                               & $0.40$                & $0.28$                           \\
    value function clip parameter                & $31.07$                   & $13.69$               \\
    value function loss coefficient              & $1.43$                   & $1.64$            \\
    entropy coefficient                          & $5.33 \cdot 10^{-3}$      & $7.94 \cdot 10^{-4}$                         \\
    learning rate                                & $8.60 \cdot 10^{-5}$       & $1.01 \cdot 10^{-5}$          \\
    lambda                                       & $0.84$                   & $0.83$            \\
    hidden neurons                               & $[256, 256]$              & $[256, 256]$      \\
    mini-batch size                              & $123$                       & $130$          \\
    gradient descent iterations       & $25$                       & $17$           \\
    batch size                                   & $2653$                    & $3643$          \\ \hline
    \end{tabular}
    \caption{PPO with random replacement configuration}
    \label{tab:ppo-hp-search-spaces}
    \end{center}
    \end{table}

    \begin{table}[h]
    \footnotesize
    \begin{center}
    \begin{tabular}{lll}
    \hline
    \multicolumn{1}{c}{\textbf{Hyperparameter}} & \multicolumn{1}{c}{\textbf{Without Obstacles}} & \multicolumn{1}{c}{\textbf{With Obstacles}} \\ \hline
    clip parameter                               & $0.35$                & $0.29$                           \\
    value function clip parameter                & $43.88$                   & $9.23$               \\
    value function loss coefficient              & $1.97$                   & $1.33$            \\
    entropy coefficient                          & $1.09 \cdot 10^{-2}$      & $3.05 \cdot 10^{-3}$                         \\
    learning rate                                & $1.28 \cdot 10^{-4}$       & $1.53 \cdot 10^{-4}$          \\
    lambda                                       & $0.86$                   & $0.88$            \\
    hidden neurons                               & $[256, 256]$              & $[128]$      \\
    mini-batch size                              & $121$                       & $83$          \\
    gradient descent iterations       & $32$                       & $18$           \\
    batch size                                   & $875$                    & $3771$          \\ \hline
    \end{tabular}
    \caption{PPO with continuous masking configuration}
    \label{tab:ppo-hp-search-spaces}
    \end{center}
    \end{table}

    \newpage
    \begin{table}[h]
    \footnotesize
    \begin{center}
    \begin{tabular}{lll}
    \hline
    \multicolumn{1}{c}{\textbf{Hyperparameter}} & \multicolumn{1}{c}{\textbf{Without Obstacles}} & \multicolumn{1}{c}{\textbf{With Obstacles}} \\ \hline
    clip parameter                               & $0.35$                & $0.32$                           \\
    value function clip parameter                & $43.88$                   & $9.5$               \\
    value function loss coefficient              & $1.97$                   & $1.10$            \\
    entropy coefficient                          & $1.09 \cdot 10^{-2}$      & $3.41$                         \\
    learning rate                                & $1.28 \cdot 10^{-4}$       & $1.59 \cdot 10^{-4}$          \\
    lambda                                       & $0.86$                   & $0.89$            \\
    hidden neurons                               & $[256, 256]$              & $[128,128]$      \\
    mini-batch size                              & $121$                       & $107$          \\
    gradient descent iterations       & $32$                       & $8$           \\
    batch size                                   & $875$                    & $3864$          \\ \hline
    \end{tabular}
    \caption{PPO with discrete masking configuration}
    \label{tab:ppo-hp-search-spaces}
    \end{center}
    \end{table}

    \begin{table}[!h]
    \footnotesize
    \begin{center}
    \begin{tabular}{llll}
    \hline
    \multicolumn{1}{c}{\textbf{Hyperparameter}} & \multicolumn{1}{c}{\textbf{Without Obstacles}} & \multicolumn{1}{c}{\textbf{With Obstacles}} \\ \hline
    policy delay                                 & $2$                              & $1$    \\
    target policy noise                          & $0.38$                             & $0.20$  \\
    target noise clip factor                     & $0.28$                             & $0.19$  \\
    hidden neurons actor                         & $[256]$          & $[256]$ \\
    hidden neurons critic                        & $[512]$          & $[512]$ \\
    learning rate actor                          & $6.30 \cdot 10^{-5}$             & $6.70 \cdot 10^{-5}$   \\
    learning rate critic                         & $8.49 \cdot 10^{-4}$             & $2.34 \cdot 10^{-4}$   \\
    tau                                          & $2.13 \cdot 10^{-3}$             & $4.38 \cdot 10^{-3}$   \\
    l2 regularization                            & $5.99 \cdot 10^{-5}$                      & $5.59 \cdot 10^{-4}$  \\
    initial noise scale                          & $16.25$                            & $9.59$  \\
    final noise scale                            & $0.56$                           & $0.40$  \\
    noise reduction time steps                   & $37184$                         & $3073$   \\
    replay buffer capacity                       & $21836$                         & $66258$   \\
    prioritized replay alpha                     & $0.67$                            & $0.55$   \\
    prioritized replay beta                      & $0.47$                            & $0.37$   \\
    target update frequency                      & $2$                                & $1$   \\
    batch size                                   & $420$                            & $438$   \\ \hline
    \end{tabular}
    \caption{TD3 configuration}
    \label{tab:td3-hp-search-spaces}
    \end{center}
    \end{table}

    \newpage
    \begin{table}[!h]
    \footnotesize
    \begin{center}
    \begin{tabular}{llll}
    \hline
    \multicolumn{1}{c}{\textbf{Hyperparameter}} & \multicolumn{1}{c}{\textbf{Without Obstacles}} & \multicolumn{1}{c}{\textbf{With Obstacles}} \\ \hline
    policy delay                                 & $1$                              & $1$    \\
    target policy noise                          & $0.09$                             & $0.40$  \\
    target noise clip factor                     & $0.32$                             & $0.25$  \\
    hidden neurons actor                         & $[256, 512]$          & $[512, 256]$ \\
    hidden neurons critic                        & $[128, 128]$          & $[512]$ \\
    learning rate actor                          & $3.54 \cdot 10^{-6}$             & $2.59 \cdot 10^{-6}$   \\
    learning rate critic                         & $4.39 \cdot 10^{-4}$             & $9.85 \cdot 10^{-5}$   \\
    tau                                          & $7.31 \cdot 10^{-4}$             & $1.52 \cdot 10^{-3}$   \\
    l2 regularization                            & $4.38 \cdot 10^{-4}$                      & $8.31 \cdot 10^{-4}$  \\
    initial noise scale                          & $25.24$                            & $15.89$  \\
    final noise scale                            & $1.30$                           & $1.67$  \\
    noise reduction time steps                   & $36192$                         & $7248$   \\
    replay buffer capacity                       & $9337$                         & $29410$   \\
    prioritized replay alpha                     & $0.31$                            & $0.60$   \\
    prioritized replay beta                      & $0.45$                            & $0.25$   \\
    target update frequency                      & $3$                                & $1$   \\
    batch size                                   & $473$                            & $318$   \\ \hline
    \end{tabular}
    \caption{TD3 with projection configuration}
    \label{tab:td3-hp-search-spaces}
    \end{center}
    \end{table}

    \begin{table}[!h]
    \footnotesize
    \begin{center}
    \begin{tabular}{llll}
    \hline
    \multicolumn{1}{c}{\textbf{Hyperparameter}} & \multicolumn{1}{c}{\textbf{Without Obstacles}} & \multicolumn{1}{c}{\textbf{With Obstacles}} \\ \hline
    policy delay                                 & $3$                              & $1$    \\
    target policy noise                          & $0.27$                             & $0.39$  \\
    target noise clip factor                     & $0.28$                             & $0.30$  \\
    hidden neurons actor                         & $[256, 256]$          & $[128, 128]$ \\
    hidden neurons critic                        & $[256]$          & $[128]$ \\
    learning rate actor                          & $8.18 \cdot 10^{-6}$             & $6.69 \cdot 10^{-5}$   \\
    learning rate critic                         & $6.46 \cdot 10^{-4}$             & $8.89 \cdot 10^{-4}$   \\
    tau                                          & $9.10 \cdot 10^{-4}$             & $1.91 \cdot 10^{-3}$   \\
    l2 regularization                            & $1.81 \cdot 10^{-5}$                      & $5.83 \cdot 10^{-4}$  \\
    initial noise scale                          & $13.29$                            & $27.32$  \\
    final noise scale                            & $0.82$                           & $2.00$  \\
    noise reduction time steps                   & $35129$                         & $37168$   \\
    replay buffer capacity                       & $56739$                         & $43248$   \\
    prioritized replay alpha                     & $0.49$                            & $0.37$   \\
    prioritized replay beta                      & $0.31$                            & $0.42$   \\
    target update frequency                      & $1$                                & $2$   \\
    batch size                                   & $499$                            & $478$   \\ \hline
    \end{tabular}
    \caption{TD3 with continuous masking configuration}
    \label{tab:td3-hp-search-spaces}
    \end{center}
    \end{table}

    \newpage
    \begin{table}[!h]
    \footnotesize
    \begin{center}
    \begin{tabular}{llll}
    \hline
    \multicolumn{1}{c}{\textbf{Hyperparameter}} & \multicolumn{1}{c}{\textbf{Without Obstacles}} & \multicolumn{1}{c}{\textbf{With Obstacles}} \\ \hline
    policy delay                                 & $1$                              & $1$    \\
    target policy noise                          & $0.08$                             & $0.21$  \\
    target noise clip factor                     & $0.36$                             & $0.26$  \\
    hidden neurons actor                         & $[256]$          & $[256, 512]$ \\
    hidden neurons critic                        & $[256, 512]$          & $[512]$ \\
    learning rate actor                          & $3.31 \cdot 10^{-5}$             & $1.03 \cdot 10^{-5}$   \\
    learning rate critic                         & $2.15 \cdot 10^{-4}$             & $6.81 \cdot 10^{-4}$   \\
    tau                                          & $3.19 \cdot 10^{-4}$             & $2.69 \cdot 10^{-3}$   \\
    l2 regularization                            & $7.43 \cdot 10^{-4}$                      & $2.64 \cdot 10^{-4}$  \\
    initial noise scale                          & $5.47$                            & $17.26$  \\
    final noise scale                            & $2.40$                           & $1.70$  \\
    noise reduction time steps                   & $17759$                         & $16524$   \\
    replay buffer capacity                       & $50433$                         & $32792$   \\
    prioritized replay alpha                     & $0.46$                            & $0.34$   \\
    prioritized replay beta                      & $0.42$                            & $0.37$   \\
    target update frequency                      & $2$                                & $2$   \\
    batch size                                   & $384$                            & $496$   \\ \hline
    \end{tabular}
    \caption{TD3 with continuous masking configuration}
    \label{tab:td3-hp-search-spaces}
    \end{center}
    \end{table}

    \begin{table}[!h]
    \footnotesize
    \begin{center}
    \begin{tabular}{lll}
    \hline
    \multicolumn{1}{c}{\textbf{Hyperparameter}} & \multicolumn{1}{c}{\textbf{Without Obstacles}} & \multicolumn{1}{c}{\textbf{With Obstacles}} \\ \hline
    learning rate                                & $5.67 \cdot 10^{-4}$        & $2.39 \cdot 10^{-4}$          \\
    sigma                                        & $0.88$                      & $1.76$         \\
    hidden neurons                               & $[256, 256]$              & $[256, 256]$      \\
    categorical distribution temperature         & $1.0$                        & $1.00$       \\
    initial epsilon                              & $0.8$                        & $0.99$       \\
    final epsilon                                & $0.01$                     & $0.05$        \\
    epsilon reduction time steps                 & $50000$                   & $14998$         \\
    replay buffer capacity                       & $50000$                   & $49766$         \\
    prioritized replay alpha                     & $0.60$                      & $0.62$         \\
    prioritized replay beta                      & $0.40$                      & $0.42$         \\
    batch size                                   & $399$                       & $211$        \\ \hline
    \end{tabular}
    \caption{DQN with discrete masking configuration}
    \label{tab:dqn-hp-search-spaces}
    \end{center}
    \end{table}

    \newpage
    \begin{table}[!h]
    \footnotesize
    \begin{center}
    \begin{tabular}{lll}
    \hline
    \multicolumn{1}{c}{\textbf{Hyperparameter}} & \multicolumn{1}{c}{\textbf{Without Obstacles}} & \multicolumn{1}{c}{\textbf{With Obstacles}} \\ \hline
    learning rate                                & $9.46 \cdot 10^{-4}$        & $3.45 \cdot 10^{-5}$          \\
    sigma                                        & $1.81$                      & $0.91$         \\
    hidden neurons                               & $[256]$              & $[64]$      \\
    categorical distribution temperature         & $1.51$                        & $0.57$       \\
    initial epsilon                              & $0.43$                        & $0.65$       \\
    final epsilon                                & $6.04 \cdot 10^{-3}$                     & $0.01$        \\
    epsilon reduction time steps                 & $27070$                   & $7157$         \\
    replay buffer capacity                       & $2398$                   & $19299$         \\
    prioritized replay alpha                     & $0.53$                      & $0.55$         \\
    prioritized replay beta                      & $0.36$                      & $0.30$         \\
    batch size                                   & $237$                       & $148$        \\ 
    parameter network hidden neurons       & $[256]$      & $[512, 56]$   \\
    parameter network learning rate              & $8.07 \cdot 10^{-4}$     & $1.36 \cdot 10^{-5}$              \\ 
    dueling                                      & $True$ & $True$                        \\ 
    double q learning                            & $True$  & $True$                       \\ \hline
    \end{tabular}
    \caption{PAM configuration}
    \label{tab:dqn-hp-search-spaces}
    \end{center}
    \end{table}

    \begin{table}[!h]
    \footnotesize
    \begin{center}
    \begin{tabular}{llll}
    \hline
    \multicolumn{1}{c}{\textbf{Hyperparameter}} & \multicolumn{1}{c}{\textbf{Without Obstacles}} & \multicolumn{1}{c}{\textbf{With Obstacles}} \\ \hline
    policy delay                                 & $2$                              & $1$    \\
    target policy noise                          & $0.05$                             & $0.32$  \\
    target noise clip factor                     & $0.21$                             & $0.24$  \\
    hidden neurons actor                         & $[256, 128]$          & $[128, 128]$ \\
    hidden neurons critic                        & $[256]$          & $[512, 256]$ \\
    learning rate actor                          & $2.70 \cdot 10^{-5}$             & $5.29 \cdot 10^{-5}$   \\
    learning rate critic                         & $8.98 \cdot 10^{-4}$             & $6.47 \cdot 10^{-4}$   \\
    tau                                          & $3.19 \cdot 10^{-4}$             & $2879 \cdot 10^{-3}$   \\
    l2 regularization                            & $2.44 \cdot 10^{-5}$                      & $3.21 \cdot 10^{-4}$  \\
    initial epsilon                              & $0.2$                   & $0.27$           \\
    final epsilon                                & $0.01$    &                $0.01$         \\
    epsilon reduction time steps                 & $15000$                & $26999$            \\
    initial noise scale                          & $8.75$                            & $18.55$  \\
    final noise scale                            & $0.12$                           & $1.59$  \\
    noise reduction time steps                   & $15792$                         & $27876$   \\
    replay buffer capacity                       & $32410$                         & $45168$   \\
    prioritized replay alpha                     & $0.61$                            & $0.30$   \\
    prioritized replay beta                      & $0.32$                            & $0.44$   \\
    target update frequency                      & $2$                                & $1$   \\
    batch size                                   & $470$                            & $101$   \\ \hline
    \end{tabular}
    \caption{MPS-TD3 configuration}
    \label{tab:td3-hp-search-spaces}
    \end{center}
    \end{table}

\chapter{Results}
The following appendix describes first the results for the agents trained without obstacles. Afterward, we continue when learning included restrictions. Note that all numbers were rounded to two decimal points.

\section{Learning Without Obstacles}

\subsection{Training}
\label{app:exp1_0}

\begin{figure}[h]
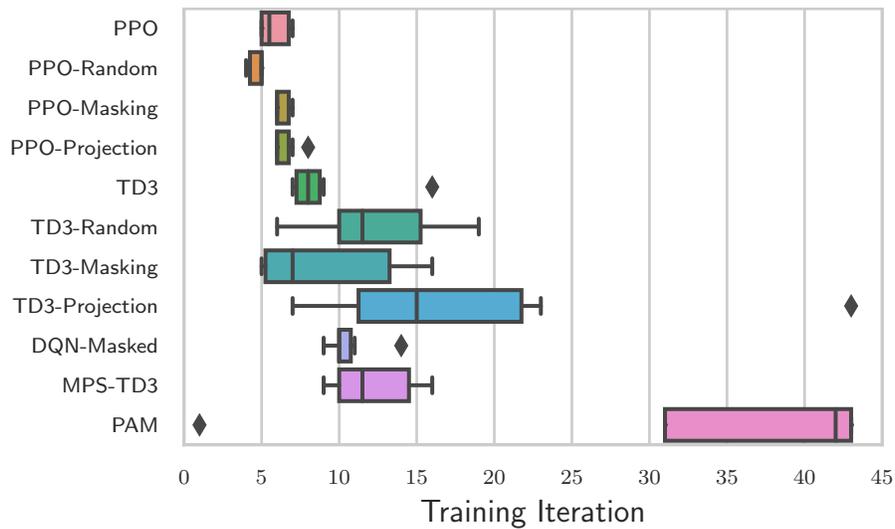

\centering
  \centering
  \importpgf{experiment_1/training}{boxplots.pgf}
\caption{Iterations until convergence (80\% solved episodes on average)}
\label{app:exp1_0_boxplots}
\end{figure}

\begin{figure}[!h]
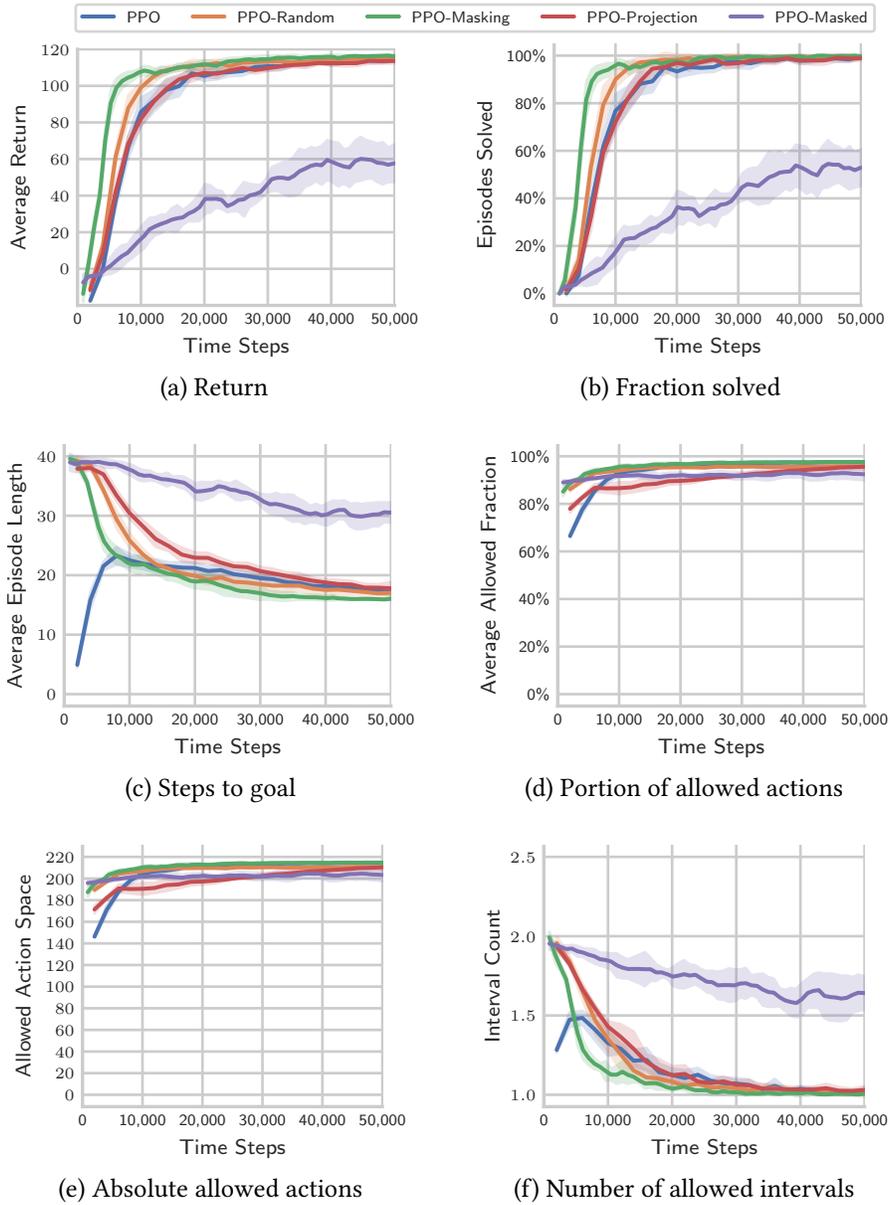

\centering
\begin{subfigure}{0.8\linewidth}
  \centering
  \importpgf{experiment_1/training/ppo}{legend.pgf}
\end{subfigure}
\begin{subfigure}{.49\linewidth}
  \centering
    \importpgf{experiment_1/training/ppo}{return_ppo.pgf}
  \caption{Return}
\end{subfigure}%
\begin{subfigure}{.49\linewidth}
  \centering
    \importpgf{experiment_1/training/ppo}{solved_ppo.pgf}
  \caption{Fraction solved}
\end{subfigure}
\\[3ex]
\begin{subfigure}{.49\linewidth}
  \centering
    \importpgf{experiment_1/training/ppo}{steps_ppo.pgf}
  \caption{Steps to goal}
  \label{app:exp1_0_episode_length_ppo}
\end{subfigure}
\begin{subfigure}{.49\linewidth}
  \centering
    \importpgf{experiment_1/training/ppo}{fraction_allowed_ppo.pgf}
  \caption{Portion of allowed actions}
\end{subfigure}
\\[3ex]
\begin{subfigure}{.49\linewidth}
  \centering
    \importpgf{experiment_1/training/ppo}{allowed_ppo.pgf}
  \caption{Absolute allowed actions}
\end{subfigure}
\begin{subfigure}{.49\linewidth}
  \centering
    \importpgf{experiment_1/training/ppo}{intervals_ppo.pgf}
  \caption{Number of allowed intervals}
  \label{app:exp1_0_count_ppo}
\end{subfigure}
\caption{On-policy training progress}
\label{app:exp1_0_ppo}
\end{figure}

\begin{figure}[!h]
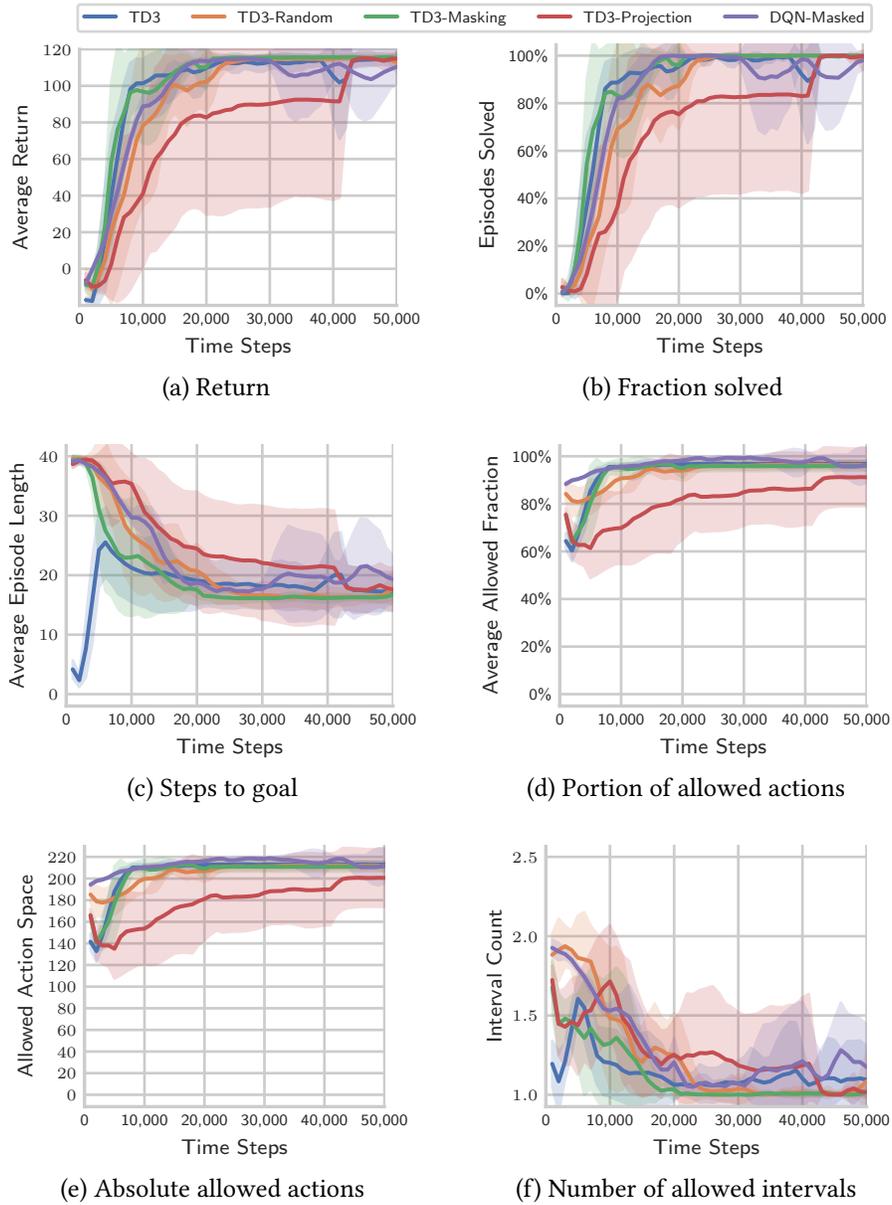

\centering
\begin{subfigure}{0.8\linewidth}
  \centering
   \importpgf{experiment_1/training/ddpg}{legend.pgf}
\end{subfigure}
\begin{subfigure}{.49\linewidth}
  \centering
  \importpgf{experiment_1/training/ddpg}{return_ddpg.pgf}
  \caption{Return}
\end{subfigure}%
\begin{subfigure}{.49\linewidth}
  \centering
  \importpgf{experiment_1/training/ddpg}{solved_ddpg.pgf}
  \caption{Fraction solved}
\end{subfigure}
\\[3ex]
\begin{subfigure}{.49\linewidth}
  \centering
  \importpgf{experiment_1/training/ddpg}{steps_ddpg.pgf}
  \caption{Steps to goal}
  \label{app:exp1_0_episode_length_td3}
\end{subfigure}
\begin{subfigure}{.49\linewidth}
  \centering
  \importpgf{experiment_1/training/ddpg}{fraction_allowed_ddpg.pgf}
  \caption{Portion of allowed actions}
\end{subfigure}
\\[3ex]
\begin{subfigure}{.49\linewidth}
  \centering
  \importpgf{experiment_1/training/ddpg}{allowed_ddpg.pgf}
  \caption{Absolute allowed actions}
  \label{app:exp1_0_allowed_action_space_td3}
\end{subfigure}
\begin{subfigure}{.49\linewidth}
  \centering
  \importpgf{experiment_1/training/ddpg}{intervals_ddpg.pgf}
  \caption{Number of allowed intervals}
  \label{app:exp1_0_count_td3}
\end{subfigure}
\caption{Off-policy training progress}
\label{app:exp1_0_td3}
\end{figure}

\begin{figure}[!h]
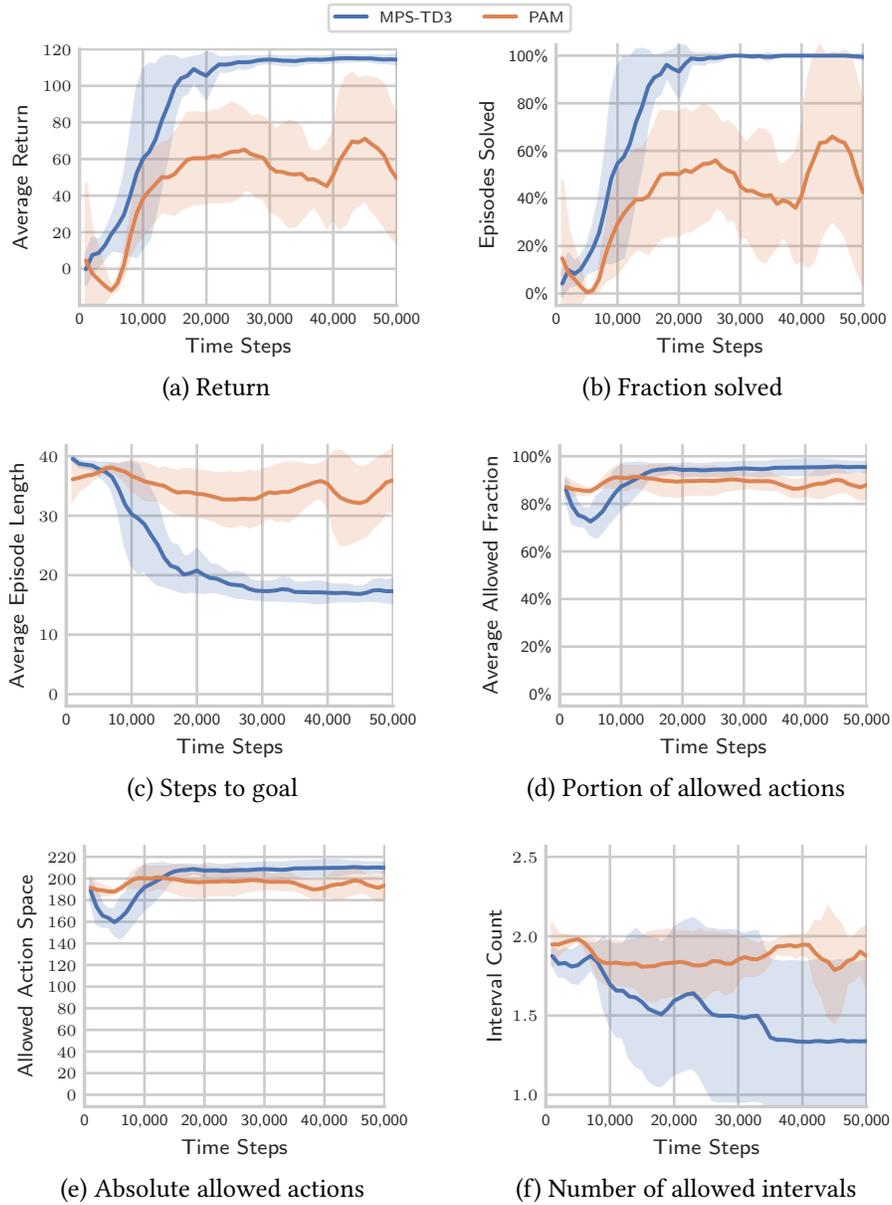

\centering
\begin{subfigure}{0.8\linewidth}
  \centering
  \importpgf{experiment_1/training/own}{legend.pgf}
\end{subfigure}
\begin{subfigure}{.49\linewidth}
  \centering
  \importpgf{experiment_1/training/own}{return_own.pgf}
  \caption{Return}
\end{subfigure}%
\begin{subfigure}{.49\linewidth}
  \centering
  \importpgf{experiment_1/training/own}{solved_own.pgf}
  \caption{Fraction solved}
\end{subfigure}
\\[3ex]
\begin{subfigure}{.49\linewidth}
  \centering
  \importpgf{experiment_1/training/own}{steps_own.pgf}
  \caption{Steps to goal}
  \label{app:exp1_0_episode_length_own}
\end{subfigure}
\begin{subfigure}{.49\linewidth}
  \centering
  \importpgf{experiment_1/training/own}{fraction_allowed_own.pgf}
  \caption{Portion of allowed actions}
\end{subfigure}
\\[3ex]
\begin{subfigure}{.49\linewidth}
  \centering
  \importpgf{experiment_1/training/own}{allowed_own.pgf}
  \caption{Absolute allowed actions}
\end{subfigure}
\begin{subfigure}{.49\linewidth}
  \centering
  \importpgf{experiment_1/training/own}{intervals_own.pgf}
  \caption{Number of allowed intervals}
  \label{app:exp1_0_count_own}
\end{subfigure}
\caption{MPS-TD3 and PAM training progress}
\label{app:exp1_0_own}
\end{figure}

\clearpage
\subsection{Evaluation Without Restrictions}
\label{app:exp1_1}

\begin{table}[!h]
\scriptsize
\centering
\begin{tabular}{llll}
\hline
\multicolumn{1}{c}{\textbf{Approach}} & \multicolumn{1}{c}{\textbf{Return}} & \multicolumn{1}{c}{\textbf{Steps}} & \multicolumn{1}{c}{\textbf{Solved}} \\ \hline

TD3                                   & $114.57 \pm 3.67$                     & $17.33 \pm 2.88$                     & $100.00\% \pm 0.00\%$                   \\
TD3-Projection                        & $91.46 \pm 55.30$                     & $17.60 \pm 4.22$                     & $83.33\% \pm 40.82\%$                  \\
TD3-Masking                           & $114.27 \pm6.21$                      & $18.83 \pm 5.53$                     & $100.00\% \pm 0.00\%$                   \\
TD3-Random                            & $114.13 \pm 4.54$                     & $18.17 \pm 3.87$                     & $100.00\% \pm 0.00\%$                   \\
DQN-Masked                            & $113.80 \pm 4.83$                     & $18.83 \pm 4.22$                     & $100.00\% \pm 0.00\%$                   \\ \hline

PPO                                   & $115.16 \pm 0.13$                     & $15.00 \pm 0.00$                     & $100.00\% \pm 0.00\%$                   \\
PPO-Projection                        & $115.12 \pm 1.03$                     & $15.33 \pm 0.52$                     & $100.00\% \pm 0.00\%$                  \\
PPO-Masking                           & $114.81 \pm 0.38$                     & $15.00 \pm 0.00$                     & $100.00\% \pm 0.00\%$                   \\
PPO-Random                            & $114.87 \pm 0.50$                     & $15.00 \pm 0.00$                     & $100.00\% \pm 0.00\%$                   \\
PPO-Masked                            & $73.06 \pm 64.29$                     & $18.00 \pm 1.83$                    & $66.67\% \pm 51.64\%$                 \\ \hline

MPS-TD3                               & $115.53 \pm 1.70$                     & $16.83 \pm 1.47$                     & $100.00\% \pm 0.00\%$                   \\
PAM                                   & $11.88 \pm 41.37$                     & $38.00 \pm 0.00$                     & $16.67\% \pm 40.8\%$                 \\ \hline
\end{tabular}
\caption{Average return, steps, and solved episodes}
\label{app:exp1_1_results}
\end{table}

\begin{table}[!h]
\scriptsize
\centering
\begin{tabular}{llll}
\hline
\multicolumn{1}{c}{\textbf{Approach}} & \multicolumn{1}{c}{\textbf{Intervals}} & \multicolumn{1}{c}{\textbf{Size}} & \multicolumn{1}{c}{\textbf{Fraction}} \\ \hline

TD3                                   & $1.01 \pm 0.02$                        & $213.07 \pm 3.04$                     & $96.85\% \pm 1.38\%$                    \\
TD3-Projection                        & $1.00 \pm 0.01$                        & $190.54 \pm 31.62$                    & $86.61\% \pm 14.37\%$                   \\
TD3-Masking                           & $1.00 \pm 0.00$                        & $212.30 \pm 3.22$                     & $96.50\% \pm 1.47\%$                    \\
TD3-Random                            & $1.01 \pm 0.03$                        & $211.58 \pm 5.93$                     & $96.17\% \pm 2.69\%$                    \\
DQN-Masked                            & $1.01 \pm 0.02$                        & $209.27 \pm 11.66$                    & $95.12\% \pm 5.30\%$                    \\ \hline

PPO                                   & $1.00 \pm 0.00$                        & $214.65 \pm 0.00$                     & $97.57\% \pm 0.00\%$                    \\
PPO-Projection                        & $1.00 \pm 0.00$                        & $213.28 \pm 3.17$                     & $96.94\% \pm 1.44\%$                    \\
PPO-Masking                           & $1.00 \pm 0.00$                        & $214.65 \pm 0.00$                     & $97.57\% \pm 0.00\%$                    \\
PPO-Random                            & $1.00 \pm 0.00$                         & $213.40 \pm 1.97$                     & $97.00\% \pm 0.89\%$                    \\
PPO-Masked                            & $1.03 \pm 0.03$                        & $200.05 \pm 16.24$                    & $90.93\% \pm 7.38\%$                    \\ \hline

MPS-TD3                               & $1.04 \pm 0.06$                        & $209.99 \pm 4.59$                     & $95.45\% \pm 2.09\%$                    \\
PAM                                   & $1.08 \pm 0.08$                        & $189.50 \pm 14.72$                    & $86.14\% \pm 6.69\%$                    \\ \hline
\end{tabular}
\caption{Average number of disjoint intervals and size of the action space}
\label{app:exp1_1_control}
\end{table}

\begin{table}[!h]
\scriptsize
\centering
\begin{tabular}{lllll}
\hline
\multicolumn{1}{c}{\textbf{Approach}} & \multicolumn{1}{c}{\textbf{Average}} & \multicolumn{1}{c}{\textbf{Minimum}} & \multicolumn{1}{c}{\textbf{Maximum}} & \multicolumn{1}{c}{\textbf{Variance}} \\ \hline

TD3 & $212.34 \pm 4.14$ & $211.80 \pm 5.21$ & $212.88 \pm 3.27$ & $40.70 \pm 99.68$ \\
TD3-Projection & $190.38 \pm 31.84$ & $190.27 \pm 32 .00$ & $190.49 \pm 31.69$ & $2.87 \pm 7.03$ \\
TD3-Masking & $212.30 \pm 3.22$ & $212.30 \pm 3.22$ & $212.30 \pm 3.22$ & $0.00 \pm 0.00$ \\
TD3-Random & $211.00 \pm 7.18$ & $210.92 \pm 7.35$ & $211.07 \pm 7.01$ & $0.76 \pm 1.87$ \\
DQN-Masked & $208.98 \pm 12.35$ & $208.71 \pm 13.00$ & $209.26 \pm 11.70$ & $11.58 \pm 28.36$ \\ \hline

PPO & $214.65 \pm 0.00$ & $214.65 \pm 0.00$ & $214.65 \pm 0.00$ & $0.00 \pm 0.00$ \\
PPO-Projection & $213.28 \pm 0.00$ & $213.28 \pm 0.00$ & $213.28 \pm 0.00$ & $0.00 \pm 0.00$ \\
PPO-Masking & $214.65 \pm 0.00$ & $214.65 \pm 0.00$ & $214.65 \pm 0.00$ & $0.00 \pm 0.00$ \\
PPO-Random & $213.40 \pm 1.97$ & $213.40 \pm 1.97$ & $213.40 \pm 1.97$ & $0.00  \pm 0.00$ \\
PPO-Masked & $198.85 \pm 17.41$ & $198.10 \pm 17.98$ & $199.60 \pm 16.86$ & $24.94 \pm 33.97$ \\ \hline

MPS-TD3 & $207.94 \pm 7.73$ & $206.64 \pm 9.74$ & $209.24 \pm 5.73$ & $48.36 \pm 75.40$ \\
PAM & $184.76 \pm 17.78$ & $182.38 \pm 19.85$ & $187.15 \pm 15.98$ & $112.01 \pm 158.66$ \\ \hline
\end{tabular}
\caption{Size of the individual intervals}
\label{app:exp1_1_single}
\end{table}

\begin{figure}[t]
\centering
\begin{subfigure}{0.8\linewidth}
  \centering
  \importpgf{experiment_1/training/ppo}{legend.pgf}
\end{subfigure}
\begin{subfigure}{.49\linewidth}
  \centering
   \importpgf{experiment_1/evaluation_0/ppo}{allowed.pgf}
  \caption{Allowed action space size}
\end{subfigure}%
\begin{subfigure}{.49\linewidth}
  \centering
  \importpgf{experiment_1/evaluation_0/ppo}{count.pgf}
  \caption{Number of intervals}
\end{subfigure}
\\[3ex]
\begin{subfigure}{.49\linewidth}
  \centering
  \importpgf{experiment_1/evaluation_0/ppo}{average.pgf}
  \caption{Average interval length}
\end{subfigure}
\begin{subfigure}{.49\linewidth}
  \centering
  \importpgf{experiment_1/evaluation_0/ppo}{minimum.pgf}
  \caption{Minimum interval length}
\end{subfigure}
\\[3ex]
\begin{subfigure}{.49\linewidth}
  \centering
  \importpgf{experiment_1/evaluation_0/ppo}{maximum.pgf}
  \caption{Maximum interval length}
\end{subfigure}
\begin{subfigure}{.49\linewidth}
  \centering
  \importpgf{experiment_1/evaluation_0/ppo}{variance.pgf}
  \caption{Interval length variance}
\end{subfigure}
\caption{On-policy control variables}
\label{app:exp1_1_control_ppo}
\end{figure}

\begin{figure}[t]
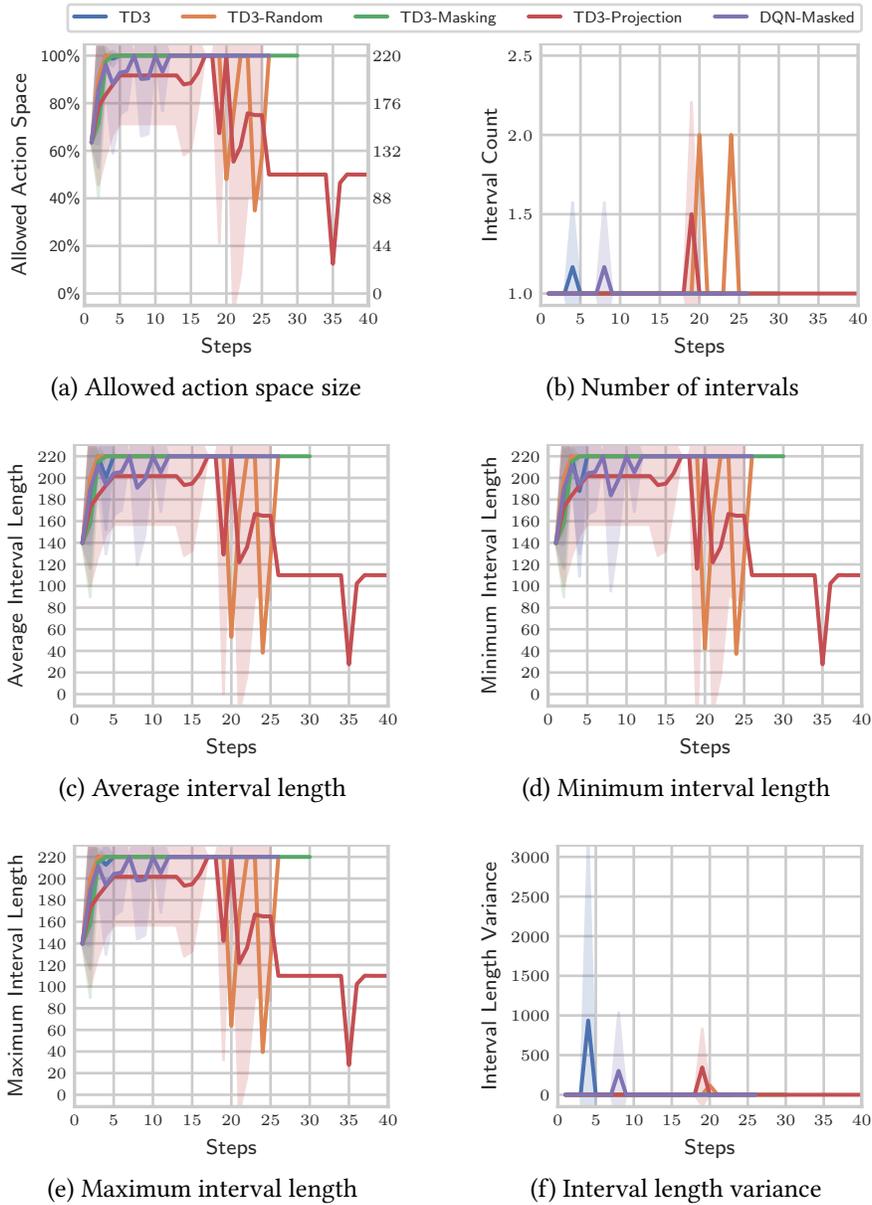

\centering
\begin{subfigure}{0.8\linewidth}
  \centering
  \importpgf{experiment_1/training/ddpg}{legend.pgf}
\end{subfigure}
\begin{subfigure}{.49\linewidth}
  \centering
   \importpgf{experiment_1/evaluation_0/td3}{allowed.pgf}
  \caption{Allowed action space size}
\end{subfigure}%
\begin{subfigure}{.49\linewidth}
  \centering
   \importpgf{experiment_1/evaluation_0/td3}{count.pgf}
  \caption{Number of intervals}
\end{subfigure}
\\[3ex]
\begin{subfigure}{.49\linewidth}
  \centering
   \importpgf{experiment_1/evaluation_0/td3}{average.pgf}
  \caption{Average interval length}
\end{subfigure}
\begin{subfigure}{.49\linewidth}
  \centering
   \importpgf{experiment_1/evaluation_0/td3}{minimum.pgf}
  \caption{Minimum interval length}
\end{subfigure}
\\[3ex]
\begin{subfigure}{.49\linewidth}
  \centering
   \importpgf{experiment_1/evaluation_0/td3}{maximum.pgf}
  \caption{Maximum interval length}
\end{subfigure}
\begin{subfigure}{.49\linewidth}
  \centering
   \importpgf{experiment_1/evaluation_0/td3}{variance.pgf}
  \caption{Interval length variance}
\end{subfigure}
\caption{Off-policy control variables}
\label{app:exp1_1_control_td3}
\end{figure}

\begin{figure}[t]
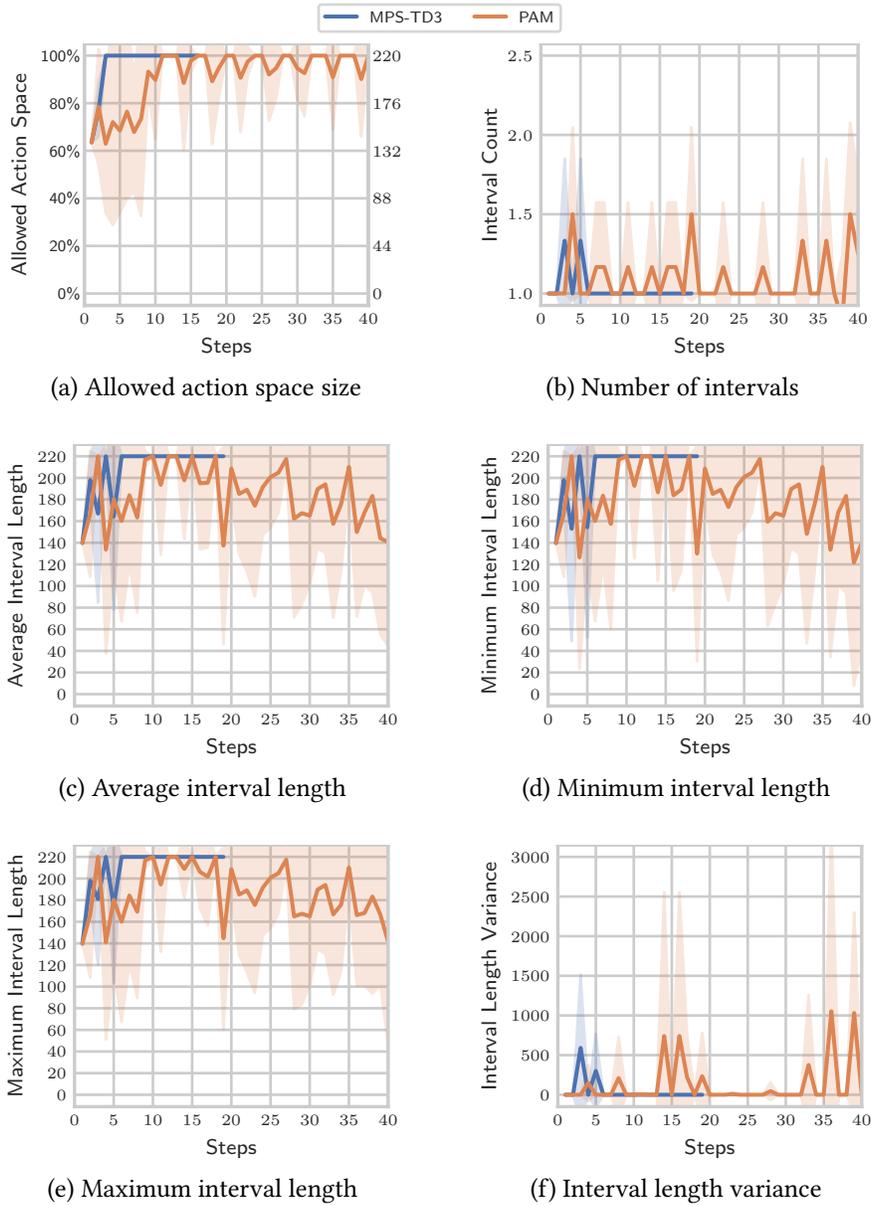

\centering
\begin{subfigure}{0.8\linewidth}
  \centering
   \importpgf{experiment_1/training/own}{legend.pgf}
\end{subfigure}
\begin{subfigure}{.49\linewidth}
  \centering
  \importpgf{experiment_1/evaluation_0/own}{allowed.pgf}
  \caption{Allowed action space size}
\end{subfigure}%
\begin{subfigure}{.49\linewidth}
  \centering
  \importpgf{experiment_1/evaluation_0/own}{count.pgf}
  \caption{Number of intervals}
\end{subfigure}
\\[3ex]
\begin{subfigure}{.49\linewidth}
  \centering
  \importpgf{experiment_1/evaluation_0/own}{average.pgf}
  \caption{Average interval length}
\end{subfigure}
\begin{subfigure}{.49\linewidth}
  \centering
  \importpgf{experiment_1/evaluation_0/own}{minimum.pgf}
  \caption{Minimum interval length}
\end{subfigure}
\\[3ex]
\begin{subfigure}{.49\linewidth}
  \centering
  \importpgf{experiment_1/evaluation_0/own}{maximum.pgf}
  \caption{Maximum interval length}
\end{subfigure}
\begin{subfigure}{.49\linewidth}
  \centering
  \importpgf{experiment_1/evaluation_0/own}{variance.pgf}
  \caption{Interval length variance}
\end{subfigure}
\caption{MPS-TD3 and PAM control variables}
\label{app:exp1_1_control_own}
\end{figure}

\begin{figure}[h]
\centering
\begin{subfigure}[b]{\linewidth}
  \centering
   \importpgf{experiment_1/evaluation_0/reward_t}{on_policy_t_reward.pgf}
  \caption{On-policy}
\end{subfigure}
\\[3ex]
\begin{subfigure}[b]{\linewidth}
  \centering
   \importpgf{experiment_1/evaluation_0/reward_t}{off_policy_t_reward.pgf}
  \caption{Off-policy}
\end{subfigure}
\caption{Welch's t-test p-values between average returns}
\label{app:exp1_1_test_reward}
\end{figure}

\begin{figure}[h]
\centering
\begin{subfigure}[b]{\linewidth}
  \centering
  \importpgf{experiment_1/evaluation_0/reward_t}{on_policy_t_steps.pgf}
  \caption{On-policy}
\end{subfigure}
\\[3ex]
\begin{subfigure}[b]{\linewidth}
  \centering
  \importpgf{experiment_1/evaluation_0/reward_t}{off_policy_t_steps.pgf}
  \caption{Off-policy}
\end{subfigure}
\caption{Welch's t-test p-values between average steps}
\label{app:exp1_1_test_steps}
\end{figure}

\clearpage
\subsection{Evaluation With Simple Restrictions}
\label{app:exp1_2}

\begin{figure}[!h]
\centering
  \centering
   \importpgf{experiment_1/evaluation}{eval_rewards.pgf}
\caption{Average Return}
\label{app:exp1_2_plot_erward}
\end{figure}

\begin{figure}[!h]
\centering
  \centering
  \importpgf{experiment_1/evaluation}{eval_solved.pgf}
\caption{Average fraction of solved environments}
\label{app:exp1_2_plot_solved}
\end{figure}

\begin{figure}[!h]
\centering
  \centering
  \importpgf{experiment_1/evaluation}{eval_steps.pgf}
\caption{Average episode length}
\label{app:exp1_2_plot_steps}
\end{figure}

\begin{table}[h]
\scriptsize
\centering
\begin{tabular}{llll}
\hline
\multicolumn{1}{c}{\textbf{Approach}} & \multicolumn{1}{c}{\textbf{Return}} & \multicolumn{1}{c}{\textbf{Steps}} & \multicolumn{1}{c}{\textbf{Solved}} \\ \hline

TD3                                   & $49.22 \pm 36.39$                     & $26.32 \pm 7.41$                     & $41.25\% \pm 34.27\%$                             \\
TD3-Projection                        & $73.11 \pm 23.50$                     & $22.56 \pm 1.01$                     & $65.50\% \pm 19.48\%$                            \\
TD3-Masking                           & $89.33 \pm 15.48$                     & $24.70 \pm 3.12$                     & $79.58\% \pm 15.03\%$                            \\
TD3-Random                            & $78.09 \pm 25.43$                     & $27.23 \pm 4.30$                     & $69.58\% \pm 26.43\%$                            \\
DQN-Masked                            & \textbf{$111.75 \pm 4.72$}            & $20.52 \pm 4.40$                     & $100.00\% \pm 0.00\%$                             \\ \hline

PPO                                   & $56.33 \pm 32.84$                     & $20.45 \pm 1.12$                     & $45.83\% \pm 32.39\%$                            \\
PPO-Projection                        & $109.43 \pm 5.16$                    & $20.54 \pm 2.24$                     & $97.50\% \pm 5.00\%$                            \\
PPO-Masking                           & $97.91 \pm 19.31$                     & \textbf{$19.25 \pm 1.22$}            & $87.08\% \pm 15.36\%$                            \\
PPO-Random                            & $102.94 \pm 2.84$                    & $22.75 \pm 0.88$                     & $94.17\% \pm 1.29\%$                            \\
PPO-Masked                            & $49.40 \pm 20.85$                     & $22.58 \pm 2.19$                     & $47.08\% \pm 19.13\%$                            \\ \hline

MPS-TD3                               & $84.01 \pm 12.23$                     & $24.85 \pm 2.98$                     & $77.50\% \pm 12.65\%$                            \\
PAM                                   & $16.76 \pm 8.53$                     & $30.34 \pm 5.03$                     & $16.88\% \pm 3.15\%$                            \\ \hline
\end{tabular}
\caption{Average return, steps, and solved episodes}
\label{app:exp1_2_results}
\end{table}

\begin{table}[t]
\scriptsize
\centering
\begin{tabular}{llll}
\hline
\multicolumn{1}{c}{\textbf{Approach}} & \multicolumn{1}{c}{\textbf{Intervals}} & \multicolumn{1}{c}{\textbf{Size}} & \multicolumn{1}{c}{\textbf{Fraction}} \\ \hline

TD3                                   & $1.14 \pm 0.05$                        & $194.47 \pm 8.80$                    & $88.40\% \pm 4.00\%$                   \\
TD3-Projection                        & $1.10 \pm 0.04$                        & $185.50 \pm 19.24$                    & $84.32\% \pm 8.75\%$                   \\
TD3-Masking                           & $1.11 \pm 0.03$                        & $191.11 \pm 6.29$                    & $86.87\% \pm 2.86\%$                   \\
TD3-Random                            & $1.13 \pm 0.04$                        & $189.21 \pm 5.14$                    & $86.00\% \pm 2.34\%$                   \\
DQN-Masked                            & $1.06 \pm 0.02$                        & $180.19	 \pm 6.55$                    & $81.90\% \pm 2.98\%$                   \\ \hline

PPO                                   & $1.14 \pm 0.02$                        & $189.32 \pm 2.34$                    & $86.05\% \pm 1.06\%$                   \\
PPO-Projection                        & $1.09 \pm 0.03$                        & $180.38 \pm 5.91$                    & $81.99\% \pm 2.69\%$                   \\
PPO-Masking                           & $1.10 \pm 0.02$                        & $183.49 \pm 6.77$                    & $83.40\% \pm 3.08\%$                   \\
PPO-Random                            & $1.12 \pm 0.02$                        & $182.43 \pm 1.10$                    & $82.92\% \pm 0.50\%$                   \\
PPO-Masked                            & $1.08 \pm 0.02$                        & $183.18 \pm 8.62$                    & $83.26\% \pm 3.92\%$                   \\ \hline

MPS-TD3                               & $1.14 \pm 0.03$                        & $187.90 \pm 5.11$                    & $85.41\% \pm 2.32\%$                   \\
PAM                                   & $1.11 \pm 0.03$                        & $183.57 \pm 6.09$                    & $83.44\% \pm 2.77\%$                   \\ \hline
\end{tabular}
\caption{Average number of disjoint intervals and size of the action space}
\label{exp1_2_control}
\end{table}

\begin{table}[h]
\scriptsize
\centering
\begin{tabular}{lllll}
\hline
\multicolumn{1}{c}{\textbf{Approach}} & \multicolumn{1}{c}{\textbf{Average}} & \multicolumn{1}{c}{\textbf{Minimum}} & \multicolumn{1}{c}{\textbf{Maximum}} & \multicolumn{1}{c}{\textbf{Variance}} \\ \hline

TD3 & $185.72 \pm 11.29$ & $180.56 \pm 12.50$ & $190.87 \pm 10.22$ & $270.04 \pm 99.67$ \\
TD3-Projection & $178.84 \pm 16.94$ & $174.97 \pm 15.42$ & $182.70 \pm 18.49$ & $188.68 \pm 88.10$ \\
TD3-Masking & $183.76 \pm 6.48$ & $179.37 \pm 6.88$ & $188.15 \pm 6.26$ & $223.50 \pm 59.92$ \\
TD3-Random & $180.78 \pm 6.86$ & $175.99 \pm 7.73$ & $185.57 \pm 6.03$ & $242.73 \pm 49.80$ \\
DQN-Masked & $176.45 \pm 6.90$ & $174.34 \pm 6.96$ & $178.57 \pm 6.90$ & $103.50 \pm 41.70$ \\ \hline

PPO & $180.56 \pm 2.61$ & $176.71 \pm 2.83$ & $184.41 \pm 2.82$ & $169.5 \pm 64.04$ \\
PPO-Projection & $174.61 \pm 4.67$ & $171.23 \pm 4.25$ & $177.99 \pm 5.29$ & $163.84 \pm 64.63$ \\
PPO-Masking & $177.55 \pm 7.54$ & $174.59 \pm 7.69$ & $180.50 \pm 7.40$ & $128.84 \pm 19.23$ \\
PPO-Random & $175.14 \pm 1.09$ & $171.75 \pm 1.30$ & $178.52 \pm 1.13$ & $150.37 \pm 22.16$ \\
PPO-Masked & $177.92 \pm 7.37$ & $175.07 \pm 6.81$ & $180.77 \pm 7.95$ & $135.52 \pm 41.77$ \\ \hline

MPS-TD3 & $179.17 \pm 6.03$ & $173.58 \pm 6.68$ & $184.77 \pm 5.61$ & $282.13 \pm 63.59$ \\
PAM & $176.92 \pm 7.11$ & $173.48 \pm 7.59$ & $180.35 \pm 6.70$ & $151.23 \pm 42.24$ \\ \hline
\end{tabular}
\caption{Size of the individual intervals}
\label{app:exp1_2_single}
\end{table}

\begin{figure}[t]
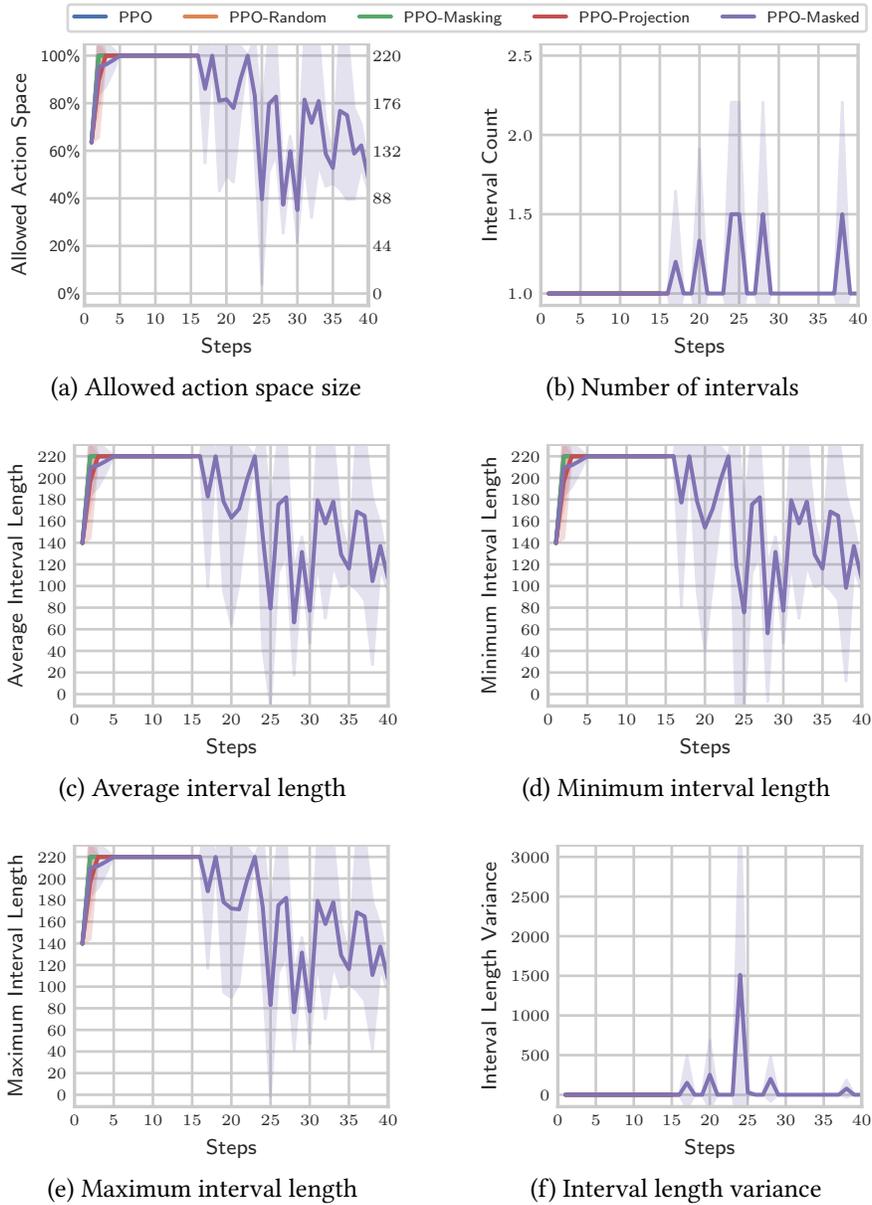

\centering
\begin{subfigure}{0.8\linewidth}
  \centering
  \importpgf{experiment_1/training/ppo}{legend.pgf}
\end{subfigure}
\begin{subfigure}{.49\linewidth}
  \centering
  \importpgf{experiment_1/evaluation/ppo}{allowed.pgf}
  \caption{Allowed action space size}
  \label{app:exp1_2_length_ppo}
\end{subfigure}%
\begin{subfigure}{.49\linewidth}
  \centering
  \importpgf{experiment_1/evaluation/ppo}{count.pgf}
  \caption{Number of intervals}
  \label{app:exp1_2_count_ppo}
\end{subfigure}
\\[3ex]
\begin{subfigure}{.49\linewidth}
  \centering
  \importpgf{experiment_1/evaluation/ppo}{average.pgf}
  \caption{Average interval length}
\end{subfigure}
\begin{subfigure}{.49\linewidth}
  \centering
  \importpgf{experiment_1/evaluation/ppo}{minimum.pgf}
  \caption{Minimum interval length}
\end{subfigure}
\\[3ex]
\begin{subfigure}{.49\linewidth}
  \centering
  \importpgf{experiment_1/evaluation/ppo}{maximum.pgf}
  \caption{Maximum interval length}
\end{subfigure}
\begin{subfigure}{.49\linewidth}
  \centering
  \importpgf{experiment_1/evaluation/ppo}{variance.pgf}
  \caption{Interval length variance}
\end{subfigure}
\caption{On-policy control variables}
\label{exp1_2_control_ppo}
\end{figure}

\begin{figure}[t]
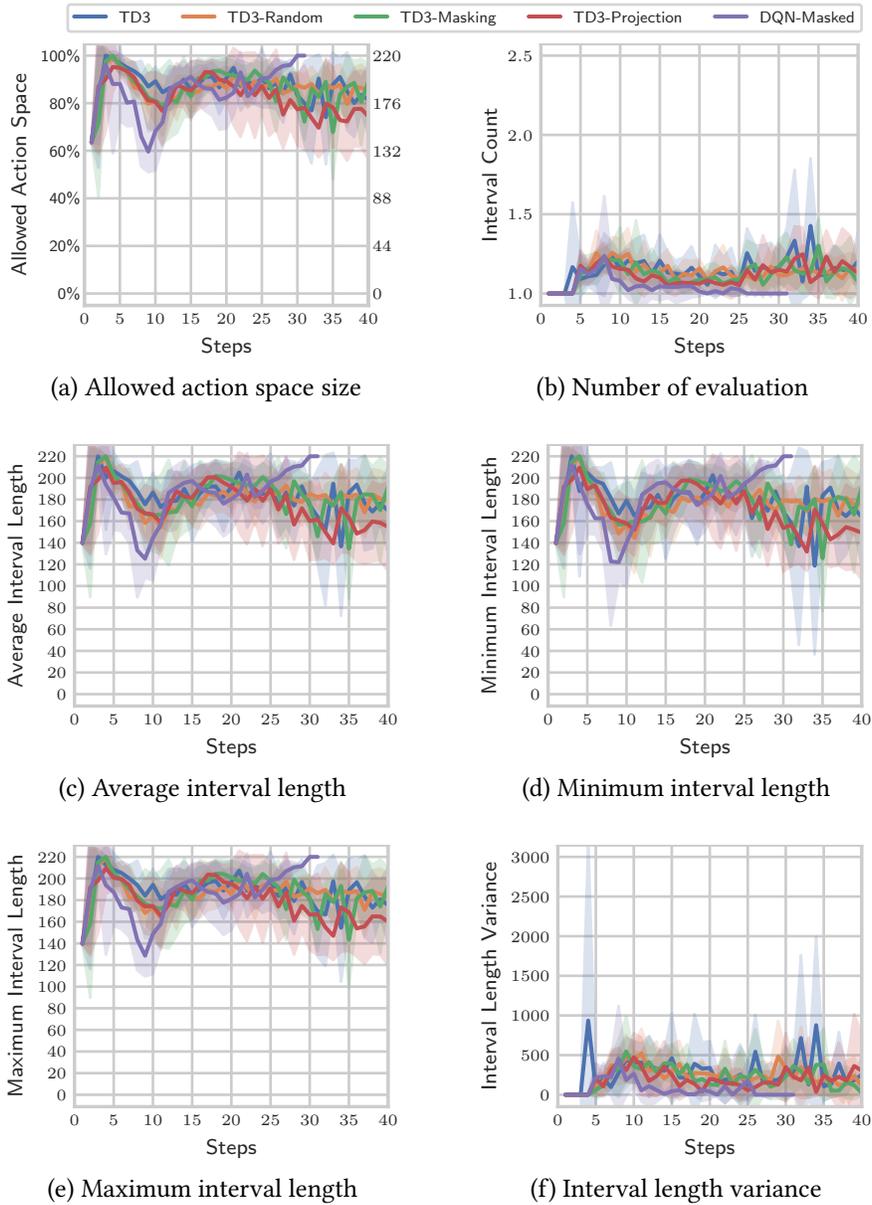

\centering
\begin{subfigure}{0.8\linewidth}
  \centering
  \importpgf{experiment_1/training/ddpg}{legend.pgf}
\end{subfigure}
\begin{subfigure}{.49\linewidth}
  \centering
   \importpgf{experiment_1/evaluation/td3}{allowed.pgf}
  \caption{Allowed action space size}
  \label{app:exp1_2_length_td3}
\end{subfigure}%
\begin{subfigure}{.49\linewidth}
  \centering
   \importpgf{experiment_1/evaluation/td3}{count.pgf}
  \caption{Number of evaluation}
  \label{app:exp1_2_count_td3}
\end{subfigure}
\\[3ex]
\begin{subfigure}{.49\linewidth}
  \centering
   \importpgf{experiment_1/evaluation/td3}{average.pgf}
  \caption{Average interval length}
\end{subfigure}
\begin{subfigure}{.49\linewidth}
  \centering
   \importpgf{experiment_1/evaluation/td3}{minimum.pgf}
  \caption{Minimum interval length}
\end{subfigure}
\\[3ex]
\begin{subfigure}{.49\linewidth}
  \centering
   \importpgf{experiment_1/evaluation/td3}{maximum.pgf}
  \caption{Maximum interval length}
\end{subfigure}
\begin{subfigure}{.49\linewidth}
  \centering
   \importpgf{experiment_1/evaluation/td3}{variance.pgf}
  \caption{Interval length variance}
\end{subfigure}
\caption{Off-policy control variables}
\label{exp1_2_control_td3}
\end{figure}

\begin{figure}[t]
\centering
\begin{subfigure}{0.8\linewidth}
  \centering
   \importpgf{experiment_1/training/own}{legend.pgf}
\end{subfigure}
\begin{subfigure}{.49\linewidth}
  \centering
  \importpgf{experiment_1/evaluation/own}{allowed.pgf}
  \caption{Allowed action space size}
  \label{app:exp1_2_length_own}
\end{subfigure}%
\begin{subfigure}{.49\linewidth}
  \centering
  \importpgf{experiment_1/evaluation/own}{count.pgf}
  \caption{Number of intervals}
  \label{app:exp1_2_count_own}
\end{subfigure}
\\[3ex]
\begin{subfigure}{.49\linewidth}
  \centering
  \importpgf{experiment_1/evaluation/own}{average.pgf}
  \caption{Average interval length}
\end{subfigure}
\begin{subfigure}{.49\linewidth}
  \centering
  \importpgf{experiment_1/evaluation/own}{minimum.pgf}
  \caption{Minimum interval length}
\end{subfigure}
\\[3ex]
\begin{subfigure}{.49\linewidth}
  \centering
  \importpgf{experiment_1/evaluation/own}{maximum.pgf}
  \caption{Maximum interval length}
\end{subfigure}
\begin{subfigure}{.49\linewidth}
  \centering
  \importpgf{experiment_1/evaluation/own}{variance.pgf}
  \caption{Interval length variance}
\end{subfigure}
\caption{MPS-TD3 and PAM control variables}
\label{app:exp1_2_control_own}
\end{figure}

\begin{figure}[t]
\centering
\begin{subfigure}{.3\linewidth}
  \centering
  \fbox{\includegraphics[width=3.5cm]{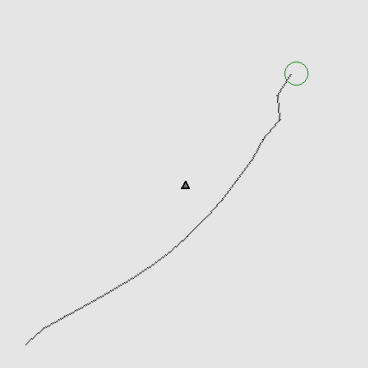}}
  \caption{Maximum reward}
\end{subfigure}%
\begin{subfigure}{.3\linewidth}
  \centering
  \fbox{\includegraphics[width=3.5cm]{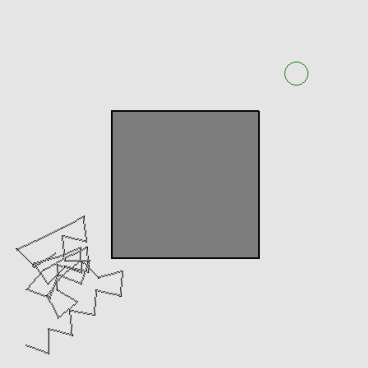}}
  \caption{Minimum reward}
\end{subfigure}
\begin{subfigure}{.3\linewidth}
  \centering
  \fbox{\includegraphics[width=3.5cm]{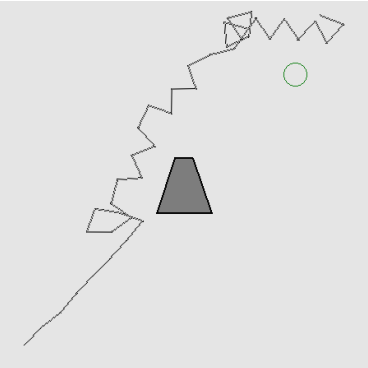}}
  \caption{Random}
\end{subfigure}
\caption{TD3 trajectory examples}
\label{app:exp1_2_trajectory_start}
\end{figure}

\begin{figure}[t]
\centering
\begin{subfigure}{.3\linewidth}
  \centering
  \fbox{\includegraphics[width=3.5cm]{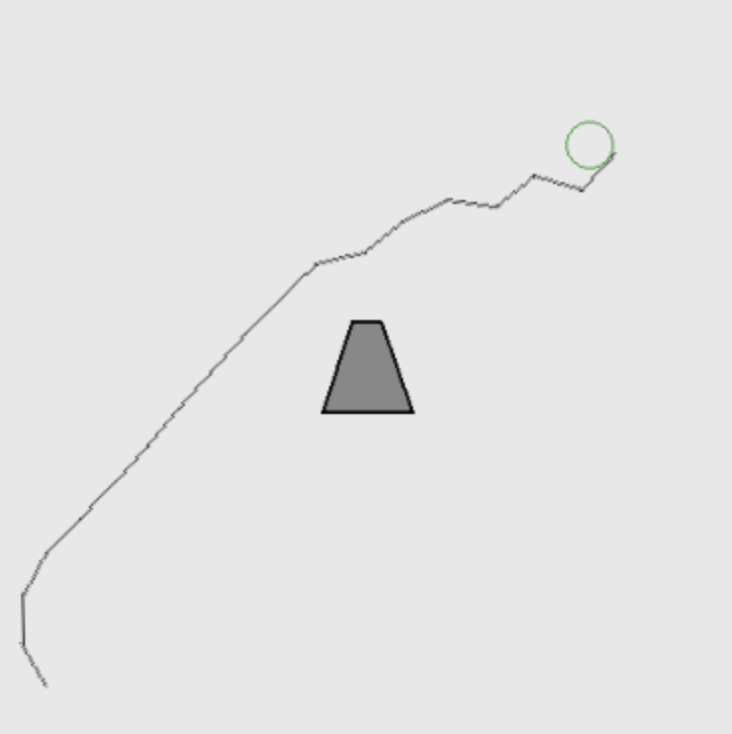}}
  \caption{Maximum reward}
\end{subfigure}%
\begin{subfigure}{.3\linewidth}
  \centering
  \fbox{\includegraphics[width=3.5cm]{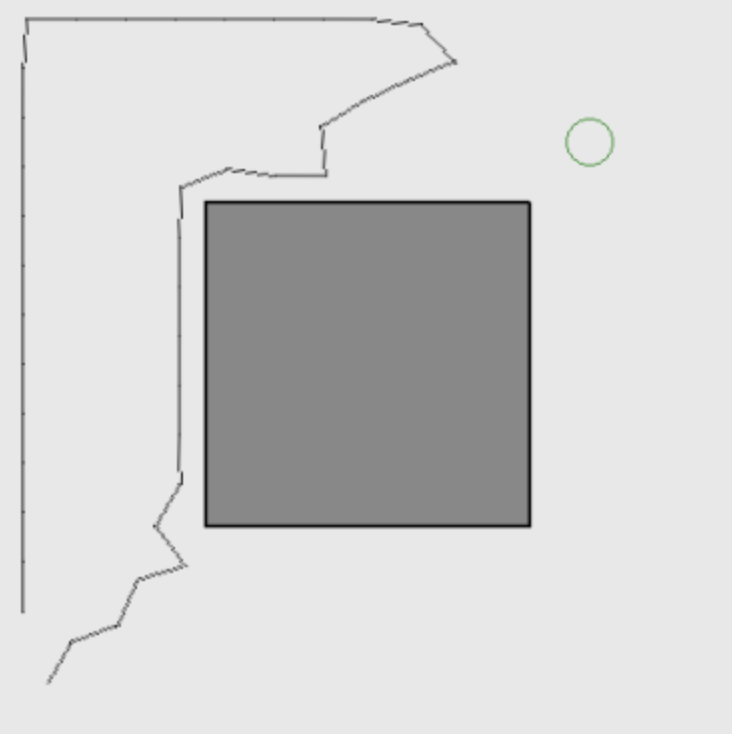}}
  \caption{Minimum reward}
\end{subfigure}
\begin{subfigure}{.3\linewidth}
  \centering
  \fbox{\includegraphics[width=3.5cm]{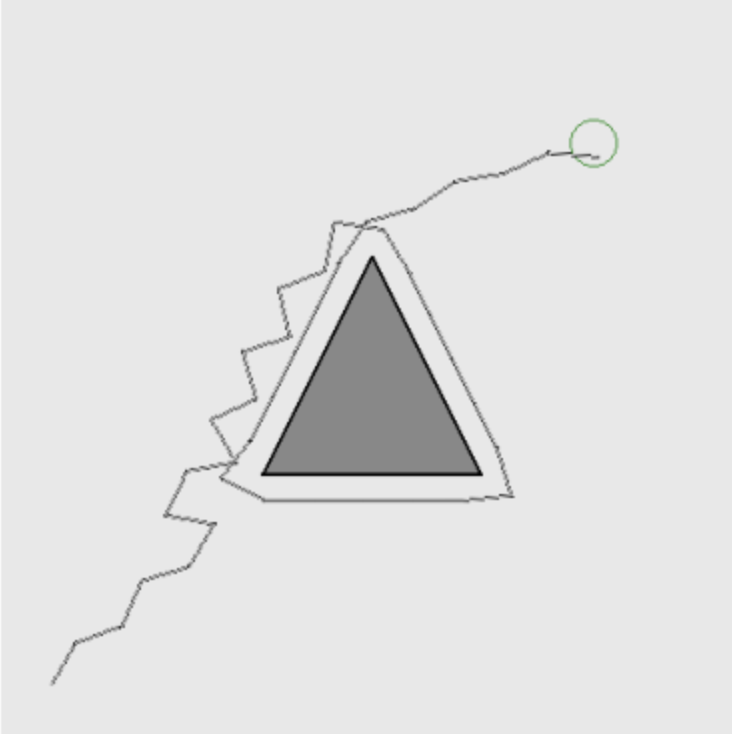}}
  \caption{Random}
\end{subfigure}
\caption{TD3 with projection trajectory examples}
\end{figure}

\begin{figure}[t]
\centering
\begin{subfigure}{.3\linewidth}
  \centering
  \fbox{\includegraphics[width=3.5cm]{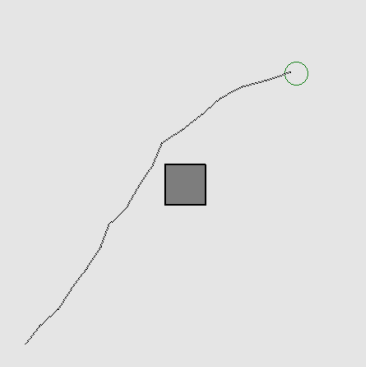}}
  \caption{Maximum reward}
\end{subfigure}%
\begin{subfigure}{.3\linewidth}
  \centering
  \fbox{\includegraphics[width=3.5cm]{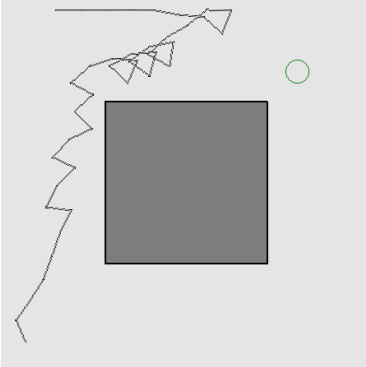}}
  \caption{Minimum reward}
\end{subfigure}
\begin{subfigure}{.3\linewidth}
  \centering
  \fbox{\includegraphics[width=3.5cm]{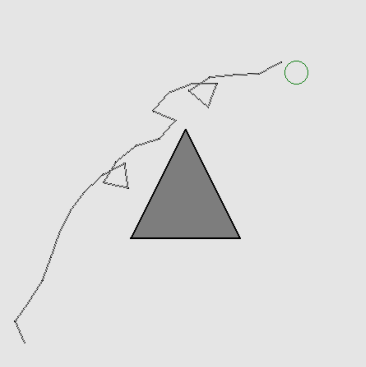}}
  \caption{Random}
\end{subfigure}
\caption{TD3 with continuous masking trajectory examples}
\end{figure}

\begin{figure}[t]
\centering
\begin{subfigure}{.3\linewidth}
  \centering
  \fbox{\includegraphics[width=3.5cm]{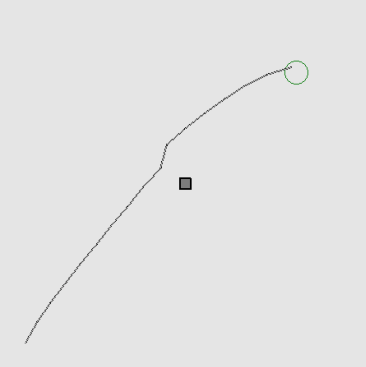}}
  \caption{Maximum reward}
\end{subfigure}%
\begin{subfigure}{.3\linewidth}
  \centering
  \fbox{\includegraphics[width=3.5cm]{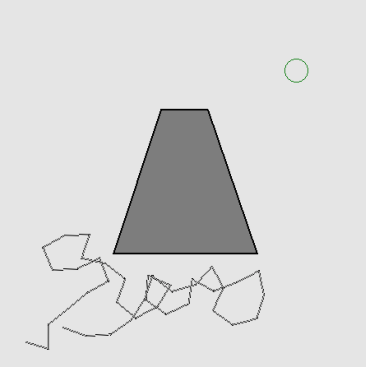}}
  \caption{Minimum reward}
\end{subfigure}
\begin{subfigure}{.3\linewidth}
  \centering
  \fbox{\includegraphics[width=3.5cm]{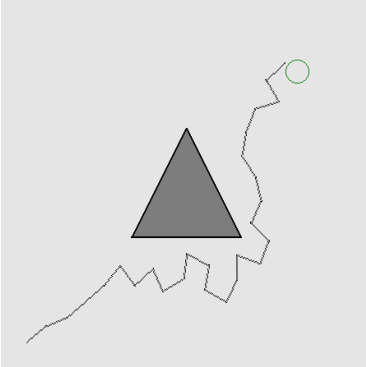}}
  \caption{Random}
\end{subfigure}
\caption{TD3 with random replacement trajectory examples}
\end{figure}

\begin{figure}[t]
\centering
\begin{subfigure}{.3\linewidth}
  \centering
  \fbox{\includegraphics[width=3.5cm]{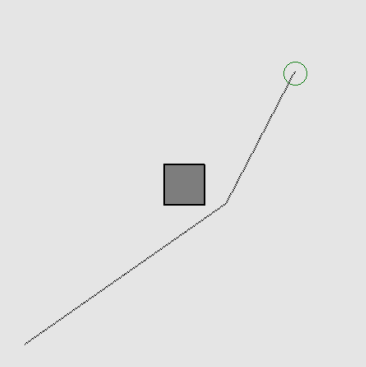}}
  \caption{Maximum reward}
\end{subfigure}%
\begin{subfigure}{.3\linewidth}
  \centering
  \fbox{\includegraphics[width=3.5cm]{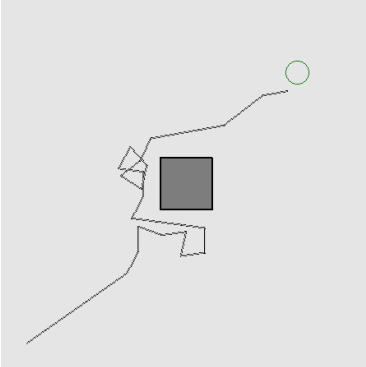}}
  \caption{Minimum reward}
\end{subfigure}
\begin{subfigure}{.3\linewidth}
  \centering
  \fbox{\includegraphics[width=3.5cm]{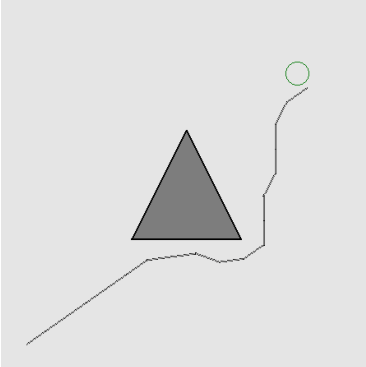}}
  \caption{Random}
\end{subfigure}
\caption{DQN with discrete masking trajectory examples}
\end{figure}

\begin{figure}[t]
\centering
\begin{subfigure}{.3\linewidth}
  \centering
  \fbox{\includegraphics[width=3.5cm]{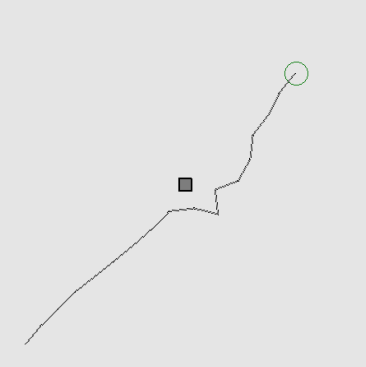}}
  \caption{Maximum reward}
\end{subfigure}%
\begin{subfigure}{.3\linewidth}
  \centering
  \fbox{\includegraphics[width=3.5cm]{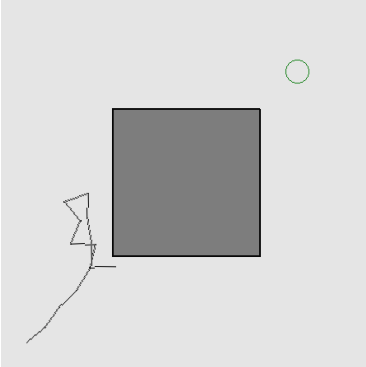}}
  \caption{Minimum reward}
\end{subfigure}
\begin{subfigure}{.3\linewidth}
  \centering
  \fbox{\includegraphics[width=3.5cm]{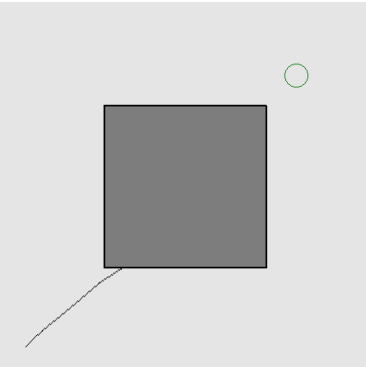}}
  \caption{Random}
\end{subfigure}
\caption{PPO trajectory examples}
\end{figure}

\begin{figure}[t]
\centering
\begin{subfigure}{.3\linewidth}
  \centering
  \fbox{\includegraphics[width=3.5cm]{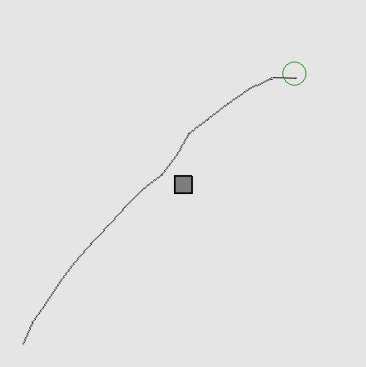}}
  \caption{Maximum reward}
\end{subfigure}%
\begin{subfigure}{.3\linewidth}
  \centering
  \fbox{\includegraphics[width=3.5cm]{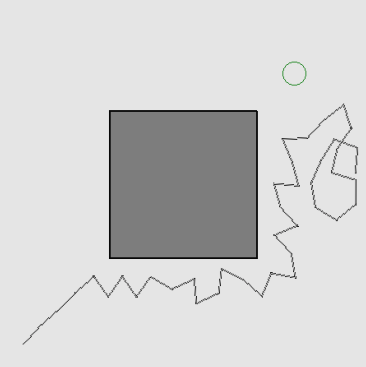}}
  \caption{Minimum reward}
\end{subfigure}
\begin{subfigure}{.3\linewidth}
  \centering
  \fbox{\includegraphics[width=3.5cm]{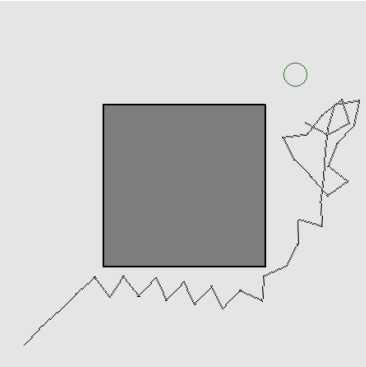}}
  \caption{Random}
\end{subfigure}
\caption{PPO with projection trajectory examples}
\end{figure}

\begin{figure}[t]
\centering
\begin{subfigure}{.3\linewidth}
  \centering
  \fbox{\includegraphics[width=3.5cm]{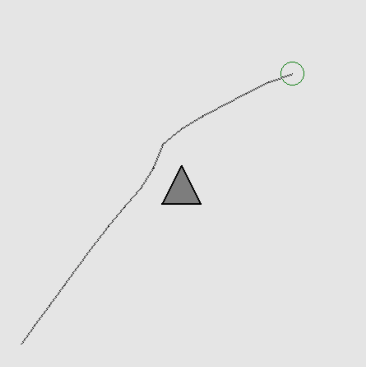}}
  \caption{Maximum reward}
\end{subfigure}%
\begin{subfigure}{.3\linewidth}
  \centering
  \fbox{\includegraphics[width=3.5cm]{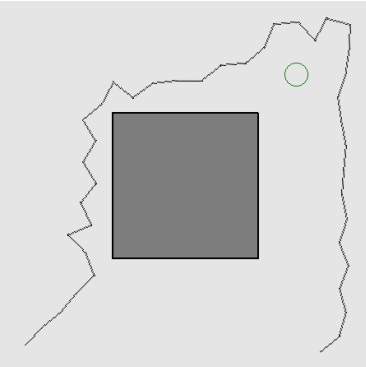}}
  \caption{Minimum reward}
\end{subfigure}
\begin{subfigure}{.3\linewidth}
  \centering
  \fbox{\includegraphics[width=3.5cm]{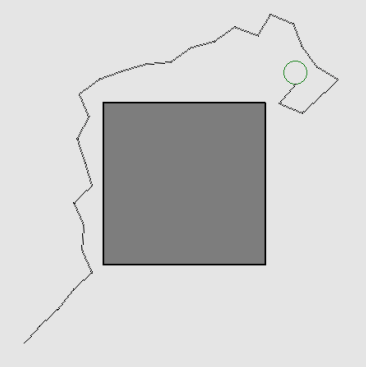}}
  \caption{Random}
\end{subfigure}
\caption{PPO with continuous masking trajectory examples}
\end{figure}

\begin{figure}[t]
\centering
\begin{subfigure}{.3\linewidth}
  \centering
  \fbox{\includegraphics[width=3.5cm]{experiment_1/trajectories/ppo/max-random.png}}
  \caption{Maximum reward}
\end{subfigure}%
\begin{subfigure}{.3\linewidth}
  \centering
  \fbox{\includegraphics[width=3.5cm]{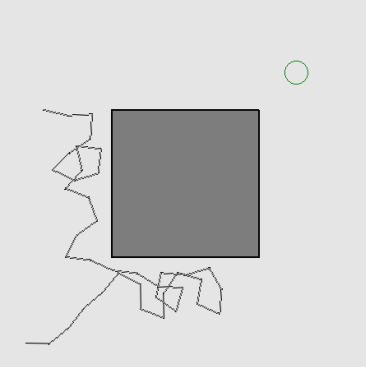}}
  \caption{Minimum reward}
\end{subfigure}
\begin{subfigure}{.3\linewidth}
  \centering
  \fbox{\includegraphics[width=3.5cm]{experiment_1/trajectories/ppo/random-random.png}}
  \caption{Random}
\end{subfigure}
\caption{PPO with random replacement trajectory examples}
\end{figure}

\begin{figure}[t]
\centering
\begin{subfigure}{.3\linewidth}
  \centering
  \fbox{\includegraphics[width=3.5cm]{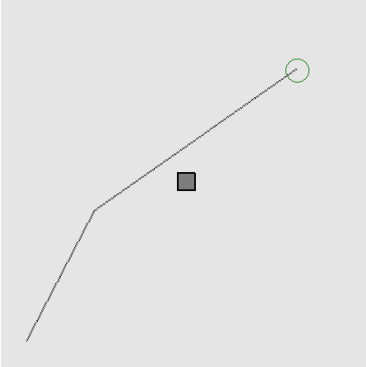}}
  \caption{Maximum reward}
\end{subfigure}%
\begin{subfigure}{.3\linewidth}
  \centering
  \fbox{\includegraphics[width=3.5cm]{experiment_1/trajectories/ppo/min-masked.png}}
  \caption{Minimum reward}
\end{subfigure}
\begin{subfigure}{.3\linewidth}
  \centering
  \fbox{\includegraphics[width=3.5cm]{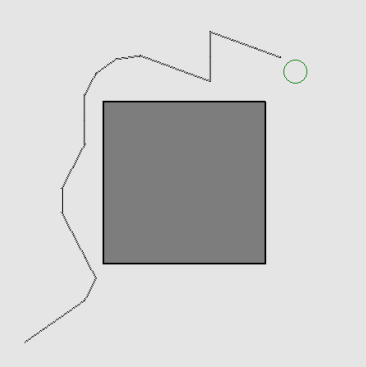}}
  \caption{Random}
\end{subfigure}
\caption{PPO with discrete masking trajectory examples}
\end{figure}

\begin{figure}[t]
\centering
\begin{subfigure}{.3\linewidth}
  \centering
  \fbox{\includegraphics[width=3.5cm]{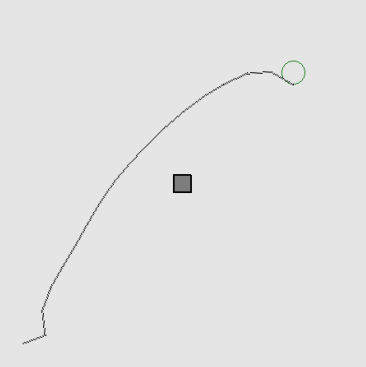}}
  \caption{Maximum reward}
\end{subfigure}%
\begin{subfigure}{.3\linewidth}
  \centering
  \fbox{\includegraphics[width=3.5cm]{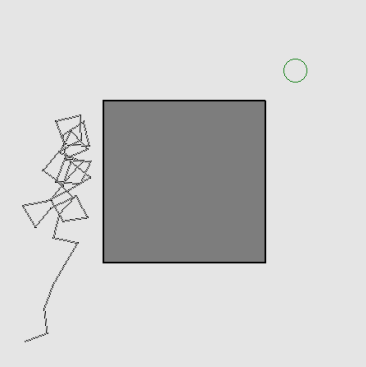}}
  \caption{Minimum reward}
\end{subfigure}
\begin{subfigure}{.3\linewidth}
  \centering
  \fbox{\includegraphics[width=3.5cm]{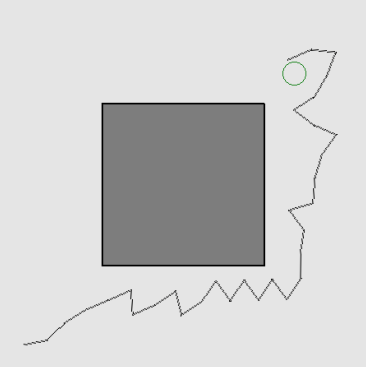}}
  \caption{Random}
\end{subfigure}
\caption{MPS-TD3 with random replacement trajectory examples}
\end{figure}

\begin{figure}[t]
\centering
\begin{subfigure}{.3\linewidth}
  \centering
  \fbox{\includegraphics[width=3.5cm]{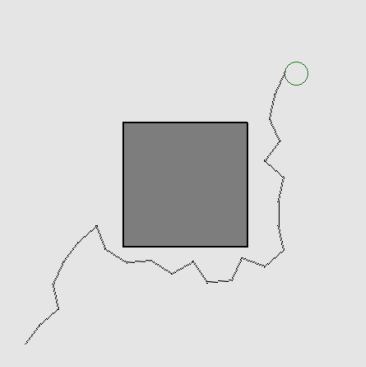}}
  \caption{Maximum reward}
\end{subfigure}%
\begin{subfigure}{.3\linewidth}
  \centering
  \fbox{\includegraphics[width=3.5cm]{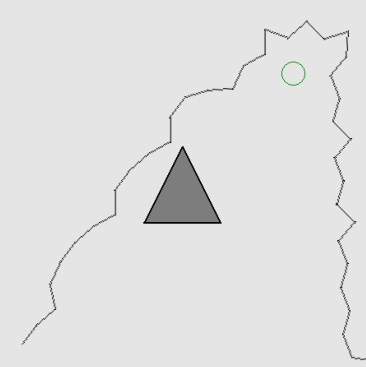}}
  \caption{Minimum reward}
\end{subfigure}
\begin{subfigure}{.3\linewidth}
  \centering
  \fbox{\includegraphics[width=3.5cm]{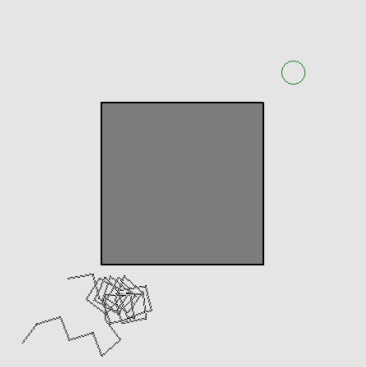}}
  \caption{Random}
\end{subfigure}
\caption{PAM trajectory examples}
\label{app:exp1_2_trajectory_end}
\end{figure}

\begin{figure}[h]
\centering
\begin{subfigure}[b]{\linewidth}
  \centering
   \importpgf{experiment_1/evaluation/reward_t}{on_policy_t_reward.pgf}
  \caption{On-policy}
\end{subfigure}
\\[3ex]
\begin{subfigure}[b]{\linewidth}
  \centering
  \importpgf{experiment_1/evaluation/reward_t}{off_policy_t_reward.pgf}
  \caption{Off-policy}
\end{subfigure}
\caption{Welch's t-test p-values between average returns}
\label{exp1_3_test_reward}
\end{figure}

\begin{figure}[h]
\centering
\begin{subfigure}[b]{\linewidth}
  \centering
  \importpgf{experiment_1/evaluation/reward_t}{on_policy_t_steps.pgf}
  \caption{On-policy}
\end{subfigure}
\\[3ex]
\begin{subfigure}[b]{\linewidth}
  \centering
  \importpgf{experiment_1/evaluation/reward_t}{off_policy_t_steps.pgf}
  \caption{Off-policy}
\end{subfigure}
\caption{Welch's t-test p-values between average steps}
\label{app:exp1_2_test_steps}
\end{figure}

\begin{figure}[h]
\centering
\begin{subfigure}[b]{\linewidth}
  \centering
  \importpgf{experiment_1/evaluation/reward_t}{on_policy_t_solved.pgf}
  \caption{On-policy}
\end{subfigure}
\\[3ex]
\begin{subfigure}[b]{\linewidth}
  \centering
  \importpgf{experiment_1/evaluation/reward_t}{off_policy_t_solved.pgf}
  \caption{Off-policy}
\end{subfigure}
\caption{Welch's t-test p-values between the fractions of solved environments}
\label{exp1_2_test_solved}
\end{figure}

\clearpage
\subsection{Evaluation With Complex Restrictions}
\label{app:exp1_3}

\begin{figure}[!h]
\centering
  \centering
  \importpgf{experiment_1/evaluation_2}{eval_rewards.pgf}
\caption{Average Return}
\label{app:exp1_3_plot_reward}
\end{figure}

\begin{figure}[!h]
\centering
  \centering
  \importpgf{experiment_1/evaluation_2}{eval_solved.pgf}
\caption{Average fraction of solved environments}
\label{app:exp_1_3_plot_solved}
\end{figure}

\begin{figure}[!h]
\centering
  \centering
  \importpgf{experiment_1/evaluation_2}{eval_steps.pgf}
\caption{Average episode length}
\label{app:exp1_3_plot_steps}
\end{figure}
    
\begin{table}[h]
\centering
\scriptsize
\begin{tabular}{llll}
\hline
\multicolumn{1}{c}{\textbf{Approach}} & \multicolumn{1}{c}{\textbf{Return}} & \multicolumn{1}{c}{\textbf{Steps}} & \multicolumn{1}{c}{\textbf{Fraction}} \\ \hline

TD3                                   & $-10.83 \pm 4.46$                     & $37.00 \pm 0.00$                     & $0.42\% \pm 1.02\%$                     \\
TD3-Projection                        & $4.36 \pm 17.70$                      & $31.89 \pm 6.21$                     & $8.33\% \pm 13.20\%$                    \\
TD3-Masking                           & $21.01 \pm 32.59$                     & $32.01 \pm 3.50$                     & $20.42\% \pm 23.42\%$                   \\
TD3-Random                            & $18.34 \pm 28.31$                     & $32.39 \pm 2.16$                     & $18.33\% \pm 22.12\%$                   \\
DQN-Masked                            & $108.31 \pm 6.47$                     & $21.51 \pm 4.47$                     & $97.50\% \pm 3.87\%$                    \\ \hline

PPO                                   & $1.27 \pm 2.13$                       & $20.00 \pm 0.00$                     & $0.42\% \pm 1.02\%$                     \\
PPO-Projection                        & $55.10 \pm 27.31$                     & $28.60 \pm 3.72$                     & $48.75\% \pm 25.29\%$                   \\
PPO-Masking                           & $52.71 \pm 13.95$                     & $25.81 \pm 1.17$                     & $45.00\% \pm 12.14\%$                   \\
PPO-Random                            & $49.73 \pm 20.90$                     & $29.65 \pm 1.70$                     & $45.00\% \pm 19.62\%$                   \\
PPO-Masked                            & $33.38 \pm 4.61$                      & $24.22 \pm 1.66$                     & $31.67\% \pm 4.65\%$                    \\ \hline

MPS-TD3                               & $5.35 \pm 11.63$                       & $35.16 \pm 3.13$                     & $5.83\% \pm 4.92\%$                     \\
PAM                                   & $10.04 \pm 24.82$                     & $28.27 \pm 1.56$                     & $12.08\% \pm 15.12\%$                   \\ \hline
\end{tabular}
\caption{Average return, steps, and solved episodes}
\label{app:exp1_3_results}
\end{table}

\begin{table}[h]
\centering
\scriptsize
\begin{tabular}{llll}
\hline
\multicolumn{1}{c}{\textbf{Approach}} & \multicolumn{1}{c}{\textbf{Intervals}} & \multicolumn{1}{c}{\textbf{Size}} & \multicolumn{1}{c}{\textbf{Fraction}} \\ \hline

TD3                                   & $1.38 \pm 0.04$                        & $165.08 \pm 3.56$                     & $75.04\% \pm 1.62\%$                    \\
TD3-Projection                        & $1.35 \pm 0.08$                        & $143.61 \pm 8.58$                     & $65.28\% \pm 3.90\%$                    \\
TD3-Masking                           & $1.42 \pm 0.09$                        & $145.85 \pm 4.04$                     & $66.29\% \pm 1.83\%$                    \\
TD3-Random                            & $1.44 \pm 0.04$                        & $148.03 \pm 7.65$                     & $67.29\% \pm 3.48\%$                   \\
DQN-Masked                            & $1.33 \pm0.04$                         & $144.09 \pm 4.85$                     & $65.49\% \pm 2.21\%$                    \\ \hline

PPO                                   & $1.32 \pm 0.03$                        & $164.34 \pm 3.07$                     & $74.70\% \pm 1.40\%$                    \\
PPO-Projection                        & $1.35 \pm 0.01$                        & $143.51	 \pm 3.05$                     & $65.23\% \pm 1.39\%$                    \\
PPO-Masking                           & $1.37 \pm 0.02$                        & $144.57 \pm 3.84$                     & $65.71\% \pm 1.74\%$                    \\
PPO-Random                            & $1.38 \pm0.02$                         & $145.62 \pm 1.99$                     & $66.19\% \pm 0.90\%$                    \\
PPO-Masked                            & $1.34 \pm 0.04$                        & $145.01 \pm 2.09$                     & $65.91\% \pm 0.95\%$                    \\ \hline

MPS-TD3                               & $1.51 \pm 0.02$                        & $142.27 \pm 4.43$                     & $64.67\% \pm 2.02\%$                    \\
PAM                                   & $1.37 \pm 0.04$                        & $151.20 \pm 4.19$                     & $68.73\% \pm 1.90\%$                    \\ \hline
\end{tabular}
\caption{Average number of disjoint intervals and size of the action space}
\label{app:exp1_3_control}
\end{table}

\begin{table}[h]
\centering
\scriptsize
\begin{tabular}{lllll}
\hline
\multicolumn{1}{c}{\textbf{Approach}} & \multicolumn{1}{c}{\textbf{Average}} & \multicolumn{1}{c}{\textbf{Minimum}} & \multicolumn{1}{c}{\textbf{Maximum}} & \multicolumn{1}{c}{\textbf{Variance}} \\ \hline

TD3 & $141.25 \pm 5.96$ & $129.71 \pm 7.40$ & $152.84 \pm 4.84$ & $525.27 \pm 118.62$ \\
TD3-Projection & $123.23 \pm 3.85$ & $113.66 \pm 2.28$ & $132.85 \pm 6.00$ & $411.42 \pm 110.39$ \\
TD3-Masking & $122.13 \pm 4.37$ & $111.07 \pm 5.96$ & $133.27 \pm 3.33$ & $473.27 \pm 75.23$ \\
TD3-Random & $122.99 \pm 6.92$ & $111.09 \pm 6.80$ & $135.00 \pm 7.17$ & $510.08 \pm 50.72$ \\
DQN-Masked & $126.31 \pm 5.18$ & $117.93 \pm 5.53$ & $134.72 \pm 4.94$ & $332.59 \pm 39.78$ \\ \hline

PPO & $144.62 \pm 3.70$ & $134.57 \pm 4.18$ & $154.69 \pm 3.30$ & $453.64 \pm 41.82$ \\
PPO-Projection & $124.21 \pm 2.50$ & $115.20 \pm 2.46$ & $133.28 \pm 2.75$ & $369.43 \pm 33.25$ \\
PPO-Masking & $124.36 \pm 4.04$ & $114.47 \pm 4.01$ & $134.32 \pm 4.11$ & $405.97 \pm 30.57$ \\
PPO-Random & $124.89 \pm 2.20$ & $115.04 \pm 2.42$ & $134.80 \pm 2.15$ & $411.82 \pm 38.29$ \\
PPO-Masked & $126.44 \pm 1.07$ & $117.91 \pm 1.39$ & $134.99 \pm 1.21$ & $341.81 \pm 26.87$ \\ \hline

MPS-TD3 & $114.46 \pm 3.91$ & $101.47 \pm 3.67$ & $127.53 \pm 4.18$ & $534.03 \pm 32.51$ \\
PAM & $130.00 \pm 5.09$ & $120.30 \pm 5.52$ & $139.75 \pm 4.83$ & $405.11 \pm 55.11$ \\ \hline
\end{tabular}
\caption{Size of the individual intervals}
\label{app:exp1_3_single}
\end{table}

\begin{figure}[t]
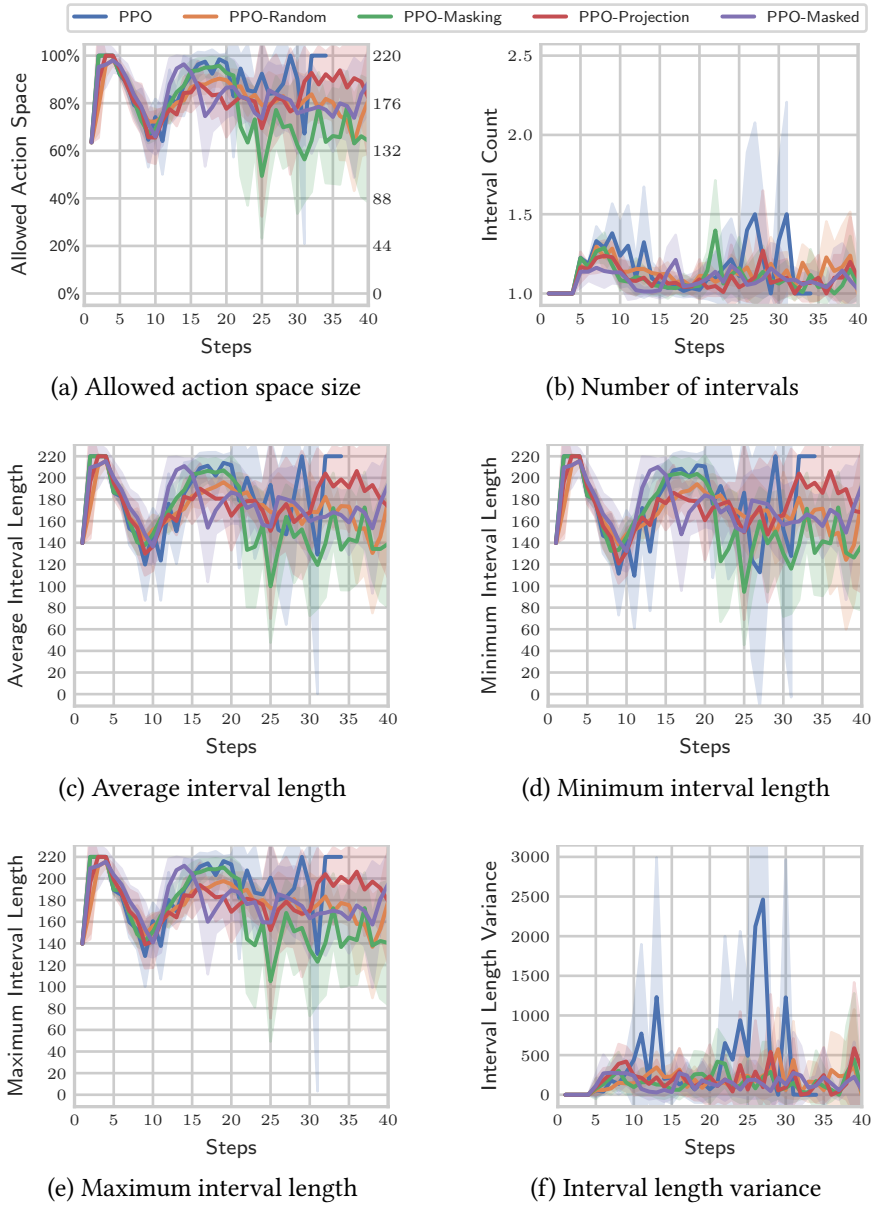

\centering
\begin{subfigure}{0.8\linewidth}
  \centering
  \importpgf{experiment_1/training/ppo}{legend.pgf}
\end{subfigure}
\begin{subfigure}{.49\linewidth}
  \centering
  \importpgf{experiment_1/evaluation_2/ppo}{allowed.pgf}
  \caption{Allowed action space size}
  \label{app:exp1_3_length_ppo}
\end{subfigure}%
\begin{subfigure}{.49\linewidth}
  \centering
  \importpgf{experiment_1/evaluation_2/ppo}{count.pgf}
  \caption{Number of intervals}
\end{subfigure}
\\[3ex]
\begin{subfigure}{.49\linewidth}
  \centering
  \importpgf{experiment_1/evaluation_2/ppo}{average.pgf}
  \caption{Average interval length}
\end{subfigure}
\begin{subfigure}{.49\linewidth}
  \centering
  \importpgf{experiment_1/evaluation_2/ppo}{minimum.pgf}
  \caption{Minimum interval length}
\end{subfigure}
\\[3ex]
\begin{subfigure}{.49\linewidth}
  \centering
  \importpgf{experiment_1/evaluation_2/ppo}{maximum.pgf}
  \caption{Maximum interval length}
\end{subfigure}
\begin{subfigure}{.49\linewidth}
  \centering
  \importpgf{experiment_1/evaluation_2/ppo}{variance.pgf}
  \caption{Interval length variance}
\end{subfigure}
\caption{On-policy control variables}
\label{app:exp1_3_control_ppo}
\end{figure}

\begin{figure}[t]
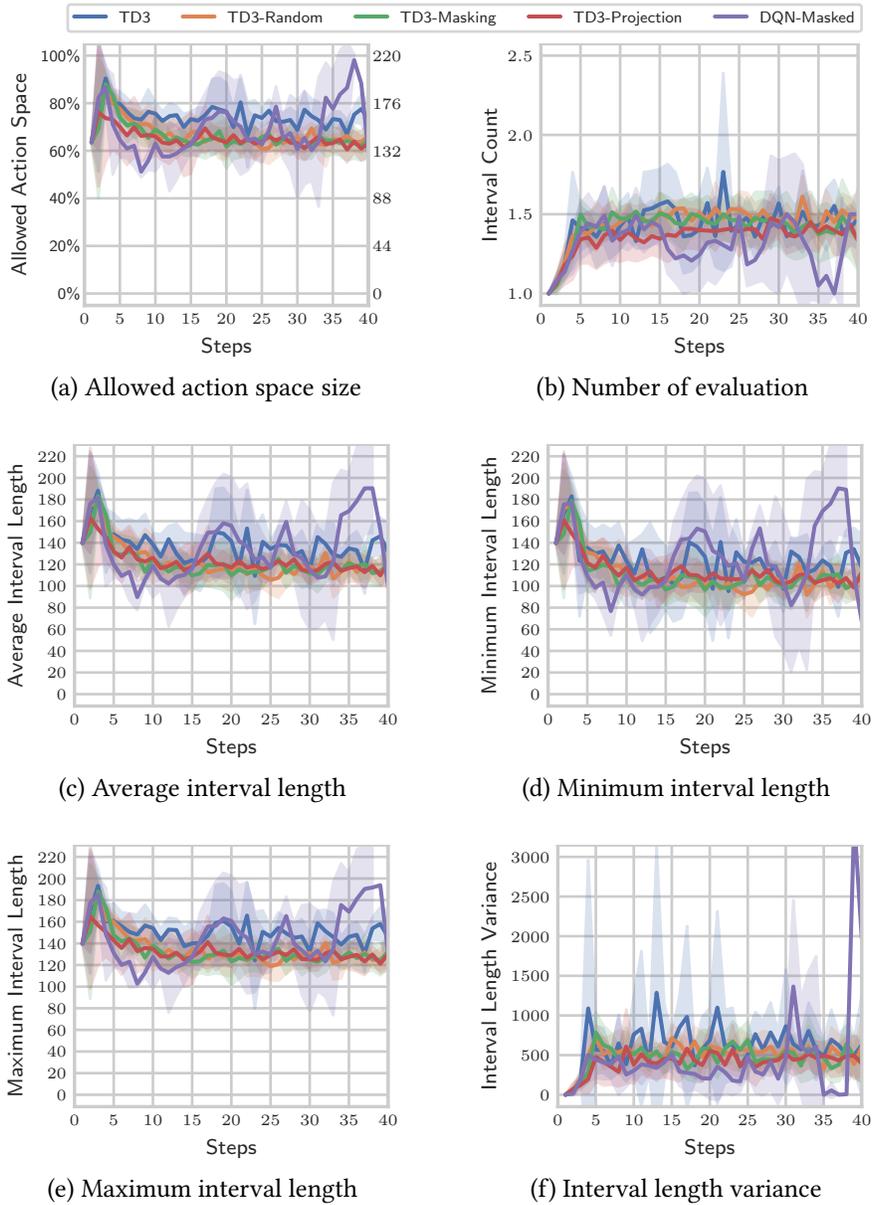

\centering
\begin{subfigure}{0.8\linewidth}
  \centering
  \importpgf{experiment_1/training/ddpg}{legend.pgf}
\end{subfigure}
\begin{subfigure}{.49\linewidth}
  \centering
  \importpgf{experiment_1/evaluation_2/td3}{allowed.pgf}
  \caption{Allowed action space size}
  \label{app:exp1_3_length_td3}
\end{subfigure}%
\begin{subfigure}{.49\linewidth}
  \centering
   \importpgf{experiment_1/evaluation_2/td3}{count.pgf}
  \caption{Number of evaluation}
\end{subfigure}
\\[3ex]
\begin{subfigure}{.49\linewidth}
  \centering
   \importpgf{experiment_1/evaluation_2/td3}{average.pgf}
  \caption{Average interval length}
\end{subfigure}
\begin{subfigure}{.49\linewidth}
  \centering
   \importpgf{experiment_1/evaluation_2/td3}{minimum.pgf}
  \caption{Minimum interval length}
\end{subfigure}
\\[3ex]
\begin{subfigure}{.49\linewidth}
  \centering
   \importpgf{experiment_1/evaluation_2/td3}{maximum.pgf}
  \caption{Maximum interval length}
\end{subfigure}
\begin{subfigure}{.49\linewidth}
  \centering
   \importpgf{experiment_1/evaluation_2/td3}{variance.pgf}
  \caption{Interval length variance}
\end{subfigure}
\caption{Off-policy control variables}
\label{app:exp1_3_control_td3}
\end{figure}

\begin{figure}[t]
\centering
\begin{subfigure}{0.8\linewidth}
  \centering
  \importpgf{experiment_1/training/own}{legend.pgf}
\end{subfigure}
\begin{subfigure}{.49\linewidth}
  \centering
  \importpgf{experiment_1/evaluation_2/own}{allowed.pgf}
  \caption{Allowed action space size}
  \label{app:exp1_3_length_own}
\end{subfigure}%
\begin{subfigure}{.49\linewidth}
  \centering
  \importpgf{experiment_1/evaluation_2/own}{count.pgf}
  \caption{Number of intervals}
\end{subfigure}
\\[3ex]
\begin{subfigure}{.49\linewidth}
  \centering
  \importpgf{experiment_1/evaluation_2/own}{average.pgf}
  \caption{Average interval length}
\end{subfigure}
\begin{subfigure}{.49\linewidth}
  \centering
  \importpgf{experiment_1/evaluation_2/own}{minimum.pgf}
  \caption{Minimum interval length}
\end{subfigure}
\\[3ex]
\begin{subfigure}{.49\linewidth}
  \centering
  \importpgf{experiment_1/evaluation_2/own}{maximum.pgf}
  \caption{Maximum interval length}
\end{subfigure}
\begin{subfigure}{.49\linewidth}
  \centering
  \importpgf{experiment_1/evaluation_2/own}{variance.pgf}
  \caption{Interval length variance}
\end{subfigure}
\caption{MPS-TD3 and PAM control variables}
\label{app:exp1_3_control_own}
\end{figure}

\begin{figure}[t]
\centering
\begin{subfigure}{.3\linewidth}
  \centering
  \fbox{\includegraphics[width=3.5cm]{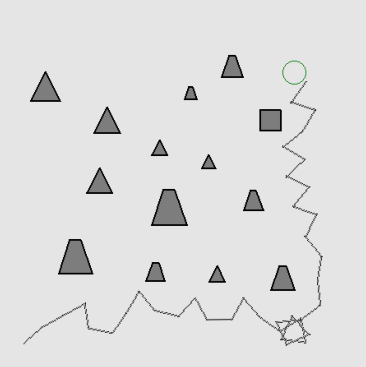}}
  \caption{Maximum reward}
\end{subfigure}%
\begin{subfigure}{.3\linewidth}
  \centering
  \fbox{\includegraphics[width=3.5cm]{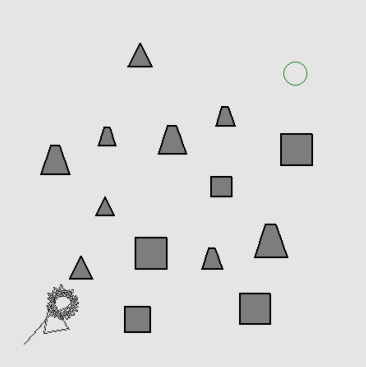}}
  \caption{Minimum reward}
\end{subfigure}
\begin{subfigure}{.3\linewidth}
  \centering
  \fbox{\includegraphics[width=3.5cm]{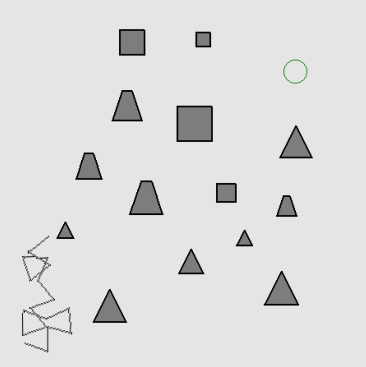}}
  \caption{Random}
\end{subfigure}
\caption{TD3 trajectory examples}
\label{app:exp1_3_trajectory_start}
\end{figure}

\begin{figure}[t]
\centering
\begin{subfigure}{.3\linewidth}
  \centering
  \fbox{\includegraphics[width=3.5cm]{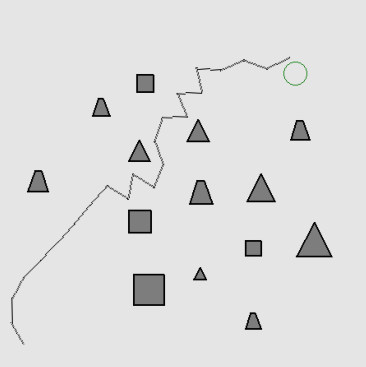}}
  \caption{Maximum reward}
\end{subfigure}%
\begin{subfigure}{.3\linewidth}
  \centering
  \fbox{\includegraphics[width=3.5cm]{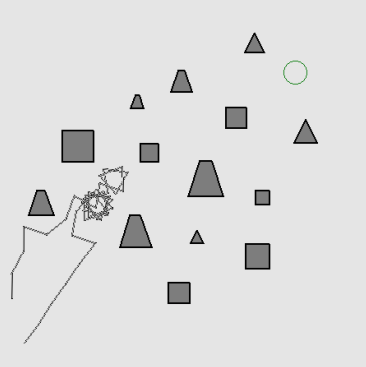}}
  \caption{Minimum reward}
\end{subfigure}
\begin{subfigure}{.3\linewidth}
  \centering
  \fbox{\includegraphics[width=3.5cm]{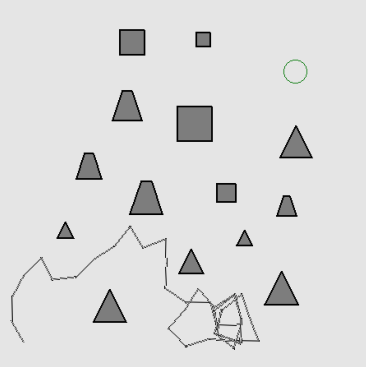}}
  \caption{Random}
\end{subfigure}
\caption{TD3 with projection trajectory examples}
\end{figure}

\begin{figure}[t]
\centering
\begin{subfigure}{.3\linewidth}
  \centering
  \fbox{\includegraphics[width=3.5cm]{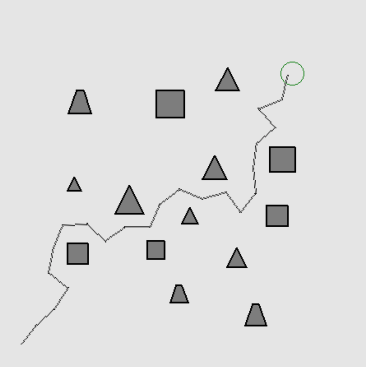}}
  \caption{Maximum reward}
\end{subfigure}%
\begin{subfigure}{.3\linewidth}
  \centering
  \fbox{\includegraphics[width=3.5cm]{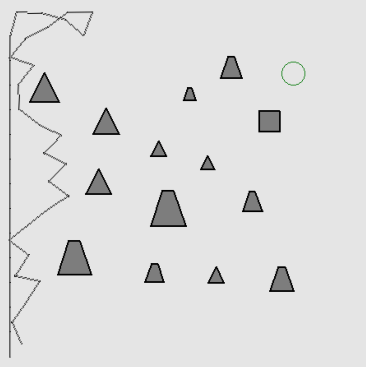}}
  \caption{Minimum reward}
\end{subfigure}
\begin{subfigure}{.3\linewidth}
  \centering
  \fbox{\includegraphics[width=3.5cm]{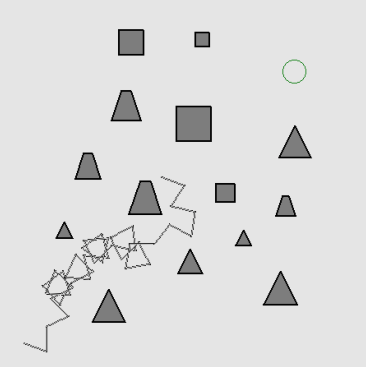}}
  \caption{Random}
\end{subfigure}
\caption{TD3 with continuous masking trajectory examples}
\end{figure}

\begin{figure}[t]
\centering
\begin{subfigure}{.3\linewidth}
  \centering
  \fbox{\includegraphics[width=3.5cm]{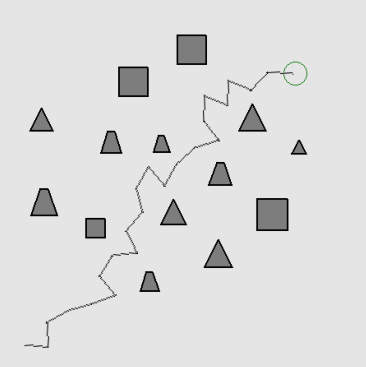}}
  \caption{Maximum reward}
\end{subfigure}%
\begin{subfigure}{.3\linewidth}
  \centering
  \fbox{\includegraphics[width=3.5cm]{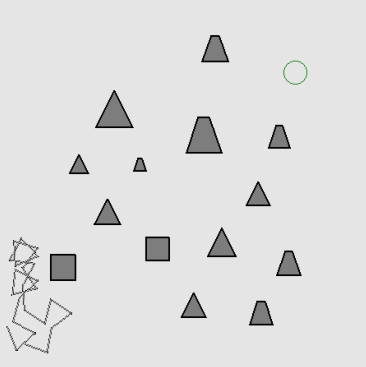}}
  \caption{Minimum reward}
\end{subfigure}
\begin{subfigure}{.3\linewidth}
  \centering
  \fbox{\includegraphics[width=3.5cm]{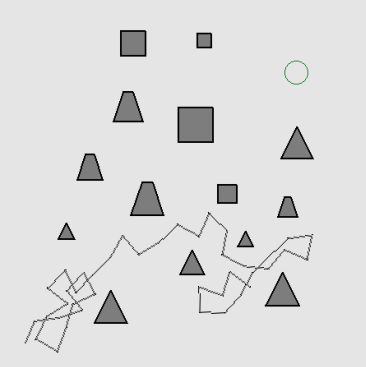}}
  \caption{Random}
\end{subfigure}
\caption{TD3 with random replacement trajectory examples}
\end{figure}

\begin{figure}[t]
\centering
\begin{subfigure}{.3\linewidth}
  \centering
  \fbox{\includegraphics[width=3.5cm]{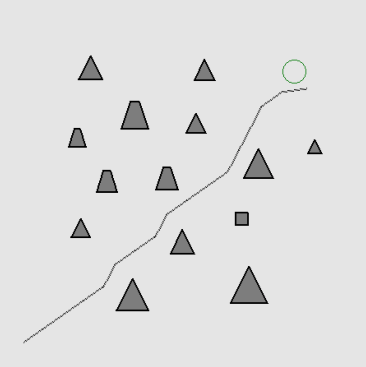}}
  \caption{Maximum reward}
\end{subfigure}%
\begin{subfigure}{.3\linewidth}
  \centering
  \fbox{\includegraphics[width=3.5cm]{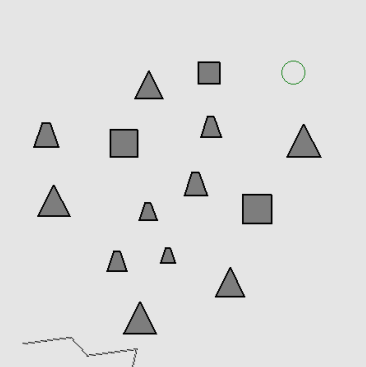}}
  \caption{Minimum reward}
\end{subfigure}
\begin{subfigure}{.3\linewidth}
  \centering
  \fbox{\includegraphics[width=3.5cm]{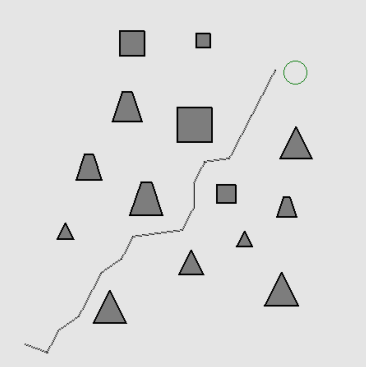}}
  \caption{Random}
\end{subfigure}
\caption{DQN with discrete masking trajectory examples}
\end{figure}

\begin{figure}[t]
\centering
\begin{subfigure}{.3\linewidth}
  \centering
  \fbox{\includegraphics[width=3.5cm]{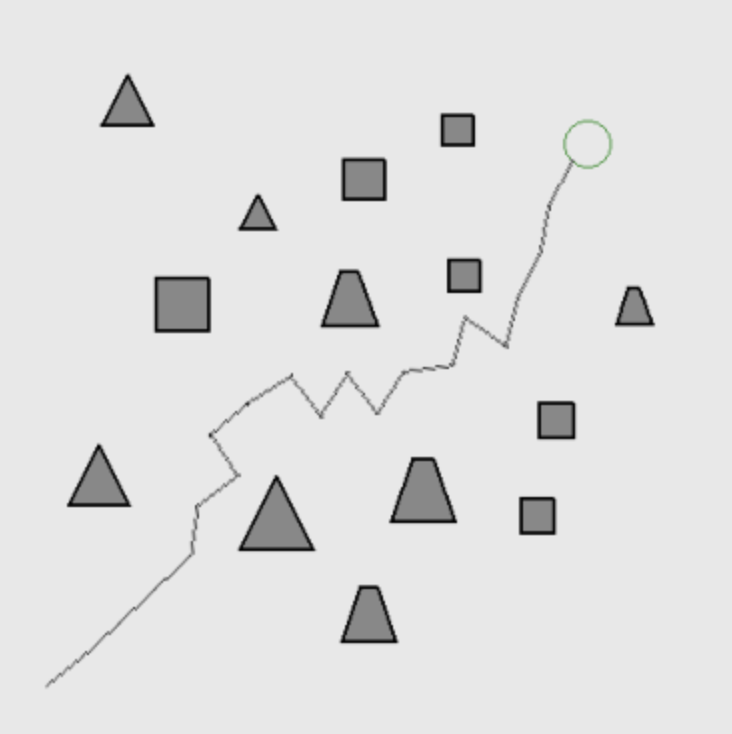}}
  \caption{Maximum reward}
\end{subfigure}%
\begin{subfigure}{.3\linewidth}
  \centering
  \fbox{\includegraphics[width=3.5cm]{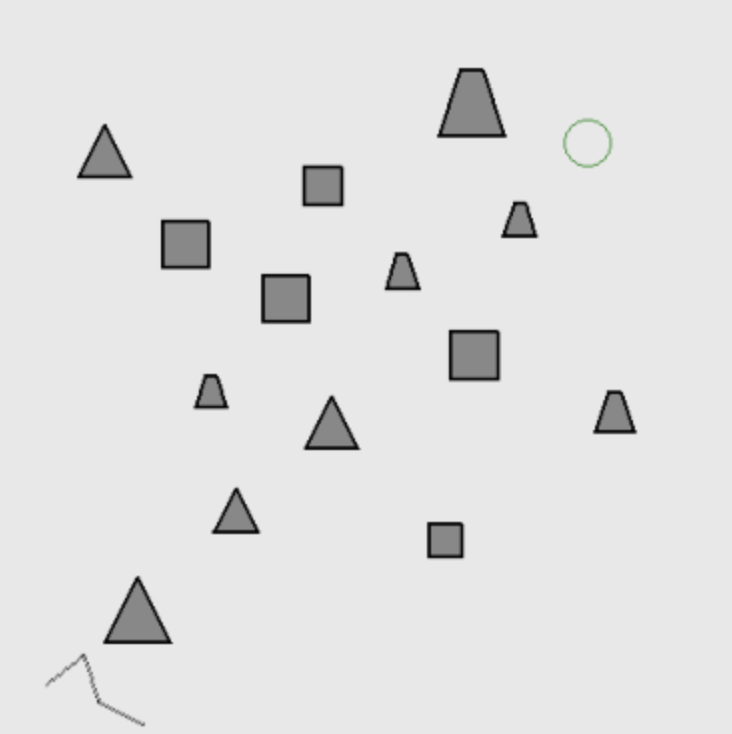}}
  \caption{Minimum reward}
\end{subfigure}
\begin{subfigure}{.3\linewidth}
  \centering
  \fbox{\includegraphics[width=3.5cm]{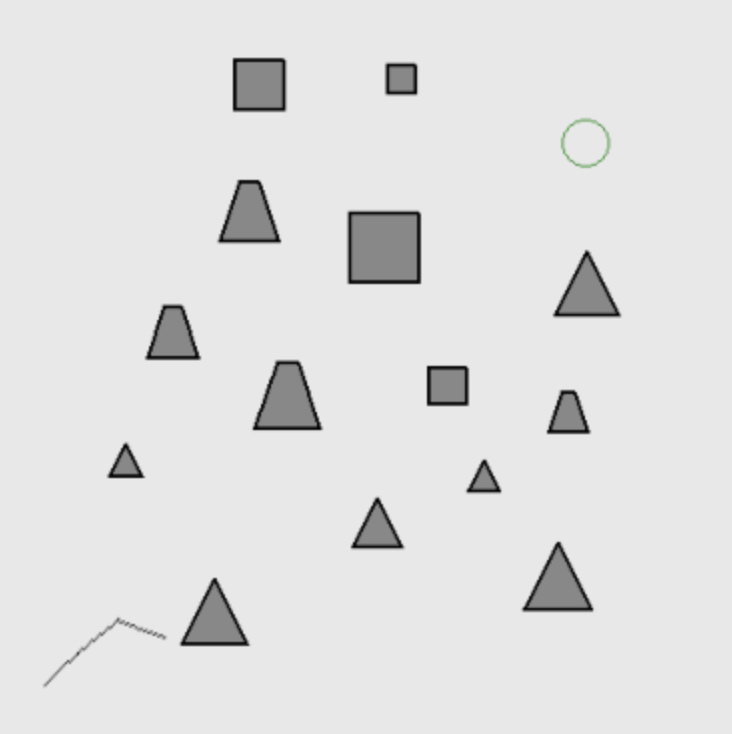}}
  \caption{Random}
\end{subfigure}
\caption{PPO trajectory examples}
\end{figure}

\begin{figure}[t]
\centering
\begin{subfigure}{.3\linewidth}
  \centering
  \fbox{\includegraphics[width=3.5cm]{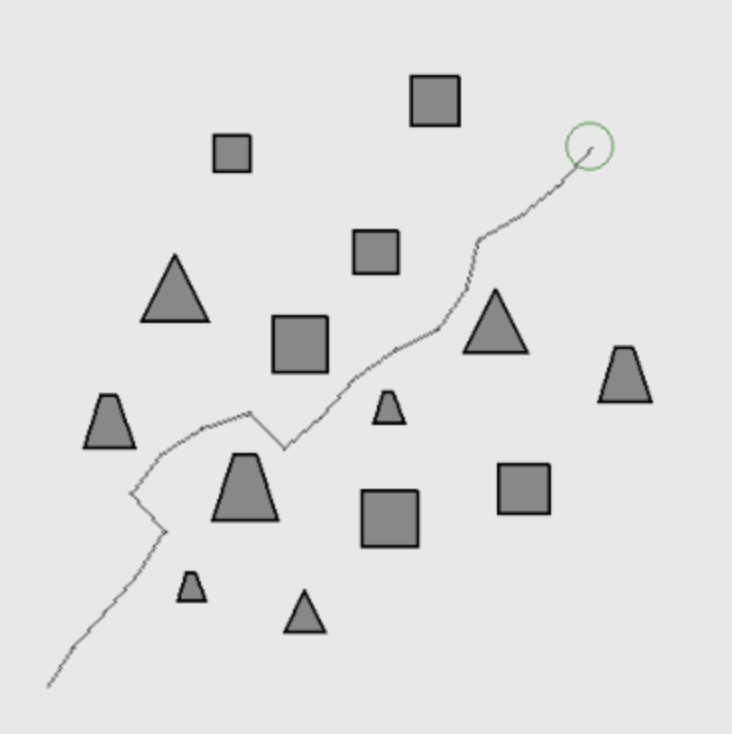}}
  \caption{Maximum reward}
\end{subfigure}%
\begin{subfigure}{.3\linewidth}
  \centering
  \fbox{\includegraphics[width=3.5cm]{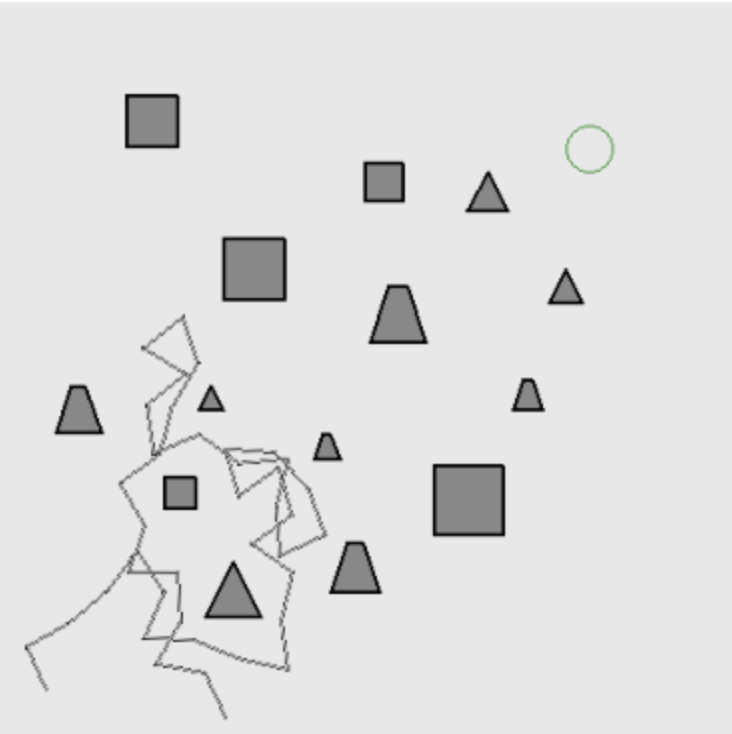}}
  \caption{Minimum reward}
\end{subfigure}
\begin{subfigure}{.3\linewidth}
  \centering
  \fbox{\includegraphics[width=3.5cm]{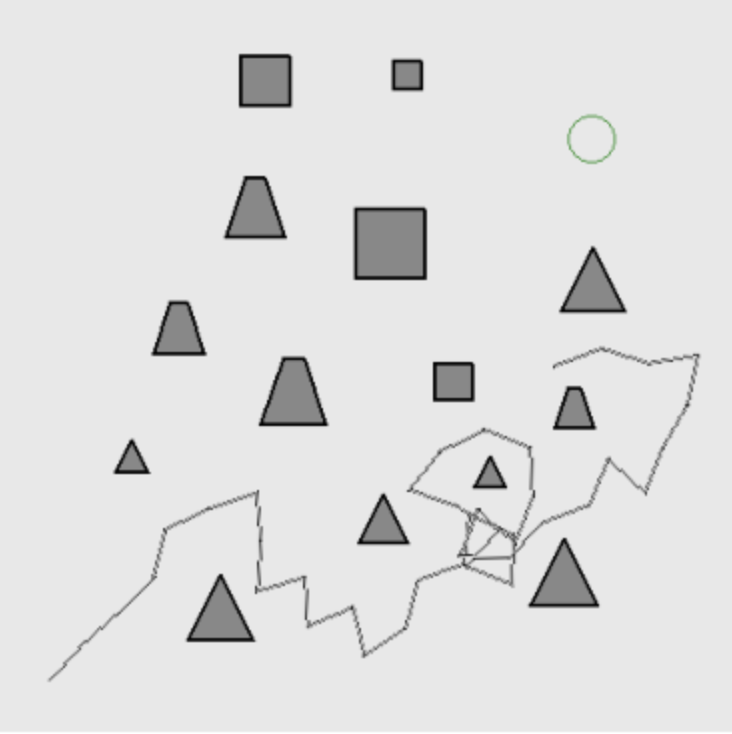}}
  \caption{Random}
\end{subfigure}
\caption{PPO with projection trajectory examples}
\end{figure}

\begin{figure}[t]
\centering
\begin{subfigure}{.3\linewidth}
  \centering
  \fbox{\includegraphics[width=3.5cm]{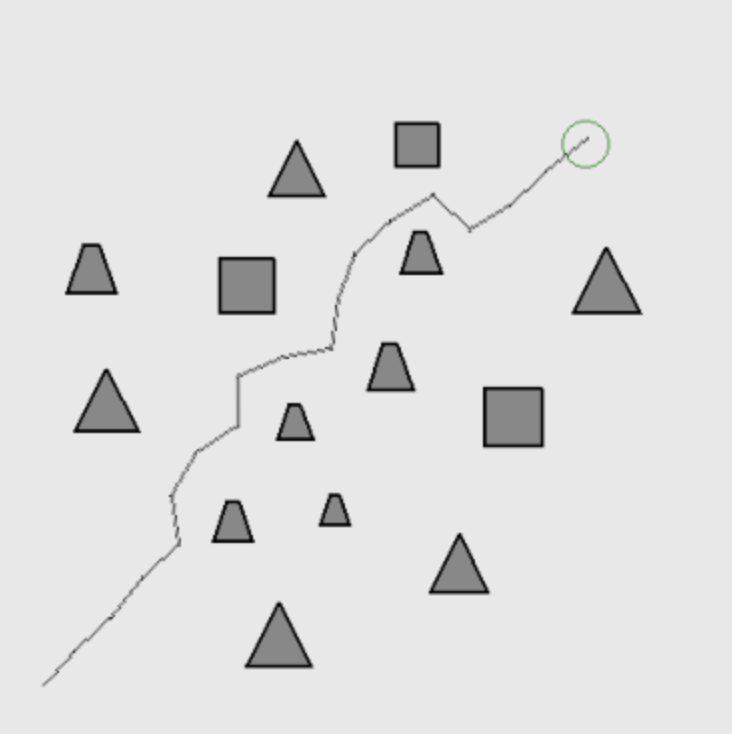}}
  \caption{Maximum reward}
\end{subfigure}%
\begin{subfigure}{.3\linewidth}
  \centering
  \fbox{\includegraphics[width=3.5cm]{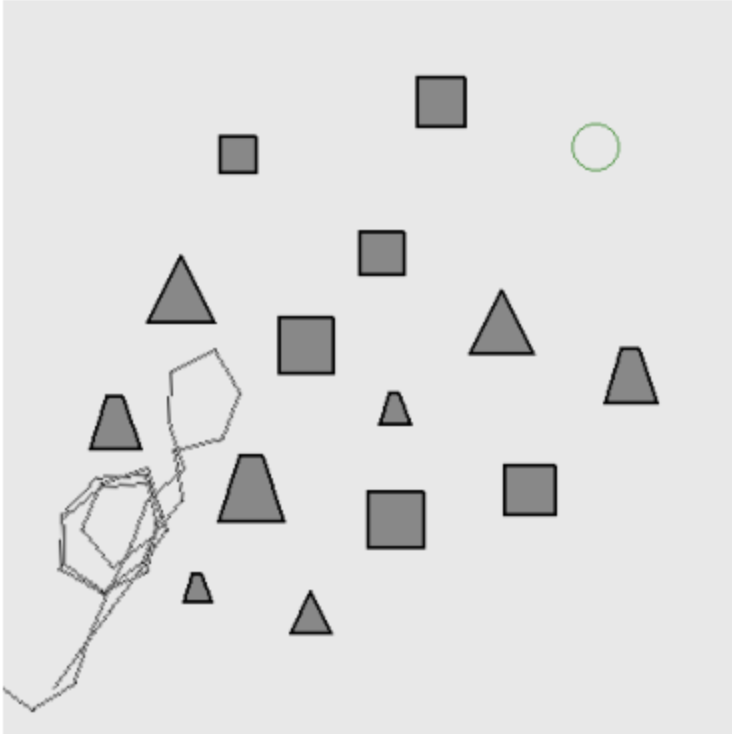}}
  \caption{Minimum reward}
\end{subfigure}
\begin{subfigure}{.3\linewidth}
  \centering
  \fbox{\includegraphics[width=3.5cm]{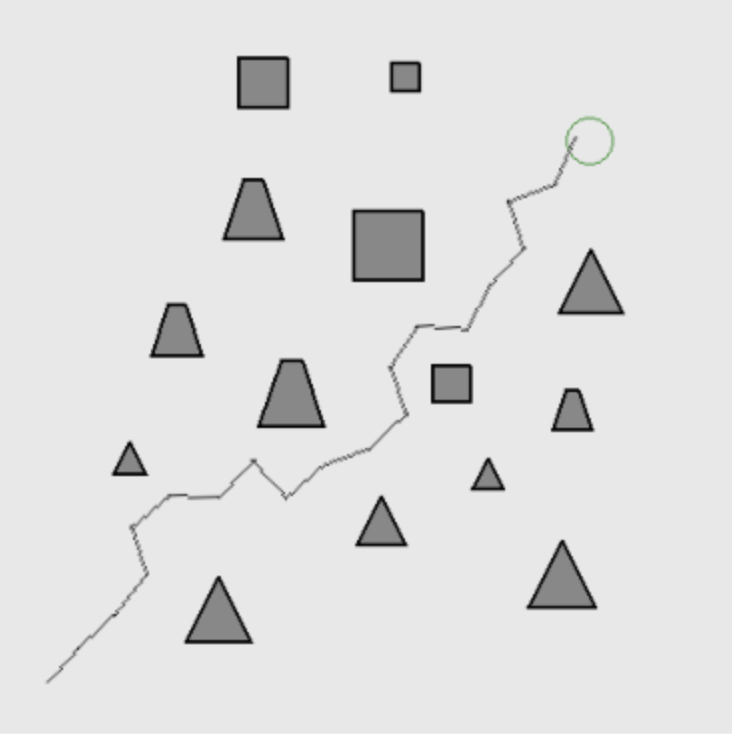}}
  \caption{Random}
\end{subfigure}
\caption{PPO with continuous masking trajectory examples}
\end{figure}

\begin{figure}[t]
\centering
\begin{subfigure}{.3\linewidth}
  \centering
  \fbox{\includegraphics[width=3.5cm]{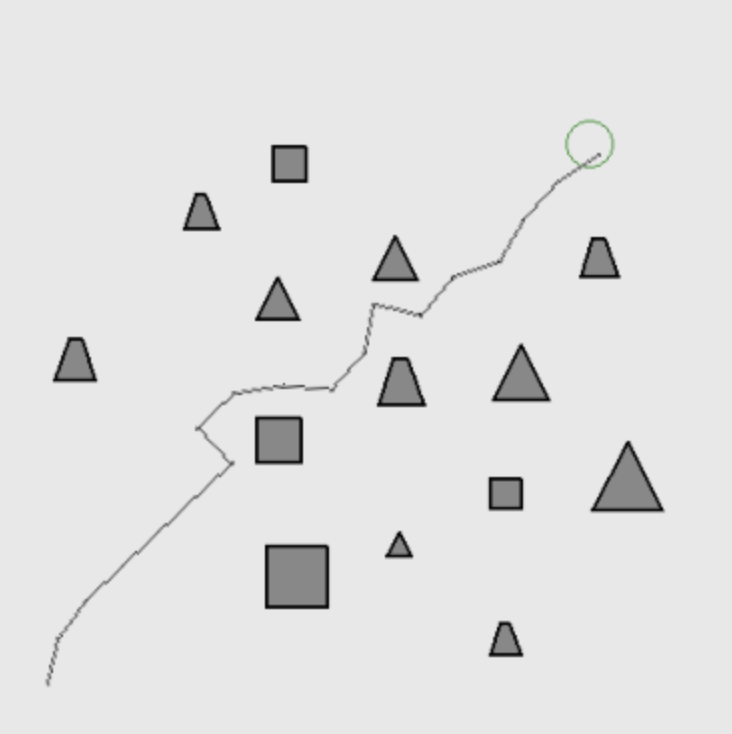}}
  \caption{Maximum reward}
\end{subfigure}%
\begin{subfigure}{.3\linewidth}
  \centering
  \fbox{\includegraphics[width=3.5cm]{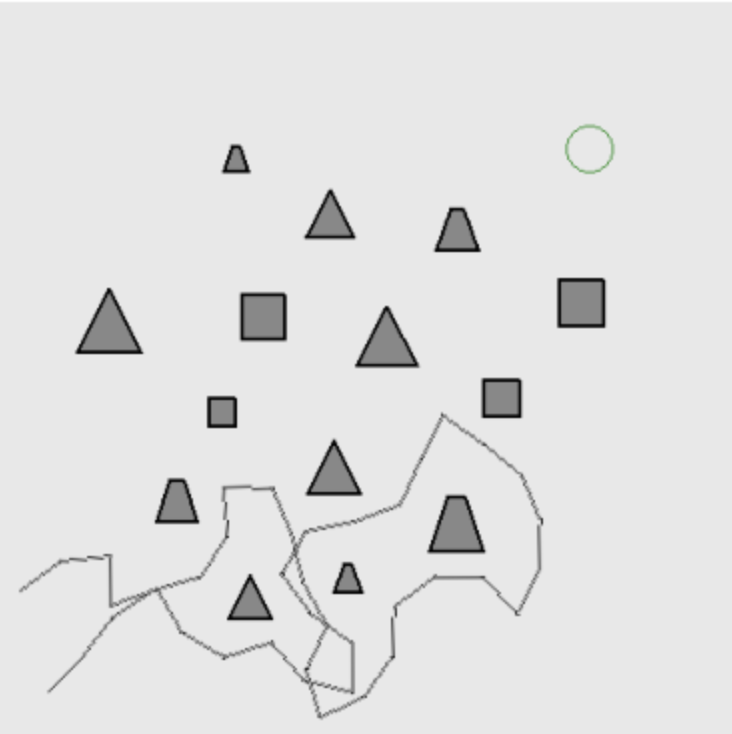}}
  \caption{Minimum reward}
\end{subfigure}
\begin{subfigure}{.3\linewidth}
  \centering
  \fbox{\includegraphics[width=3.5cm]{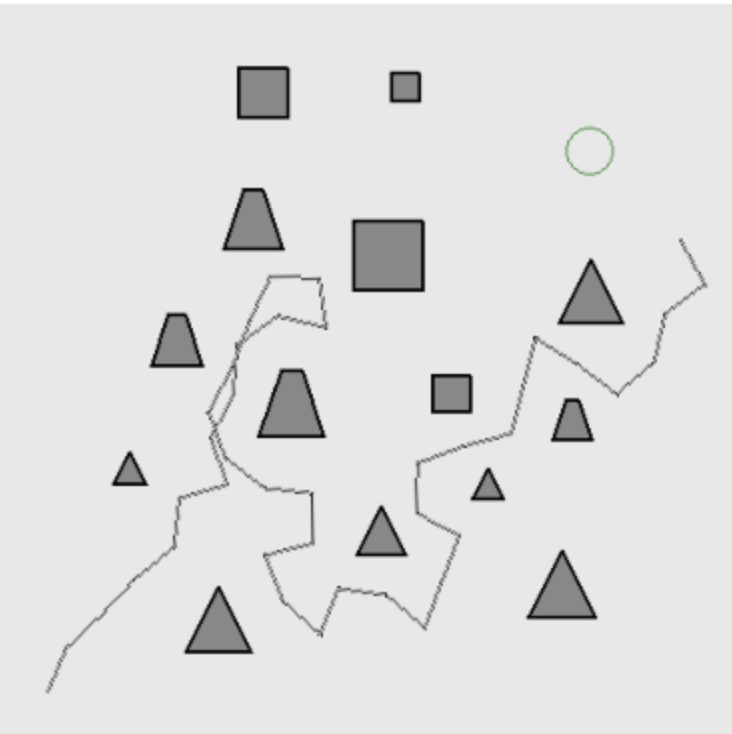}}
  \caption{Random}
\end{subfigure}
\caption{PPO with random replacement trajectory examples}
\end{figure}

\begin{figure}[t]
\centering
\begin{subfigure}{.3\linewidth}
  \centering
  \fbox{\includegraphics[width=3.5cm]{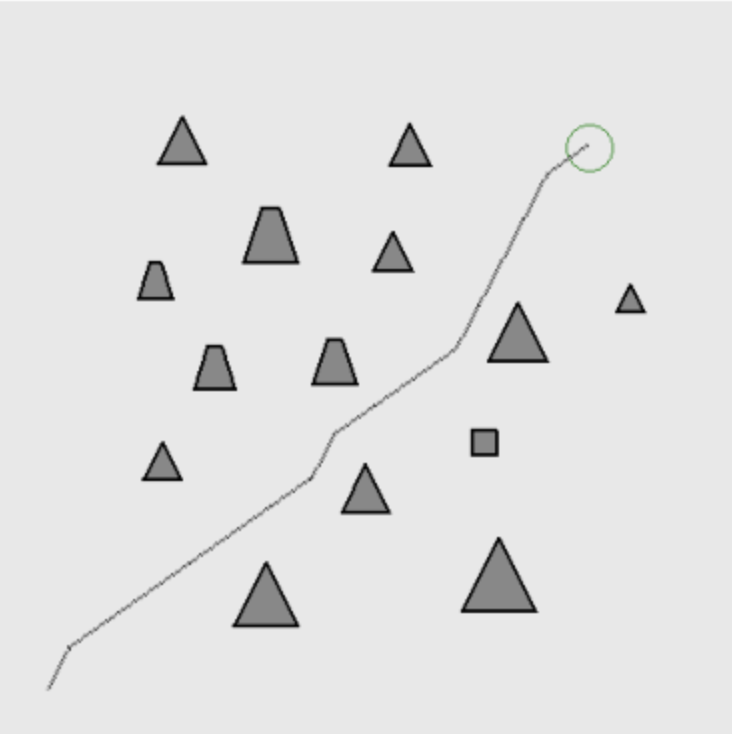}}
  \caption{Maximum reward}
\end{subfigure}%
\begin{subfigure}{.3\linewidth}
  \centering
  \fbox{\includegraphics[width=3.5cm]{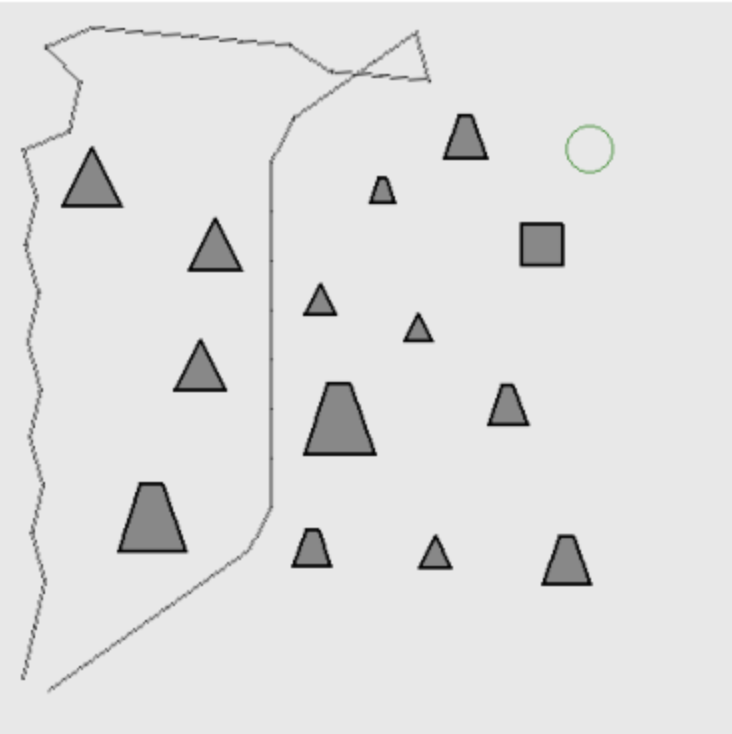}}
  \caption{Minimum reward}
\end{subfigure}
\begin{subfigure}{.3\linewidth}
  \centering
  \fbox{\includegraphics[width=3.5cm]{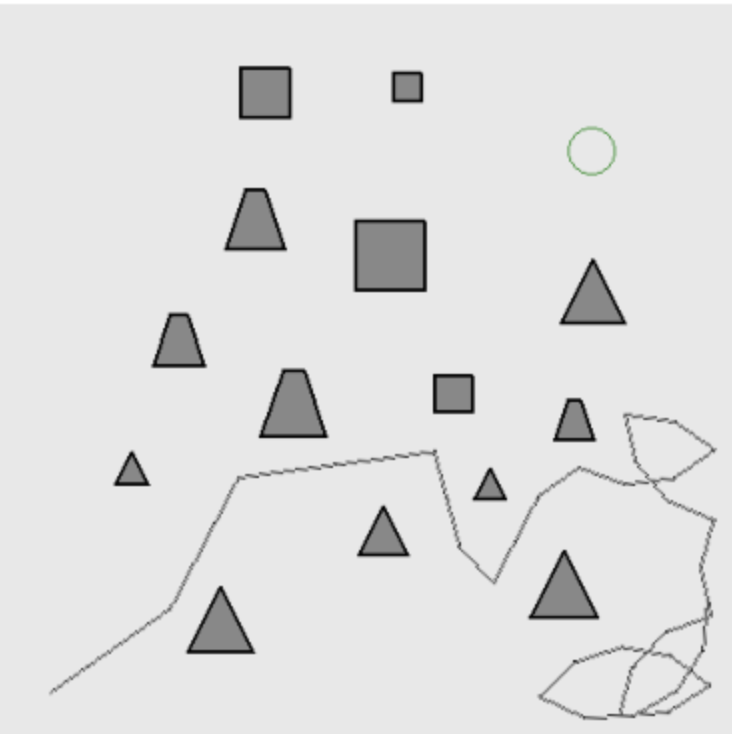}}
  \caption{Random}
\end{subfigure}
\caption{PPO with discrete masking trajectory examples}
\end{figure}

\begin{figure}[t]
\centering
\begin{subfigure}{.3\linewidth}
  \centering
  \fbox{\includegraphics[width=3.5cm]{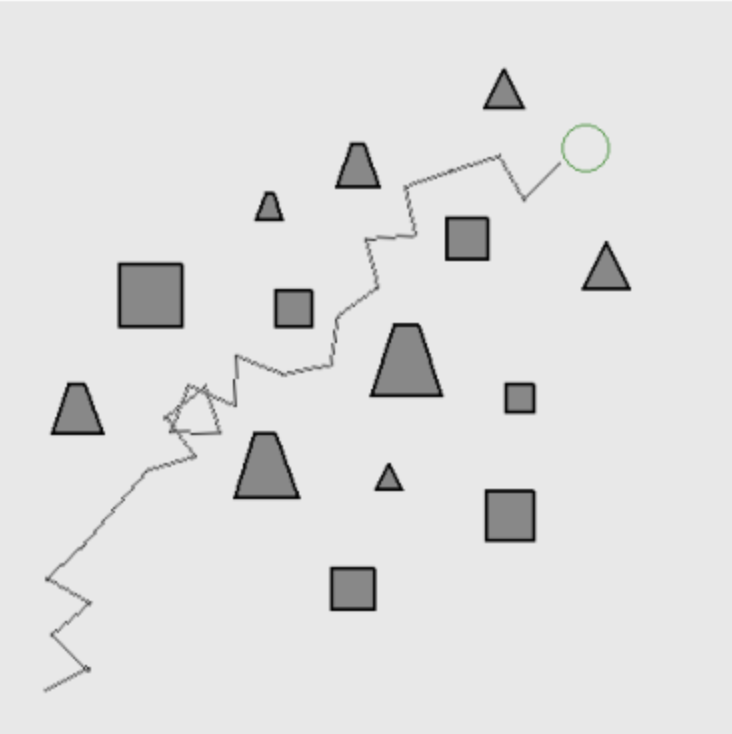}}
  \caption{Maximum reward}
\end{subfigure}%
\begin{subfigure}{.3\linewidth}
  \centering
  \fbox{\includegraphics[width=3.5cm]{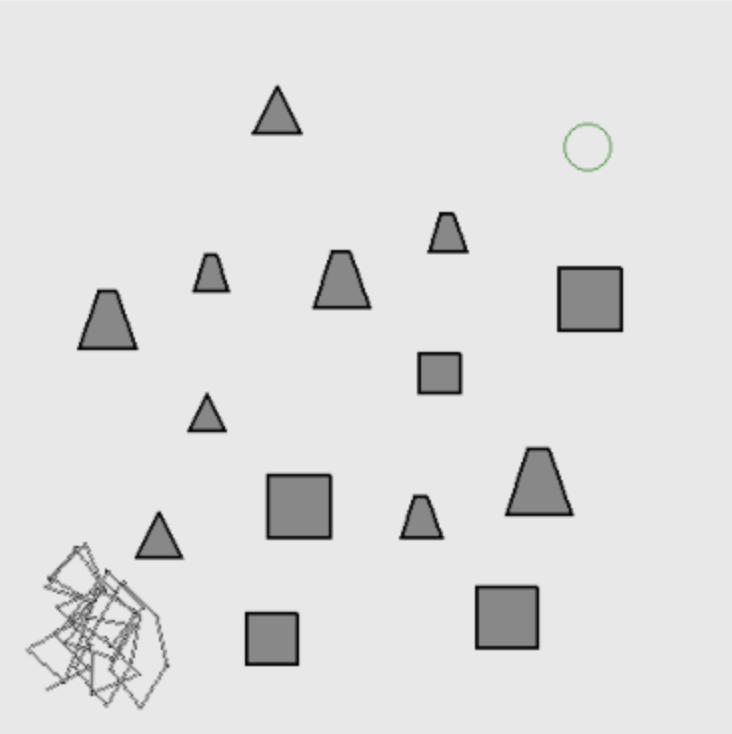}}
  \caption{Minimum reward}
\end{subfigure}
\begin{subfigure}{.3\linewidth}
  \centering
  \fbox{\includegraphics[width=3.5cm]{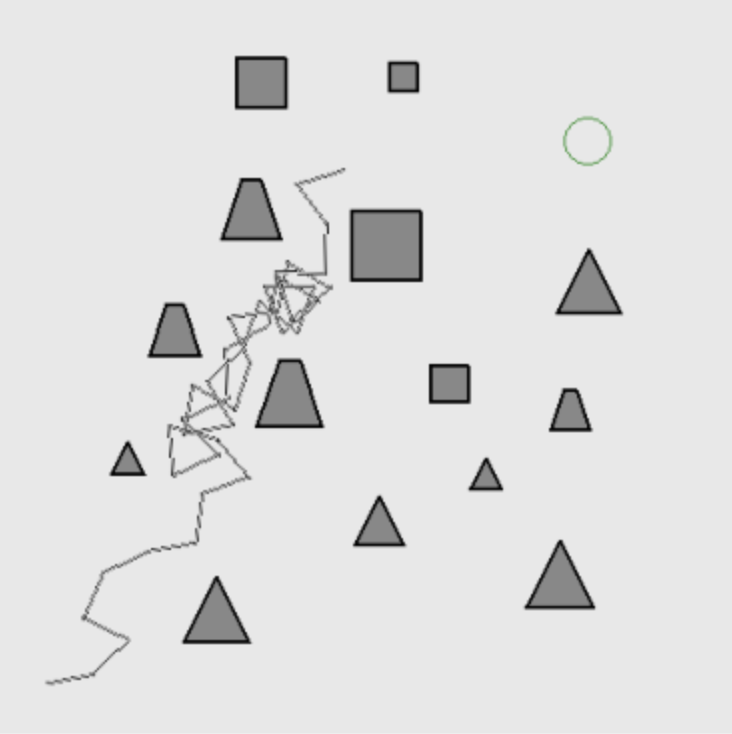}}
  \caption{Random}
\end{subfigure}
\caption{MPS-TD3 with random replacement trajectory examples}
\end{figure}

\begin{figure}[t]
\centering
\begin{subfigure}{.3\linewidth}
  \centering
  \fbox{\includegraphics[width=3.5cm]{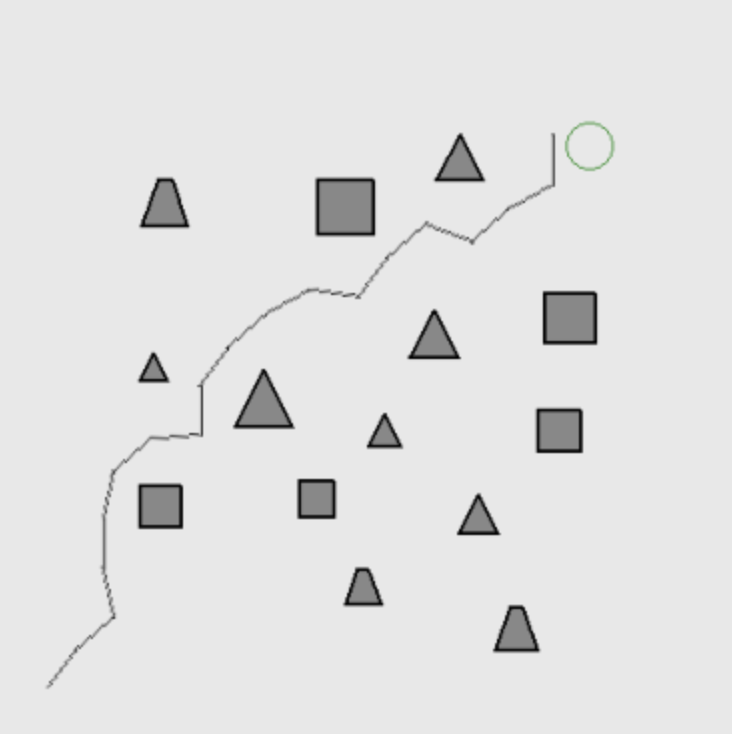}}
  \caption{Maximum reward}
\end{subfigure}%
\begin{subfigure}{.3\linewidth}
  \centering
  \fbox{\includegraphics[width=3.5cm]{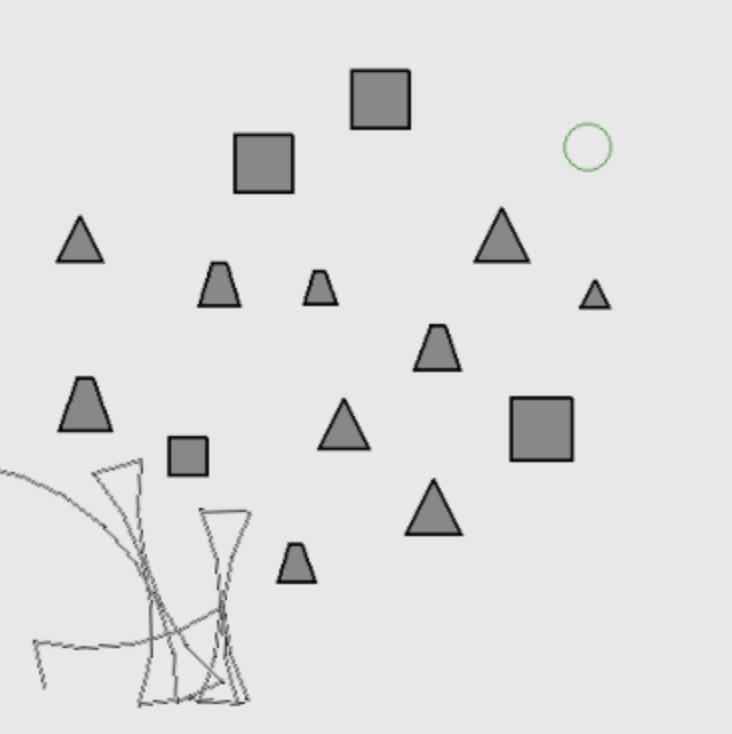}}
  \caption{Minimum reward}
\end{subfigure}
\begin{subfigure}{.3\linewidth}
  \centering
  \fbox{\includegraphics[width=3.5cm]{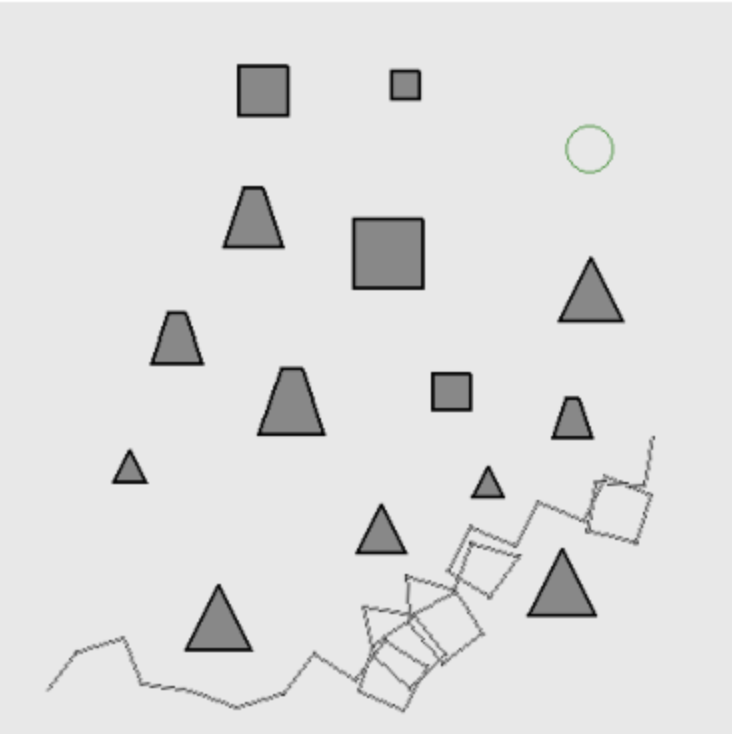}}
  \caption{Random}
\end{subfigure}
\caption{PAM trajectory examples}
\label{app:exp1_3_trajectory_end}
\end{figure}

\begin{figure}[h]
\centering
\begin{subfigure}[b]{\linewidth}
  \centering
  \importpgf{experiment_1/evaluation_2/reward_t}{on_policy_t_reward.pgf}
  \caption{On-policy}
\end{subfigure}
\\[3ex]
\begin{subfigure}[b]{\linewidth}
  \centering
   \importpgf{experiment_1/evaluation_2/reward_t}{off_policy_t_reward.pgf}
  \caption{Off-policy}
\end{subfigure}
\caption{Welch's t-test p-values between average returns}
\label{app:exp1_3_test_reward}
\end{figure}

\begin{figure}[h]
\centering
\begin{subfigure}[b]{\linewidth}
  \centering
   \importpgf{experiment_1/evaluation_2/reward_t}{on_policy_t_steps.pgf}
  \caption{On-policy}
\end{subfigure}
\\[3ex]
\begin{subfigure}[b]{\linewidth}
  \centering
   \importpgf{experiment_1/evaluation_2/reward_t}{off_policy_t_steps.pgf}
  \caption{Off-policy}
\end{subfigure}
\caption{Welch's t-test p-values between average steps}
\label{app:exp1_3_test_steps}
\end{figure}

\begin{figure}[h]
\centering
\begin{subfigure}[b]{\linewidth}
  \centering
  \importpgf{experiment_1/evaluation_2/reward_t}{on_policy_t_solved.pgf}
  \caption{On-policy}
\end{subfigure}
\\[3ex]
\begin{subfigure}[b]{\linewidth}
  \centering
  \importpgf{experiment_1/evaluation_2/reward_t}{off_policy_t_solved.pgf}
  \caption{Off-policy}
\end{subfigure}
\caption{Welch's t-test p-values between the fractions of solved environments}
\label{app:exp1_3_test_solved}
\end{figure}

\clearpage
\section{Learning With Dynamic Obstacles}

\subsection{Training}
\label{app:exp2_0}

\begin{figure}[h]
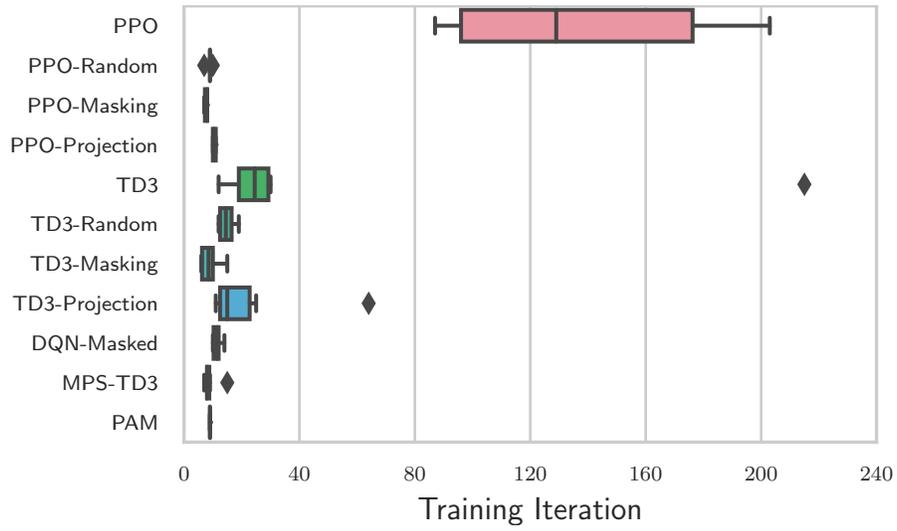

\centering
  \centering
  \importpgf{experiment_2/training}{boxplots.pgf}
\caption{Iterations until convergence (80\% solved episodes on average)}
\label{app:exp2_0_boxplots}
\end{figure}

\begin{figure}[!h]
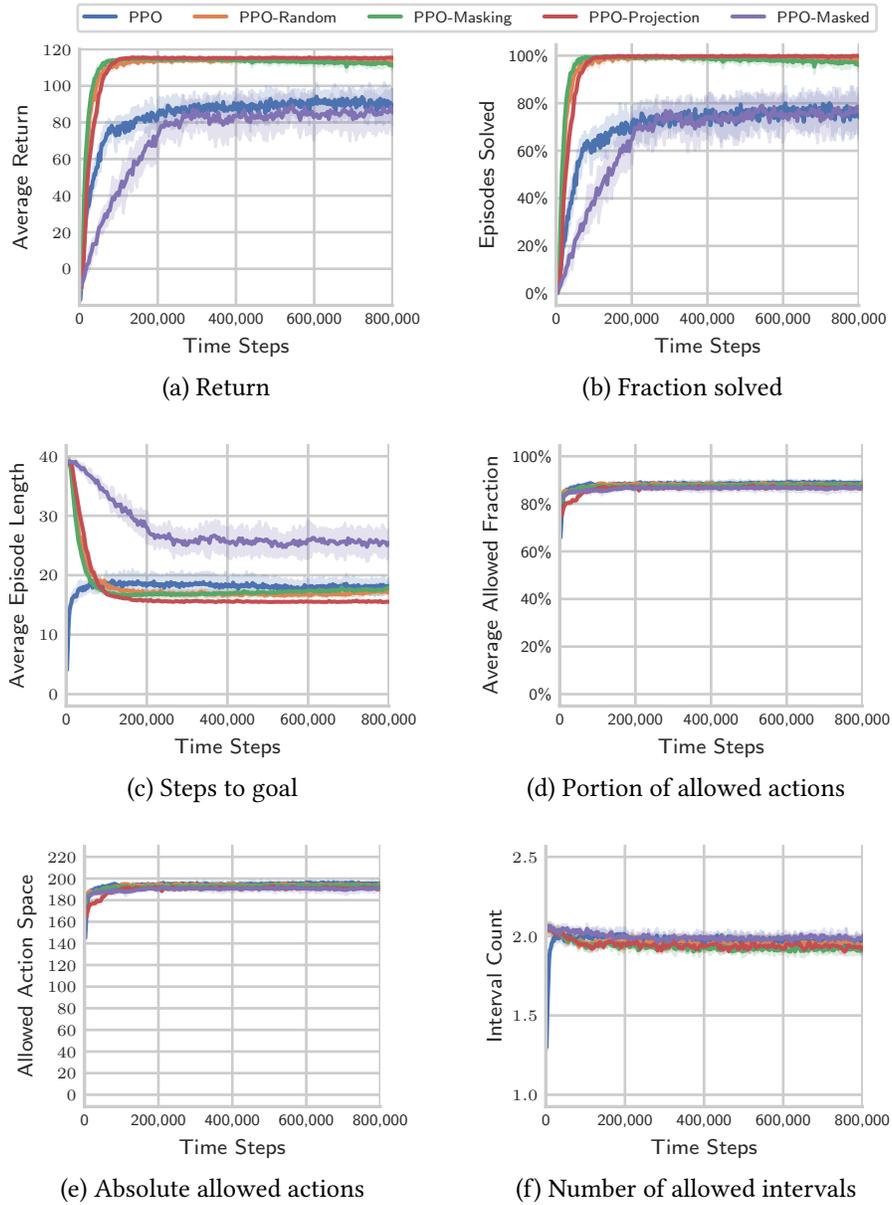

\centering
\begin{subfigure}{0.8\linewidth}
  \centering
  \importpgf{experiment_1/training/ppo}{legend.pgf}
\end{subfigure}
\begin{subfigure}{.49\linewidth}
  \centering
  \importpgf{experiment_2/training/ppo}{return_ppo.pgf}
  \caption{Return}
\end{subfigure}%
\begin{subfigure}{.49\linewidth}
  \centering
  \importpgf{experiment_2/training/ppo}{solved_ppo.pgf}
  \caption{Fraction solved}
\end{subfigure}
\\[3ex]
\begin{subfigure}{.49\linewidth}
  \centering
  \importpgf{experiment_2/training/ppo}{steps_ppo.pgf}
  \caption{Steps to goal}
  \label{app:exp2_0_steps_ppo}
\end{subfigure}
\begin{subfigure}{.49\linewidth}
  \centering
  \importpgf{experiment_2/training/ppo}{fraction_allowed_ppo.pgf}
  \caption{Portion of allowed actions}
\end{subfigure}
\\[3ex]
\begin{subfigure}{.49\linewidth}
  \centering
  \importpgf{experiment_2/training/ppo}{allowed_ppo.pgf}
  \caption{Absolute allowed actions}
\end{subfigure}
\begin{subfigure}{.49\linewidth}
  \centering
  \importpgf{experiment_2/training/ppo}{intervals_ppo.pgf}
  \caption{Number of allowed intervals}
  \label{app:exp2_0_count_ppo}
\end{subfigure}
\caption{On-policy training progress}
\label{app:exp2_0_ppo}
\end{figure}

\begin{figure}[!h]
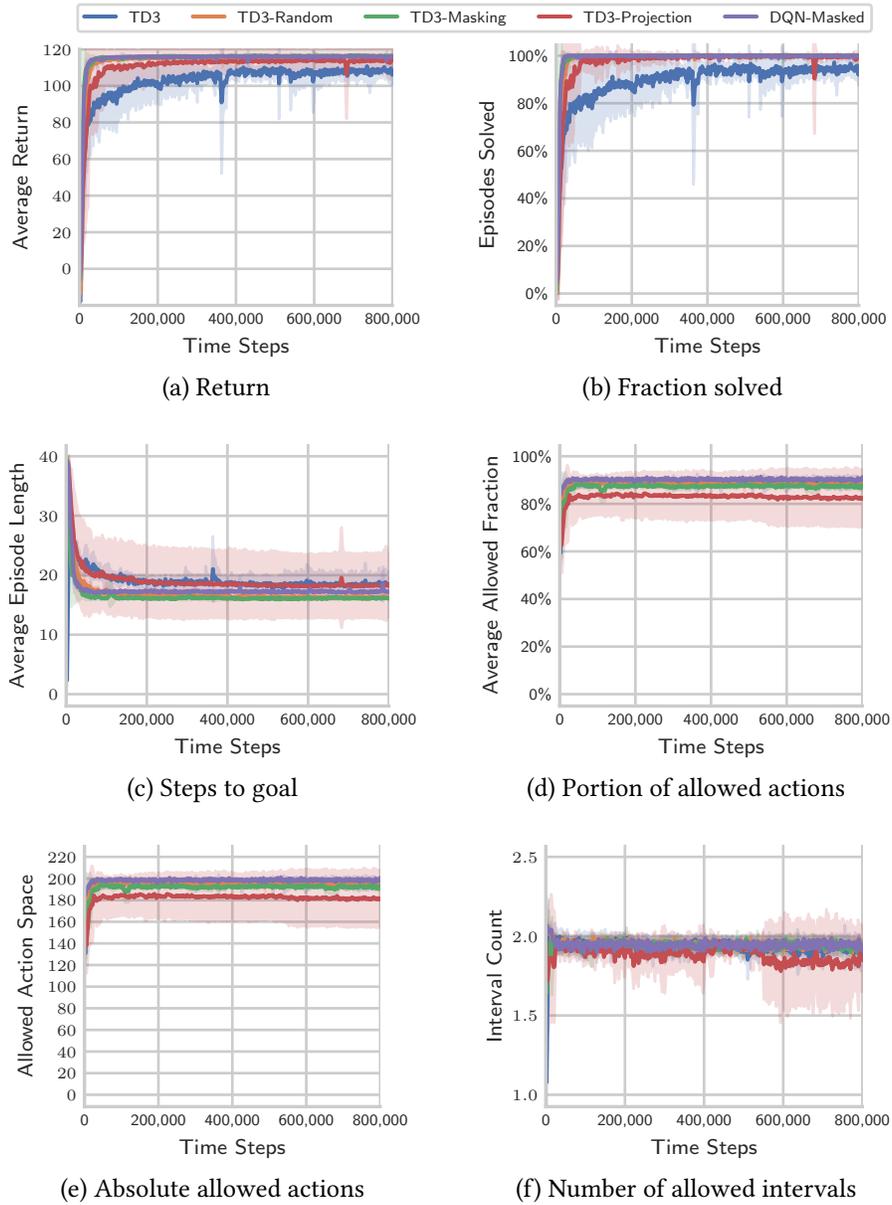

\centering
\begin{subfigure}{0.8\linewidth}
  \centering
  \importpgf{experiment_1/training/ddpg}{legend.pgf}
\end{subfigure}
\begin{subfigure}{.49\linewidth}
  \centering
  \importpgf{experiment_2/training/ddpg}{return_ddpg.pgf}
  \caption{Return}
\end{subfigure}%
\begin{subfigure}{.49\linewidth}
  \centering
  \importpgf{experiment_2/training/ddpg}{solved_ddpg.pgf}
  \caption{Fraction solved}
\end{subfigure}
\\[3ex]
\begin{subfigure}{.49\linewidth}
  \centering
  \importpgf{experiment_2/training/ddpg}{steps_ddpg.pgf}
  \caption{Steps to goal}
  \label{app:exp2_0_steps_td3}
\end{subfigure}
\begin{subfigure}{.49\linewidth}
  \centering
  \importpgf{experiment_2/training/ddpg}{fraction_allowed_ddpg.pgf}
  \caption{Portion of allowed actions}
\end{subfigure}
\\[3ex]
\begin{subfigure}{.49\linewidth}
  \centering
  \importpgf{experiment_2/training/ddpg}{allowed_ddpg.pgf}
  \caption{Absolute allowed actions}
\end{subfigure}
\begin{subfigure}{.49\linewidth}
  \centering
  \importpgf{experiment_2/training/ddpg}{intervals_ddpg.pgf}
  \caption{Number of allowed intervals}
  \label{app:exp2_0_count_td3}
\end{subfigure}
\caption{Off-policy training progress}
\label{app:exp2_0_td3}
\end{figure}

\begin{figure}[!h]
\centering
\begin{subfigure}{0.8\linewidth}
  \centering
  \importpgf{experiment_2/training/own}{legend.pgf}
\end{subfigure}
\begin{subfigure}{.49\linewidth}
  \centering
  \importpgf{experiment_1training/own}{return_own.pgf}
  \caption{Return}
\end{subfigure}%
\begin{subfigure}{.49\linewidth}
  \centering
  \importpgf{experiment_2/training/own}{solved_own.pgf}
  \caption{Fraction solved}
  \label{app:exp2_0_solved_own}
\end{subfigure}
\\[3ex]
\begin{subfigure}{.49\linewidth}
  \centering
  \importpgf{experiment_2/training/own}{steps_own.pgf}
  \caption{Steps to goal}
\end{subfigure}
\begin{subfigure}{.49\linewidth}
  \centering
  \importpgf{experiment_2/training/own}{fraction_allowed_own.pgf}
  \caption{Portion of allowed actions}
\end{subfigure}
\\[3ex]
\begin{subfigure}{.49\linewidth}
  \centering
  \importpgf{experiment_2/training/own}{allowed_own.pgf}
  \caption{Absolute allowed actions}
\end{subfigure}
\begin{subfigure}{.49\linewidth}
  \centering
  \importpgf{experiment_2/training/own}{intervals_own.pgf}
  \caption{Number of allowed intervals}
  \label{app:exp2_0_count_own}
\end{subfigure}
\caption{MPS-TD3 and PAM training progress}
\label{app:exp2_0_own}
\end{figure}

\clearpage
\subsection{Evaluation Without Restrictions}
\label{app:exp2_1}

\begin{table}[!h]
\scriptsize
\centering
\begin{tabular}{llll}
\hline
\multicolumn{1}{c}{\textbf{Approach}} & \multicolumn{1}{c}{\textbf{Return}} & \multicolumn{1}{c}{\textbf{Steps}} & \multicolumn{1}{c}{\textbf{Solved}} \\ \hline
TD3                                   & $115.02 \pm 2.18$                   & $16.50 \pm 1.76$                   & $100.00\% \pm 0.00\%$               \\
TD3-Projection                        & $114.15 \pm 6.11$                   & $17.83 \pm 5.49$                   & $100.00\% \pm 0.00\%$               \\
TD3-Masking                           & $117.54 \pm 0.27$                   & $16.00 \pm 0.00$                   & $100.00\% \pm 0.00\%$               \\
TD3-Random                            & $115.56 \pm 5.11$                   & $17.50 \pm 3.67$                   & $100.00\% \pm 0.00\%$               \\
DQN-Masked                            & $117.61 \pm 0.09$                   & $16.00 \pm 0.00$                   & $100.00\% \pm 0.00\%$               \\ \hline

PPO                                   & $114.82 \pm 1.07$                   & $15.17 \pm 0.41$                   & $100.00\% \pm 0.00\%$               \\
PPO-Projection                        & $115.17 \pm 0.08$                   & $15.00 \pm 0.00$                   & $100.00\% \pm 0.00\%$               \\
PPO-Masking                           & $115.73 \pm 1.91$                   & $16.00 \pm 0.00$                   & $100.00\% \pm 0.00\%$               \\
PPO-Random                            & $115.03 \pm 0.23$                   & $15.00 \pm 0.00$                   & $100.00\% \pm 0.00\%$               \\
PPO-Masked                            & $115.04 \pm 1.54$                   & $17.00 \pm 0.89$                   & $100.00\% \pm 0.00\%$               \\ \hline

MPS-TD3                               & $115.76 \pm 1.68$                   & $15.67 \pm 0.52$                   & $100.00\% \pm 0.00\%$               \\
PAM                                   & $0.20 \pm 13.78$                    & $-$                                & $0.00\% \pm 0.00\%$                 \\ \hline
\end{tabular}
\caption{Average return, steps, and solved episodes}
\label{app:exp2_1_results}
\end{table}

\begin{table}[!h]
\scriptsize
\centering
\begin{tabular}{llll}
\hline
\multicolumn{1}{c}{\textbf{Approach}} & \multicolumn{1}{c}{\textbf{Intervals}} & \multicolumn{1}{c}{\textbf{Size}} & \multicolumn{1}{c}{\textbf{Fraction}} \\ \hline

TD3                                   & $1.00 \pm 0.00$                        & $211.56 \pm 5.93$                   & $96.16\% \pm 2.70\%$                  \\
TD3-Projection                        & $1.00 \pm 0.00$                        & $200.64 \pm 34.81$                  & $91.20\% \pm 15.82\%$                 \\
TD3-Masking                           & $1.00 \pm 0.00$                        & $214.81 \pm 0.44$                   & $97.64\% \pm 0.20\%$                  \\
TD3-Random                            & $1.00 \pm 0.00$                        & $210.05 \pm 12.09$                  & $95.48\% \pm 5.50\%$                  \\
DQN-Masked                            & $1.00 \pm 0.00$                        & $214.98 \pm 0.00$                   & $97.72\% \pm 0.00\%$                  \\ \hline

PPO                                   & $1.00 \pm 0.00$                        & $214.58 \pm 0.16$                   & $97.54\% \pm 0.07\%$                  \\
PPO-Projection                        & $1.00 \pm 0.00$                        & $214.65 \pm 0.00$                   & $97.57\% \pm 0.00\%$                  \\
PPO-Masking                           & $1.00 \pm 0.00$                        & $214.98 \pm 0.00$                   & $97.72\% \pm 0.00\%$                  \\
PPO-Random                            & $1.00 \pm 0.00$                        & $214.65 \pm 0.00$                   & $97.57\% \pm 0.00\%$                  \\
PPO-Masked                            & $1.00 \pm 0.00$                        & $213.85 \pm 2.30$                   & $97.21\% \pm 1.05\%$                  \\ \hline

MPS-TD3                               & $1.00 \pm 0.00$                        & $211.69 \pm 1.50$                   & $96.22\% \pm 0.68\%$                  \\
PAM                                   & $1.05 \pm 0.05$                        & $201.33 \pm 14.69$                  & $91.51\% \pm 6.68\%$                  \\ \hline
\end{tabular}
\caption{Average number of disjoint intervals and size of the action space}
\label{app:exp2_1_control}
\end{table}

\begin{table}[!h]
\scriptsize
\centering
\begin{tabular}{lllll}
\hline
\multicolumn{1}{c}{\textbf{Approach}} & \multicolumn{1}{c}{\textbf{Average}} & \multicolumn{1}{c}{\textbf{Minimum}} & \multicolumn{1}{c}{\textbf{Maximum}} & \textbf{Variance} \\ \hline

TD3 & $211.56 \pm 5.93$ & $211.56 \pm 5.93$ & $211.56 \pm 5.93$ & $0.00 \pm 0.00$ \\
TD3-Projection & $200.64 \pm 34.81$ & $200.64 \pm 34.81$ & $200.64 \pm 34.81$ & $0.00 \pm 0.00$ \\
TD3-Masking & $214.81 \pm 0.44$ & $214.81 \pm 0.44$ & $214.81 \pm 0.44$ & $0.00 \pm 0.00$ \\
TD3-Random & $210.05 \pm 12.09$ & $210.05 \pm 12.09$ & $210.05 \pm 12.09$ & $0.00 \pm 0.00$ \\
DQN-Masked & $214.98 \pm 0.00$ & $214.98 \pm 0.00$ & $214.98 \pm 0.00$ & $0.00 \pm 0.00$ \\ \hline

PPO & $214.58 \pm 0.16$ & $214.58 \pm 0.16$ & $214.58 \pm 0.16$ & $0.00 \pm 0.00$ \\
PPO-Projection & $214.65 \pm 0.00$ & $214.65 \pm 0.00$ & $214.65 \pm 0.00$ & $0.00 \pm 0.00$ \\
PPO-Masking & $214.98 \pm 0.00$ & $214.98 \pm 0.00$ & $214.98 \pm 0.00$ & $0.00 \pm 0.00$ \\
PPO-Random & $214.65 \pm 0.00$ & $214.65 \pm 0.00$ & $214.65 \pm 0.00$ & $0.00 \pm 0.00$ \\
PPO-Masked & $213.85 \pm 2.30$ & $213.85 \pm 2.30$ & $213.85 \pm 2.30$ & $0.00 \pm 0.00$ \\ \hline

MPS-TD3 & $211.69 \pm 1.50$ & $211.69 \pm 1.50$ & $211.69 \pm 1.50$ & $0.00 \pm 0.00$ \\
PAM & $198.99 \pm 17.84$ & $198.19 \pm 18.92$ & $199.8 \pm 16.78$ & $28.27 \pm 51.06$ \\ \hline
\end{tabular}
\caption{Size of the individual intervals}
\label{exp2_1_single}
\end{table}

\begin{figure}[t]
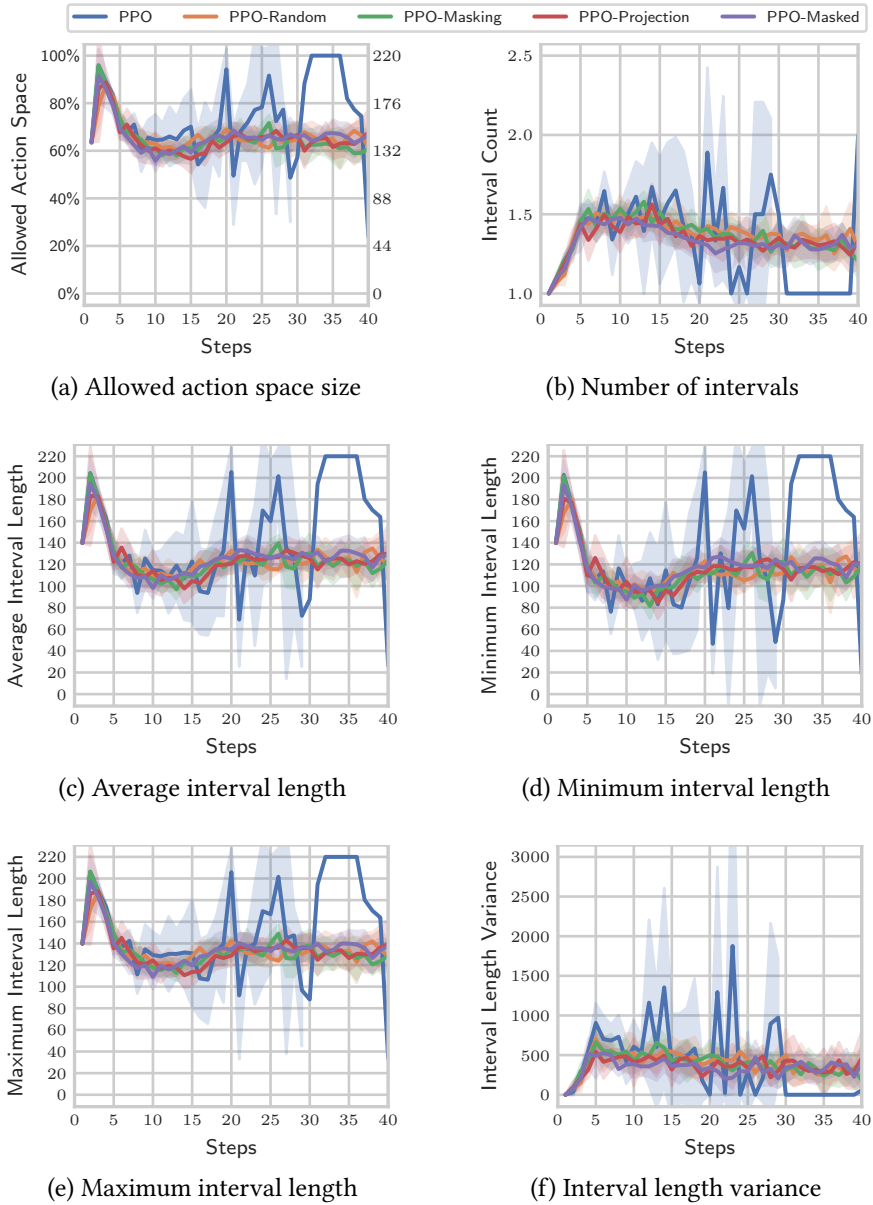

\centering
\begin{subfigure}{0.8\linewidth}
  \centering
  \importpgf{experiment_1/training/ppo}{legend.pgf}
\end{subfigure}
\begin{subfigure}{.49\linewidth}
  \centering
  \importpgf{experiment_2/evaluation_0/ppo}{allowed.pgf}
  \caption{Allowed action space size}
\end{subfigure}%
\begin{subfigure}{.49\linewidth}
  \centering
  \importpgf{experiment_2/evaluation_0/ppo}{count.pgf}
  \caption{Number of intervals}
  \label{app:exp2_1_count_ppo}
\end{subfigure}
\\[3ex]
\begin{subfigure}{.49\linewidth}
  \centering
  \importpgf{experiment_2/evaluation_0/ppo}{average.pgf}
  \caption{Average interval length}
\end{subfigure}
\begin{subfigure}{.49\linewidth}
  \centering
  \importpgf{experiment_2/evaluation_0/ppo}{minimum.pgf}
  \caption{Minimum interval length}
\end{subfigure}
\\[3ex]
\begin{subfigure}{.49\linewidth}
  \centering
  \importpgf{experiment_2/evaluation_0/ppo}{maximum.pgf}
  \caption{Maximum interval length}
\end{subfigure}
\begin{subfigure}{.49\linewidth}
  \centering
  \importpgf{experiment_2/evaluation_0/ppo}{variance.pgf}
  \caption{Interval length variance}
\end{subfigure}
\caption{On-policy control variables}
\label{app:exp2_1_control_ppo}
\end{figure}

\begin{figure}[t]
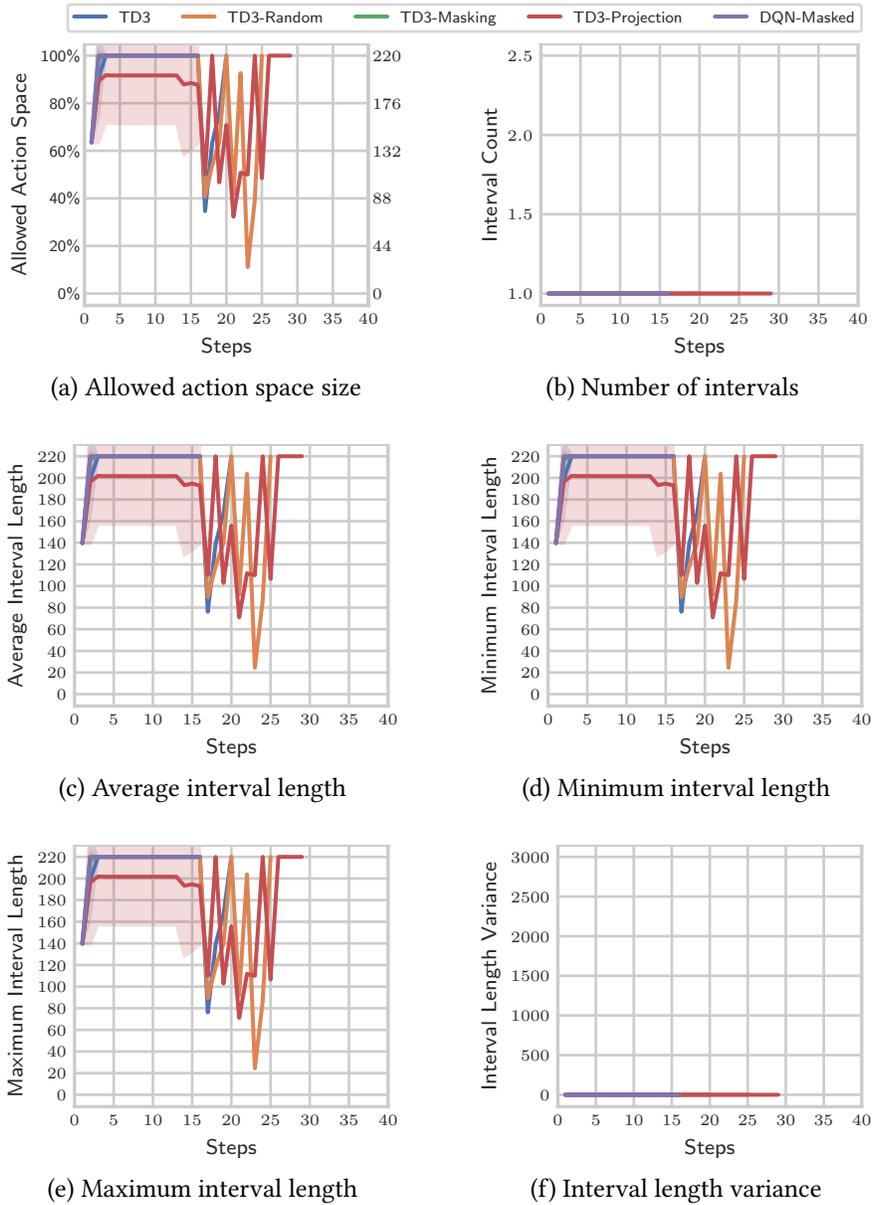

\centering
\begin{subfigure}{0.8\linewidth}
  \centering
  \importpgf{experiment_1/training/ddpg}{legend.pgf}
\end{subfigure}
\begin{subfigure}{.49\linewidth}
  \centering
  \importpgf{experiment_2/evaluation_0/td3}{allowed.pgf}
  \caption{Allowed action space size}
\end{subfigure}%
\begin{subfigure}{.49\linewidth}
  \centering
  \importpgf{experiment_2/evaluation_0/td3}{count.pgf}
  \caption{Number of intervals}
  \label{app:exp2_1_count_td3}
\end{subfigure}
\\[3ex]
\begin{subfigure}{.49\linewidth}
  \centering
  \importpgf{experiment_2/evaluation_0/td3}{average.pgf}
  \caption{Average interval length}
\end{subfigure}
\begin{subfigure}{.49\linewidth}
  \centering
  \importpgf{experiment_2/evaluation_0/td3}{minimum.pgf}
  \caption{Minimum interval length}
\end{subfigure}
\\[3ex]
\begin{subfigure}{.49\linewidth}
  \centering
  \importpgf{experiment_2/evaluation_0/td3}{maximum.pgf}
  \caption{Maximum interval length}
\end{subfigure}
\begin{subfigure}{.49\linewidth}
  \centering
  \importpgf{experiment_2/evaluation_0/td3}{variance.pgf}
  \caption{Interval length variance}
\end{subfigure}
\caption{Off-policy control variables}
\label{app:exp2_1_control_td3}
\end{figure}

\begin{figure}[t]
\centering
\begin{subfigure}{0.8\linewidth}
  \centering
  \importpgf{experiment_1/training/own}{legend.pgf}
\end{subfigure}
\begin{subfigure}{.49\linewidth}
  \centering
   \importpgf{experiment_2/evaluation_0/own}{allowed.pgf}
  \caption{Allowed action space size}
\end{subfigure}%
\begin{subfigure}{.49\linewidth}
  \centering
  \importpgf{experiment_2/evaluation_0/own}{count.pgf}
  \caption{Number of intervals}
  \label{app:exp2_1_count_own}
\end{subfigure}
\\[3ex]
\begin{subfigure}{.49\linewidth}
  \centering
  \importpgf{experiment_2/evaluation_0/own}{average.pgf}
  \caption{Average interval length}
\end{subfigure}
\begin{subfigure}{.49\linewidth}
  \centering
  \importpgf{experiment_2/evaluation_0/own}{minimum.pgf}
  \caption{Minimum interval length}
\end{subfigure}
\\[3ex]
\begin{subfigure}{.49\linewidth}
  \centering
  \importpgf{experiment_2/evaluation_0/own}{maximum.pgf}
  \caption{Maximum interval length}
\end{subfigure}
\begin{subfigure}{.49\linewidth}
  \centering
  \importpgf{experiment_2/evaluation_0/own}{variance.pgf}
  \caption{Interval length variance}
\end{subfigure}
\caption{MPS-TD3 and PAM control variables}
\label{app:exp2_1_control_own}
\end{figure}

\begin{figure}[h]
\centering
\begin{subfigure}[b]{\linewidth}
  \centering
  \importpgf{experiment_2/evaluation_0/reward_t}{on_policy_t_reward.pgf}
  \caption{On-policy}
\end{subfigure}
\\[3ex]
\begin{subfigure}[b]{\linewidth}
  \centering
  \importpgf{experiment_2/evaluation_0/reward_t}{off_policy_t_reward.pgf}
  \caption{Off-policy}
\end{subfigure}
\caption{Welch's t-test p-values between average returns}
\label{app:exp2_1_test_reward}
\end{figure}

\begin{figure}[h]
\centering
\begin{subfigure}[b]{\linewidth}
  \centering
  \importpgf{experiment_2/evaluation_0/reward_t}{on_policy_t_steps.pgf}
  \caption{On-policy}
\end{subfigure}
\\[3ex]
\begin{subfigure}[b]{\linewidth}
  \centering
  \importpgf{experiment_2/evaluation_0/reward_t}{off_policy_t_steps.pgf}
  \caption{Off-policy}
\end{subfigure}
\caption{Welch's t-test p-values between average steps}
\label{app:exp2_1_test_steps}
\end{figure}

\clearpage
\subsection{Evaluation With Simple Restrictions}
\label{app:exp2_2}

\begin{figure}[!h]
\centering
  \centering
  \importpgf{experiment_2/evaluation}{eval_rewards.pgf}
\caption{Average Return}
\label{app:exp2_2_plot_reward}
\end{figure}

\begin{figure}[!h]
\centering
  \centering
  \importpgf{experiment_2/evaluation}{eval_solved.pgf}
\caption{Average fraction of solved environments}
\label{app:exp2_2_plot_solved}
\end{figure}

\begin{figure}[!h]
\centering
  \centering
  \importpgf{experiment_2/evaluation}{eval_steps.pgf}
\caption{Average episode length}
\label{app:exp2_2_plot_steps}
\end{figure}

\begin{table}[h]
\scriptsize
\centering
\begin{tabular}{llll}
\hline
\multicolumn{1}{c}{\textbf{Approach}} & \multicolumn{1}{c}{\textbf{Return}} & \multicolumn{1}{c}{\textbf{Steps}} & \multicolumn{1}{c}{\textbf{Solved}} \\ \hline

TD3                                   & $107.03 \pm 9.62$                   & $19.10 \pm 1.58$                   & $94.00\% \pm 8.22\%$                \\
TD3-Projection                        & $114.23 \pm 1.98$                   & $17.23 \pm 0.33$                   & $99.00\% \pm 2.24\%$                \\
TD3-Masking                           & $110.53 \pm 6.44$                   & $17.53 \pm 0.16$                   & $95.83\% \pm 5.63\%$                \\
TD3-Random                            & $112.22 \pm 2.94$                   & $18.09 \pm 0.45$                   & $98.50\% \pm 2.24\%$                \\
DQN-Masked                            & $115.34 \pm 0.49$                   & $17.90 \pm 0.36$                   & $100.00\% \pm 0.00\%$               \\ \hline

PPO                                   & $90.77 \pm 15.85$                   & $19.61 \pm 1.27$                   & $78.33\% \pm 15.47\%$               \\
PPO-Projection                        & $90.73 \pm 6.75$                    & $16.39 \pm 0.25$                   & $80.00\% \pm 4.47\%$                \\
PPO-Masking                           & $111.61 \pm 2.63$                   & $17.12 \pm 0.41$                   & $97.08\% \pm 2.92\%$                \\
PPO-Random                            & $104.83 \pm 1.27$                   & $19.42 \pm 1.02$                   & $91.67\% \pm 2.04\%$                \\
PPO-Masked                            & $95.76 \pm 9.99$                    & $20.92 \pm 0.84$                   & $85.42\% \pm 9.14\%$                \\ \hline

MPS-TD3                               & $110.82 \pm 5.27$                   & $19.99 \pm 0.97$                   & $97.92\% \pm 5.10\%$                \\
PAM                                   & $2.67 \pm 11.72$                    & $-$                                & $0.00\% \pm 0.00\%$                 \\ \hline
\end{tabular}
\caption{Average return, steps, and solved episodes}
\label{app:exp2_2_results}
\end{table}

\begin{table}[h]
\scriptsize
\centering
\begin{tabular}{llll}
\hline
\multicolumn{1}{c}{\textbf{Approach}} & \multicolumn{1}{c}{\textbf{Intervals}} & \multicolumn{1}{c}{\textbf{Size}} & \multicolumn{1}{c}{\textbf{Fraction}} \\ \hline

TD3                                   & $1.06 \pm 0.02$                        & $191.65 \pm 10.88$                  & $87.11\% \pm 4.94\%$                  \\
TD3-Projection                        & $1.06 \pm 0.00$                        & $179.27	 \pm 4.73$                  & $81.48\% \pm 2.15\%$                  \\
TD3-Masking                           & $1.06 \pm 0.01$                        & $176.59 \pm 4.08$                   & $80.27\% \pm 1.85\%$                  \\
TD3-Random                            & $1.06 \pm 0.01$                        & $185.03	 \pm 4.24$                   & $84.11\% \pm 1.93\%$                  \\
DQN-Masked                            & $1.05 \pm0.01$                         & $183.71 \pm 3.19$                   & $83.50\% \pm 1.45\%$                  \\ \hline

PPO                                   & $1.08 \pm 0.02$                        & $192.99 \pm 3.73$                   & $87.72\% \pm 1.70\%$                  \\
PPO-Projection                        & $1.06 \pm0.00$                         & $165.93 \pm 5.13$                   & $75.42\% \pm 2.33\%$                  \\
PPO-Masking                           & $1.04 \pm 0.01$                        & $189.26 \pm 6.30$                   & $86.03\% \pm 2.86\%$                  \\
PPO-Random                            & $1.08 \pm 0.01$                        & $180.81 \pm 0.94$                   & $82.19\% \pm 0.43\%$                  \\
PPO-Masked                            & $1.07 \pm 0.01$                        & $180.42 \pm 3.66$                   & $82.01\% \pm 1.66\%$                  \\ \hline

MPS-TD3                               & $1.07 \pm 0.02$                        & $190.53 \pm 4.32$                   & $86.60\% \pm 1.97\%$                  \\
PAM                                   & $1.10 \pm 0.03$                        & $192.07 \pm 8.97$                   & $87.30\% \pm 4.08\%$                  \\ \hline
\end{tabular}
\caption{Average number of disjoint intervals and size of the action space}
\label{exp2_2_control}
\end{table}

\begin{table}[h]
\scriptsize
\centering
\begin{tabular}{lllll}
\hline
\multicolumn{1}{c}{\textbf{Approach}} & \multicolumn{1}{c}{\textbf{Average}} & \multicolumn{1}{c}{\textbf{Minimum}} & \multicolumn{1}{c}{\textbf{Maximum}} & \multicolumn{1}{c}{\textbf{Variance}} \\ \hline

TD3 & $187.48 \pm 11.46$ & $184.77 \pm 11.62$ & $190.18 \pm 11.33$ & $150.57 \pm 39.10$ \\
TD3-Projection & $175.27 \pm 4.83$ & $173.54 \pm 4.66$ & $177.00 \pm 5.07$ & $65.24 \pm 31.99$ \\
TD3-Masking & $172.52 \pm 4.41$ & $171.05 \pm 4.43$ & $173.99 \pm 4.45$ & $61.92 \pm 30.05$ \\
TD3-Random & $180.56 \pm 4.37$ & $179.43 \pm 4.39$ & $181.70 \pm 4.40$ & $39.07 \pm 19.09$ \\
DQN-Masked & $180.01 \pm 3.46$ & $178.16 \pm 3.13$ & $181.86 \pm 3.91$ & $95.43 \pm 64.22$ \\ \hline

PPO & $187.94 \pm 4.44$ & $185.29 \pm 5.09$ & $190.58 \pm 4.07$ & $117.63 \pm 84.64$ \\
PPO-Projection & $162.15 \pm 5.00$ & $161.00 \pm 5.09$ & $163.30 \pm 4.95$ & $32.70 \pm 16.90$ \\
PPO-Masking & $186.49 \pm 6.58$ & $184.44 \pm 6.62$ & $188.55 \pm 6.55$ & $123.00 \pm 19.83$ \\
PPO-Random & $175.44 \pm 0.97$ & $173.63 \pm 1.00$ & $177.24 \pm 1.07$ & $76.70 \pm 19.81$ \\
PPO-Masked & $175.87 \pm 3.41$ & $173.68 \pm 3.57$ & $178.07 \pm 3.32$ & $98.82 \pm 27.54$ \\ \hline

MPS-TD3 & $185.30 \pm 4.54$ & $182.89 \pm 4.46$ & $187.71 \pm 4.89$ & $127.62 \pm 71.50$ \\
PAM & $186.81 \pm 10.60$ & $184.50 \pm 11.08$ & $189.12 \pm 10.19$ & $98.49 \pm 38.23$ \\ \hline
\end{tabular}
\caption{Size of the individual intervals}
\label{exp2_2_single}
\end{table}

\begin{figure}[t]
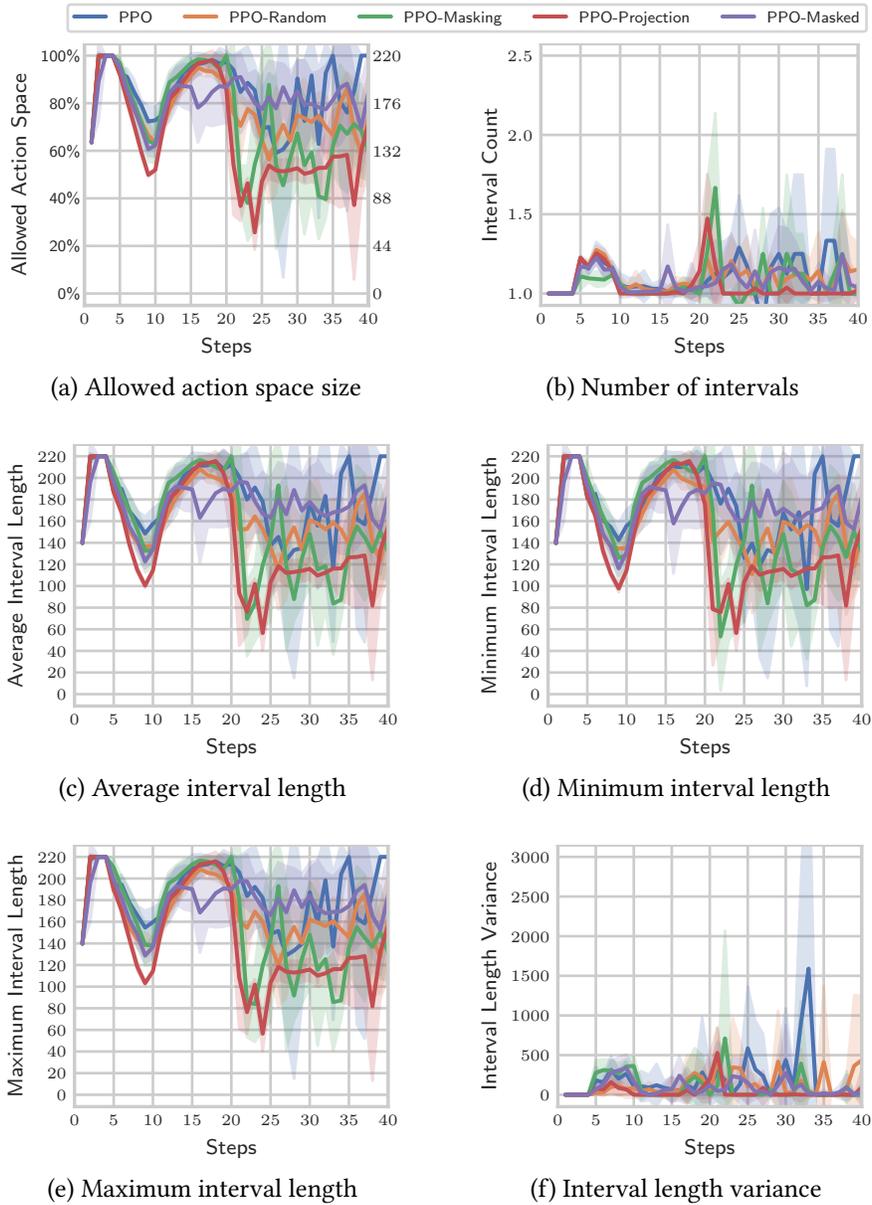

\centering
\begin{subfigure}{0.8\linewidth}
  \centering
  \importpgf{experiment_1/training/ppo}{legend.pgf}
\end{subfigure}
\begin{subfigure}{.49\linewidth}
  \centering
  \importpgf{experiment_2/evaluation/ppo}{allowed.pgf}
  \caption{Allowed action space size}
\end{subfigure}%
\begin{subfigure}{.49\linewidth}
  \centering
   \importpgf{experiment_2/evaluation/ppo}{count.pgf}
  \caption{Number of intervals}
\end{subfigure}
\\[3ex]
\begin{subfigure}{.49\linewidth}
  \centering
   \importpgf{experiment_2/evaluation/ppo}{average.pgf}
  \caption{Average interval length}
\end{subfigure}
\begin{subfigure}{.49\linewidth}
  \centering
   \importpgf{experiment_2/evaluation/ppo}{minimum.pgf}
  \caption{Minimum interval length}
\end{subfigure}
\\[3ex]
\begin{subfigure}{.49\linewidth}
  \centering
   \importpgf{experiment_2/evaluation/ppo}{maximum.pgf}
  \caption{Maximum interval length}
\end{subfigure}
\begin{subfigure}{.49\linewidth}
  \centering
   \importpgf{experiment_2/evaluation/ppo}{variance.pgf}
  \caption{Interval length variance}
  \label{app:exp2_2_variance_ppo}
\end{subfigure}
\caption{On-policy control variables}
\label{app:exp2_2_control_ppo}
\end{figure}

\begin{figure}[t]
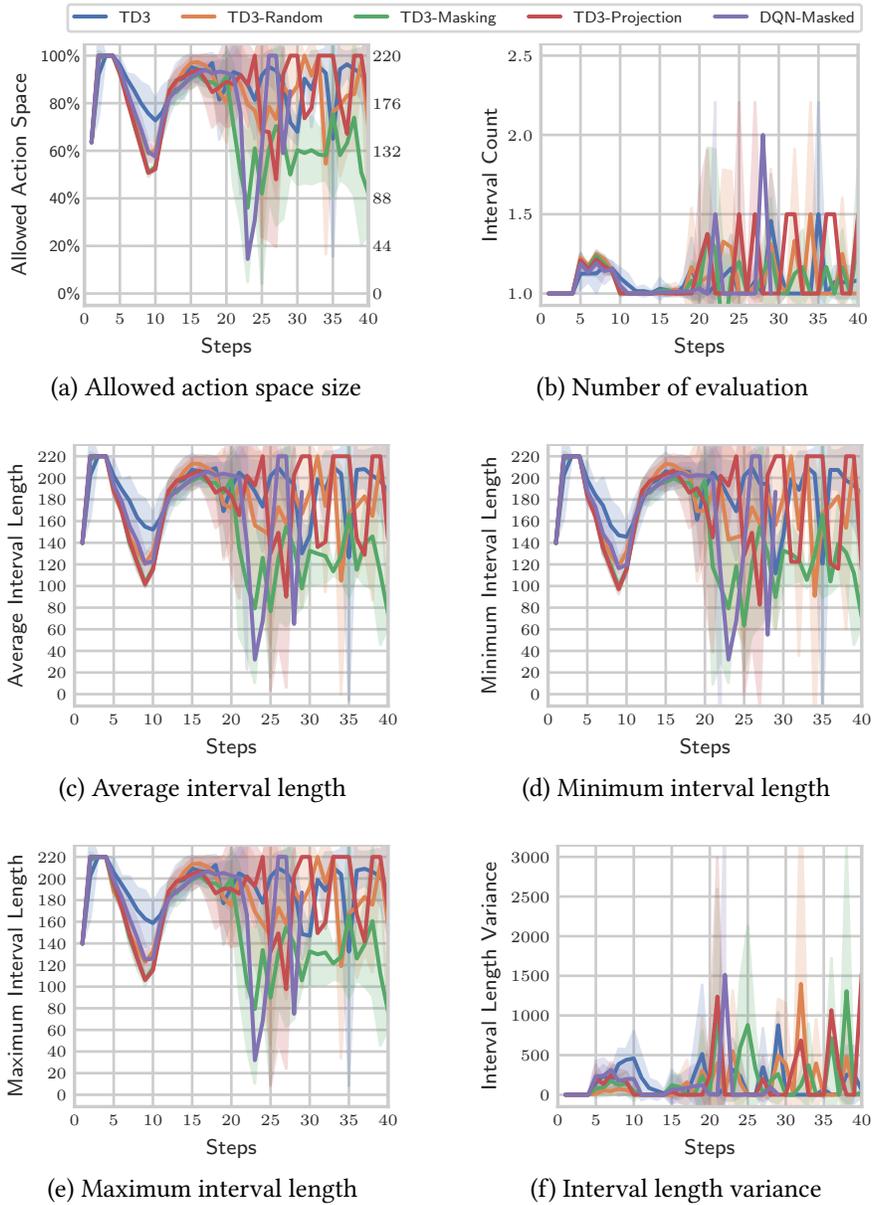

\centering
\begin{subfigure}{0.8\linewidth}
  \centering
 \importpgf{experiment_1/training/ddpg}{legend.pgf}
\end{subfigure}
\begin{subfigure}{.49\linewidth}
  \centering
  \importpgf{experiment_2/evaluation/td3}{allowed.pgf}
  \caption{Allowed action space size}
\end{subfigure}%
\begin{subfigure}{.49\linewidth}
  \centering
  \importpgf{experiment_2/evaluation/td3}{count.pgf}
  \caption{Number of evaluation}
\end{subfigure}
\\[3ex]
\begin{subfigure}{.49\linewidth}
  \centering
  \importpgf{experiment_2/evaluation/td3}{average.pgf}
  \caption{Average interval length}
\end{subfigure}
\begin{subfigure}{.49\linewidth}
  \centering
  \importpgf{experiment_2/evaluation/td3}{minimum.pgf}
  \caption{Minimum interval length}
\end{subfigure}
\\[3ex]
\begin{subfigure}{.49\linewidth}
  \centering
  \importpgf{experiment_2/evaluation/td3}{maximum.pgf}
  \caption{Maximum interval length}
\end{subfigure}
\begin{subfigure}{.49\linewidth}
  \centering
  \importpgf{experiment_2/evaluation/td3}{variance.pgf}
  \caption{Interval length variance}
  \label{app:exp2_2_variance_td3}
\end{subfigure}
\caption{Off-policy control variables}
\label{app:exp2_2_control_td3}
\end{figure}

\begin{figure}[t]
\centering
\begin{subfigure}{0.8\linewidth}
  \centering
   \importpgf{experiment_1/training/own}{legend.pgf}
\end{subfigure}
\begin{subfigure}{.49\linewidth}
  \centering
   \importpgf{experiment_2/evaluation/own}{allowed.pgf}
  \caption{Allowed action space size}
\end{subfigure}%
\begin{subfigure}{.49\linewidth}
  \centering
  \importpgf{experiment_2/evaluation/own}{count.pgf}
  \caption{Number of intervals}
\end{subfigure}
\\[3ex]
\begin{subfigure}{.49\linewidth}
  \centering
  \importpgf{experiment_2/evaluation/own}{average.pgf}
  \caption{Average interval length}
\end{subfigure}
\begin{subfigure}{.49\linewidth}
  \centering
  \importpgf{experiment_2/evaluation/own}{minimum.pgf}
  \caption{Minimum interval length}
\end{subfigure}
\\[3ex]
\begin{subfigure}{.49\linewidth}
  \centering
  \importpgf{experiment_2/evaluation/own}{maximum.pgf}
  \caption{Maximum interval length}
\end{subfigure}
\begin{subfigure}{.49\linewidth}
  \centering
  \importpgf{experiment_2/evaluation/own}{variance.pgf}
  \caption{Interval length variance}
  \label{app:exp2_2_variance_own}
\end{subfigure}
\caption{MPS-TD3 and PAM control variables}
\label{app:exp2_2_control_own}
\end{figure}

\begin{figure}[t]
\centering
\begin{subfigure}{.3\linewidth}
  \centering
  \fbox{\includegraphics[width=3.5cm]{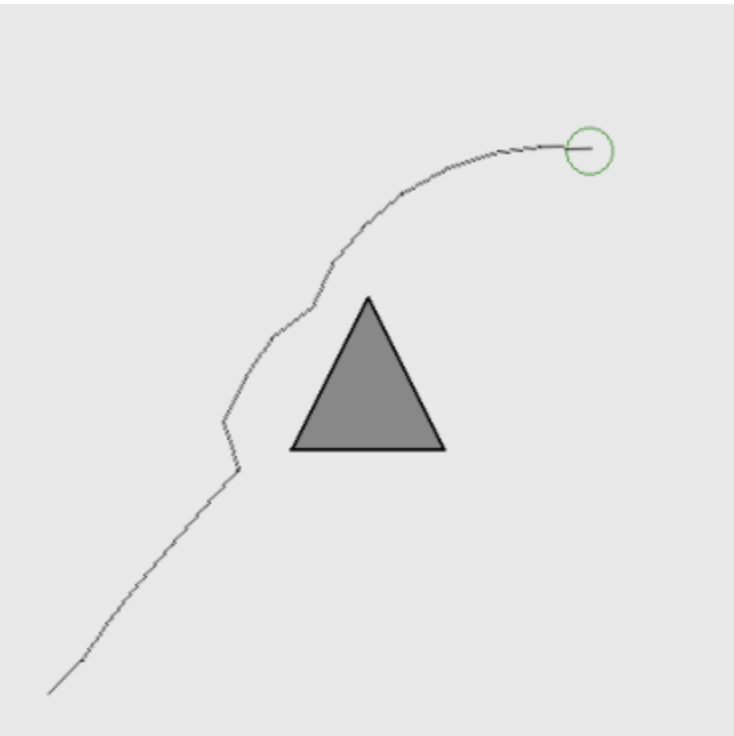}}
  \caption{Maximum reward}
\end{subfigure}%
\begin{subfigure}{.3\linewidth}
  \centering
  \fbox{\includegraphics[width=3.5cm]{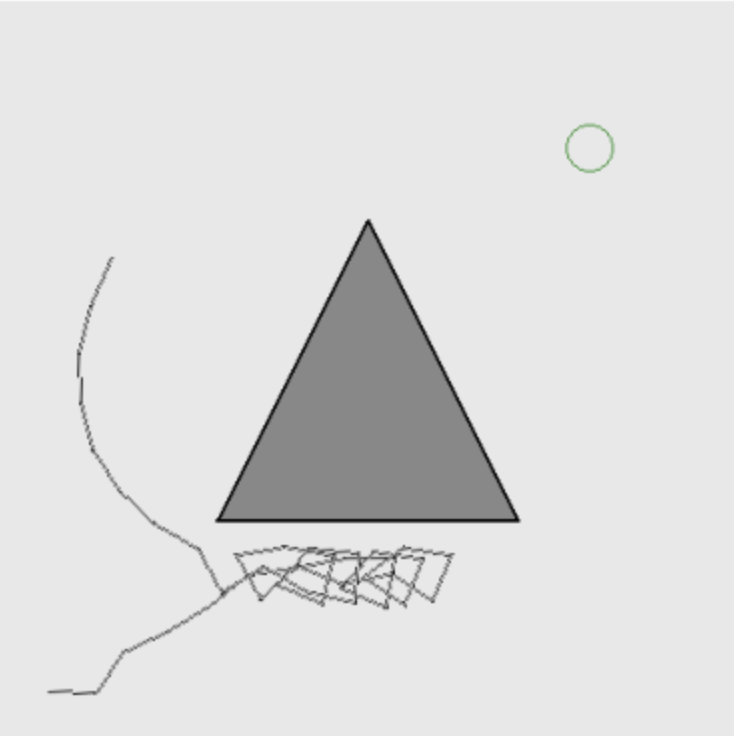}}
  \caption{Minimum reward}
\end{subfigure}
\begin{subfigure}{.3\linewidth}
  \centering
  \fbox{\includegraphics[width=3.5cm]{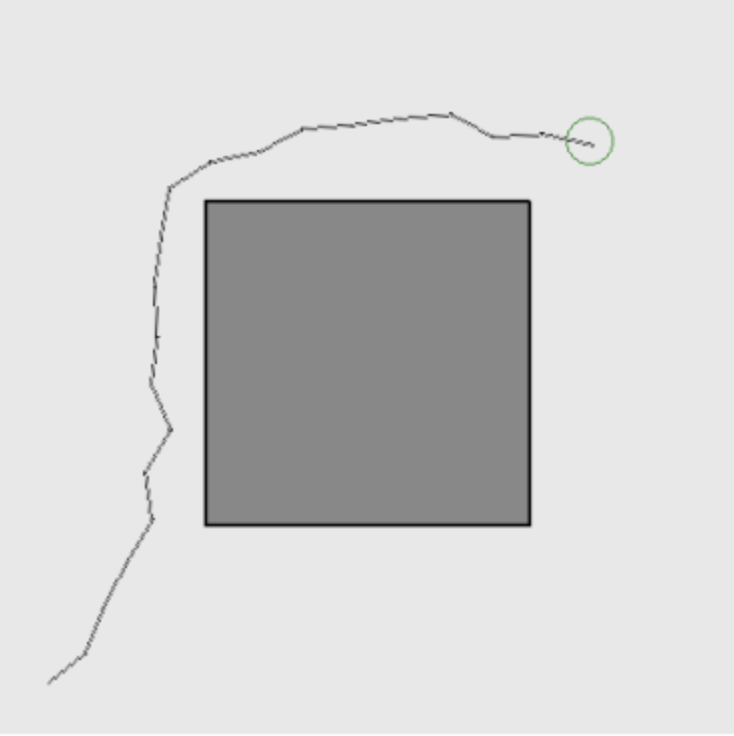}}
  \caption{Random}
\end{subfigure}
\caption{TD3 trajectory examples}
\label{app:exp2_2_trajectory_start}
\end{figure}

\begin{figure}[t]
\centering
\begin{subfigure}{.3\linewidth}
  \centering
  \fbox{\includegraphics[width=3.5cm]{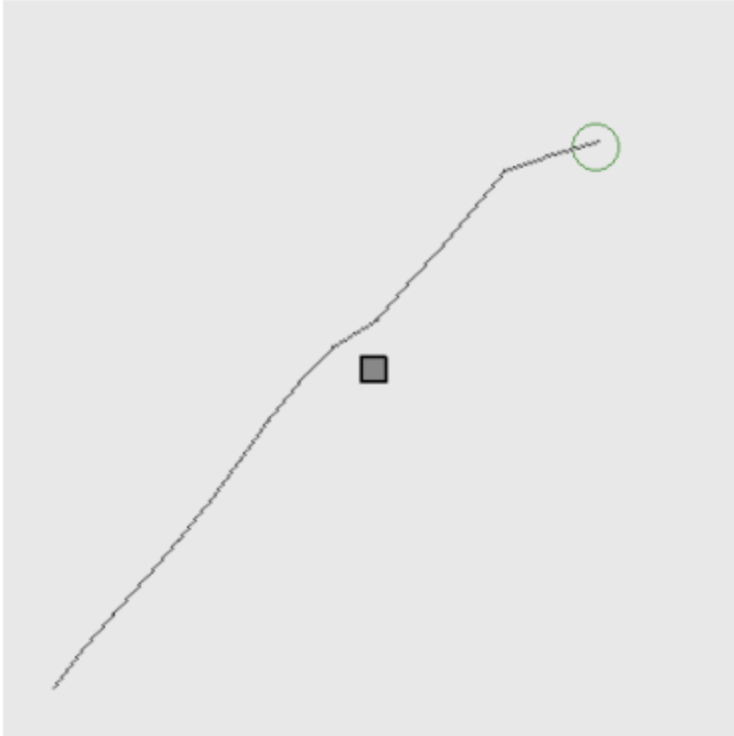}}
  \caption{Maximum reward}
\end{subfigure}%
\begin{subfigure}{.3\linewidth}
  \centering
  \fbox{\includegraphics[width=3.5cm]{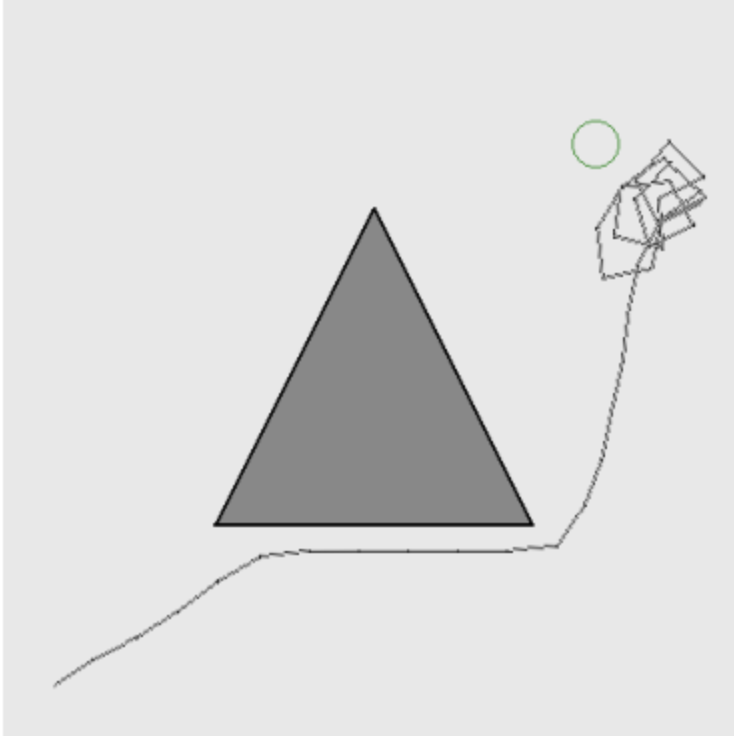}}
  \caption{Minimum reward}
\end{subfigure}
\begin{subfigure}{.3\linewidth}
  \centering
  \fbox{\includegraphics[width=3.5cm]{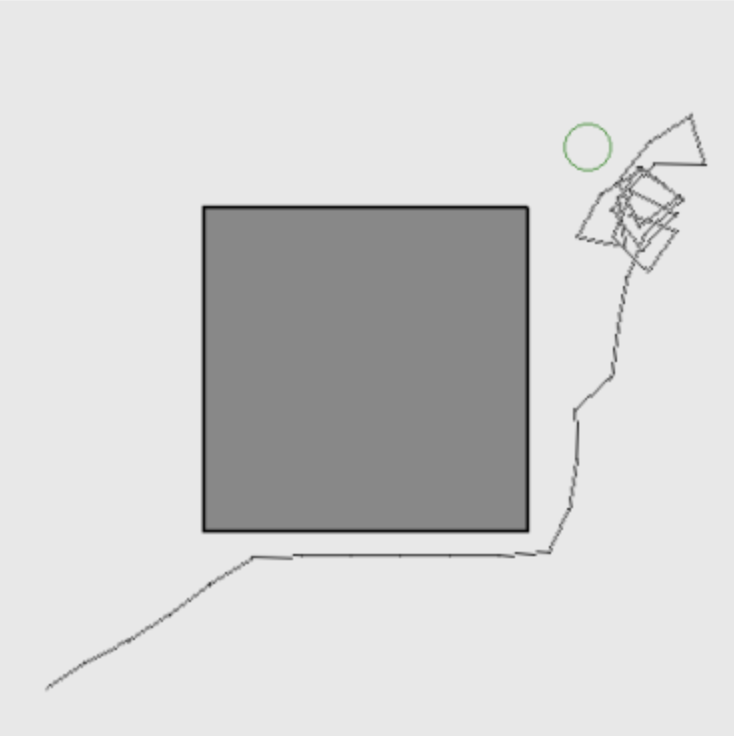}}
  \caption{Random}
\end{subfigure}
\caption{TD3 with projection trajectory examples}
\end{figure}

\begin{figure}[t]
\centering
\begin{subfigure}{.3\linewidth}
  \centering
  \fbox{\includegraphics[width=3.5cm]{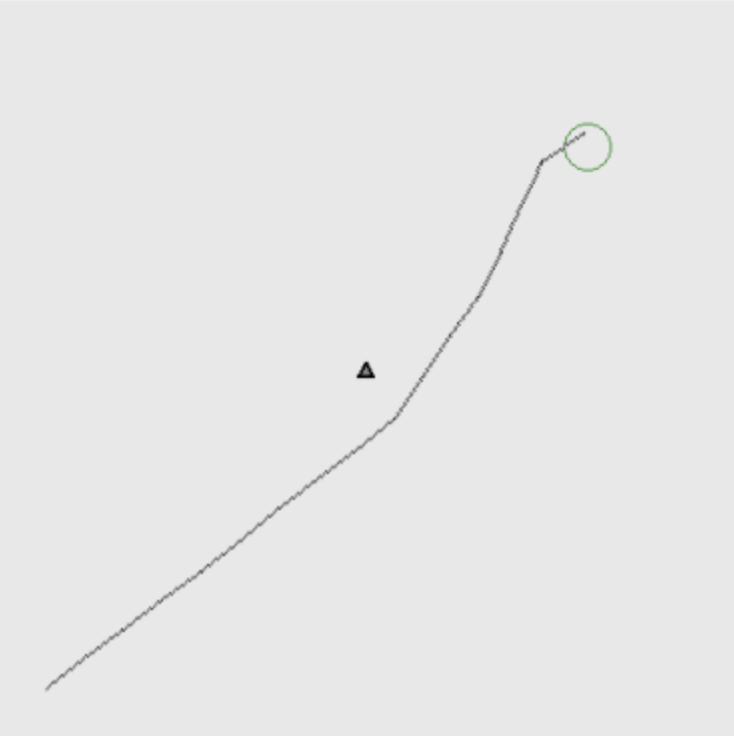}}
  \caption{Maximum reward}
\end{subfigure}%
\begin{subfigure}{.3\linewidth}
  \centering
  \fbox{\includegraphics[width=3.5cm]{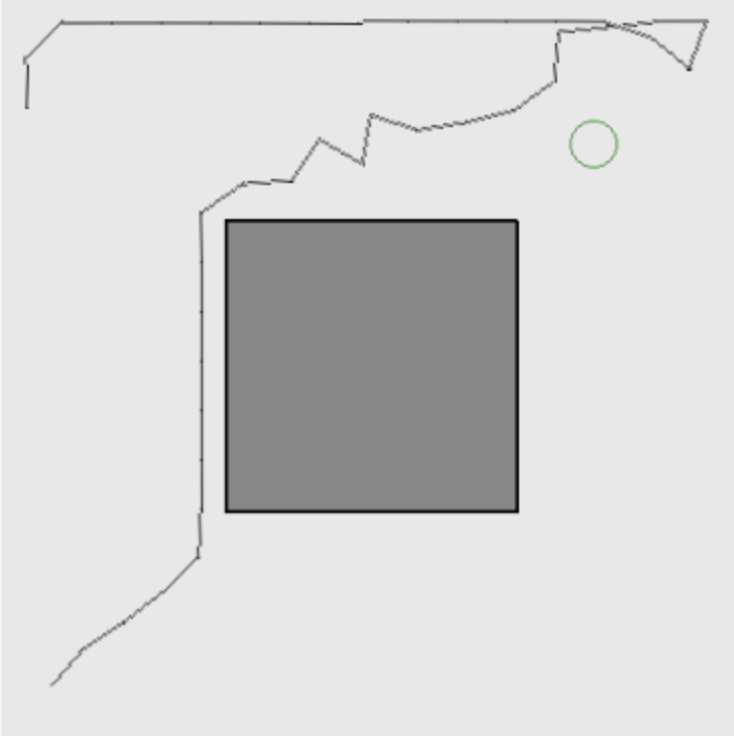}}
  \caption{Minimum reward}
\end{subfigure}
\begin{subfigure}{.3\linewidth}
  \centering
  \fbox{\includegraphics[width=3.5cm]{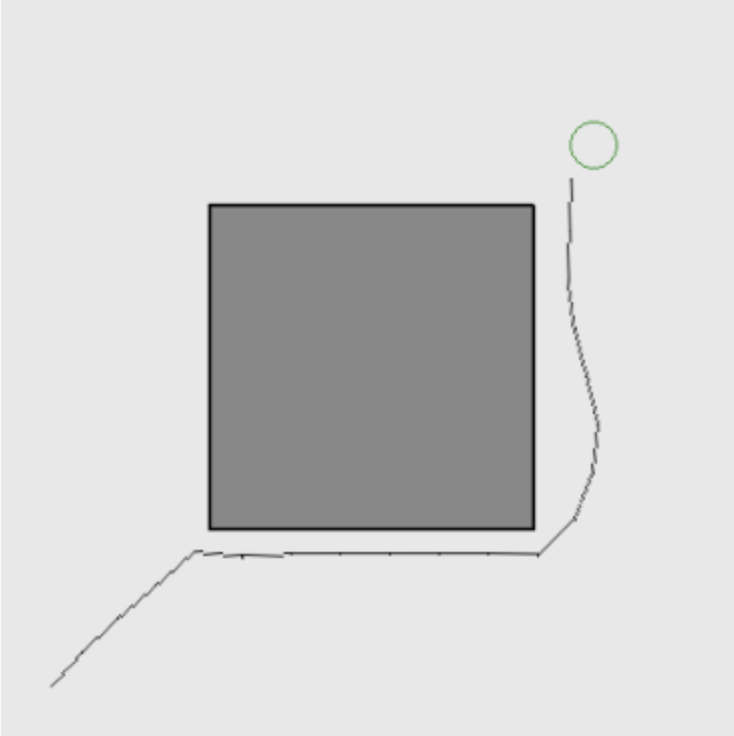}}
  \caption{Random}
\end{subfigure}
\caption{TD3 with continuous masking trajectory examples}
\end{figure}

\begin{figure}[t]
\centering
\begin{subfigure}{.3\linewidth}
  \centering
  \fbox{\includegraphics[width=3.5cm]{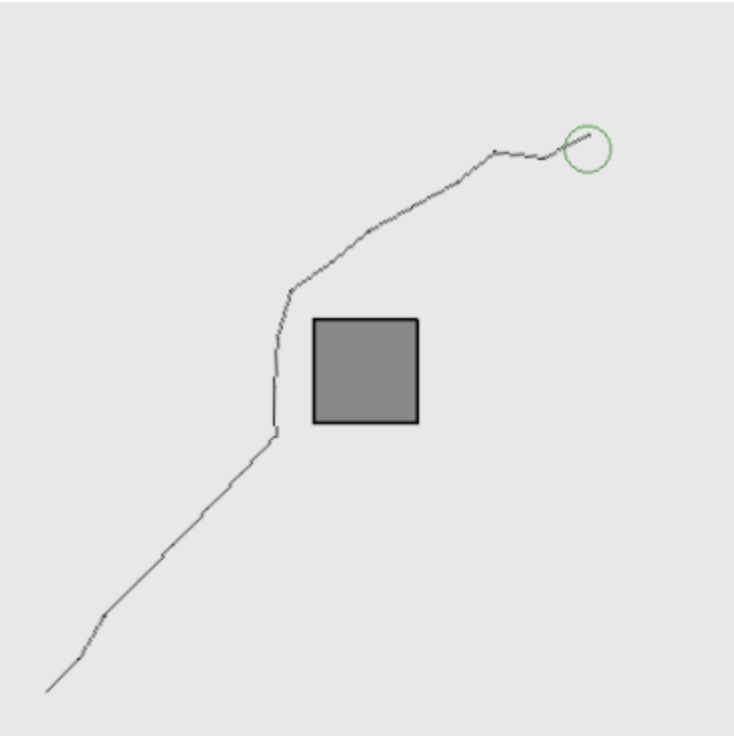}}
  \caption{Maximum reward}
\end{subfigure}%
\begin{subfigure}{.3\linewidth}
  \centering
  \fbox{\includegraphics[width=3.5cm]{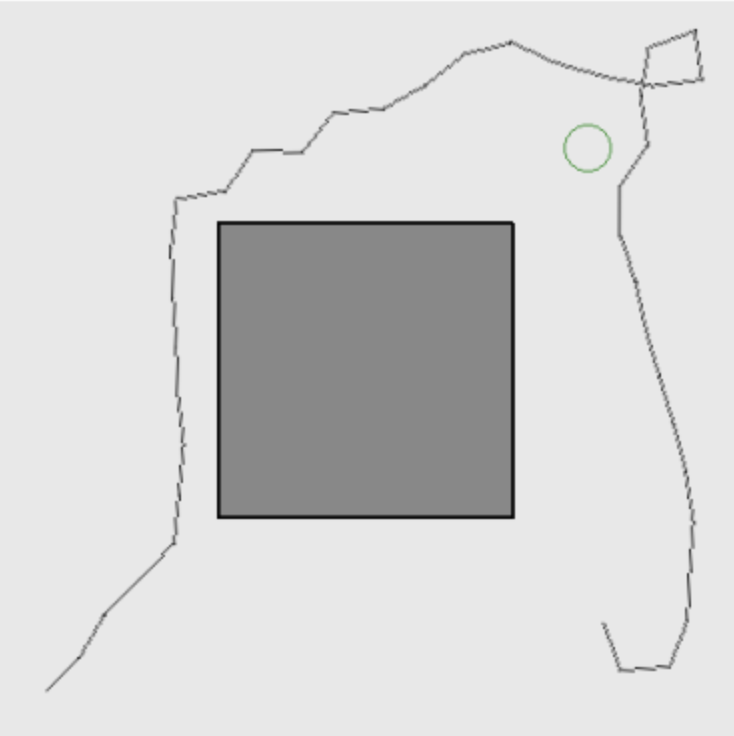}}
  \caption{Minimum reward}
\end{subfigure}
\begin{subfigure}{.3\linewidth}
  \centering
  \fbox{\includegraphics[width=3.5cm]{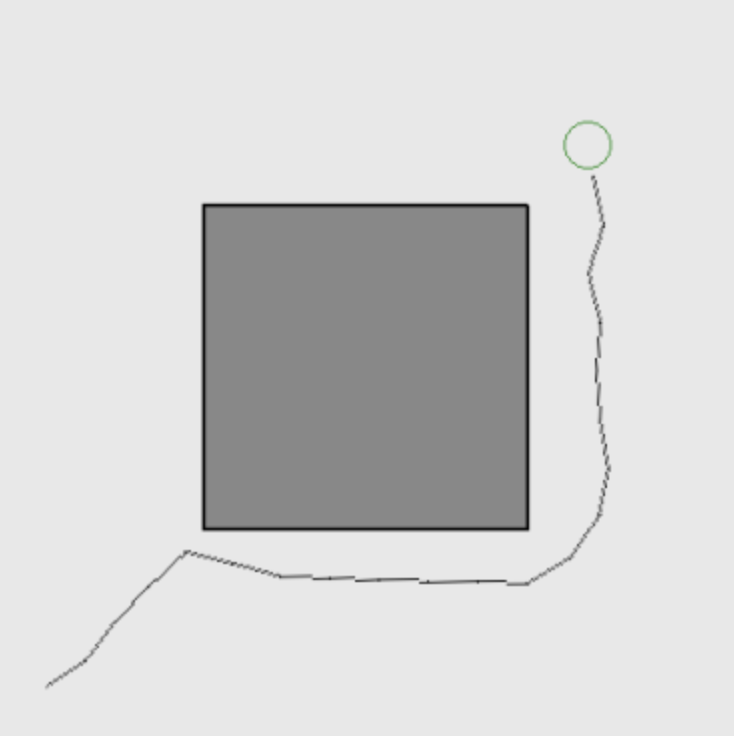}}
  \caption{Random}
\end{subfigure}
\caption{TD3 with random replacement trajectory examples}
\end{figure}

\begin{figure}[t]
\centering
\begin{subfigure}{.3\linewidth}
  \centering
  \fbox{\includegraphics[width=3.5cm]{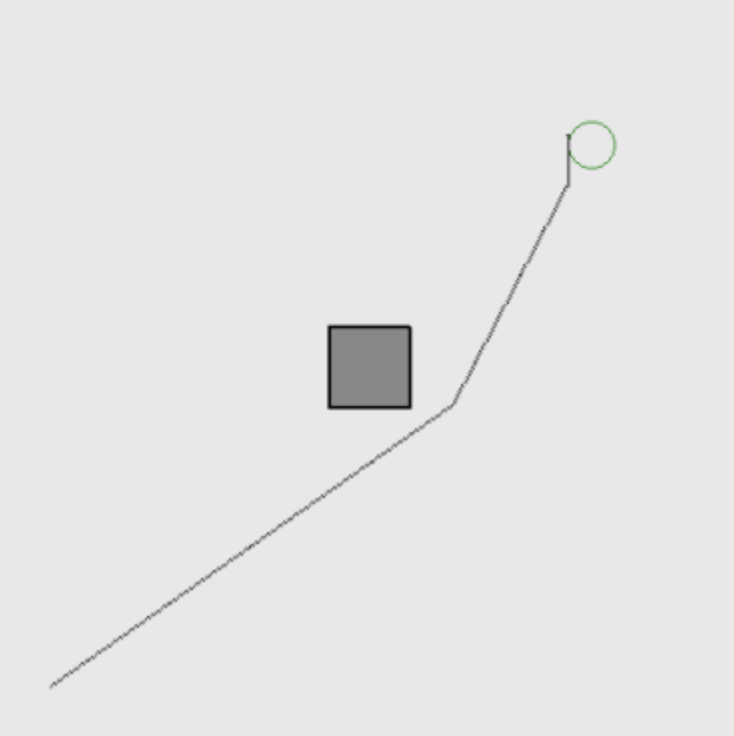}}
  \caption{Maximum reward}
\end{subfigure}%
\begin{subfigure}{.3\linewidth}
  \centering
  \fbox{\includegraphics[width=3.5cm]{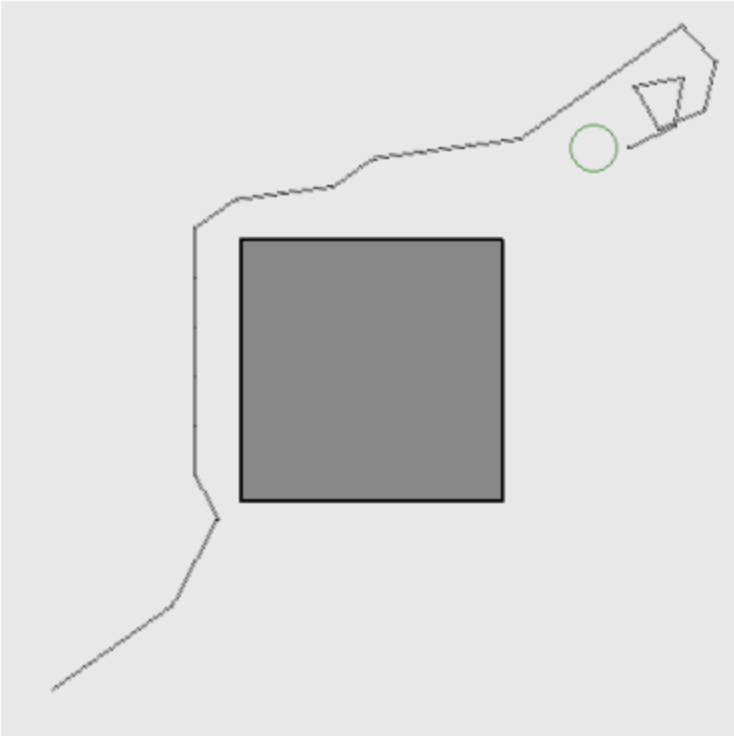}}
  \caption{Minimum reward}
\end{subfigure}
\begin{subfigure}{.3\linewidth}
  \centering
  \fbox{\includegraphics[width=3.5cm]{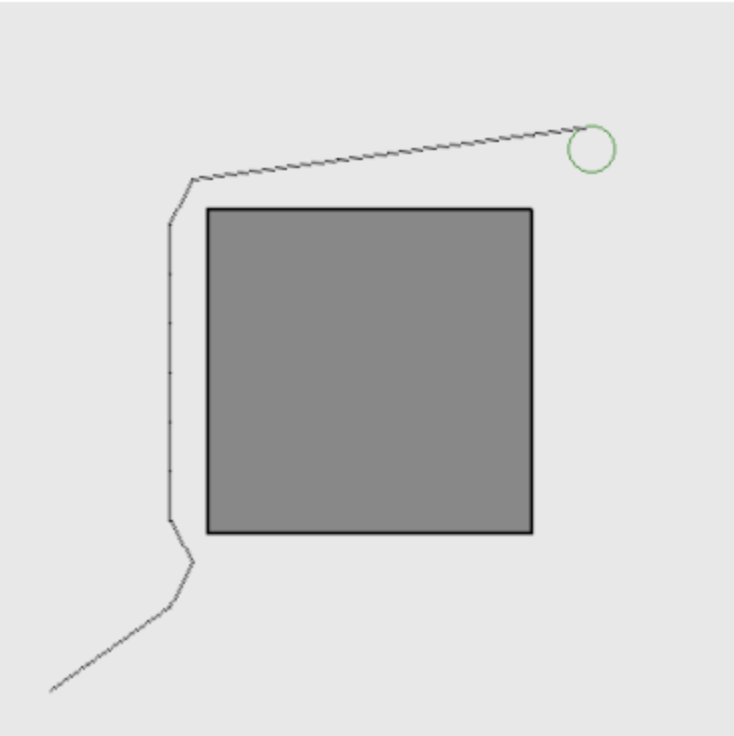}}
  \caption{Random}
\end{subfigure}
\caption{DQN with discrete masking trajectory examples}
\end{figure}

\begin{figure}[t]
\centering
\begin{subfigure}{.3\linewidth}
  \centering
  \fbox{\includegraphics[width=3.5cm]{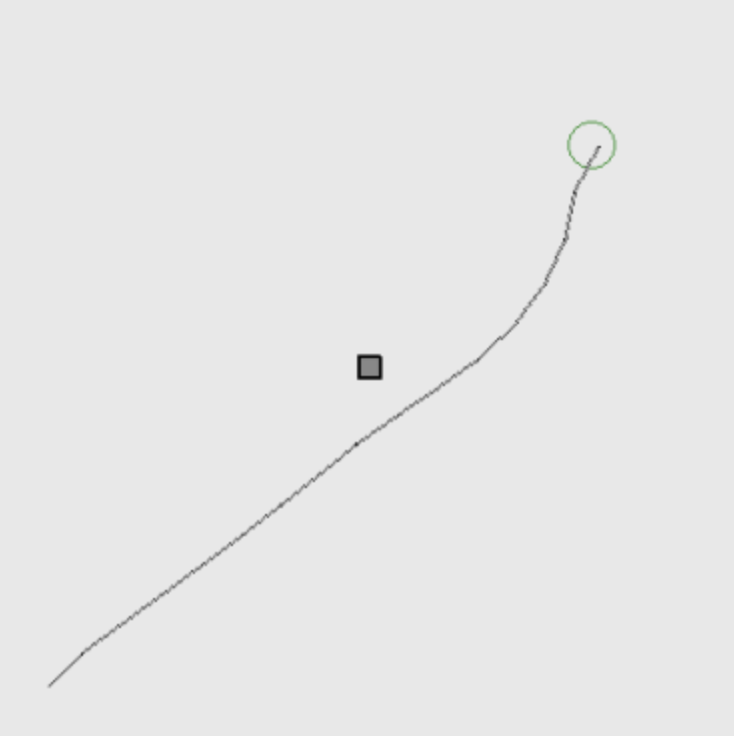}}
  \caption{Maximum reward}
\end{subfigure}%
\begin{subfigure}{.3\linewidth}
  \centering
  \fbox{\includegraphics[width=3.5cm]{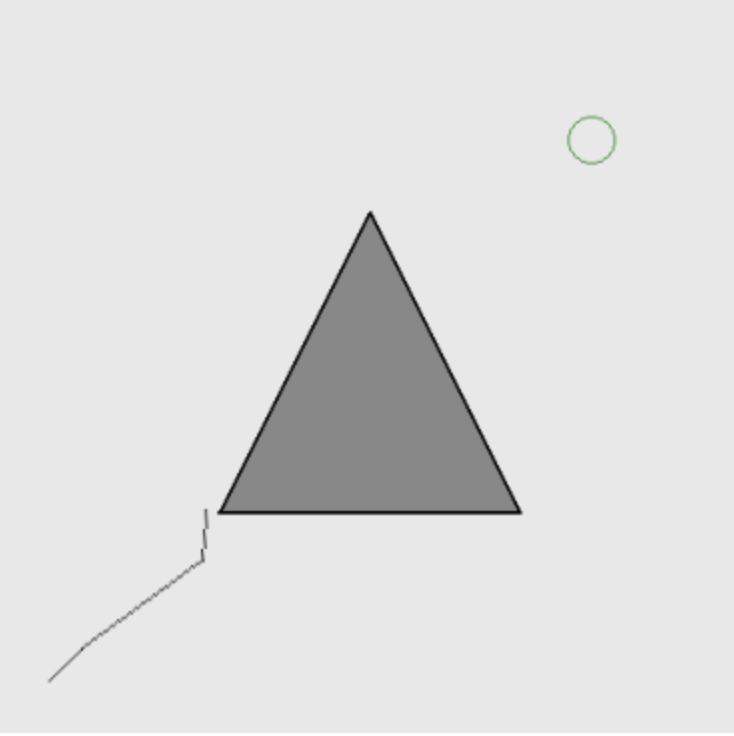}}
  \caption{Minimum reward}
\end{subfigure}
\begin{subfigure}{.3\linewidth}
  \centering
  \fbox{\includegraphics[width=3.5cm]{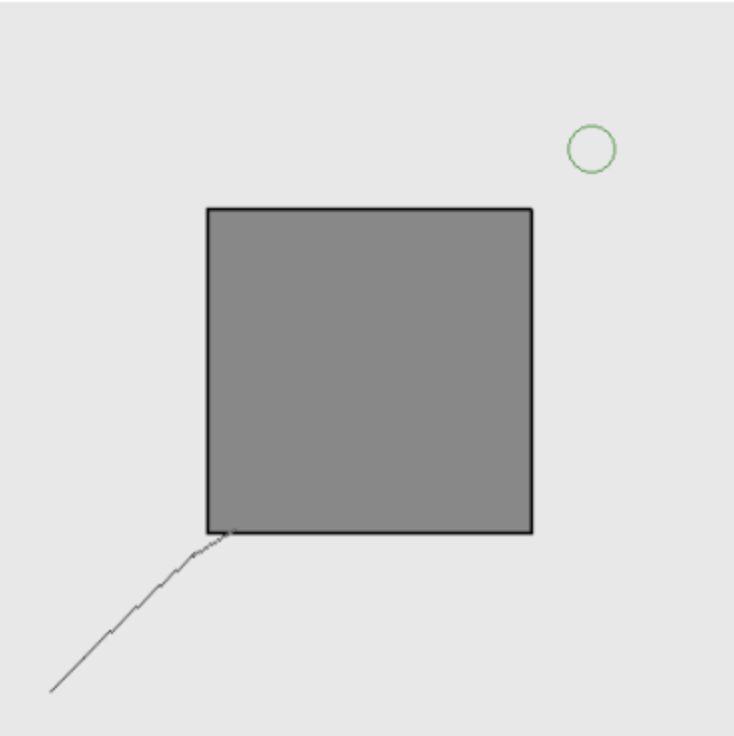}}
  \caption{Random}
\end{subfigure}
\caption{PPO trajectory examples}
\end{figure}

\begin{figure}[t]
\centering
\begin{subfigure}{.3\linewidth}
  \centering
  \fbox{\includegraphics[width=3.5cm]{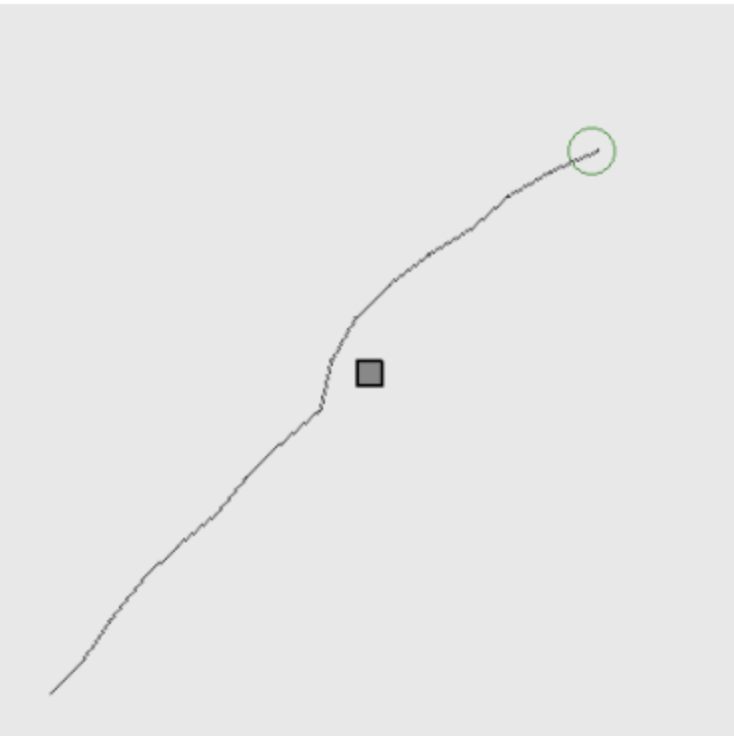}}
  \caption{Maximum reward}
\end{subfigure}%
\begin{subfigure}{.3\linewidth}
  \centering
  \fbox{\includegraphics[width=3.5cm]{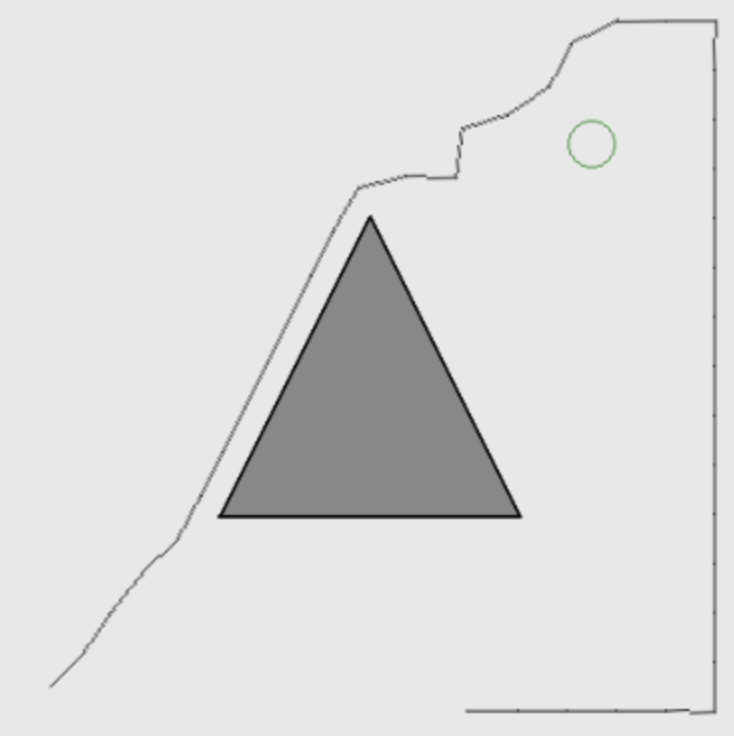}}
  \caption{Minimum reward}
\end{subfigure}
\begin{subfigure}{.3\linewidth}
  \centering
  \fbox{\includegraphics[width=3.5cm]{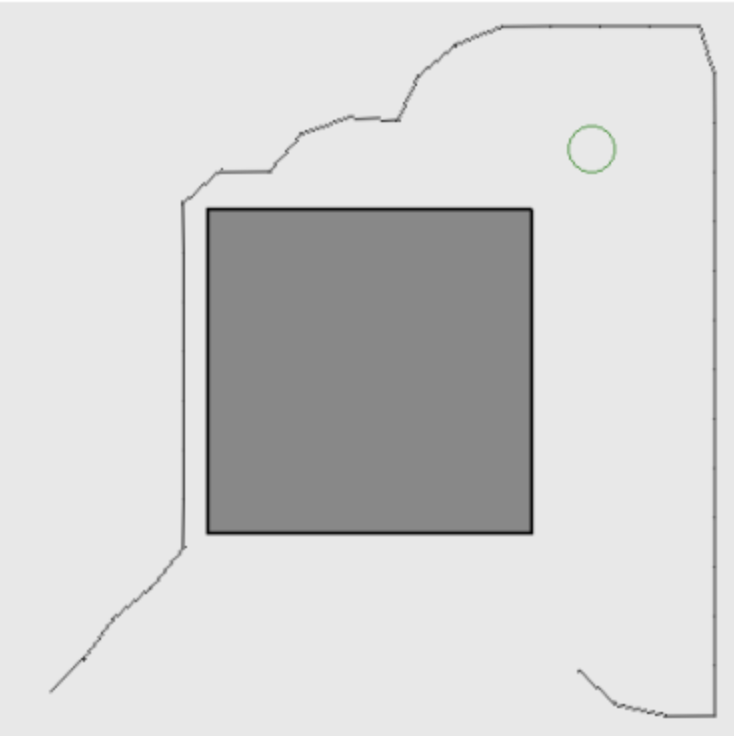}}
  \caption{Random}
\end{subfigure}
\caption{PPO with projection trajectory examples}
\end{figure}

\begin{figure}[t]
\centering
\begin{subfigure}{.3\linewidth}
  \centering
  \fbox{\includegraphics[width=3.5cm]{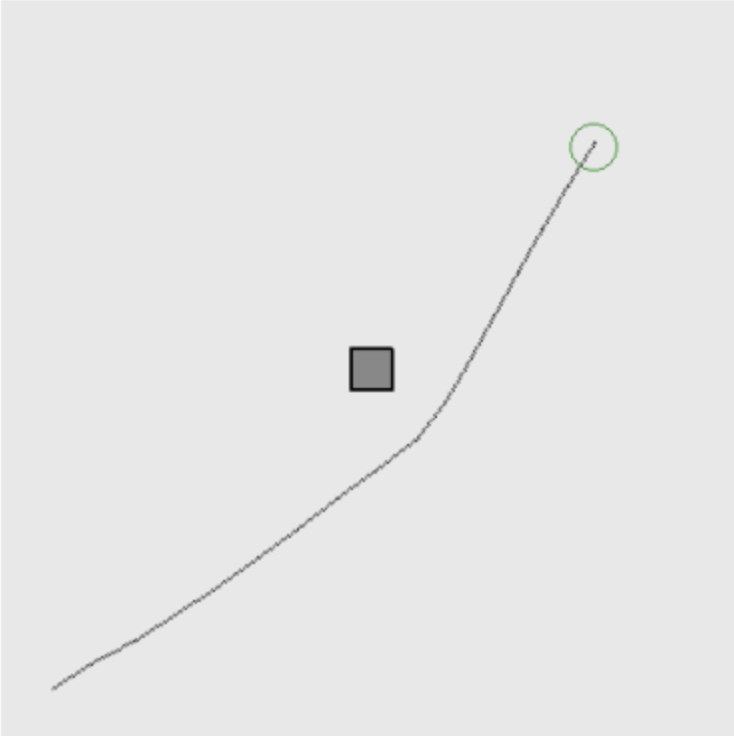}}
  \caption{Maximum reward}
\end{subfigure}%
\begin{subfigure}{.3\linewidth}
  \centering
  \fbox{\includegraphics[width=3.5cm]{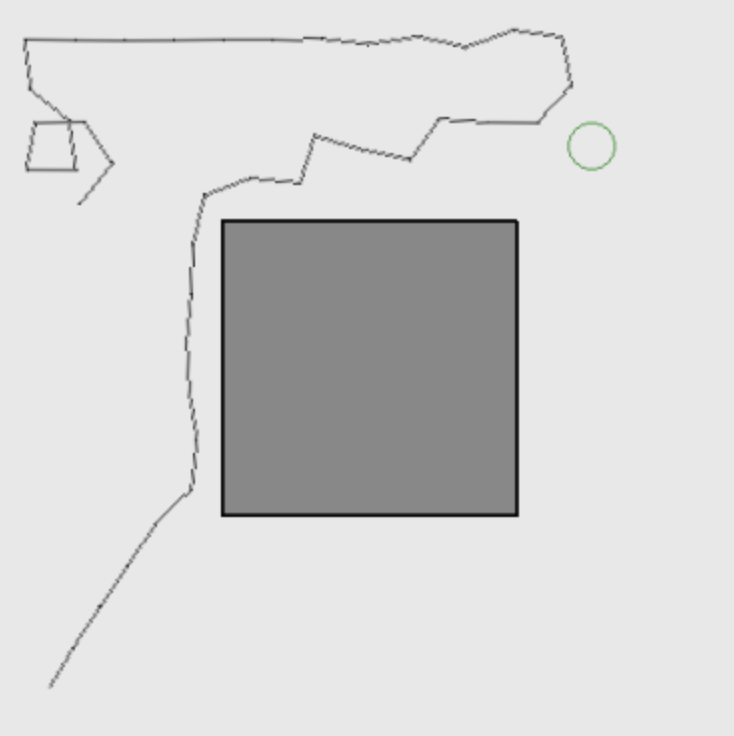}}
  \caption{Minimum reward}
\end{subfigure}
\begin{subfigure}{.3\linewidth}
  \centering
  \fbox{\includegraphics[width=3.5cm]{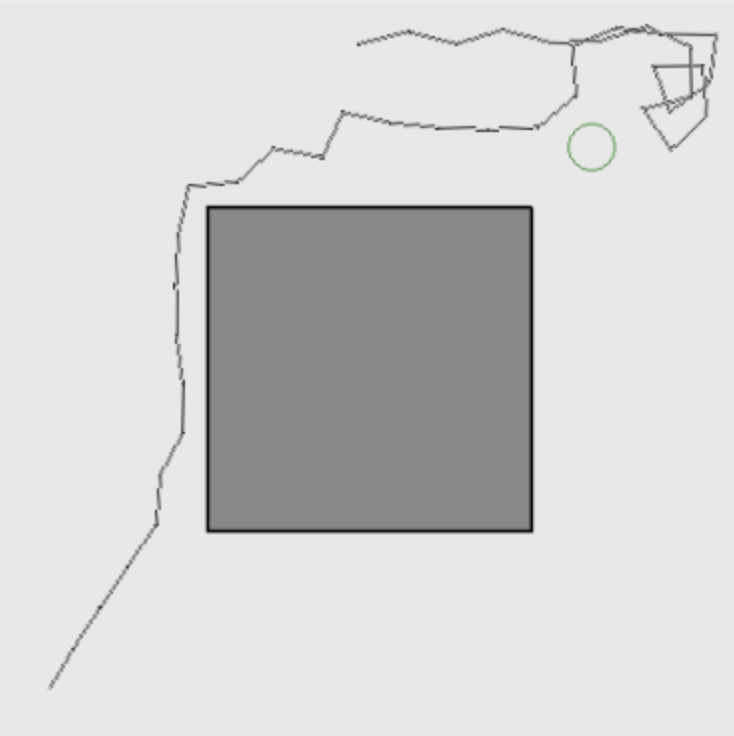}}
  \caption{Random}
\end{subfigure}
\caption{PPO with continuous masking trajectory examples}
\end{figure}

\begin{figure}[t]
\centering
\begin{subfigure}{.3\linewidth}
  \centering
  \fbox{\includegraphics[width=3.5cm]{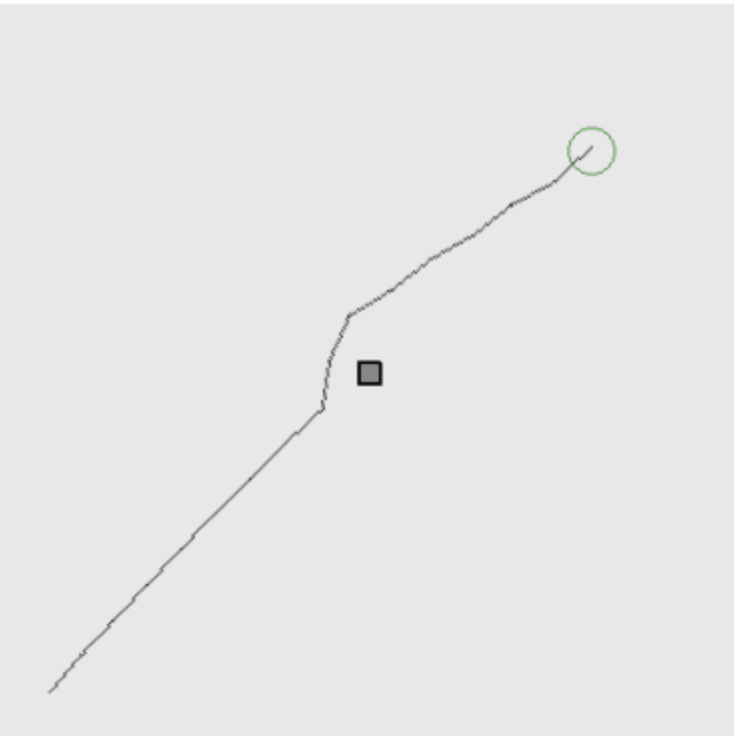}}
  \caption{Maximum reward}
\end{subfigure}%
\begin{subfigure}{.3\linewidth}
  \centering
  \fbox{\includegraphics[width=3.5cm]{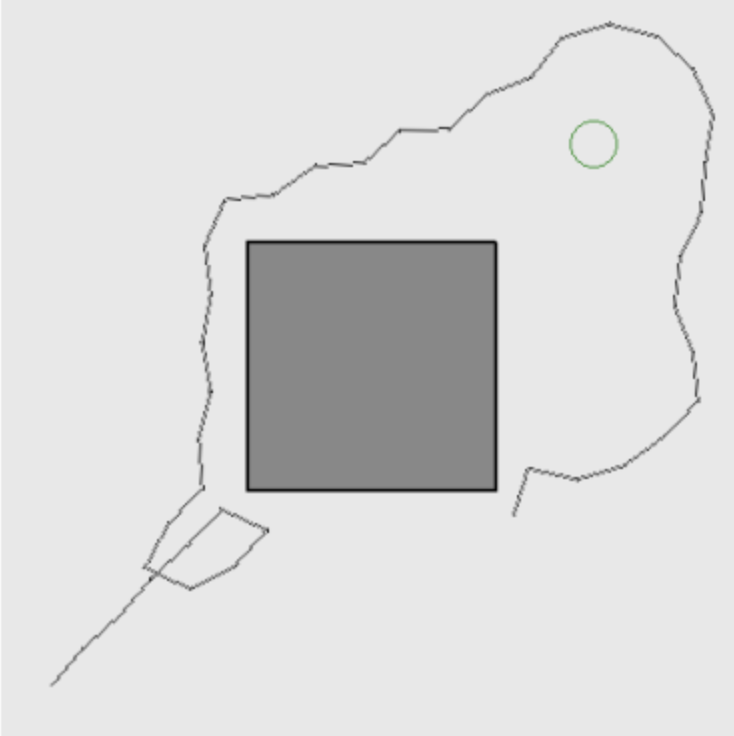}}
  \caption{Minimum reward}
\end{subfigure}
\begin{subfigure}{.3\linewidth}
  \centering
  \fbox{\includegraphics[width=3.5cm]{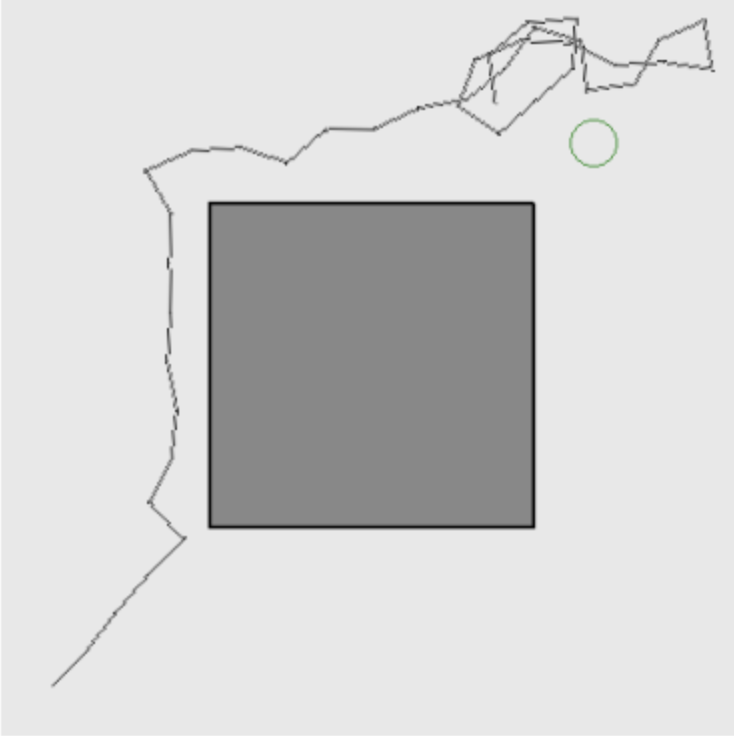}}
  \caption{Random}
\end{subfigure}
\caption{PPO with random replacement trajectory examples}
\end{figure}

\begin{figure}[t]
\centering
\begin{subfigure}{.3\linewidth}
  \centering
  \fbox{\includegraphics[width=3.5cm]{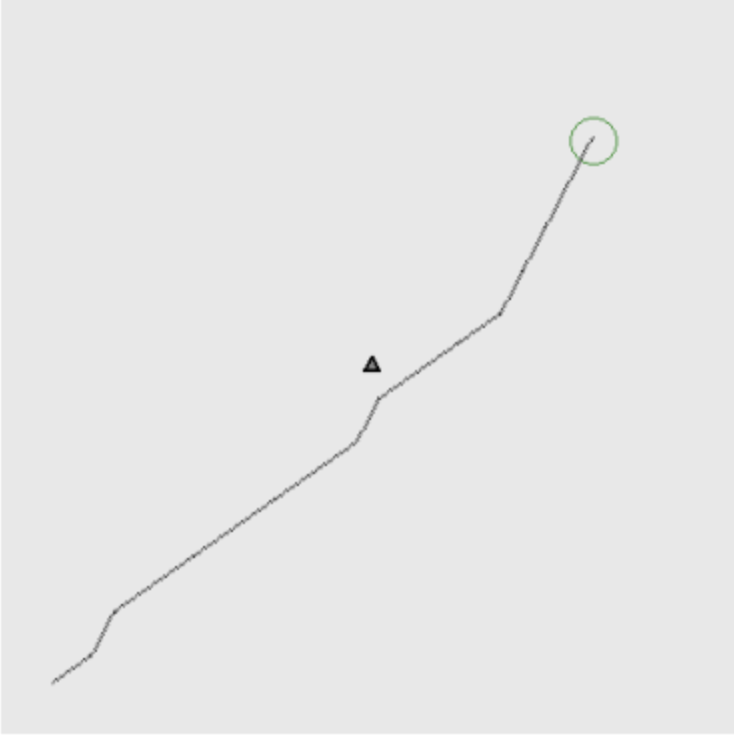}}
  \caption{Maximum reward}
\end{subfigure}%
\begin{subfigure}{.3\linewidth}
  \centering
  \fbox{\includegraphics[width=3.5cm]{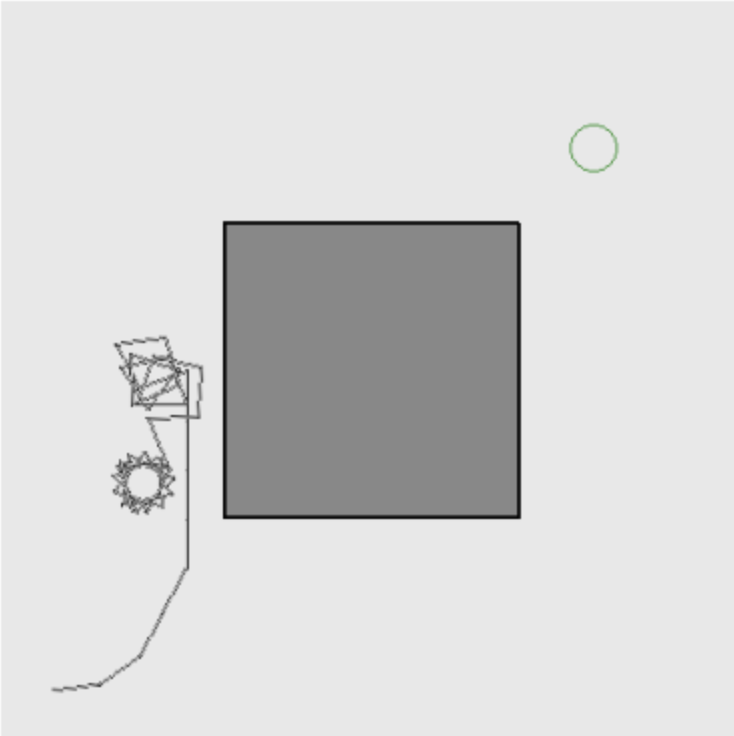}}
  \caption{Minimum reward}
\end{subfigure}
\begin{subfigure}{.3\linewidth}
  \centering
  \fbox{\includegraphics[width=3.5cm]{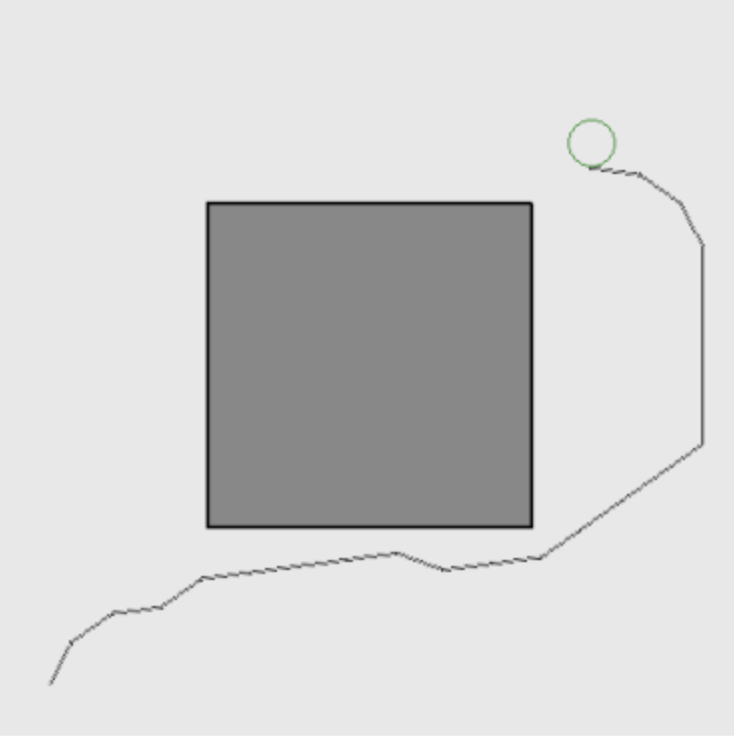}}
  \caption{Random}
\end{subfigure}
\caption{PPO with discrete masking trajectory examples}
\end{figure}

\begin{figure}[t]
\centering
\begin{subfigure}{.3\linewidth}
  \centering
  \fbox{\includegraphics[width=3.5cm]{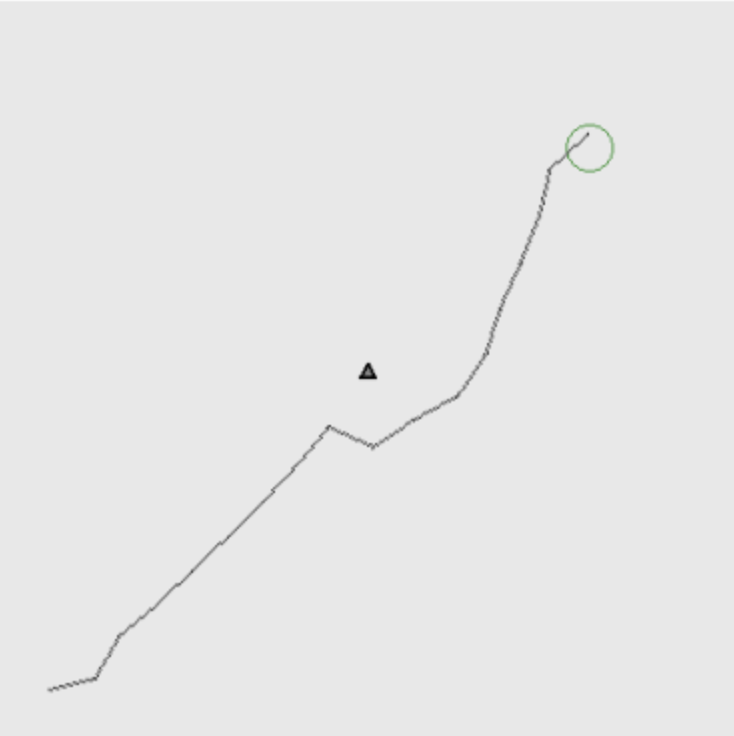}}
  \caption{Maximum reward}
\end{subfigure}%
\begin{subfigure}{.3\linewidth}
  \centering
  \fbox{\includegraphics[width=3.5cm]{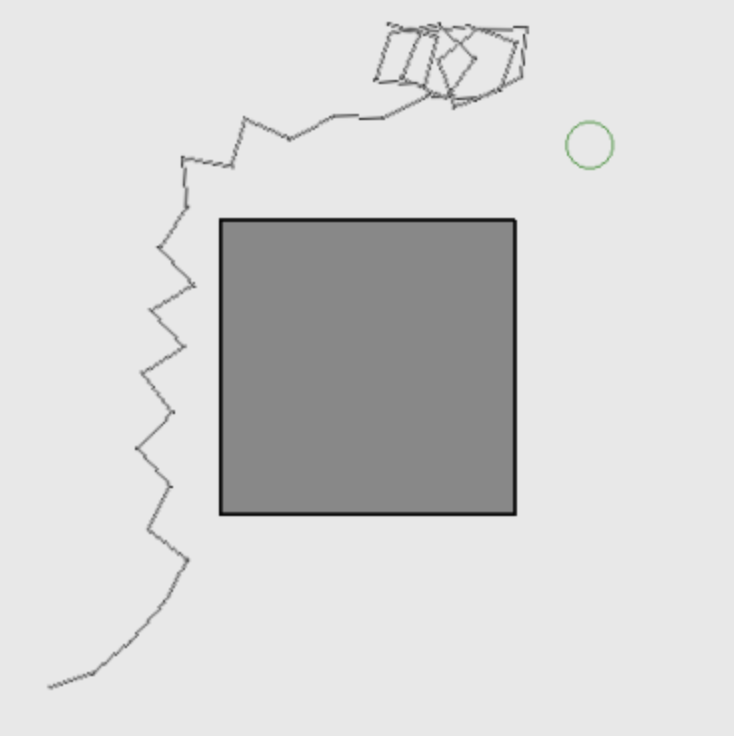}}
  \caption{Minimum reward}
\end{subfigure}
\begin{subfigure}{.3\linewidth}
  \centering
  \fbox{\includegraphics[width=3.5cm]{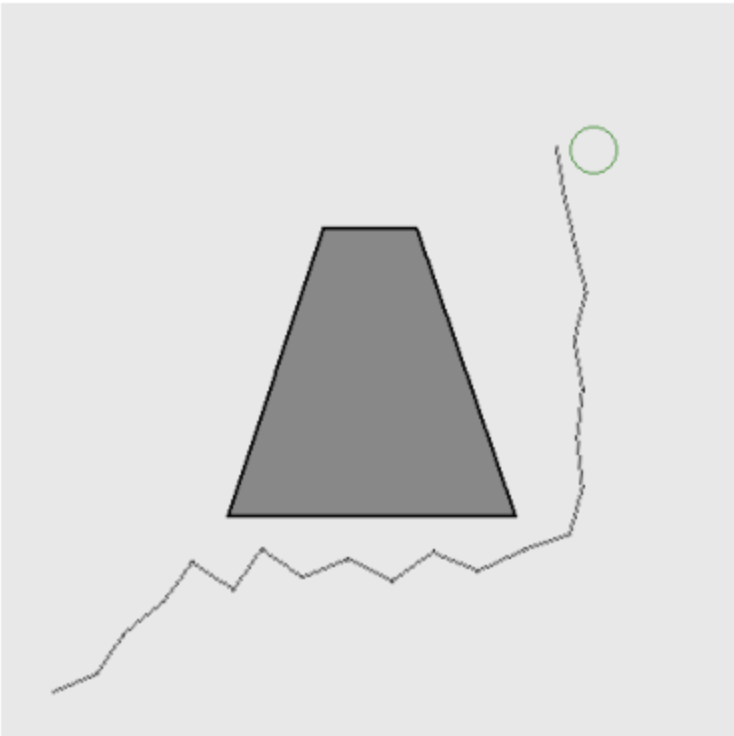}}
  \caption{Random}
\end{subfigure}
\caption{MPS-TD3 with random replacement trajectory examples}
\end{figure}

\begin{figure}[t]
\centering
\begin{subfigure}{.3\linewidth}
  \centering
  \fbox{\includegraphics[width=3.5cm]{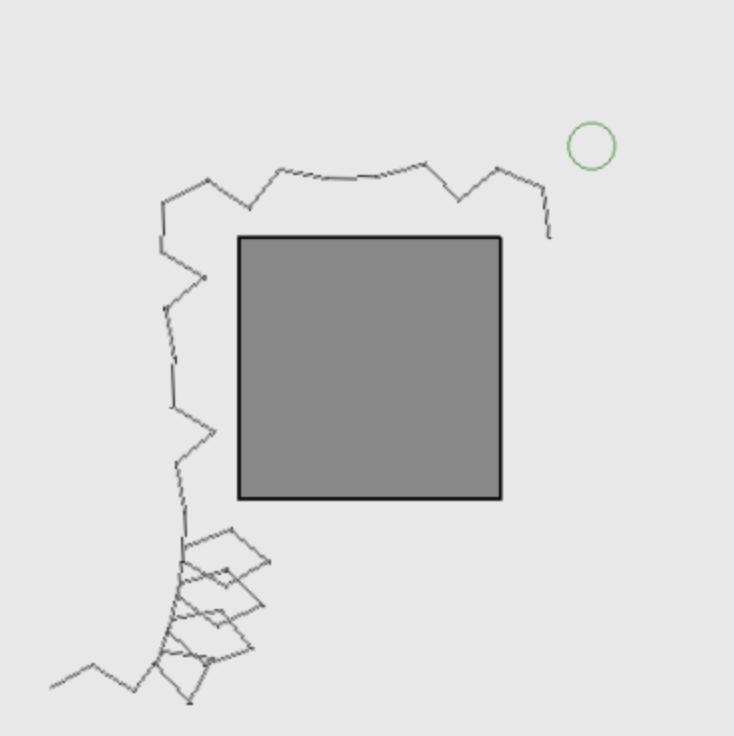}}
  \caption{Maximum reward}
\end{subfigure}%
\begin{subfigure}{.3\linewidth}
  \centering
  \fbox{\includegraphics[width=3.5cm]{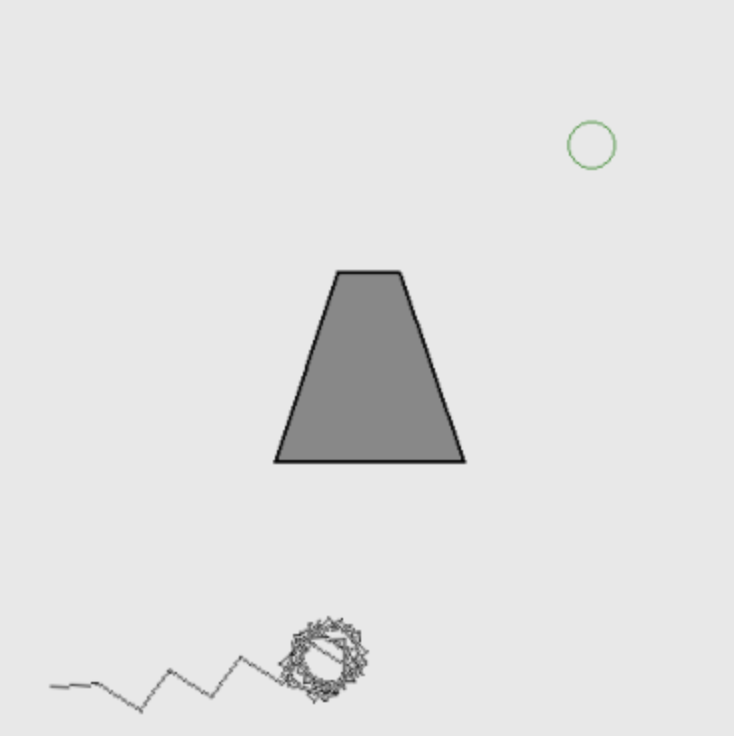}}
  \caption{Minimum reward}
\end{subfigure}
\begin{subfigure}{.3\linewidth}
  \centering
  \fbox{\includegraphics[width=3.5cm]{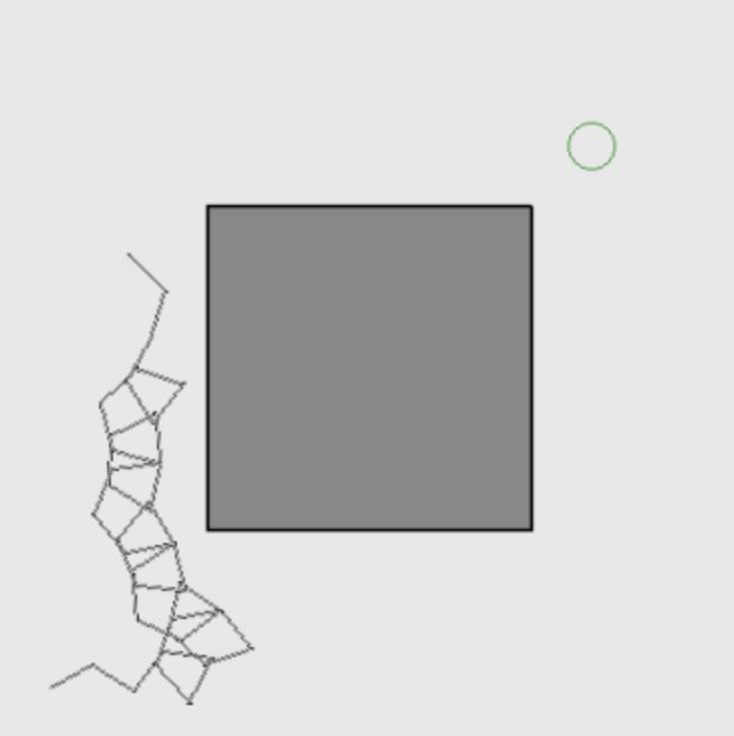}}
  \caption{Random}
\end{subfigure}
\caption{PAM trajectory examples}
\label{app:exp2_2_trajectory_end}
\end{figure}

\begin{figure}[h]
\centering
\begin{subfigure}[b]{\linewidth}
  \centering
   \importpgf{experiment_2/evaluation/reward_t}{on_policy_t_reward.pgf}
  \caption{On-policy}
\end{subfigure}
\\[3ex]
\begin{subfigure}[b]{\linewidth}
  \centering
   \importpgf{experiment_2/evaluation/reward_t}{off_policy_t_reward.pgf}
  \caption{Off-policy}
\end{subfigure}
\caption{Welch's t-test p-values between average returns}
\label{exp2_2_test_reward}
\end{figure}

\begin{figure}[h]
\centering
\begin{subfigure}[b]{\linewidth}
  \centering
   \importpgf{experiment_2/evaluation/reward_t}{on_policy_t_steps.pgf}
  \caption{On-policy}
\end{subfigure}
\\[3ex]
\begin{subfigure}[b]{\linewidth}
  \centering
   \importpgf{experiment_2/evaluation/reward_t}{off_policy_t_steps.pgf}
  \caption{Off-policy}
\end{subfigure}
\caption{Welch's t-test p-values between average steps}
\label{app:exp2_2_test_steps}
\end{figure}

\begin{figure}[h]
\centering
\begin{subfigure}[b]{\linewidth}
  \centering
   \importpgf{experiment_2/evaluation/reward_t}{on_policy_t_solved.pgf}
  \caption{On-policy}
\end{subfigure}
\\[3ex]
\begin{subfigure}[b]{\linewidth}
  \centering
   \importpgf{experiment_2/evaluation/reward_t}{off_policy_t_solved.pgf}
  \caption{Off-policy}
\end{subfigure}
\caption{Welch's t-test p-values between the fractions of solved environments}
\label{exp2_2_test_solved}
\end{figure}

\clearpage
\subsection{Evaluation With Complex Restrictions}
\label{app:exp2_3}

\begin{figure}[!h]
\centering
  \centering
  \importpgf{experiment_2/evaluation_2}{eval_rewards.pgf}
\caption{Average Return}
\label{app:exp2_3_plot_rerward}
\end{figure}

\begin{figure}[!h]
\centering
  \centering
  \importpgf{experiment_2/evaluation_2}{eval_solved.pgf}
\caption{Average fraction of solved environments}
\label{app:exp2_3_plot_solved}
\end{figure}

\begin{figure}[!h]
\centering
  \centering
  \importpgf{experiment_2/evaluation_2}{eval_steps.pgf}
\caption{Average episode length}
\label{app:exp2_3_plot_steps}
\end{figure}

\begin{table}[h]
\scriptsize
\centering
\begin{tabular}{llll}
\hline
\multicolumn{1}{c}{\textbf{Approach}} & \multicolumn{1}{c}{\textbf{Return}} & \multicolumn{1}{c}{\textbf{Steps}} & \multicolumn{1}{c}{\textbf{Solved}} \\ \hline

TD3                                   & $43.94 \pm 19.87$                   & $22.89 \pm 2.17$                   & $33.75\% \pm 17.23\%$               \\
TD3-Projection                        & $107.36 \pm 6.57$                   & $19.80 \pm 4.90$                   & $95.00\% \pm 4.47\%$                \\
TD3-Masking                           & $113.88 \pm 1.28$                   & $17.59 \pm 0.81$                   & $99.17\% \pm 1.29\%$                \\
TD3-Random                            & $106.85 \pm 4.65$                   & $21.36 \pm 2.34$                   & $95.42\% \pm 4.31\%$                \\
DQN-Masked                            & $112.14 \pm 2.40$                   & $17.88 \pm 0.28$                   & $96.25\% \pm 2.62\%$                \\ \hline

PPO                                   & $12.70 \pm 7.96$                    & $22.02 \pm 5.80$                   & $6.67\% \pm 4.92\%$                 \\
PPO-Projection                        & $110.61 \pm 3.06$                   & $16.45 \pm 0.23$                   & $95.00\% \pm2.74\%$                 \\
PPO-Masking                           & $106.80 \pm 4.92$                   & $18.15 \pm 0.75$                   & $86.03\% \pm 2.86\%$                \\
PPO-Random                            & $94.56 \pm 4.39$                    & $21.83 \pm 1.61$                   & $83.33\% \pm 4.38\%$                \\
PPO-Masked                            & $67.56 \pm 1.45$                    & $22.86 \pm 1.76$                   & $82.01\% \pm 1.66\%$                \\ \hline

MPS-TD3                               & $110.73 \pm 1.42$                   & $21.10 \pm 0.10$                   & $99.17\% \pm 2.04\%$                \\
PAM                                   & $-10.87 \pm 3.94$                   & $-$                                & $0.00\% \pm 0.00\%$                 \\ \hline
\end{tabular}
\caption{Average return, steps, and solved episodes}
\label{app:exp2_3_results}
\end{table}

\begin{table}[h]
\scriptsize
\centering
\begin{tabular}{llll}
\hline
\multicolumn{1}{c}{\textbf{Approach}} & \multicolumn{1}{c}{\textbf{Intervals}} & \multicolumn{1}{c}{\textbf{Size}} & \multicolumn{1}{c}{\textbf{Fraction}} \\ \hline

TD3                                   & $1.39 \pm 0.08$                        & $151.03 \pm 7.59$                   & $68.65\% \pm 3.45\%$                  \\
TD3-Projection                        & $1.28 \pm 0.10$                        & $140.90 \pm 10.16$                  & $64.04\% \pm 4.62\%$                  \\
TD3-Masking                           & $1.32 \pm 0.01$                        & $147.72 \pm 1.30$                   & $67.15\% \pm 0.59\%$                  \\
TD3-Random                            & $1.33 \pm 0.01$                        & $149.42 \pm 1.43$                   & $67.92\% \pm 0.65\%$                  \\
DQN-Masked                            & $1.34 \pm0.01$                         & $147.99 \pm 0.61$                   & $67.27\% \pm 0.28\%$                  \\ \hline

PPO                                   & $1.39 \pm 0.02$                        & $160.93\pm 4.73$                    & $73.15\% \pm 2.15\%$                  \\
PPO-Projection                        & $1.32 \pm 0.02$                        & $144.10 \pm 1.31$                   & $65.50\% \pm 0.59\%$                  \\
PPO-Masking                           & $1.34 \pm 0.01$                        & $145.96 \pm1.23$                    & $66.35\% \pm 0.56\%$                  \\
PPO-Random                            & $1.34 \pm 0.01$                        & $145.39 \pm 0.65$                   & $66.08\% \pm 0.30\%$                  \\
PPO-Masked                            & $1.36 \pm 0.02$                        & $144.46 \pm 2.11$                   & $65.67\% \pm 0.96\%$                  \\ \hline

MPS-TD3                               & $1.39 \pm 0.03$                        & $144.93 \pm 1.31$                   & $65.88\% \pm 0.60\%$                  \\
PAM                                   & $1.37 \pm 0.03$                        & $153.85 \pm 3.87$                   & $69.93\% \pm 1.76\%$                  \\ \hline
\end{tabular}
\caption{Average number of disjoint intervals and size of the action space}
\label{app:exp2_3_control}
\end{table}

\begin{table}[h]
\scriptsize
\centering
\begin{tabular}{lllll}
\hline
\multicolumn{1}{c}{\textbf{Approach}} & \multicolumn{1}{c}{\textbf{Average}} & \multicolumn{1}{c}{\textbf{Minimum}} & \multicolumn{1}{c}{\textbf{Maximum}} & \multicolumn{1}{c}{\textbf{Variance}} \\ \hline

TD3 & $129.19 \pm 11.02$ & $119.01 \pm 12.69$ & $139.44 \pm 9.35$ & $431.86 \pm 66.18$ \\
TD3-Projection & $126.09 \pm 4.88$ & $119.3 \pm 2.67$ & $132.94 \pm 7.21$ & $261.44 \pm 84.26$ \\
TD3-Masking & $130.65 \pm 1.69$ & $122.46 \pm 1.99$ & $138.91 \pm 1.41$ & $319.12 \pm 27.2$ \\
TD3-Random & $131.62 \pm 1.49$ & $123.24 \pm 1.65$ & $140.04 \pm 1.49$ & $331.53 \pm 25.68$ \\
DQN-Masked & $129.89 \pm 1.28$ & $121.46 \pm 1.64$ & $138.32 \pm 1.07$ & $332.14 \pm 31.81$ \\ \hline

PPO & $137.77 \pm 4.57$ & $126.66 \pm 4.76$ & $148.97 \pm 4.42$ & $467.54 \pm 25.27$ \\
PPO-Projection & $126.64 \pm 1.10$ & $118.44 \pm 1.34$ & $134.89 \pm 1.05$ & $320.97 \pm 24.38$ \\
PPO-Masking & $127.55 \pm 0.99$ & $118.85 \pm 1.16$ & $136.30 \pm 0.99$ & $353.00 \pm 33.03$ \\
PPO-Random & $126.87 \pm 0.81$ & $118.27 \pm 0.98$ & $135.53 \pm 0.80$ & $339.53 \pm 17.15$ \\
PPO-Masked & $125.10 \pm 2.72$ & $116.26 \pm 2.88$ & $133.96 \pm 2.57$ & $353.42 \pm 15.28$ \\ \hline

MPS-TD3 & $124.01 \pm 2.39$ & $113.57 \pm 3.23$ & $134.53 \pm 1.75$ & $446.62 \pm 57.25$ \\
PAM & $133.34 \pm 2.92$ & $123.84 \pm 2.55$ & $142.90 \pm 3.68$ & $405.65 \pm 56.05$ \\ \hline
\end{tabular}
\caption{Size of the individual intervals}
\label{app:exp2_3_single}
\end{table}

\begin{figure}[t]
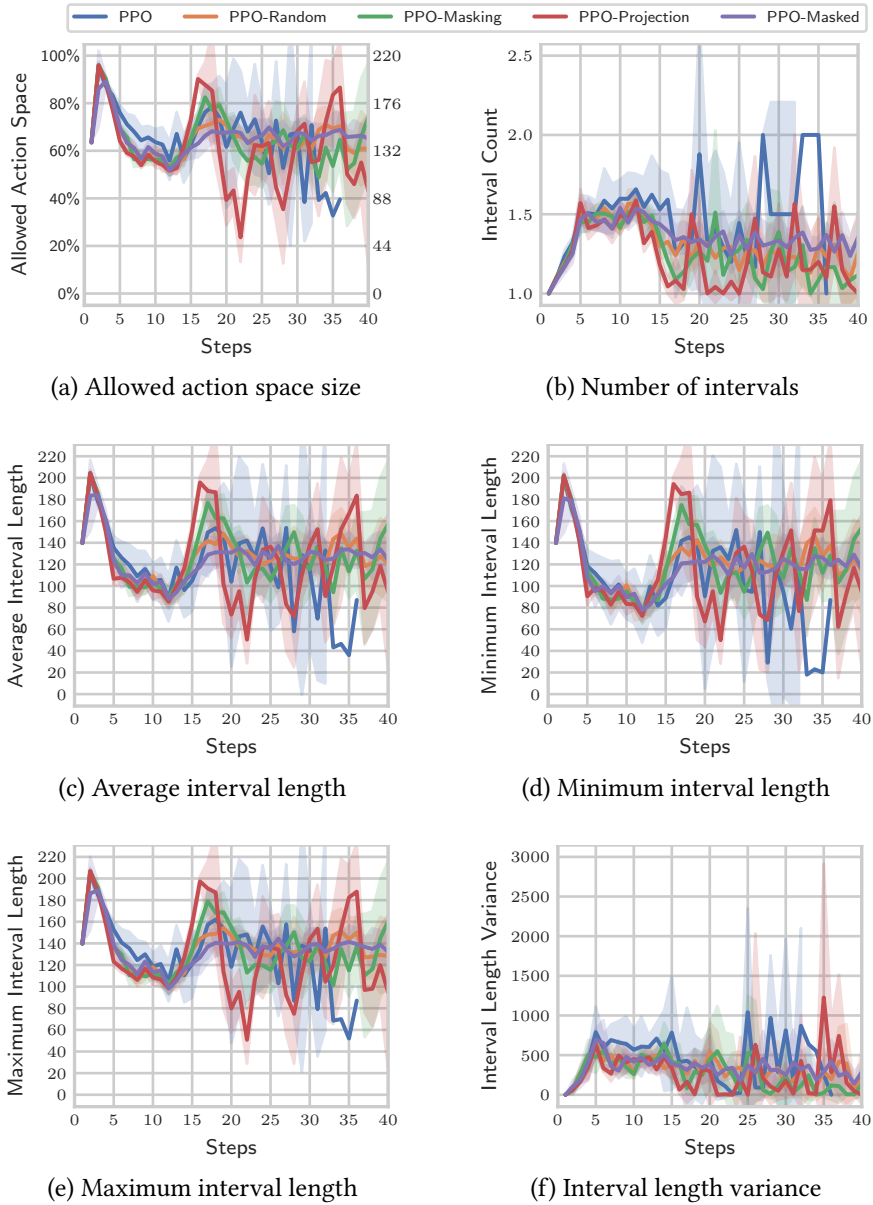

\centering
\begin{subfigure}{0.8\linewidth}
  \centering
  \importpgf{experiment_1/training/ppo}{legend.pgf}
\end{subfigure}
\begin{subfigure}{.49\linewidth}
  \centering
  \importpgf{experiment_2/evaluation_2/ppo}{allowed.pgf}
  \caption{Allowed action space size}
\end{subfigure}%
\begin{subfigure}{.49\linewidth}
  \centering
  \importpgf{experiment_2/evaluation_2/ppo}{count.pgf}
  \caption{Number of intervals}
\end{subfigure}
\\[3ex]
\begin{subfigure}{.49\linewidth}
  \centering
  \importpgf{experiment_2/evaluation_2/ppo}{average.pgf}
  \caption{Average interval length}
\end{subfigure}
\begin{subfigure}{.49\linewidth}
  \centering
  \importpgf{experiment_2/evaluation_2/ppo}{minimum.pgf}
  \caption{Minimum interval length}
\end{subfigure}
\\[3ex]
\begin{subfigure}{.49\linewidth}
  \centering
  \importpgf{experiment_2/evaluation_2/ppo}{maximum.pgf}
  \caption{Maximum interval length}
\end{subfigure}
\begin{subfigure}{.49\linewidth}
  \centering
  \importpgf{experiment_2/evaluation_2/ppo}{variance.pgf}
  \caption{Interval length variance}
\end{subfigure}
\caption{On-policy control variables}
\label{app:exp2_3_control_ppo}
\end{figure}

\begin{figure}[t]
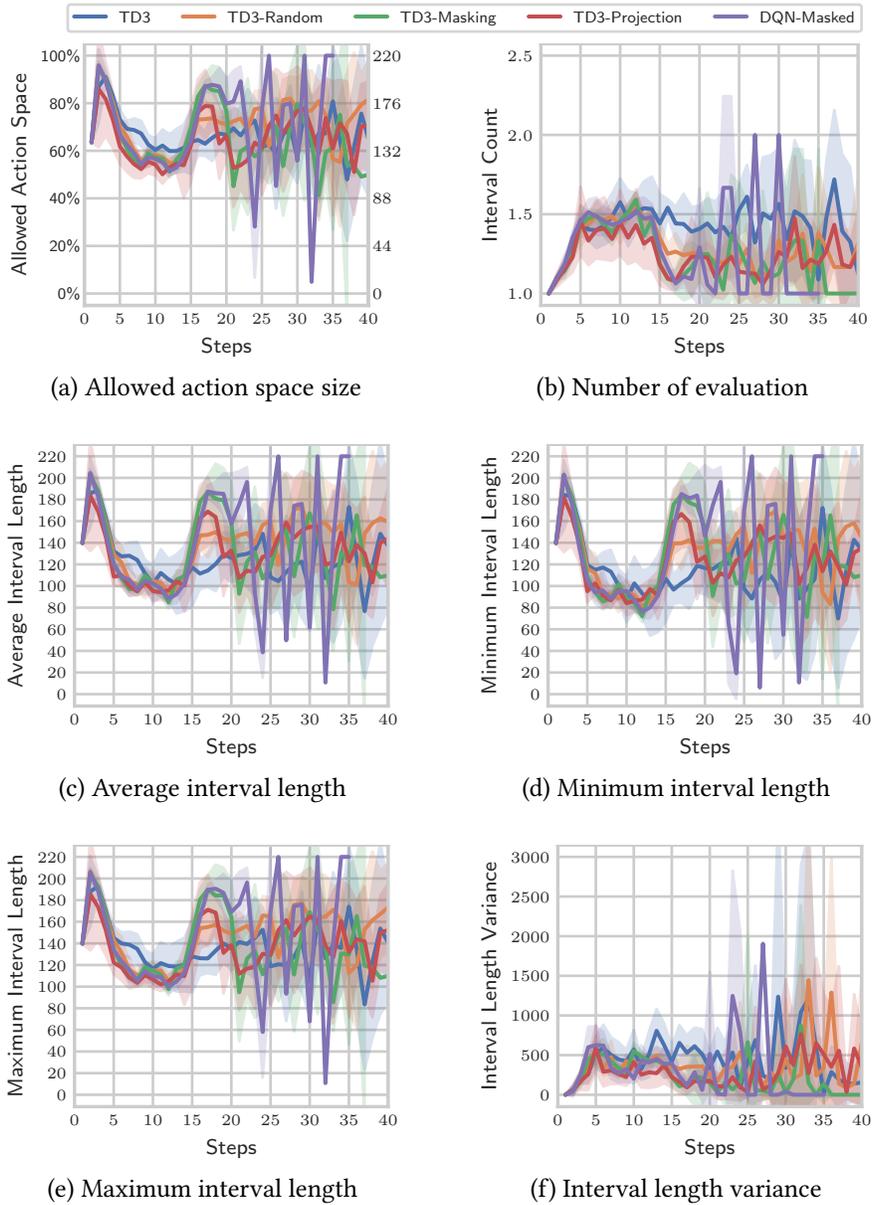

\centering
\begin{subfigure}{0.8\linewidth}
  \centering
  \importpgf{experiment_1/training/ddpg}{legend.pgf}
\end{subfigure}
\begin{subfigure}{.49\linewidth}
  \centering
   \importpgf{experiment_2/evaluation_2/td3}{allowed.pgf}
  \caption{Allowed action space size}
\end{subfigure}%
\begin{subfigure}{.49\linewidth}
  \centering
   \importpgf{experiment_2/evaluation_2/td3}{count.pgf}
  \caption{Number of evaluation}
\end{subfigure}
\\[3ex]
\begin{subfigure}{.49\linewidth}
  \centering
   \importpgf{experiment_2/evaluation_2/td3}{average.pgf}
  \caption{Average interval length}
\end{subfigure}
\begin{subfigure}{.49\linewidth}
  \centering
   \importpgf{experiment_2/evaluation_2/td3}{minimum.pgf}
  \caption{Minimum interval length}
\end{subfigure}
\\[3ex]
\begin{subfigure}{.49\linewidth}
  \centering
   \importpgf{experiment_2/evaluation_2/td3}{maximum.pgf}
  \caption{Maximum interval length}
\end{subfigure}
\begin{subfigure}{.49\linewidth}
  \centering
   \importpgf{experiment_2/evaluation_2/td3}{variance.pgf}
  \caption{Interval length variance}
\end{subfigure}
\caption{Off-policy control variables}
\label{app:exp2_3_control_td3}
\end{figure}

\begin{figure}[t]
\centering
\begin{subfigure}{0.8\linewidth}
  \centering
 \importpgf{experiment_1/training/own}{legend.pgf}
\end{subfigure}
\begin{subfigure}{.49\linewidth}
  \centering
  \importpgf{experiment_2/evaluation_2/own}{allowed.pgf}
  \caption{Allowed action space size}
\end{subfigure}%
\begin{subfigure}{.49\linewidth}
  \centering
  \importpgf{experiment_2/evaluation_2/own}{count.pgf}
  \caption{Number of intervals}
\end{subfigure}
\\[3ex]
\begin{subfigure}{.49\linewidth}
  \centering
  \importpgf{experiment_2/evaluation_2/own}{average.pgf}
  \caption{Average interval length}
\end{subfigure}
\begin{subfigure}{.49\linewidth}
  \centering
  \importpgf{experiment_2/evaluation_2/own}{minimum.pgf}
  \caption{Minimum interval length}
\end{subfigure}
\\[3ex]
\begin{subfigure}{.49\linewidth}
  \centering
  \importpgf{experiment_2/evaluation_2/own}{maximum.pgf}
  \caption{Maximum interval length}
\end{subfigure}
\begin{subfigure}{.49\linewidth}
  \centering
  \importpgf{experiment_2/evaluation_2/own}{variance.pgf}
  \caption{Interval length variance}
\end{subfigure}
\caption{MPS-TD3 and PAM control variables}
\label{app:exp2_3_control_own}
\end{figure}

\begin{figure}[t]
\centering
\begin{subfigure}{.3\linewidth}
  \centering
  \fbox{\includegraphics[width=3.5cm]{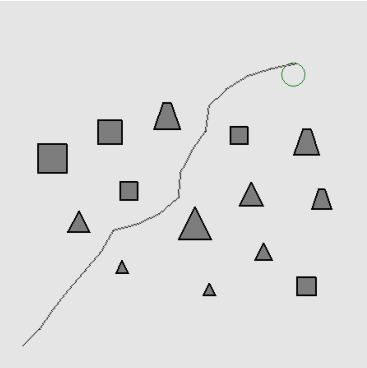}}
  \caption{Maximum reward}
\end{subfigure}%
\begin{subfigure}{.3\linewidth}
  \centering
  \fbox{\includegraphics[width=3.5cm]{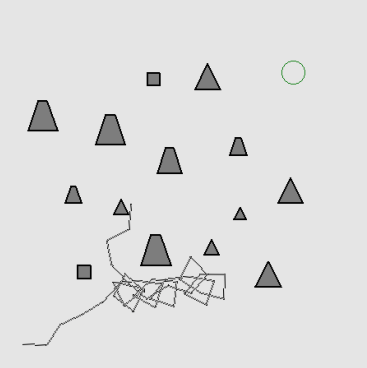}}
  \caption{Minimum reward}
\end{subfigure}
\begin{subfigure}{.3\linewidth}
  \centering
  \fbox{\includegraphics[width=3.5cm]{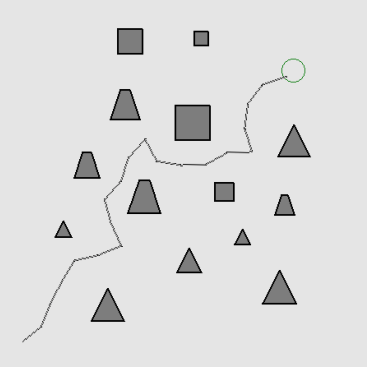}}
  \caption{Random}
\end{subfigure}
\caption{TD3 trajectory examples}
\label{app:exp2_3_trajectory_start}
\end{figure}

\begin{figure}[t]
\centering
\begin{subfigure}{.3\linewidth}
  \centering
  \fbox{\includegraphics[width=3.5cm]{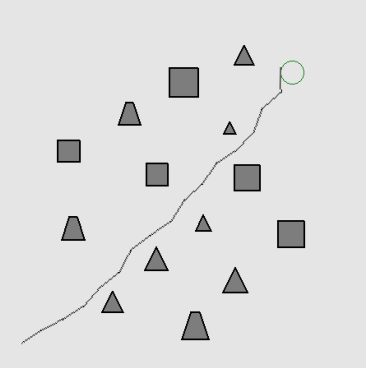}}
  \caption{Maximum reward}
\end{subfigure}%
\begin{subfigure}{.3\linewidth}
  \centering
  \fbox{\includegraphics[width=3.5cm]{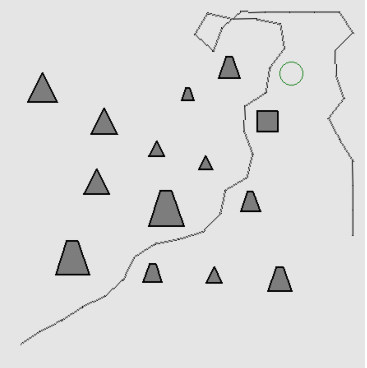}}
  \caption{Minimum reward}
\end{subfigure}
\begin{subfigure}{.3\linewidth}
  \centering
  \fbox{\includegraphics[width=3.5cm]{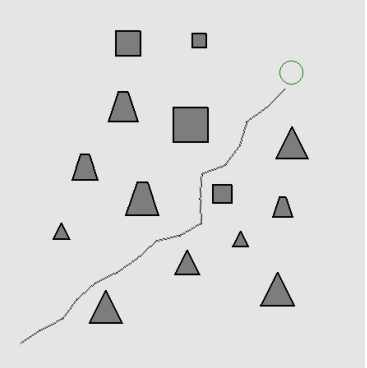}}
  \caption{Random}
\end{subfigure}
\caption{TD3 with projection trajectory examples}
\end{figure}

\begin{figure}[t]
\centering
\begin{subfigure}{.3\linewidth}
  \centering
  \fbox{\includegraphics[width=3.5cm]{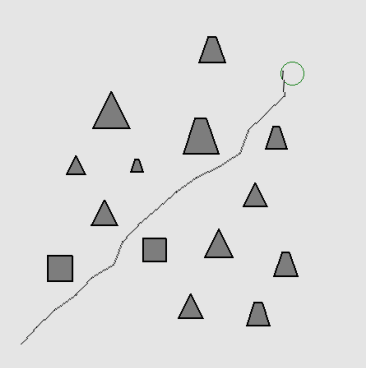}}
  \caption{Maximum reward}
\end{subfigure}%
\begin{subfigure}{.3\linewidth}
  \centering
  \fbox{\includegraphics[width=3.5cm]{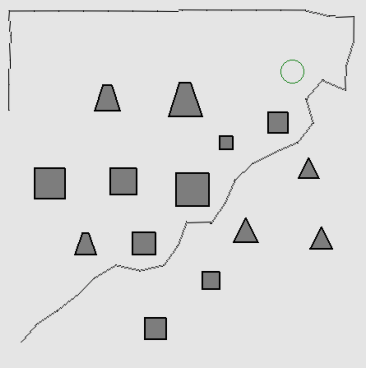}}
  \caption{Minimum reward}
\end{subfigure}
\begin{subfigure}{.3\linewidth}
  \centering
  \fbox{\includegraphics[width=3.5cm]{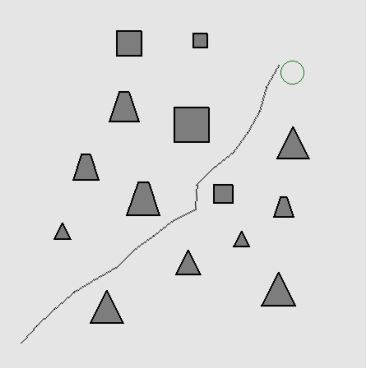}}
  \caption{Random}
\end{subfigure}
\caption{TD3 with continuous masking trajectory examples}
\end{figure}

\begin{figure}[t]
\centering
\begin{subfigure}{.3\linewidth}
  \centering
  \fbox{\includegraphics[width=3.5cm]{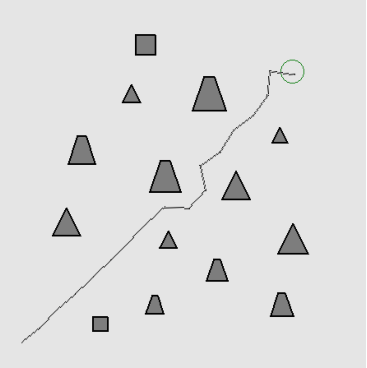}}
  \caption{Maximum reward}
\end{subfigure}%
\begin{subfigure}{.3\linewidth}
  \centering
  \fbox{\includegraphics[width=3.5cm]{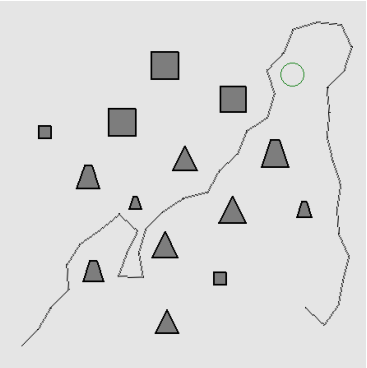}}
  \caption{Minimum reward}
\end{subfigure}
\begin{subfigure}{.3\linewidth}
  \centering
  \fbox{\includegraphics[width=3.5cm]{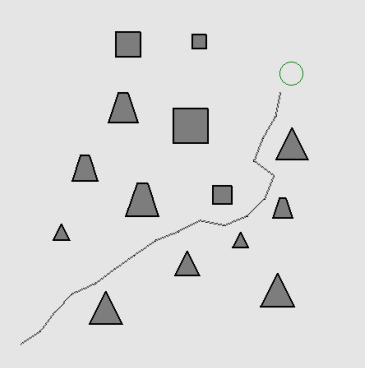}}
  \caption{Random}
\end{subfigure}
\caption{TD3 with random replacement trajectory examples}
\end{figure}

\begin{figure}[t]
\centering
\begin{subfigure}{.3\linewidth}
  \centering
  \fbox{\includegraphics[width=3.5cm]{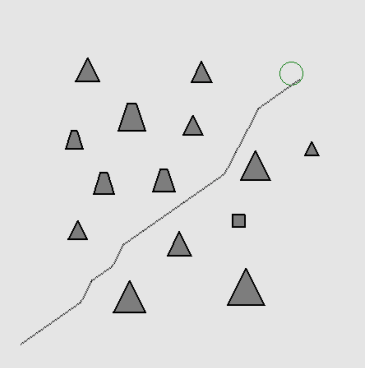}}
  \caption{Maximum reward}
\end{subfigure}%
\begin{subfigure}{.3\linewidth}
  \centering
  \fbox{\includegraphics[width=3.5cm]{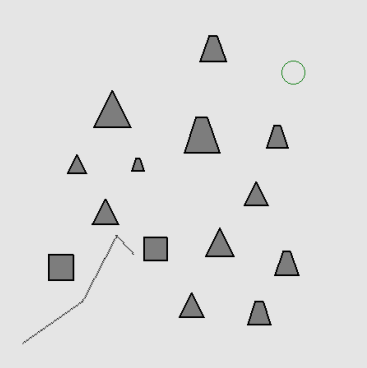}}
  \caption{Minimum reward}
\end{subfigure}
\begin{subfigure}{.3\linewidth}
  \centering
  \fbox{\includegraphics[width=3.5cm]{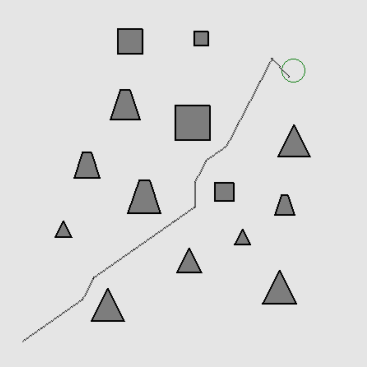}}
  \caption{Random}
\end{subfigure}
\caption{DQN with discrete masking trajectory examples}
\end{figure}

\begin{figure}[t]
\centering
\begin{subfigure}{.3\linewidth}
  \centering
  \fbox{\includegraphics[width=3.5cm]{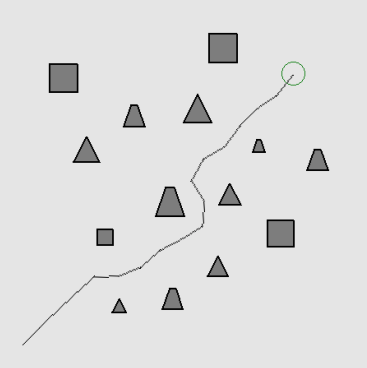}}
  \caption{Maximum reward}
\end{subfigure}%
\begin{subfigure}{.3\linewidth}
  \centering
  \fbox{\includegraphics[width=3.5cm]{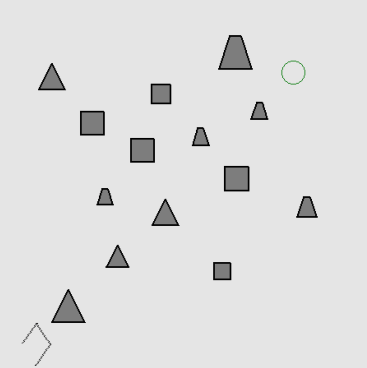}}
  \caption{Minimum reward}
\end{subfigure}
\begin{subfigure}{.3\linewidth}
  \centering
  \fbox{\includegraphics[width=3.5cm]{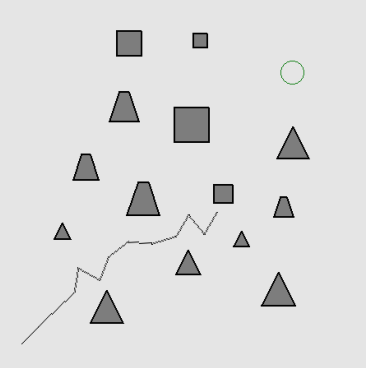}}
  \caption{Random}
\end{subfigure}
\caption{PPO trajectory examples}
\end{figure}

\begin{figure}[t]
\centering
\begin{subfigure}{.3\linewidth}
  \centering
  \fbox{\includegraphics[width=3.5cm]{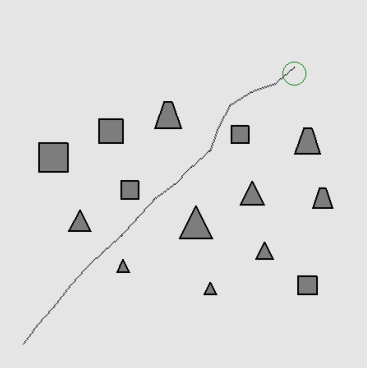}}
  \caption{Maximum reward}
\end{subfigure}%
\begin{subfigure}{.3\linewidth}
  \centering
  \fbox{\includegraphics[width=3.5cm]{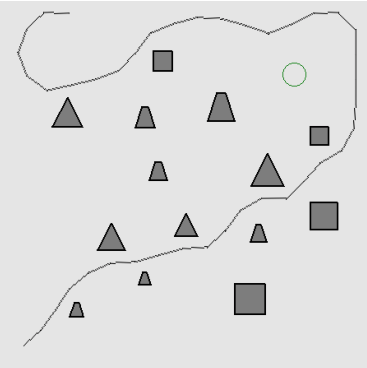}}
  \caption{Minimum reward}
\end{subfigure}
\begin{subfigure}{.3\linewidth}
  \centering
  \fbox{\includegraphics[width=3.5cm]{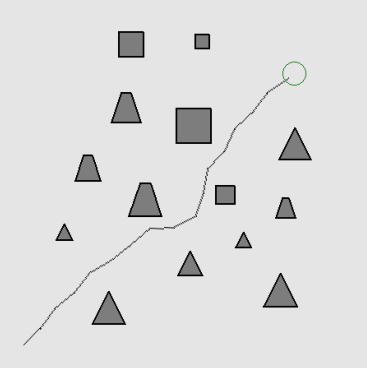}}
  \caption{Random}
\end{subfigure}
\caption{PPO with projection trajectory examples}
\end{figure}

\begin{figure}[t]
\centering
\begin{subfigure}{.3\linewidth}
  \centering
  \fbox{\includegraphics[width=3.5cm]{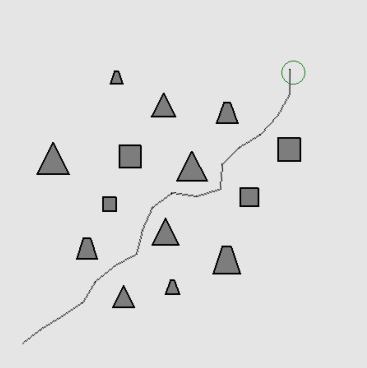}}
  \caption{Maximum reward}
\end{subfigure}%
\begin{subfigure}{.3\linewidth}
  \centering
  \fbox{\includegraphics[width=3.5cm]{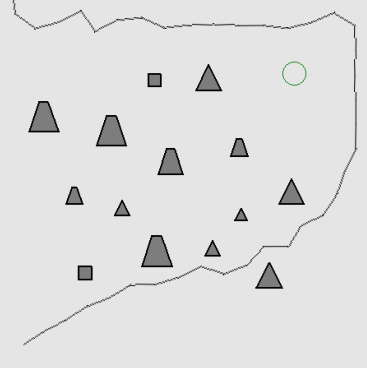}}
  \caption{Minimum reward}
\end{subfigure}
\begin{subfigure}{.3\linewidth}
  \centering
  \fbox{\includegraphics[width=3.5cm]{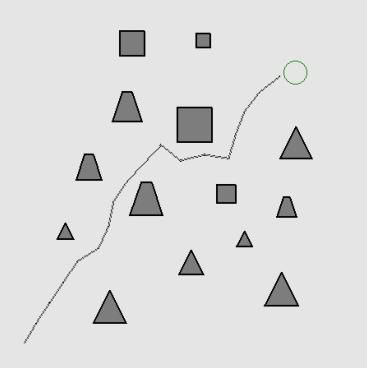}}
  \caption{Random}
\end{subfigure}
\caption{PPO with continuous masking trajectory examples}
\end{figure}

\begin{figure}[t]
\centering
\begin{subfigure}{.3\linewidth}
  \centering
  \fbox{\includegraphics[width=3.5cm]{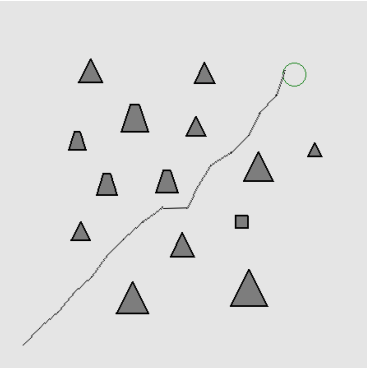}}
  \caption{Maximum reward}
\end{subfigure}%
\begin{subfigure}{.3\linewidth}
  \centering
  \fbox{\includegraphics[width=3.5cm]{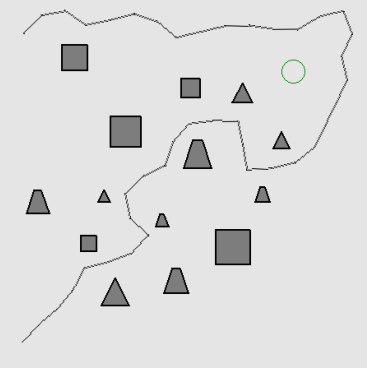}}
  \caption{Minimum reward}
\end{subfigure}
\begin{subfigure}{.3\linewidth}
  \centering
  \fbox{\includegraphics[width=3.5cm]{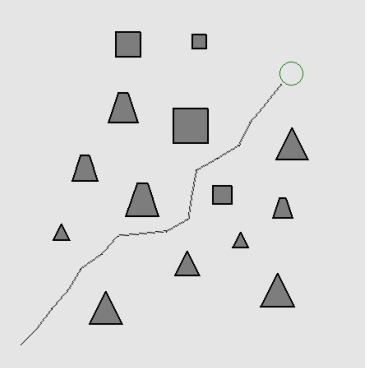}}
  \caption{Random}
\end{subfigure}
\caption{PPO with random replacement trajectory examples}
\end{figure}

\begin{figure}[t]
\centering
\begin{subfigure}{.3\linewidth}
  \centering
  \fbox{\includegraphics[width=3.5cm]{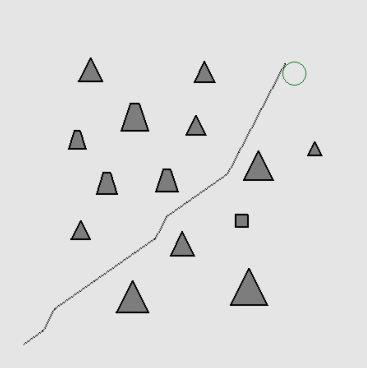}}
  \caption{Maximum reward}
\end{subfigure}%
\begin{subfigure}{.3\linewidth}
  \centering
  \fbox{\includegraphics[width=3.5cm]{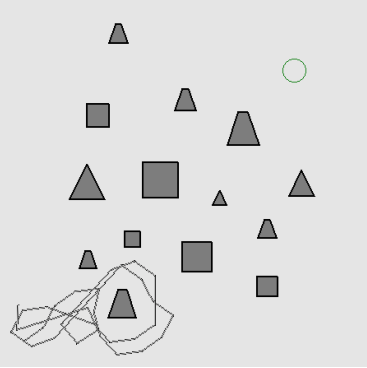}}
  \caption{Minimum reward}
\end{subfigure}
\begin{subfigure}{.3\linewidth}
  \centering
  \fbox{\includegraphics[width=3.5cm]{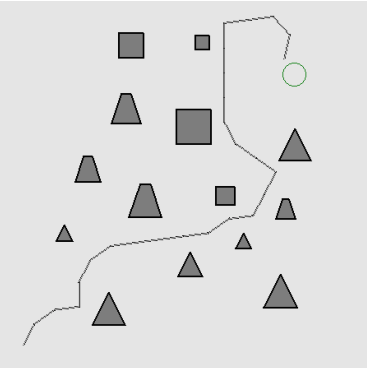}}
  \caption{Random}
\end{subfigure}
\caption{PPO with discrete masking trajectory examples}
\end{figure}

\begin{figure}[t]
\centering
\begin{subfigure}{.3\linewidth}
  \centering
  \fbox{\includegraphics[width=3.5cm]{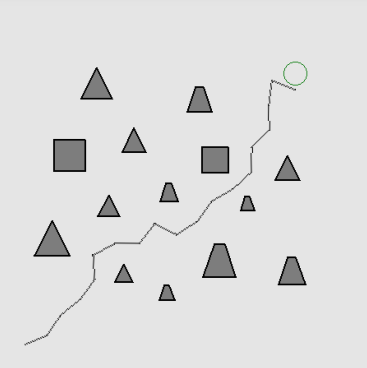}}
  \caption{Maximum reward}
\end{subfigure}%
\begin{subfigure}{.3\linewidth}
  \centering
  \fbox{\includegraphics[width=3.5cm]{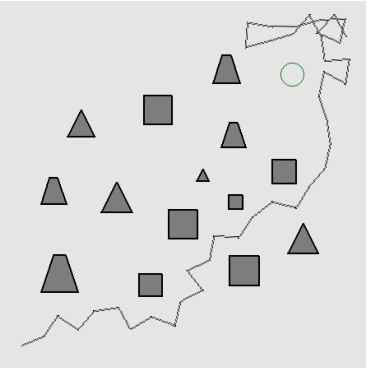}}
  \caption{Minimum reward}
\end{subfigure}
\begin{subfigure}{.3\linewidth}
  \centering
  \fbox{\includegraphics[width=3.5cm]{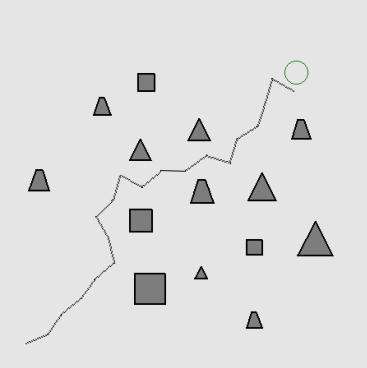}}
  \caption{Random}
\end{subfigure}
\caption{MPS-TD3 with random replacement trajectory examples}
\end{figure}

\begin{figure}[t]
\centering
\begin{subfigure}{.3\linewidth}
  \centering
  \fbox{\includegraphics[width=3.5cm]{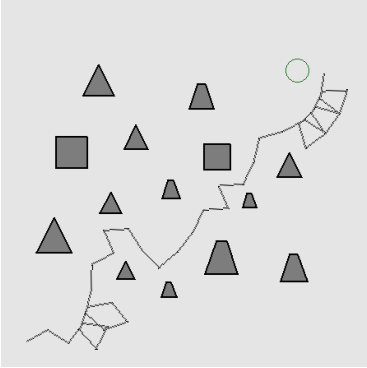}}
  \caption{Maximum reward}
\end{subfigure}%
\begin{subfigure}{.3\linewidth}
  \centering
  \fbox{\includegraphics[width=3.5cm]{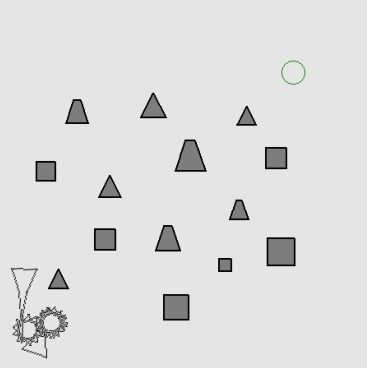}}
  \caption{Minimum reward}
\end{subfigure}
\begin{subfigure}{.3\linewidth}
  \centering
  \fbox{\includegraphics[width=3.5cm]{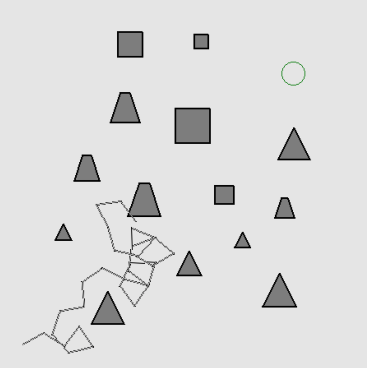}}
  \caption{Random}
\end{subfigure}
\caption{PAM trajectory examples}
\label{app:exp2_3_trajectory_end}
\end{figure}

\begin{figure}[h]
\centering
\begin{subfigure}[b]{\linewidth}
  \centering
   \importpgf{experiment_2/evaluation_2/reward_t}{on_policy_t_reward.pgf}
  \caption{On-policy}
\end{subfigure}
\\[3ex]
\begin{subfigure}[b]{\linewidth}
  \centering
  \importpgf{experiment_2/evaluation_2/reward_t}{on_policy_t_reward.pgf}
  \caption{Off-policy}
\end{subfigure}
\caption{Welch's t-test p-values between average returns}
\label{app:exp2_3_test_reward}
\end{figure}

\begin{figure}[h]
\centering
\begin{subfigure}[b]{\linewidth}
  \centering
  \importpgf{experiment_2/evaluation_2/reward_t}{on_policy_t_steps.pgf}
  \caption{On-policy}
\end{subfigure}
\\[3ex]
\begin{subfigure}[b]{\linewidth}
  \centering
  \importpgf{experiment_2/evaluation_2/reward_t}{off_policy_t_steps.pgf}
  \caption{Off-policy}
\end{subfigure}
\caption{Welch's t-test p-values between average steps}
\label{app:exp2_3_test_steps}
\end{figure}

\begin{figure}[h]
\centering
\begin{subfigure}[b]{\linewidth}
  \centering
  \importpgf{experiment_2/evaluation_2/reward_t}{on_policy_t_solved.pgf}
  \caption{On-policy}
\end{subfigure}
\\[3ex]
\begin{subfigure}[b]{\linewidth}
  \centering
  \importpgf{experiment_2/evaluation_2/reward_t}{off_policy_t_solved.pgf}
  \caption{Off-policy}
\end{subfigure}
\caption{Welch's t-test p-values between the fractions of solved environments}
\label{app:exp2_3_test_solved}
\end{figure}

\clearpage

\pagestyle{empty}

\section*{Statement of Authorship}
I hereby declare that the paper presented is my own work and that I have not called upon the help of a third
party. In addition, I declare that neither I nor anybody else has submitted this paper or parts of it to obtain
credits elsewhere before. I have clearly marked and acknowledged all quotations or references that have been taken from the works of others. All secondary literature and other sources are marked and listed in the bibliography. The same applies to all charts, diagrams and illustrations as well as to all Internet resources. Moreover, I consent to my paper being electronically stored and sent anonymously in order to be checked for plagiarism. I am aware that if this declaration is not made, the paper may not be graded.
\\
\\

\noindent
Mannheim, 03.04.2023 \hspace{4cm} Signature

\end{document}